\pdfoutput=1

\documentclass[11pt]{article}

\usepackage[preprint]{acl}

\usepackage{times}
\usepackage{latexsym}
\usepackage{comment}
\usepackage[T1]{fontenc}

\usepackage[utf8]{inputenc}

\usepackage{microtype}

\usepackage{inconsolata}

\usepackage{graphicx}
\usepackage{multirow}
\usepackage{booktabs}
\usepackage[most]{tcolorbox}

\definecolor{lightgray}{gray}{0.95}

\usepackage{etoolbox}
\AtBeginEnvironment{tcolorbox}{\small}

\definecolor{DR}{HTML}{B3CDE0} 
\definecolor{FI}{HTML}{CCEBC5} 
\definecolor{EK}{HTML}{FBB4AE} 

\definecolor{DDR}{HTML}{639CB4} 
\definecolor{DFI}{HTML}{74B288} 
\definecolor{DEK}{HTML}{F06A6A} 

\newcommand{\drc}[1]{\textcolor{DDR}{#1}}
\newcommand{\fic}[1]{\textcolor{DFI}{#1}}
\newcommand{\ekc}[1]{\textcolor{DEK}{#1}}

\usepackage{subcaption}

\usepackage{framed}
\usepackage{enumitem}

\usepackage{color,soul}

\DeclareRobustCommand{\hldr}[1]{{\sethlcolor{DR}\hl{#1}}}
\DeclareRobustCommand{\hlfi}[1]{{\sethlcolor{FI}\hl{#1}}}
\DeclareRobustCommand{\hlek}[1]{{\sethlcolor{EK}\hl{#1}}}

\newcommand{\llama}{Llama3.2 }
\newcommand{\gemini}{Gemini }
\newcommand{\gpt}{GPT-4o }
\newcommand{\gptm}{GPT-4o-mini }

\providecolor{YellowGreen}{rgb}{0.6, 0.8, 0.2}
\providecolor{YellowOrange}{rgb}{1.0, 0.75, 0.0}

\title{AI Analyst: Framework and Comprehensive Evaluation of Large Language Models for Financial Time Series Report Generation}

\author{
 \textbf{Elizabeth Fons\textsuperscript{*,1}},
 \textbf{Elena Kochkina\textsuperscript{*,1}},
 \textbf{Rachneet Kaur\textsuperscript{1}},
 \textbf{Zhen Zeng\textsuperscript{1}},
\\
 \textbf{Berowne Hlavaty\textsuperscript{2}},
 \textbf{Charese Smiley\textsuperscript{1}},
 \textbf{Svitlana Vyetrenko\textsuperscript{1}},
 \textbf{Manuela Veloso\textsuperscript{1}},
\\
 \textsuperscript{1}J.P. Morgan AI Research,
 \textsuperscript{2}J.P. Morgan Chase
 \\
 \textsuperscript{*}\textit{Equal Contribution}
}

\begin{document}
\maketitle
\begin{abstract}
This paper explores the potential of large language models (LLMs) to generate financial reports from time series data. We propose a framework encompassing prompt engineering, model selection, and evaluation. We introduce an automated highlighting system to categorize information within the generated reports, differentiating between insights derived directly from time series data, stemming from financial reasoning, and those reliant on external knowledge. This approach aids in evaluating the factual grounding and reasoning capabilities of the models.  Our experiments, utilizing both data from the real stock market indices and synthetic time series, demonstrate the capability of LLMs to produce coherent and informative financial reports. 
\end{abstract}

\section{Introduction}

In today’s fast-moving markets, stakeholders such as investors, portfolio managers, and financial analysts rely on financial reports to interpret trends, risks, and opportunities and make informed investment decisions.  However, writing timely, high-quality financial reports is labor-intensive and requires deep expertise in financial analysis, making it an ideal candidate for the application of Natural Language Processing (NLP) to foster automation.  While, recent studies demonstrate the proficiency of large language models (LLMs)~\citep{phi3,openai2023gpt4,team2023gemini,llama2} in processing financial texts and time series data,  \citep{fons2024evaluating}, the task of generating reports from time series remains understudied within NLP. This paper focuses on the financial domain, however, the techniques detailed here can be applied to other fields that use time series such as healthcare, transportation, and climate science.
\begin{figure}[t]
    \centering
    \includegraphics[width=0.95\columnwidth]{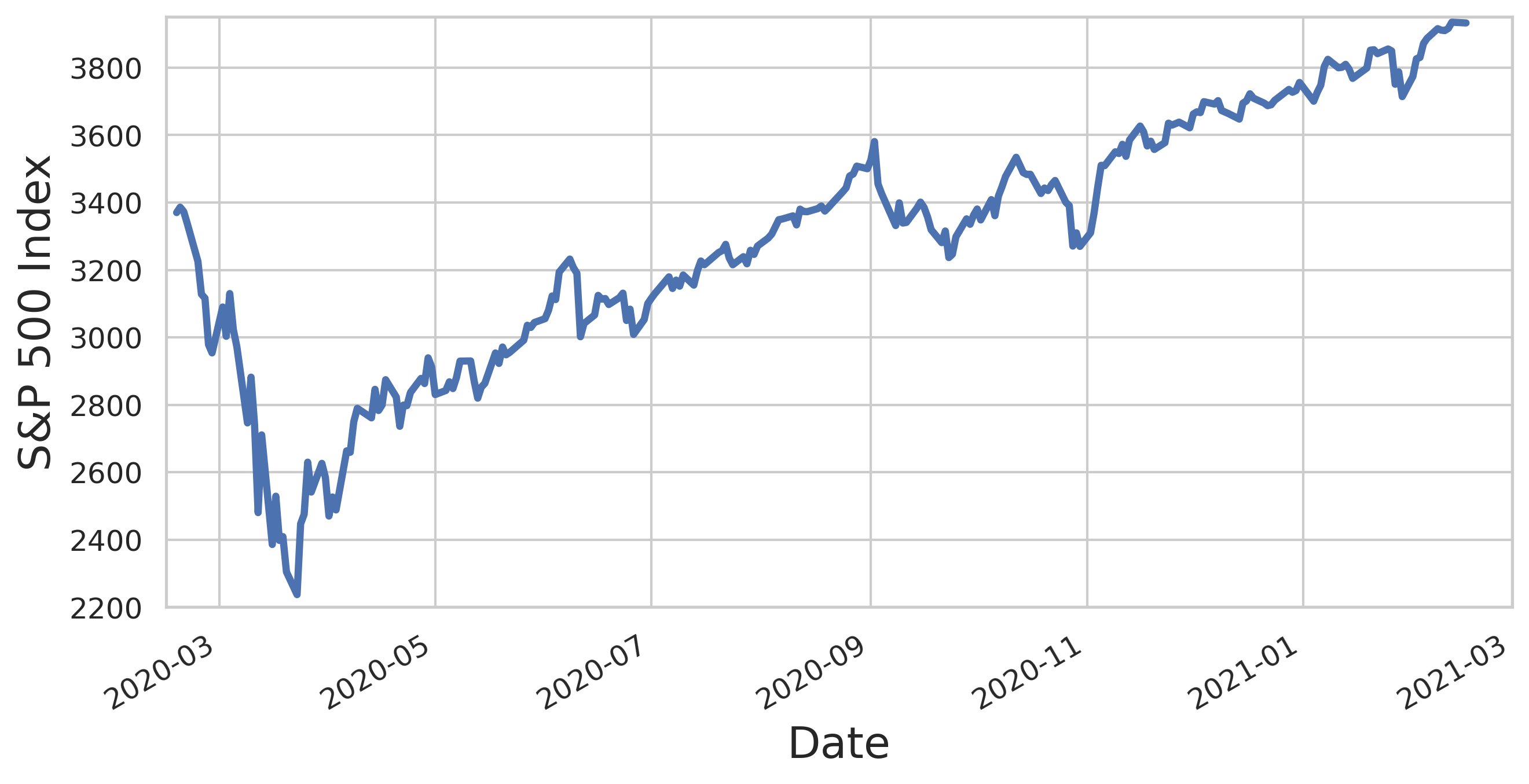}
    \begin{tcolorbox}[colframe=black, colback=white, boxrule=0.5pt, width=\columnwidth, arc=1mm, auto outer arc]
        \scriptsize
        \textbf{Short report}: 
        \hldr{The period from February 2020 to February 2021} \hlfi{marks a tumultuous yet ultimately bullish trend for the S\&P 500. At the beginning of this interval, the index was at a relatively high point}, \hldr{reaching levels just under 3400}. However, 
        \hlek{with the onset of the COVID-19 pandemic}, \hldr{the index experienced a significant and rapid decline}, \hldr{hitting a low of 2237.4 on March 23, 2020}. 
        This represented \hldr{a precipitous drop of approximately 33\% from its high observed in February 2020}. \hlfi{This sharp decline highlights a period of intense volatility, reflecting market uncertainty} \hlek{as investors grappled with the global implications of the pandemic}. \hldr{Following the March low, the S\&P 500 embarked on a robust recovery}, \hlek{seemingly buoyed by a mix of fiscal stimulus, accommodative monetary policy, and optimism around economic recovery.}
    \end{tcolorbox}
    \vspace{-0.5em}
    \caption{Example of time series and short report generated based on it, with highlights indicating \textbf{\drc{Direct Reference (blue)}}, \textbf{\fic{Financial Interpretation (green)}}, and \textbf{\ekc{External Knowledge (red)}} using our proposed highlighting system.}
    \label{fig:timeseries_report}
    \vspace{-1.5em}
\end{figure}

In this paper, we propose AI Analyst, a framework that leverages LLMs to generate financial reports from time series data. Specifically, experimenting with data from major stock indices—S\&P 500, Nasdaq, Dow Jones Industrial Average, and Nikkei 225 \cite{fred_indices_2024}.
Our framework focuses on prompt engineering and model selection by leveraging automated evaluation metrics. 
As the absence of clear ground truth poses a unique challenge, to assess the quality of the generated reports we employ G-Eval (Liu et al., 2023), an LLM-based evaluation method that has been shown to correlate well with human judgment. While G-Eval has been used in other NLG tasks, our work applies it to the financial domain, where its effectiveness is yet to be verified. As such, we compare the performance of G-Eval to human evaluations, analyzing its effectiveness in assessing consistency, coherence, and fluency in this domain.
Further, we introduce a novel automated highlighting system that categorizes insights based on their source, whether directly from the time series data, derived from financial reasoning, or reliant on external knowledge. This categorization helps to evaluate the factual grounding and reasoning capabilities of the generated reports.
Our contributions are the following:
\begin{itemize}[noitemsep, topsep=0pt, leftmargin=*]
\item We introduce an end-to-end framework for generating financial reports from time series data. 
\item We propose an automated segment source classification and  highlighting system that enhances the interpretability of the generated reports by distinguishing between the different types of reasoning used.
\item We provide a systematic assessment of the proposed framework and its components under different settings (model, prompt and report types) using human evaluations.  
\item We perform extensive linguistic and temporal analysis of the generated reports. We discuss common pitfalls to highlight the areas to be addressed in the future.
\end{itemize}

\begin{figure*}[t]
  \centering
    \includegraphics[width=1\linewidth]
    {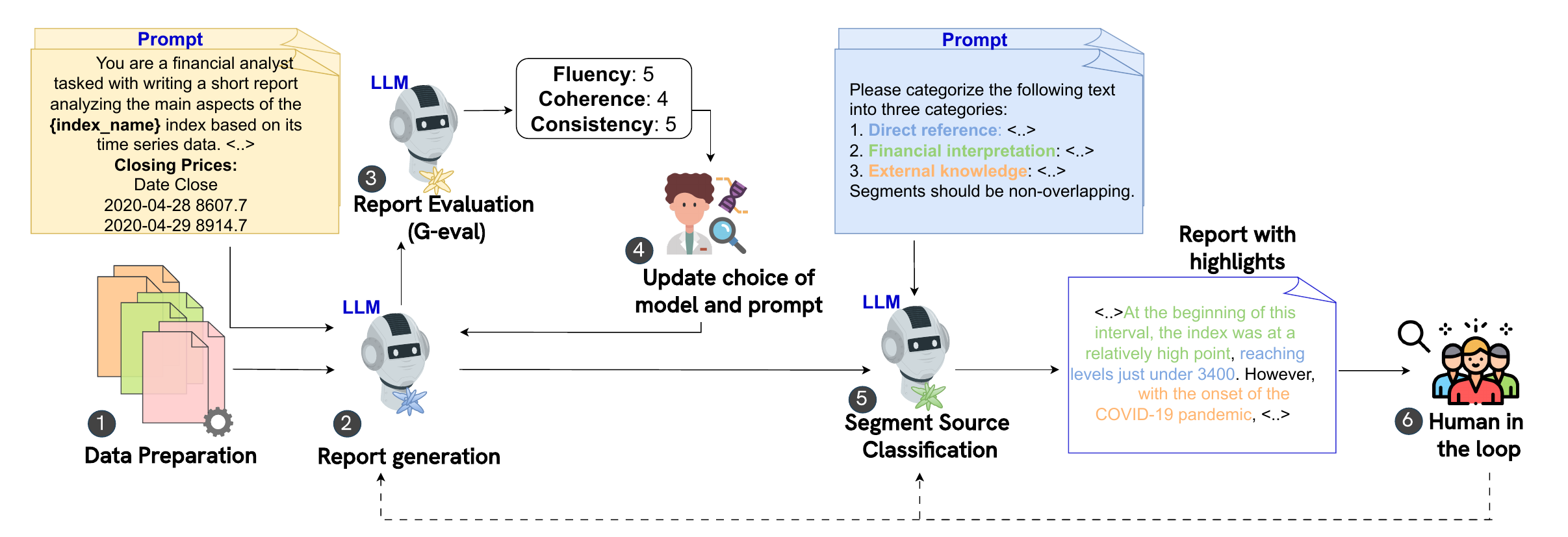}
    \caption{Proposed framework for financial report generation based on time series data}
    \label{fig:diagram}
\end{figure*}
\section{Related Work}

\noindent \textbf{Automated Report Generation}
While generative tasks based on textual data is an active research area~\citep{nishida2023headline,assis2024analysis,liu2024findabenchbenchmarkingfinancialdata}, report generation based on time series data remains underexplored. \citet{liu2024findabenchbenchmarkingfinancialdata} address this by generating professional data analysis plans from chart data in Chinese, which are then manually evaluated by experts. However, their task differs from ours as it focuses on creating analysis plans rather than reports based on chart analysis.
Several studies employ LLMs for stock market predictions. For example, \citet{yang2023fingpt} introduced FinGPT Forecaster,
which uses market news and basic financial data to predict stock price movements and provide analysis summaries. This approach, however, still relies heavily on textual data. Similarly, \citet{li2024alphafin} use LLMs for stock trend prediction and generate reports to justify these predictions, with human annotators evaluating the model's outputs.
In our work, we focus on using time series data to leverage LLMs for generating informative reports for human users, rather than making decisions or predictions. We also explore evaluation strategies in the absence of gold standard answers.

\noindent \textbf{Generation Evaluation}
Evaluating the quality of generations is a critical aspect of natural language generation (NLG) research. 
The majority of works on report generation~\cite{messina2022survey,nishida2023headline,sloan2024automated,assis2024analysis} apply traditional evaluation metrics such as BLEU~\citep{papineni2002bleu}, ROUGE~\citep{lin2004rouge}, and METEOR~\citep{banerjee2005meteor}. These metrics, while well-established and easy to compute, require access to ground truth answers, which are not available to us in this work.  
Moreover, they may also lack the sensitivity to account for the specific requirements of financial reporting, such as factual accuracy and domain-specific terminology. 
Human evaluation is another major approach~\citep{messina2022survey,li2024alphafin,chiang2024badge} involving human judges who assess various aspects of the generated reports, such as fluency, factual accuracy, and grammatical correctness. While this is considered the gold standard, it is time-consuming and expensive, requiring significant human resources. Due to this, here we provide human-evaluation on a subset of the data.

Recently, LLM-based evaluation methods have been proposed as a suitable alternative in the absence of ground truth~\citep{gao2024llm}. 
\citet{liu2023g} introduced G-eval, an LLM-based metric that aligns well with human judgment. This approach provides a scalable and consistent evaluation compared to traditional approaches. Therefore we adopt it in this work.

\noindent \textbf{LLMs in Finance.}
The use of LLMs in Finance is a growing area of research, aiding automation of a wide range of tasks such as  text classification tasks, time series, financial reasoning, and agent-based modelling~\cite{nie2024survey,li2023chatgpt}. 
For example, \citet{callanan2023can} showed that GPT-4 is likely to pass a professional CFA exam. \citet{aguda2024large} demonstrated that LLMs can be used for financial data annotation outperforming untrained crowd-workers. BloombergGPT~\cite{wu2023bloomberggpt} enabled a finance-specialized LLM by training an LLM on extensive financial data. 
With the increased interest in their multimodal capabilities, LLMs are rapidly being applied to time series analysis, particularly in finance.
Recent works explores using LLMs for forecasting~\citep{xue2023promptcast, yu2023temporal} as well as imputation and classification~\citep{zhou2023one}. \citet{kawarada2024prompting} have demonstrated prompting LLMs with time-series information to obtain market comments, while \citet{fons2024evaluating} evaluate LLM understanding on time series across a taxonomy of time series features.

\section{Methodology}

Our proposed framework for generating financial reports from time series is shown in  Figure \ref{fig:diagram}.

\subsection{Data Preparation} 
The first step of the framework is the data preparation. We test two types of data: real and synthetic. 

\noindent \textbf{Real indices} We compile a comprehensive dataset spanning five years (2019--2024), encompassing the S\&P 500, Nasdaq, Dow Jones Industrial Average, and Nikkei 225 indices. The data is divided into overlapping one-year windows with a one-month stride, facilitating the analysis of both short-term and longer-term trends. For each one-year window, we compute a set of standard technical indicators, including the 50-day Simple Moving Average (SMA), 50-day Exponential Moving Average (EMA), Volatility rolling window, Relative Strength Index (RSI)~\cite{murphy1999technical}, Moving Average Convergence Divergence (MACD)~\cite{macd}, Bollinger Bands~\cite{murphy1999technical} and Fibonacci Retracement levels~\cite{Malkiel1973}. 

\noindent \textbf{Synthetic indices}
We generate synthetic time series data using Geometric Brownian Motion  that simulate five years of daily stock prices for two distinct periods: 2019-2024, aligning with the period of real-world data, and 2024-2029, extending beyond the LLMs' training data. This entirely unseen data enables a robust evaluation of the models' generalization and time series processing capabilities.

\subsection{Report Generation}
Given financial time series that contain the close price, and/or technical indicators, the LLM is prompted to describe and summarize the patterns and trends observed in the data in an analyst report style. We generate two types of reports: (1) \textbf{short}, a 1-paragraph high level summary of the main trends given the close price over the observed period, and (2) \textbf{technical indicator reports (TI)}, 2-3 paragraphs covering the main patterns as well as conclusions made from analyzing the close price and technical indicators provided.  

As shown in Figure \ref{fig:diagram} report generation depends on the choice of the model and prompt formulation and can be determined through the iterations that incorporate feedback from either or both automated evaluation and human users or experts. In this work we present experiments with a set of various prompt formats and LLMs from different families.  

\noindent \textbf{Prompt Engineering:}
We developed two main types of prompts tailored to the short and TI reports (Appendix~\ref{sec:app-prompts-gen}). 
The prompt for generating short reports includes a time series of  daily close price and emphasizes a high-level overview of the time series data, guiding the model to focus on key trends, major price movements, and volatility. 

For TI reports, the prompt is designed to incorporate both time series data and technical indicators requiring a more detailed description. We explore two methods for integrating data into the prompt: (1) text-based time series and (2) data plot images.

The resulting report should provide insights into how these indicators signal trends or shifts in the index's performance, thus offering a more technical perspective on the market movements. 
\noindent \textbf{Models:} We evaluate our framework with GPT-4o, GPT-4o-mini~\cite{openai2023gpt4}, Gemini~\cite{team2023gemini}, LLama3.2-Instruct~\cite{dubey2024llama} and Phi-3~\cite{phi3} models. 
These models are chosen for their multimodal abilities, allowing integration of text and visuals like technical indicator plots. The model selection offers diversity in model size and family.

\subsection{Report Evaluation}
We evaluate LLM-generated reports using an automated approach and human experts. This feedback can be used to improve the generation parameters as shown in stages 3, 4 and 6 in Figure~\ref{fig:diagram}.

\noindent \textbf{Automated Evaluation}
In the absence of ground truth reports, we adapt the G-Eval framework for automated evaluation~\citep{liu2023g}, using GPT-4o as the evaluator. Originally, G-Eval assesses summaries against source texts. We adapt it for financial report generation from time series data, using the data itself as the "source". GPT-4o receives a prompt with task and evaluation criteria definitions (see Appendix \ref{sec:app-geval-prompts}), and scores the report on a 1-5 scale based on three key criteria below. 

\begin{itemize}[noitemsep, topsep=0pt, leftmargin=*]
    \item \textbf{Consistency}: ensures factual accuracy by comparing the report to the original time series data, penalizing discrepancies.
    \item \textbf{Coherence}: assesses logical flow and clarity, focusing on structure and transitions. 
    \item \textbf{Fluency}: evaluates grammar, clarity, and readability. 
\end{itemize}
 
\begin{table*}[t]
\centering
\resizebox{\linewidth}{!}{
\begin{tabular}{ll|ccc|ccc||ccc|ccc}
\toprule
& & \multicolumn{6}{c||}{\textbf{Real Data}} & \multicolumn{6}{c}{\textbf{Synthetic Data}} \\
\midrule
& & \multicolumn{3}{c|}{\textit{G-Eval}} & \multicolumn{3}{c||}{\textit{Human scores}} & \multicolumn{3}{c|}{\textit{G-Eval}} & \multicolumn{3}{c}{\textit{Human scores}}\\
\textbf{\shortstack{Report \\ type}} & \textbf{Model} & \textbf{Con} & \textbf{Coh} & \textbf{Flu}  & \textbf{Con} & \textbf{Coh} & \textbf{Flu}
& \textbf{Con} & \textbf{Coh} & \textbf{Flu}  & \textbf{Con} & \textbf{Coh} & \textbf{Flu}\\ 
\midrule
\multirow{4}{*}{Short} 
    & GPT-4o       & \textbf{3.77 $\pm$ 0.46} & \textbf{4.12 $\pm$ 0.15} & \textbf{4.99 $\pm$ 0.05} 
    &   3.83 ± 0.43 &   \textbf{4.33 ± 0.54} &     \textbf{5.0 ± 0.0}
    & \textbf{3.66 $\pm$ 0.42} & \textbf{4.06 $\pm$ 0.15} & 4.96 $\pm$ 0.12 
    & 4.33 $\pm$ 0.47 & \textbf{4.83 $\pm$ 0.33} & \textbf{5.0 $\pm$ 0.0} 
    \\
    & GPT-4-mini   & 3.47 $\pm$ 0.46 & 4.01 $\pm$ 0.16 & 4.98 $\pm$ 0.05  
    &   \textbf{3.96 ± 0.86} &   4.08 ± 0.57 &   4.92 ± 0.17  
    & 3.44 $\pm$ 0.41 & 3.95 $\pm$ 0.18 & \textbf{4.97 $\pm$ 0.09}
     & 4.0 $\pm$ 0.86 & 4.58 $\pm$ 0.63 & \textbf{5.0 $\pm$ 0.0}
    \\
    & Gemini       & 3.40 $\pm$ 0.44 & 3.94 $\pm$ 0.24 & \textbf{4.99 $\pm$ 0.06}  
    &    3.5 ± 0.79 &   4.17 ± 0.19 &   4.75 ± 0.17
    & 3.30 $\pm$ 0.45 & 3.87 $\pm$ 0.24 & 4.96 $\pm$ 0.07
    & \textbf{4.75 $\pm$ 0.17} & 4.58 $\pm$ 0.32 & \textbf{5.0 $\pm$ 0.0} 
    \\
    & Phi-3        & 2.46 $\pm$ 0.40 & 2.94 $\pm$ 0.42 & 4.48 $\pm$ 0.42 
    &    2.5 ± 0.43 &    2.5 ± 0.19 &   3.75 ± 0.32 
    & 2.28 $\pm$ 0.37 & 2.60 $\pm$ 0.47 & 4.25 $\pm$ 0.55 
    & 2.0 $\pm$ 0.0 & 2.08 $\pm$ 0.16 & \textbf{5.0 $\pm$ 0.0} 
    \\
    & \llama       & 2.57 $\pm$ 0.60 & 3.27 $\pm$ 0.64 & 4.71 $\pm$ 0.58 
    &    3.66 ± 0.40 &    4.5 ± 0.28 &   4.91 ± 0.14 
    & 2.75 $\pm$ 0.47 & 3.45 $\pm$ 0.44 & 4.77 $\pm$ 0.46
    & 3.33 $\pm$ 0.47 & 3.25 $\pm$ 0.32 & 4.75 $\pm$ 0.17 \\
\midrule
\multirow{4}{*}{TI}          
    & GPT-4o       & \textbf{3.46 $\pm$ 0.37} & \textbf{4.03 $\pm$ 0.13} & 4.96 $\pm$ 0.11 
    &   4.04 ± 0.72 &    \textbf{4.58 ± 0.50} &     \textbf{5.0 ± 0.0}  
    & \textbf{3.38 $\pm$ 0.32} & \textbf{4.00 $\pm$ 0.11} & \textbf{4.97 $\pm$ 0.10}
    & \textbf{4.42 $\pm$ 0.95} & \textbf{4.83 $\pm$ 0.33} & \textbf{5.0 $\pm$ 0.0} 
    \\ 
    & GPT-4-mini   & 3.19 $\pm$ 0.46 & 3.95 $\pm$ 0.15 & 4.98 $\pm$ 0.04 
    &   \textbf{4.17 ± 0.19} &   \textbf{4.58 ± 0.17} &   4.83 ± 0.33  
    & 3.06 $\pm$ 0.36 & 3.90 $\pm$ 0.20 & 4.94 $\pm$ 0.10 
    & 4.25 $\pm$ 0.56 & 4.5 $\pm$ 0.19 & \textbf{5.0 $\pm$ 0.0} 
    \\
    & Gemini       & 2.83 $\pm$ 0.39 & 3.87 $\pm$ 0.26 & \textbf{5.00 $\pm$ 0.01} 
    &   3.04 ± 0.64 &   \textbf{4.58 ± 0.42} &   4.92 ± 0.17  
    & 2.79 $\pm$ 0.36 & 3.72 $\pm$ 0.29 & \textbf{4.97 $\pm$ 0.07} 
    & 4.25 $\pm$ 0.42 & 4.67 $\pm$ 0.0 & \textbf{5.0 $\pm$ 0.0} 
    \\
    & \llama       & 2.38 $\pm$ 0.55 & 3.21 $\pm$ 0.46 & 4.68 $\pm$ 0.51 
    &    3.41 ± 0.49 &   4.08 ± 0.54 &   4.25 ± 0.43  
    & 2.55 $\pm$ 0.41 & 3.32 $\pm$ 0.39 & 4.66 $\pm$ 0.63
    & 3.25 $\pm$ 0.42 & 3.75 $\pm$ 0.56 & 4.58 $\pm$ 0.17
    \\  
\midrule
\multirow{4}{*}{\shortstack{TI \\ (plots)}} 
    & GPT-4o       & \textbf{3.48 $\pm$ 0.36
    } & \textbf{4.00 $\pm$ 0.11} & 4.97 $\pm$ 0.08  
    &   3.83 ± 0.58 &   \textbf{4.92 ± 0.17} &   4.92 ± 0.17  
    & \textbf{3.39 $\pm$ 0.38} & \textbf{3.95 $\pm$ 0.15} & \textbf{4.98$\pm$ 0.04} 
    & 4.25 $\pm$ 1.01 & \textbf{4.83 $\pm$ 0.33} & \textbf{5.0$\pm$ 0.0}
    \\ 
    & GPT-4-mini   & 3.09 $\pm$ 0.43 & 3.91 $\pm$ 0.16 & 4.99 $\pm$ 0.04
     &    3.92 ± 0.5 &   4.75 ± 0.32 &   4.92 ± 0.17
     & 3.10 $\pm$ 0.40 & 3.80 $\pm$ 0.22 & 4.93 $\pm$ 0.15  
     & \textbf{4.58 $\pm$ 0.42} & 4.42 $\pm$ 0.22 & \textbf{5.0 $\pm$ 0.0} \\
    & Gemini       & 3.10 $\pm$ 0.43 & 3.86 $\pm$ 0.23 & \textbf{5.00 $\pm$ 0.01} 
    &   3.75 ± 0.63 &   4.67 ± 0.27 &     \textbf{5.0 ± 0.0}
    & 2.96 $\pm$ 0.45 & 3.74 $\pm$ 0.26 & 4.96 $\pm$ 0.10  
    & 4.0 $\pm$ 0.14 & 4.58 $\pm$ 0.17 & \textbf{5.0 $\pm$ 0.0}
    \\
    & \llama       & 2.20 $\pm$ 0.53 & 2.83 $\pm$ 0.51 & 4.24 $\pm$ 0.74
    &   \textbf{4.0 ± 0.62} &  4.33 ± 0.47 &   4.41 ± 0.43 
    & 2.24 $\pm$ 0.50 & 2.82 $\pm$ 0.47 & 4.24 $\pm$ 0.72 
    & 3.58 $\pm$ 0.50 & 3.33 $\pm$ 0.98 & 4.16 $\pm$ 0.64 
    \\ 
\bottomrule
\end{tabular}
}
\caption{Comparative analysis of report quality (for both real and synthetic data reports) across diverse models and report types using G-Eval scores, which assess Consistency (Con), Coherence (Coh), and Fluency (Flu), alongside corresponding Human Evaluation scores from expert annotators. The best model for each report type is highlighted in \textbf{bold}.}
\label{tab:geval_scores_real_top}
\end{table*}
\paragraph{Human Evaluation}
To verify the reliability of our automated evaluation approach and address potential biases of using GPT-4o as evaluator, we conduct human evaluation of a sample of reports, limited by the cost of financial experts. 
Human evaluation follows G-Eval's dimensions, with annotators scoring each report from 1 to 5 for consistency, coherence, and fluency. Each report is assessed by 3 annotators.
For the real data reports, we focus on the S\&P 500 index, as it has the best G-Eval scores from Section~\ref{res:geval}, sampling 4 reports per model and report/prompt type. For the synthetic data reports, we evaluate reports based on both of the synthetic indices (`past' and `future' time spans), sampling 2 reports per model and report/prompt type.  
We then check the alignment between the scores assigned by annotators with those assigned by GPT-4o as part of the G-Eval process.

\subsection{Segment Source Classification}
As the final step in our framework, we suggest categorizing report segments based on the type of information or reasoning used to generate them. We begin by dividing each report into sentences, which are then processed by an LLM (GPT-4o) using a prompt (Appendix~\ref{app:highlightprompt}) to identify sentence segments and their respective categories. We define three categories:
\begin{itemize}[noitemsep, topsep=0pt, leftmargin=*]
    \item 
    \textbf{Direct Reference (DR):} segments that directly cite specific data points or trends in the time series, like index values, dates, or changes.
    \item 
    \textbf{Financial Interpretation (FI):} segments offering analysis or inference on financial data without relying on external input, such as market trends or fluctuation explanations.
    \item 
    \textbf{External Knowledge (EK):} segments referencing information outside the time series, e.g. economic factors, geopolitical or industry events.
\end{itemize}
Figure \ref{fig:timeseries_report} presents an example of a highlighted report.
This highlighting system aims to facilitate the revision process for end users. It enables the quick identification of areas where the generated report may deviate from the factual basis of the input time series, thus promoting greater reliability of the final output.

\paragraph{Evaluation}
To evaluate the quality of segmentation and segment categorization, we  sample a set of 52 reports for manual evaluation with one expert annotator. 
The annotator then goes through each segment in the report and assigns a correct label to the segment. This allows us to the compute overall and per-class performance metrics such as accuracy, precision and recall. The annotator also indicates if the given segment should be split into two or more categories as it contains several labels. 

\begin{table}[t]
\centering
\resizebox{0.6\columnwidth}{!}{
    \begin{tabular}{l|ccc}
        \toprule
        & \textbf{Con} & \textbf{Coh} & \textbf{Flu} \\
        \midrule
        \textbf{Spearman ($\rho$)} & 0.33 & 0.57 & 0.22 \\
        \textbf{Kendall-Tau ($\tau$)} & 0.23 & 0.44 & 0.18 \\
        \bottomrule
    \end{tabular}
    }
\caption{Spearman and Kendall-Tau, capturing the alignment between G-Eval and Human Evaluation.}
\label{tab:correlations}
\end{table}
\section{Results and Analysis}
We show sample reports generated by different models in Appendix~\ref{app:repexampels} (Figure~\ref{fig:qual_real} and Figure~\ref{fig:qual_syn}).
\subsection{Report Quality}
\paragraph{G-Eval and Human Evaluation}
\label{res:geval}
Table~\ref{tab:geval_scores_real_top} shows the evaluation scores for various models (GPT-4o, GPT-4-mini, Gemini, \llama, and Phi-3) across all report types (Short, TI, and TI (plots)), using both real and synthetic indices. The evaluations were conducted through two methods: G-Eval and Human Evaluation.
Table~\ref{tab:correlations} shows correlation metrics (Spearman and Kendall-Tau) between the two evaluations.

\gpt consistently generated the highest quality reports across all report types and evaluation metrics. 
This model demonstrated superior ability to analyze time series data and produce comprehensive, informative financial reports.  In contrast, Phi-3 exhibited the weakest performance, with reports often lacking in factual accuracy and coherence.  This suggests limitations in Phi-3's capacity to handle complex financial data. Models like \gptm and Gemini generally performed well, with \gptm often slightly outperforming Gemini.  \llama also demonstrated strong performance, particularly in Fluency, indicating its ability to generate readable and grammatically correct reports.

Along G-Eval, Human Evaluation consistently rate GPT-4o as the highest-performing model across all dimensions. 
Despite the generally strong performance of most models in Fluency, human evaluation scores showed higher variability compared to G-Eval, especially for Consistency and Coherence. This discrepancy suggests a degree of subjectivity in human interpretations of these criteria within the financial domain.  While a moderate correlation was observed between G-Eval and human evaluations, with Coherence showing the strongest alignment, the lower correlations for Consistency and Fluency highlight a divergence between automated and subjective human assessments. It is important to acknowledge a potential evaluation bias in G-Eval since it uses GPT-4o as the evaluator. Prior studies have shown that LLM-based evaluation tends to favor generations from the same model family, leading to systematically higher scores for GPT-4o-generated reports \cite{liu2023g}. While human evaluation largely aligns with G-Eval’s ranking, this potential bias should be considered when interpreting the results.

Phi-3's poor performance, particularly its tendency to generate repetitive prose and errors in accurately describing specific data points, led to its exclusion from further experiments. 
This underscores the need for models to synthesize a holistic understanding of the data and generate insightful reports, not just isolated descriptions of specific data points. 
Another issue observed in our evaluation is related to Llama3.2’s context length limitations. The maximum number of tokens that can be generated is 8092, but our TI reports required longer outputs. In these cases, \llama occasionally produced word salad—unstructured and incoherent text—due to truncation issues. To ensure a fair evaluation, we manually filtered out these extreme cases before scoring the outputs. This highlights a key limitation when using models with strict context-length constraints for financial reporting tasks that require longer textual generations.

Further analysis revealed a decline in the Consistency score of short reports for both GPT models for the period after the cutoff date of the models training data, suggesting a potential challenge in accurately interpreting data beyond their training period. We present additional analysis of the evolution of these metrics over time in Appendix \ref{sec:app:geval-time}.

\paragraph{Financial reports from synthetic time series}

Table \ref{tab:geval_scores_real_top} (right) presents the G-Eval scores for synthetic data. Similar to the real data results, GPT-4o consistently achieves the highest scores in consistency and coherence, demonstrating its ability to generate high-quality reports even with unseen financial indices. However, absolute scores for both GPT-4o and GPT-4o-mini are generally lower with synthetic data, indicating a potential challenge in interpreting unfamiliar time series. This is not the case for \llama, whose scores improve on reports generated from synthetic time series.

Interestingly, the performance gap between GPT-4o and GPT-4o-mini is less pronounced with synthetic data, particularly for consistency. This suggests that the larger model's advantages may be less pronounced when analyzing unfamiliar financial patterns. Furthermore, Gemini exhibits a more noticeable performance drop with synthetic data, especially in consistency, implying a potential sensitivity to the nature of the input time series and a stronger reliance on real-world data for robust analysis. The lower overall scores and reduced inter-model differences may be attributed to the challenges of interpreting hypothetical future trends within the synthetic data, which requires a deeper understanding of market dynamics that may be harder to extract from simulated data.

\begin{table*}[]
    \centering
\resizebox{0.99\textwidth}{!}{
\begin{tabular}{llcccccccc|ccccc}
\toprule
                           \textbf{Report type} &        \textbf{Model} &  \textbf{Rep. len} &  \textbf{Sent. len} &  \textbf{TTR (W)} &  \textbf{TTR (A)} &  \textbf{Polarity} &  \textbf{Subjectivity} &  \textbf{Terms} &  \textbf{Readability} &  \textbf{MA} &  \textbf{RSI} &  \textbf{MACD} &  \textbf{BB} &  \textbf{Retracemt.} \\
\midrule
                                 Short &       \gpt&        226.6 &          27.6 &         0.79 &         0.07 &      0.11 &          0.42 &   0.03 &         33.4 &               0.00 &        0.00 &         0.00 &       0.00 &              0.00 \\
                                  Short &  \gpt-mini &        204.7 &          25.8 &         0.80 &         0.06 &      0.11 &          0.42 &   0.03 &         35.8 &               0.01 &        0.00 &         0.00 &       0.00 &              0.00 \\
                                  Short &       \gemini &        150.1 &          24.7 &         0.82 &         0.07 &      0.13 &          0.47 &   0.04 &         38.4 &               0.01 &        0.00 &         0.00 &       0.00 &              0.00 \\
                                 Short &         Phi-3 &        199.0 &          23.7 &         0.53 &         0.02 &      0.24 &          0.52 &   0.01 &         62.5 &               0.00 &        0.00 &         0.00 &       0.00 &              0.00 \\
                                  Short &       \llama &        216.5 &          19.5 &         0.73 &         0.10 &      0.13 &          0.43 &   0.02 &         47.2 &               0.02 &        0.00 &         0.00 &       0.00 &              0.00 \\ \hline
                  TI &       \gpt&        354.2 &          27.2 &         0.73 &         0.06 &      0.09 &          0.41 &   0.05 &         28.7 &               0.87 &        1.00 &         1.00 &       0.57 &              0.00 \\
                   TI &  \gpt-mini &        328.7 &          27.0 &         0.76 &         0.06 &      0.07 &          0.42 &   0.06 &         26.6 &               0.99 &        1.00 &         0.96 &       0.20 &              0.00 \\
                   TI &       \gemini &        291.7 &          22.8 &         0.70 &         0.05 &      0.09 &          0.44 &   0.06 &         35.6 &               0.93 &        1.00 &         1.00 &       0.52 &              0.00 \\
                   TI &        \llama &        452.0 &          11.3 &         0.60 &         0.07 &      0.10 &          0.42 &   0.04 &         42.0 &               0.91 &        0.94 &         0.87 &       0.21 &              0.00 \\ \hline
             TI (plots) &       \gpt&        334.1 &          27.8 &         0.75 &         0.06 &      0.09 &          0.41 &   0.04 &         28.8 &               1.00 &        1.00 &         0.02 &       0.00 &              0.00 \\
             TI (plots) &  \gpt-mini &        320.4 &          26.3 &         0.76 &         0.06 &      0.08 &          0.41 &   0.04 &         29.2 &               1.00 &        1.00 &         0.01 &       0.00 &              0.00 \\
            TI (plots) &       Gemini &        275.7 &          23.5 &         0.71 &         0.05 &      0.10 &          0.44 &   0.07 &         39.6 &               0.94 &        1.00 &         0.94 &       0.84 &              0.21 \\
             TI (plots) &        \llama &        293.4 &          14.8 &         0.71 &         0.12 &      0.08 &          0.41 &   0.04 &         45.2 &               0.52 &        0.61 &         0.45 &       0.58 &              0.28 \\ 
\bottomrule
\end{tabular}
   }
   \caption{Linguistic analysis of all report types: short, technical indicator (TI), technical indicator reports generated using time series and plots (TI (plots)) for real indexes. Table presents average report lengths (Rep. len), sentence lengths (Sent. len), Type-Token Ratio within each report (TTR (w)) and across reports (TTR (a)), sentiment polarity, report subjectivity, proportion of financial terms in the report, Flesch reading ease score (Readability), proportion of reports mentioning each of the technical indicators (MA, RSI, MACD, BB and Retracement).}
\label{tab:linguistic}
\vspace{-1em}
\end{table*}
\subsection{Linguistic Analysis}
We analyze linguistic properties of the generated financial reports, our findings for reports based on real indices are shown in Table~\ref{tab:linguistic} (see full Table with synthetic indices in Appendix~\ref{app:linguist}).  
These are common features that help evaluate the generations by identifying relevant linguistic phenomena, allowing us to detect anomalies like excessive sentiment or subjectivity, which is crucial for maintaining professionalism in formal reports.

While models were instructed to produce 1-2 paragraphs for short reports and 2-3 paragraphs for TI, we observe that models exhibited varying interpretations of ``paragraph length''. GPT-4o and GPT-4o-mini generate the most verbose reports and Gemini producing the most concise. This difference in verbosity was also reflected in average sentence length. \llama is an outlier, producing lengthy reports with shorter sentences. Overall, this variation in report length among the models is  reasonable and aligns with expectations.

To evaluate lexical diversity, we calculated two metrics: Type-Token Ratio for individual reports (TTR (W)), which measures the average proportion of unique words within each report, and Type-Token Ratio across all reports of a given type (TTR (A)), which measures the overall proportion of unique words. While individual reports, especially shorter ones, showed high TTR (W), the diversity across reports (TTR (A)) was consistently low for all models and report types, indicating a tendency towards formulaic, template-like language. This suggests that while each report benefits from rich language, consistency across various reports is prioritized for reliability. Notably, synthetic reports had higher TTR (A) than real reports, likely due to the distinct linguistic features of the two types of synthetic reports (``past'' and ``future'').

Sentiment analysis, conducted using TextBlob\footnote{The polarity score is a float in the range [-1.0, 1.0]. The subjectivity is a float in the range [0.0, 1.0] where 0.0 is very objective and 1.0 is very subjective.}, revealed that most models produced neutral reports. Phi-3 was a notable exception, generating notably more positive and subjective reports.

We use a financial terminology lexicon\footnote{\url{ https://www.iotafinance.com/en/Glossary-of-Financial-Terms.html}} to estimate the amount of terms used in the reports, as an average percentage of terms used out of the total number of words. We found it to be a rather low percentage, though higher for TI reports compared to short reports, as expected.

Readability was evaluated using the Flesch Reading Ease score\footnote{\url{https://pypi.org/project/textstat/}}. Report scores are rather low, reflecting the complex and specialized nature of the content. As anticipated, shorter reports are generally more readable than TI. Short reports by Phi-3 are an exception, achieving a "standard" difficulty rating despite being repetitive and having the lowest quality scores in both human and automated evaluations. This shows the limitations of automated readability assessments. Synthetic reports had slightly higher readability compared to those based on real indices, which may indicate a simplification of the reports based on unseen data.

Finally, we examined the proportion of reports that mention the names of technical indicators provided in the prompt (Table~\ref{tab:linguistic}, right side). Notably, the Moving Average (MA), Relative Strength Index (RSI), and MACD were consistently mentioned across all TI report types, indicating their widespread recognition and usage. We even observe a small percentage of short reports mentioning MA. Bollinger Bands were frequently referenced in TI reports but were less prevalent in reports with plots, demonstrating the influence of report format on indicator mention. Interestingly, Fibonacci retracement levels were scarcely mentioned, appearing only in Gemini and Llama TI reports with plots, highlighting better instruction following by these models. This distribution may reflect the commonality and perceived importance of these indicators in financial analysis, as well as their positioning in the prompt.

\begin{figure}[t]
  \centering
  \begin{subfigure}{0.49\columnwidth}
    \includegraphics[width=\linewidth]{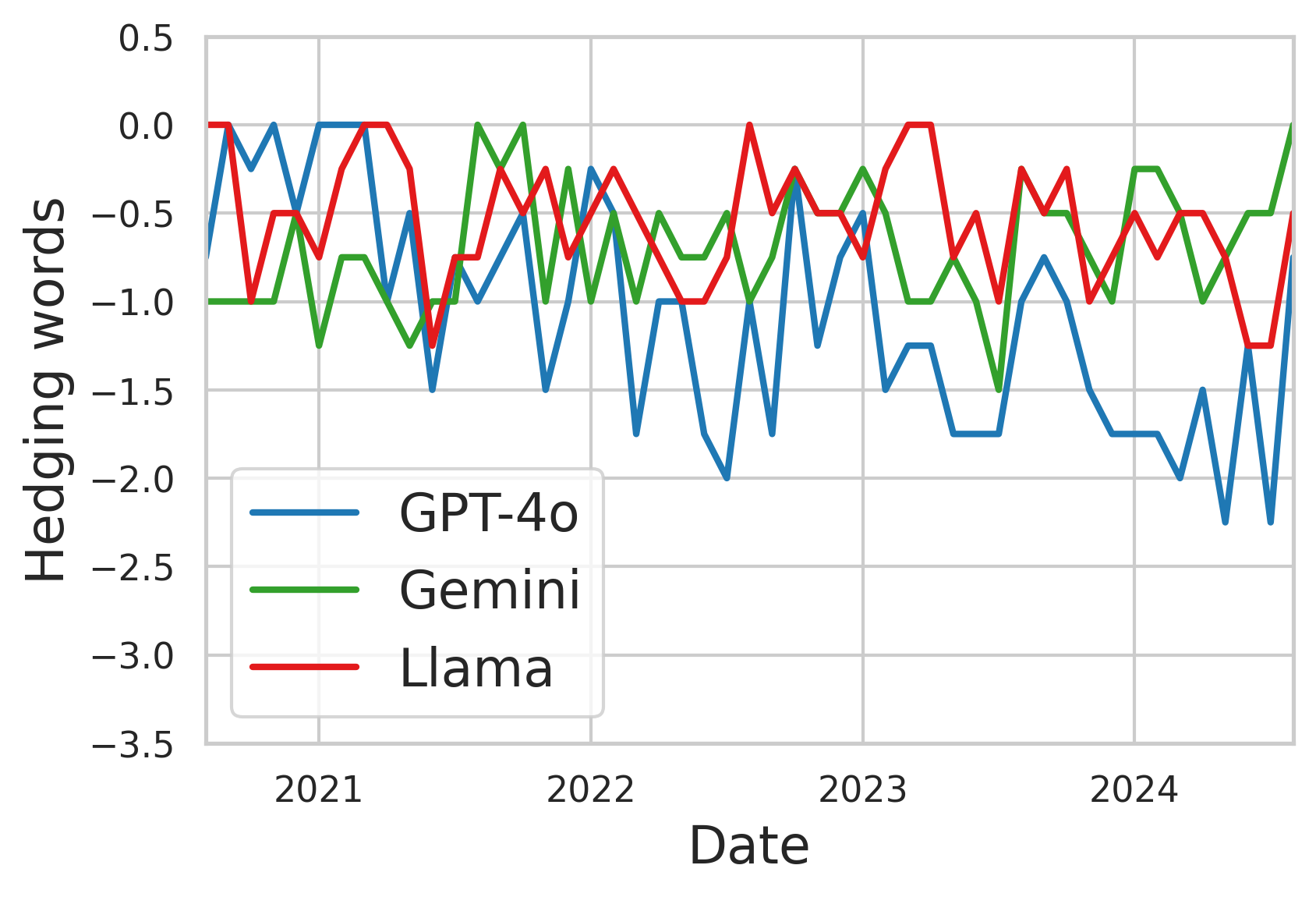}
    \caption{Short reports}
    \vspace{-0.1cm}
  \end{subfigure}
  \vspace{-0.1cm}
  \begin{subfigure}{0.49\columnwidth}
    \includegraphics[width=\linewidth]{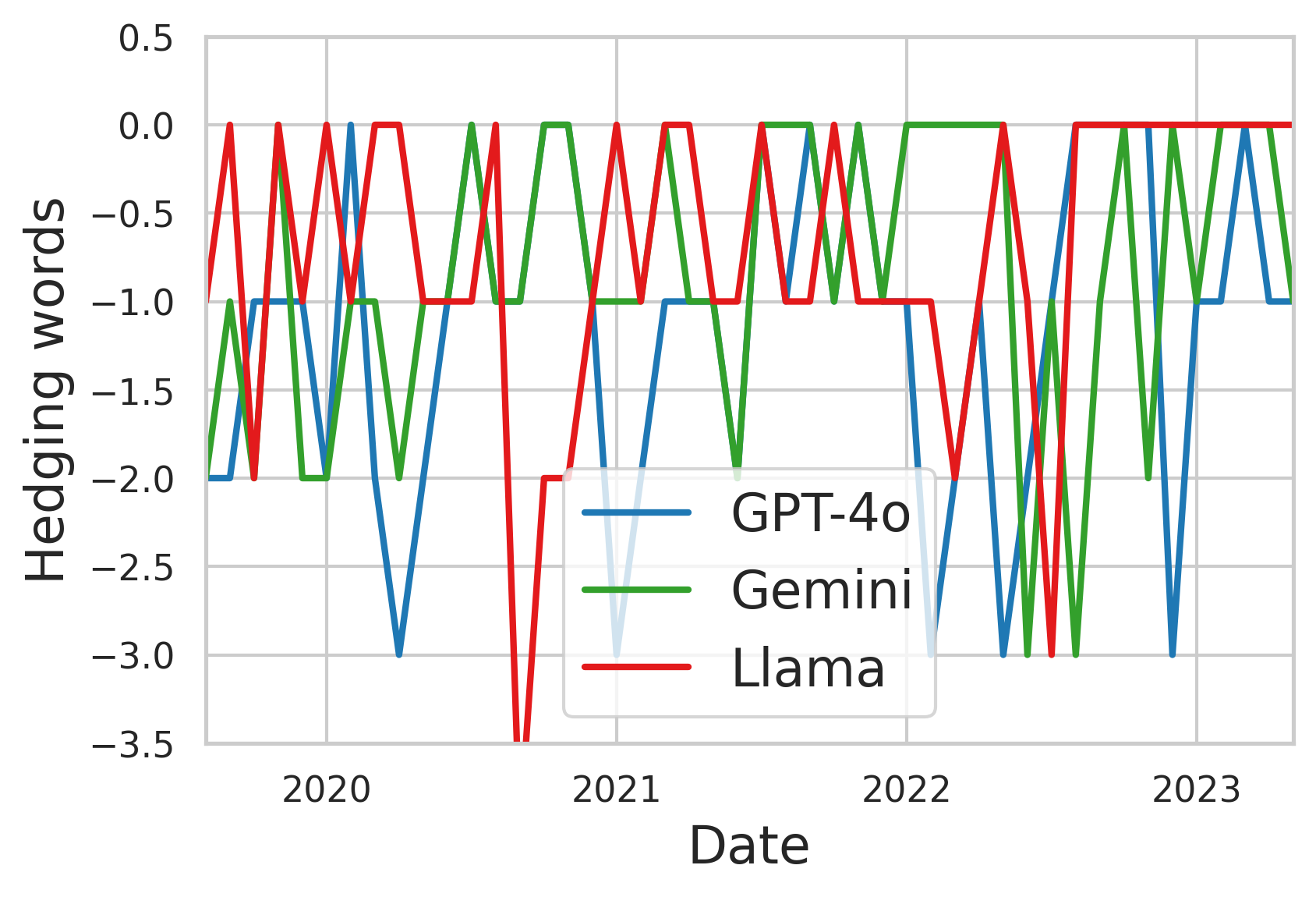}
    \caption{Synthetic short reports}
    \vspace{-0.1cm}
  \end{subfigure}
  \caption{The evolution of hedging words over time for Close Price reports generated by GPT-4o, Gemini and \llama.}
  \label{fig:hedge_pair_plots}
  \vspace{-1em}
\end{figure}

\paragraph{Hedging Language Usage}
We define a lexicon of hedging words that indicate speaker uncertainty, such as "potentially" and "possibly" (see Appendix~\ref{app:uncertainlex}). We track the usage of these words in reports over time, with each word contributing a $-1$ to the count. The average count of hedge words is calculated across indices. Figure~\ref{fig:hedge_pair_plots} illustrates the evolution of hedging words in short reports for GPT-4o, Gemini, and \llama (additional plots in Appendix~\ref{app:uncertainlexplots}).
In subplot (a), based on real indices, we observe an increase in hedging words over time for GPT-4o. As references to external knowledge approach and surpass the knowledge cut-off date, uncertainty rises, suggesting the model may be extrapolating or speculating on trends. This pattern is somewhat seen in \llama but not in Gemini.
For reports generated from synthetic data (subplot (b)), no clear trend is observed in any model. These trends are consistent across other report types.

\subsection{Error Analysis}
In this section, we discuss key issues identified during human evaluation of report quality:

\paragraph{Inclusion of Non-Input Data:} Aside from including confident references to external real-world events, which are not necessarily undesirable or incorrect, reports occasionally referenced technical indicators, such as the 200-day moving average, which were not part of the input data. This can be explained by the fact that this type of TI is very common for the long-term financial analysis.  

\paragraph{Temporal Inconsistencies:} Some reports described events in a sequence, which contradicts the natural flow of time. For example, it might mention a peak in mid-December 2023, followed by a downward trend in late September and early October 2023, which is temporally impossible. We observed occasional mentions of months not included in the input time series. The time series was sometimes split into intervals that were not meaningful, such as at points that were not significant changes. 

\paragraph{Inconsistent Report Content:} The content varied across reports, suggesting a need for more prescriptive prompt design to ensure consistency.

\paragraph{Bias Towards Positive Reporting:} There appeared to be a bias towards reporting positive or above-average trends, avoiding low points,  This observation aligns with findings by \citet{mantion2024analysis}, who noted a hawkish bias in ChatGPT when processing FedSpeak. However quantifying the extent of this is left for future work. 

\paragraph{Use of Informal Language:} Occasional use of slang was noted, such as the term "flirting" in the context of technical indicators: ``The Relative Strength Index (RSI) occasionally flirted with overbought levels.''.

\subsection{Segment Source Classification Analysis and Evaluation}
\label{sec:span-over-time}
\begin{table}[]
\centering
\resizebox{0.9\columnwidth}{!}{%
\begin{tabular}{llllllll}
\hline
 & \begin{tabular}[c]{@{}l@{}}Pred. \\ DR\end{tabular} & \begin{tabular}[c]{@{}l@{}}Pred.\\ FI\end{tabular} & \multicolumn{1}{l|}{\begin{tabular}[c]{@{}l@{}}Pred.\\ EK\end{tabular}} & P & R & F1 & Supp. \\ \hline
DR & 352 & 85 & \multicolumn{1}{l|}{20} & 0.95 & 0.77 & 0.85 & 457 \\
FI & 12 & 261 & \multicolumn{1}{l|}{20} & 0.30 & 0.46 & 0.36 & 37 \\
EK & 7 & 13 & \multicolumn{1}{l|}{17} & 0.73 & 0.89 & 0.80 & 293 \\ \hline
Acc. &  &  &  &  &  & 0.80 & 787 \\
Macro &  &  &  & 0.66 & 0.71 & 0.67 & 787 \\ \hline
\end{tabular}%
}
\caption{Segment categorization performance per class for Direct Reference (DR), Financial Interpretation (FI), and External Knowledge (EK): Confusion matrix (left), Precision (P), Recall (R), F1-score (F1) and Support (Supp.) (right), Accuracy and Macro-averages (bottom).}
\label{tab:highlightseval}
\end{table}

\paragraph{Evaluation}
Table~\ref{tab:highlightseval} shows the confusion matrix and performance metrics of the segment source classification approach. 
The overall accuracy of 80\% is reasonably high, supporting the method's reliability in aiding users with report parsing.
Examining per-class performance, EK emerges as the most challenging category, likely due to limited support. The confusion matrix indicates that most misclassifications occur when DR is predicted as FI. The FI category is more broadly defined compared to the other two, and segments summarizing overall trends and volatility might be classified as either DR or FI. Additionally, segments suggesting hypotheses about potential events leading to observed trends, using hedging language, might be classified as either FI or EK.
Other common errors involve categorizing descriptions of Technical Indicators as FI, as the model fails to recognize that TIs are part of the `time series' mentioned in the prompt. Regarding segmentation, only 12 out of 787 annotated segments were identified as needing division into two categories.

\begin{figure}[t]
  \centering
  \begin{subfigure}{0.49\columnwidth}
    \includegraphics[width=\linewidth]{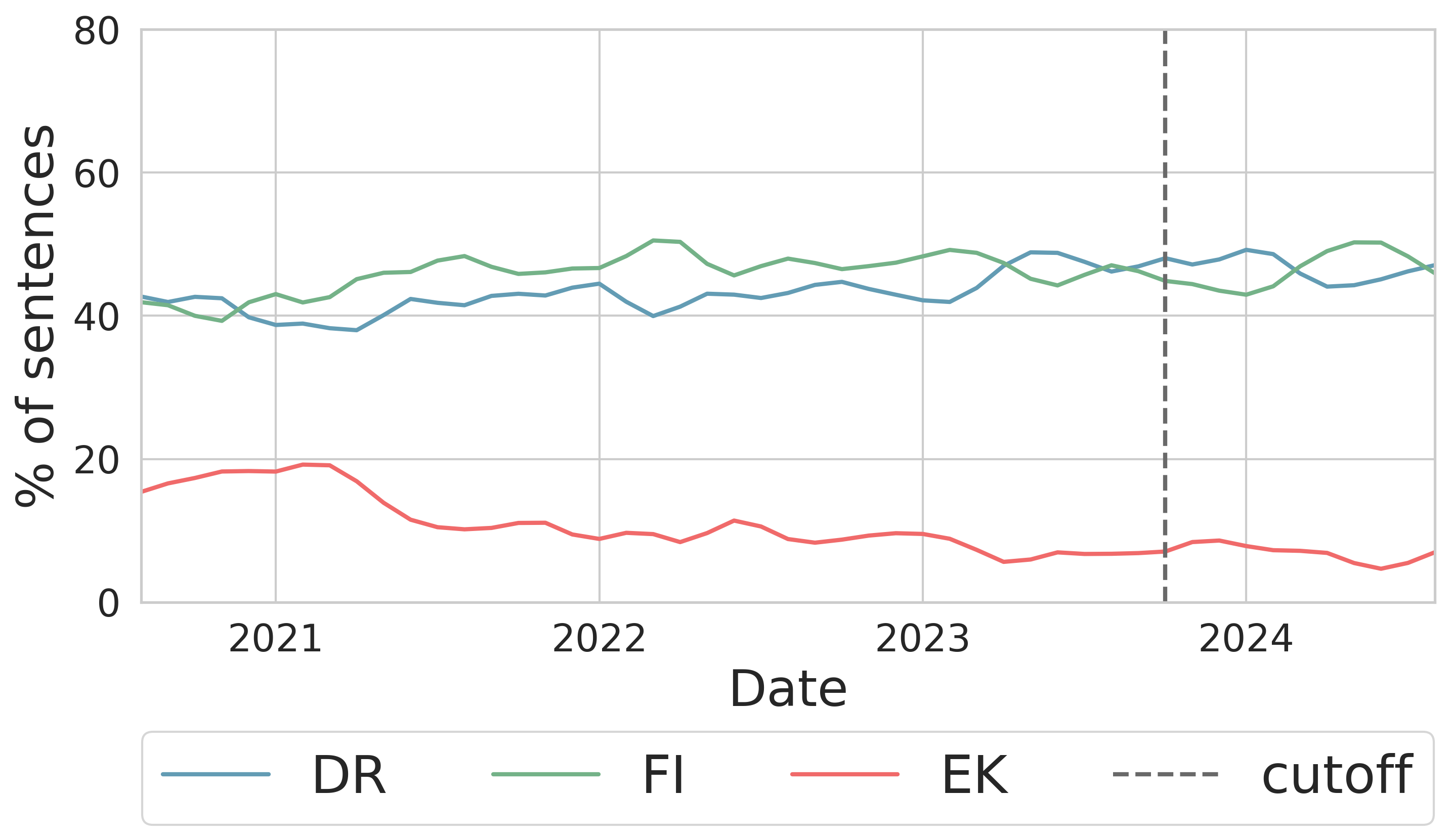}
    \caption{GPT-4o short reports}
    \vspace{-0.1cm}
  \end{subfigure}
  \vspace{-0.1cm}
  \begin{subfigure}{0.49\columnwidth}
    \includegraphics[width=\linewidth]{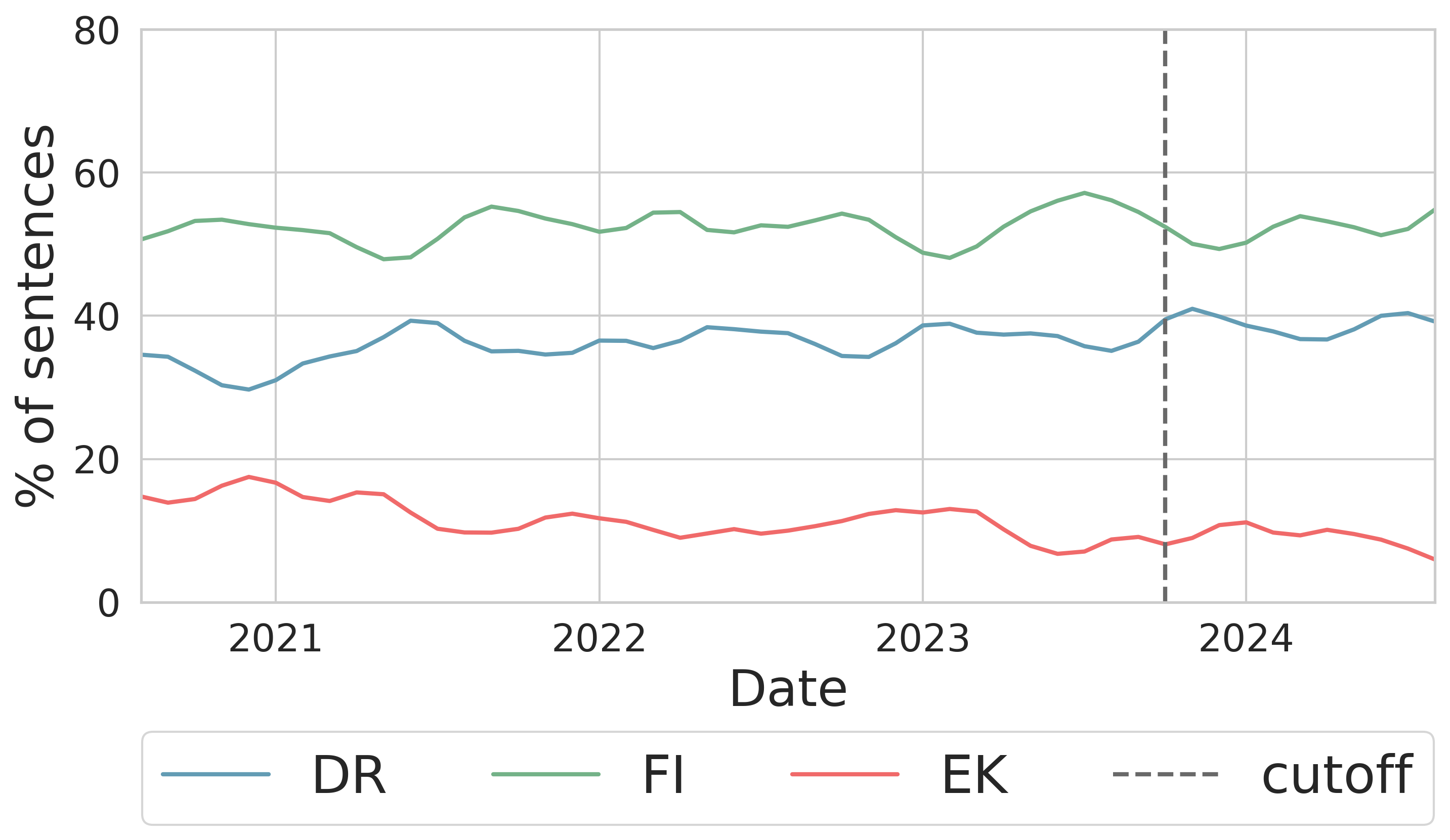}
    \caption{GPT-4o TI (plots) reports}
    \vspace{-0.1cm}
  \end{subfigure}
  \caption{The evolution of information categories (DR, FI, EK) in financial reports generated by the GPT-4 model over time. The vertical dashed line indicates the model's training data cutoff date.}
  \label{fig:highlightsovertime}
  \vspace{-1em}
\end{figure}

\paragraph{Temporal Analysis}
Figure \ref{fig:highlightsovertime} presents the evolution of the proportions of different information categories (DR, FI, EK) in reports generated by the GPT-4o over time for two types of reports (plots for all models and report types in Appendix~\ref{app:seganalysis}). EK category has the lowest proportion of segments, with reports mentioning only very few world events directly, the specific ones are COVID-19 pandemic, vaccination rollouts, US presedential election, and more general ones are Federal Reserve policy shifts and positive corporate earnings reports.  The most notable observation for both types of reports is the pronounced decrease in the proportion of EK (External Knowledge) as it approaches and past the model's training data cutoff date. This suggests that when GPT-4o encounters data beyond its training period, it relies more heavily on the information explicitly present in the time series data and its internal financial reasoning capabilities, rather than drawing upon external knowledge. The corresponding increase in DR (Direct Reference) and FI (Financial Interpretation) post-cutoff further supports this observation, indicating a shift towards a more data-driven and analytical approach in the absence of readily available external context. This analysis underscores the dynamic interplay between the model's training data, its inherent reasoning abilities, and its capacity to integrate external knowledge in generating financial reports. 
We further observe that in short reports the levels of FI and DR are very similar, while reports that are using technical indicators produce more financial interpretation of the given inputs compared to direct references.

\section{Conclusion}
Our study has demonstrated that LLMs, equipped with a robust framework of prompt engineering, model selection, and evaluation metrics, can effectively utilize time series data from prominent stock market indices to produce consistent and fluent financial reports. The introduction of an automated highlighting system further enhances the utility of these models by efficiently categorizing information into insights derived from direct data analysis, financial reasoning, or external knowledge. This categorization not only facilitates a clearer understanding of the models' outputs but also provides a nuanced evaluation of their reasoning and factual grounding capabilities. This framework can assist in generating technical reports that provide insights into market trends and dynamics.

\section{Limitations}
This study highlights several limitations that suggest areas for future research. 

One of the main limitations is the absence of definitive ground truth in the evaluation process, and the potential bias in the automated evaluations towards models from GPT family as GPT-4o is used as the evaluator model. It is partially mitigated by verifying and aligning scores with human annotations, but it  still remains an open area for future research.

The current framework also struggles to fully mitigate the risk of generating plausible but incorrect information—a critical concern in the financial domain requiring stringent human oversight to maintain report integrity. Future efforts should aim to enhance data robustness, improve classification accuracy, and refine prompt engineering to reduce hallucinations and advance the practical application of LLMs in financial analysis.

Moreover, the effectiveness of large language models (LLMs)  is heavily dependent on the precision of prompt engineering and model selection, which may not fully capture the nuances of financial data, leading to discrepancies in outputs. 

\section*{Disclosure}

Disclaimer: This paper was prepared for informational purposes by the Artificial Intelligence Research group of JPMorgan Chase \& Co. and its affiliates (``JP Morgan'') and is not a product of the Research Department of JP Morgan. JP Morgan makes no representation and warranty whatsoever and disclaims all liability, for the completeness, accuracy or reliability of the information contained herein. This document is not intended as investment research or investment advice, or a recommendation, offer or solicitation for the purchase or sale of any security, financial instrument, financial product or service, or to be used in any way for evaluating the merits of participating in any transaction, and shall not constitute a solicitation under any jurisdiction or to any person, if such solicitation under such jurisdiction or to such person would be unlawful.
\bibliography{custom}

\appendix
\onecolumn 

\section{Prompts}
\label{sec:app-prompts}
We present the prompts used for the following tasks below.

\subsection{Report generation prompts}
\label{sec:app-prompts-gen}

    \begin{tcolorbox}[title=Task: Short report generation]
    
    \vspace{0.2cm}
    "You are a financial analyst tasked with writing a short report analyzing the main aspects of the \texttt{\{index\_name\}} index based on its time series data. The report should be concise, focusing on key trends, volatility, and any notable price patterns observed in the data. Your report should be one or two paragraphs long, summarizing the overall performance and recent movements."

    \vspace{0.2cm}
    Closing Prices:
    \begin{verbatim}
    Date       Close
    2020-04-28 8607.7
    2020-04-29 8914.7
    2020-04-30 8889.6
    2020-05-01 8605.0
    \end{verbatim}

    \end{tcolorbox}
    \begin{tcolorbox}[title=Task: Long report generation with numerical technical indicators]

    "You are a financial analyst tasked with writing a short report analyzing the main aspects of the \texttt{\{index\_name\}} index based on its time series data and technical indicators. Focus on key trends, volatility, notable price patterns, and significant changes in the technical indicators such as moving averages or RSI. Summarize the overall performance and recent movements in two or three paragraphs."\\
    \vspace{0.2cm}
    Time Series Data with Technical Indicators:\\
    \texttt{
    Date       Close    SMA\_50  RSI   MACD    Volatility \\
    2020-04-28 8607.7   8210.1   68.1  175.7   0.416     \\
    2020-04-29 8914.7   8193.7   69.7  200.0   0.425     \\
    2020-04-30 8889.6   8175.2   67.9  214.8   0.381     \\
    2020-05-01 8605.0   8152.2   59.0  201.2   0.408    \\
    }

    \vspace{0.2cm}
    \end{tcolorbox}
\clearpage

    \begin{tcolorbox}[title=Task: Long report generation with plots of technical indicators]

    \vspace{0.2cm}
    "You are a financial analyst tasked with writing a short report analyzing the main aspects of the \texttt{\{index\_name\}} index based on its time series data and technical indicator plots. The report should focus on key trends, volatility, and any notable price patterns observed in the data and the indicator plots. Your report should be two or three paragraphs long, summarizing the overall performance and recent movements."\\
    The plots show the main technical indicators and this is the Time Series Data:\\

    \begin{verbatim}
    Date       Close
    2020-04-28 8607.7
    2020-04-29 8914.7
    2020-04-30 8889.6
    2020-05-01 8605.0
    \end{verbatim}

    \texttt{Technical Indicators:}\\
    \centering
    \includegraphics[width=0.8\linewidth]{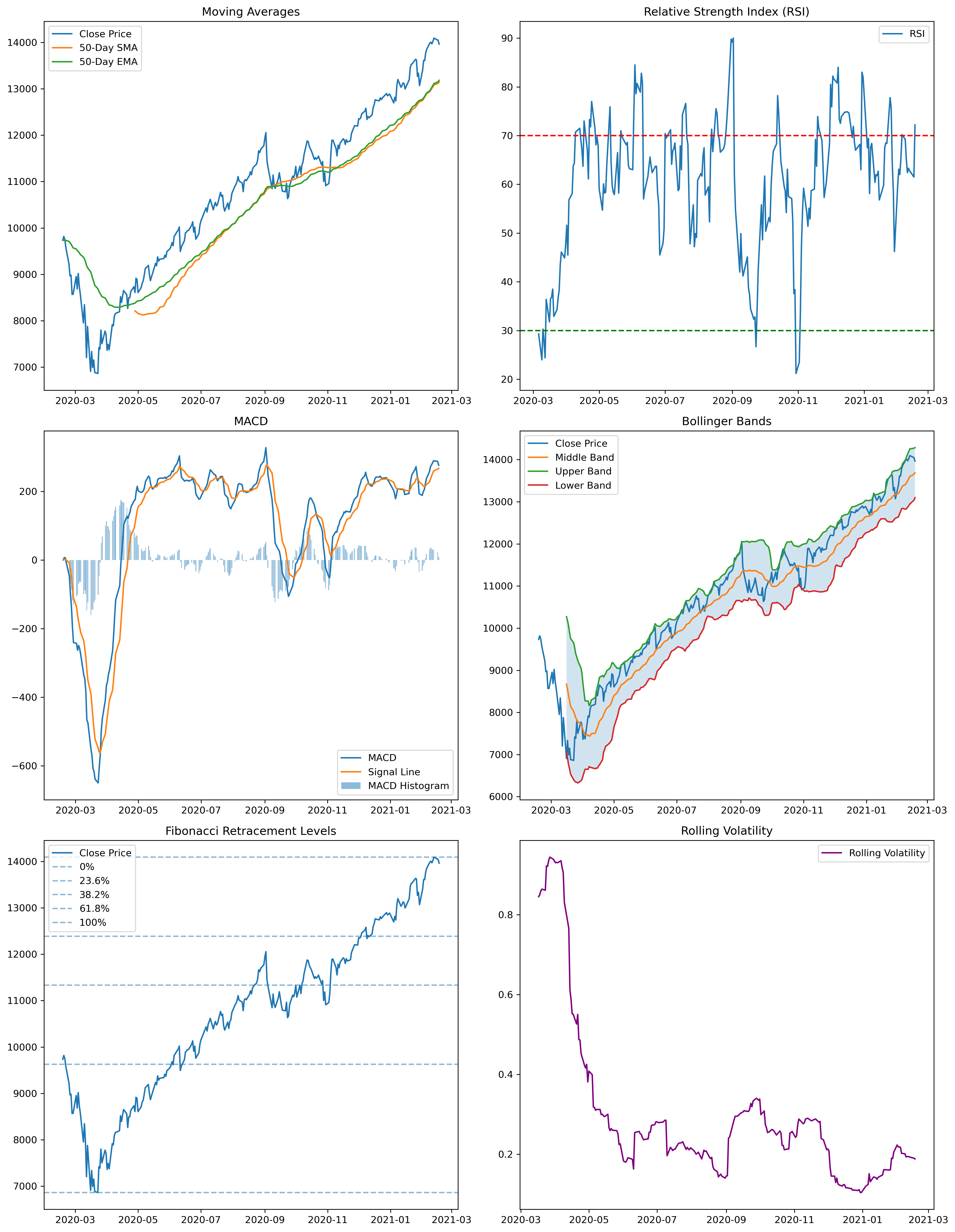}  

    \end{tcolorbox}

\clearpage

\subsection{G-eval}
\label{sec:app-geval-prompts}
    \begin{tcolorbox}[title=G-Eval: Consistency Evaluation Prompt]
    \textbf{Task:}\\
    Your task is to rate the report on one metric.

    \vspace{0.2cm}
    \textbf{Evaluation Criteria:}\\
    \texttt{Consistency (1-5) - The factual alignment between the financial report and the time series data. A factually consistent report accurately reflects the trends, values, and key events present in the time series without introducing information not supported by it. Reports that contain hallucinated facts (i.e., statements that introduce or infer information not present in the time series) should be penalized.}

    \vspace{0.2cm}
    \textbf{Evaluation Steps:}
    \begin{enumerate}
        \item \texttt{Read the Time Series: Examine the time series data to understand the key facts, trends, and details it presents.}
        \item \texttt{Read the Financial Report: Review the report and compare its content to the time series data. Identify any statements that do not align with the data or introduce unsupported information.}
        \item \texttt{Assign a score for consistency based on the Evaluation Criteria.}
    \end{enumerate}
    
    \vspace{0.2cm}
    \textbf{Input:}\\
    \texttt{Time series data: }\\
    \begin{verbatim}
    Date       Close
    2020-04-28 8607.7
    2020-04-29 8914.7
    2020-04-30 8889.6
    2020-05-01 8605.0
    \end{verbatim}
    \texttt{Technical Indicators \textbf{(if analyzing reports with technical indicators)}}\\
    \vspace{1em}
    \texttt{
    Date        SMA\_50  RSI   MACD    Volatility \\
    2020-04-28  8210.1   68.1  175.7   0.416     \\
    2020-04-29  8193.7   69.7  200.0   0.425     \\
    2020-04-30  8175.2   67.9  214.8   0.381     \\
    2020-05-01  8152.2   59.0  201.2   0.408    \\
    }

    \vspace{0.2cm}

    \texttt{Financial report:}\\
     \{\{Report\}\}\\
    \vspace{0.2cm}
    \textbf{Evaluation Form (Scores ONLY):}\\
    \vspace{0.2cm}
    \texttt{- Consistency:}
    \end{tcolorbox}

\clearpage

    \begin{tcolorbox}[title=G-Eval: Coherence Evaluation Prompt]
    \textbf{Task:}\\
    Your task is to rate the report on one metric.

    \vspace{0.2cm}
    \textbf{Evaluation Criteria:}\\
    \texttt{Coherence (1-5) - The degree to which the report is logically organized and well-structured. The report should clearly present the insights from both the time series data and the technical indicators in a way that builds sentence by sentence into a coherent body of information. The report should not feel like a disjointed collection of statements but should present a logical progression of ideas and insights, where each sentence and paragraph naturally follows from the previous ones.}

    \vspace{0.2cm}
    \textbf{Evaluation Steps:}
    \begin{enumerate}
        \item \texttt{Examine the Time Series and Technical Indicators: Carefully review both the time series data and the technical indicators. Identify the main trends, signals, and key points in the data.}
        \item \texttt{Read the Financial Report: Read the financial report and assess its logical flow and structure. Check if the report covers the key trends and points from the time series and technical indicators in a clear, organized, and logical manner. Look for a smooth progression of information, where each insight follows naturally from the previous one.}
        \item \texttt{Assign a score for coherence on a scale of 1 to 5, where 1 is the lowest and 5 is the highest based on the Evaluation Criteria.}
    \end{enumerate}
    
    \vspace{0.2cm}
    \textbf{Input:}\\
    \texttt{Time series data: }\\
    \begin{verbatim}
    Date       Close
    2020-04-28 8607.7
    2020-04-29 8914.7
    2020-04-30 8889.6
    2020-05-01 8605.0
    \end{verbatim}
    
    \vspace{0.2cm}
    \texttt{Technical Indicators \textbf{(if analyzing reports with technical indicators)}}\\
    \vspace{0.2cm}
    \texttt{
    Date        SMA\_50  RSI   MACD    Volatility \\
    2020-04-28  8210.1   68.1  175.7   0.416     \\
    2020-04-29  8193.7   69.7  200.0   0.425     \\
    2020-04-30  8175.2   67.9  214.8   0.381     \\
    2020-05-01  8152.2   59.0  201.2   0.408    \\
    }

    \vspace{0.2cm}

    \texttt{Financial report:}\\
     \{\{Report\}\}\\
    \vspace{0.2cm}
    \textbf{Evaluation Form (Scores ONLY):}\\
    \vspace{0.2cm}
    \texttt{- Coherence:}
    \end{tcolorbox}

\clearpage

    \begin{tcolorbox}[title=G-Eval: Fluency Evaluation Prompt]
    \textbf{Task:}\\
    Your task is to evaluate the report on one metric.

    \vspace{0.2cm}
    \textbf{Evaluation Criteria:}\\
    \texttt{Fluency (1-5) - The readability and naturalness of the language used in the report. A fluent report should be free from grammatical errors, awkward phrasing, and unnatural language. It should read smoothly and be easy to understand.}

    \vspace{0.2cm}
    \textbf{Score Breakdown:}\\
    \texttt{
    - 1 = The report is highly unnatural with significant grammar and phrasing issues.\\
    - 2 = The report has major fluency problems, with noticeable awkwardness and errors.\\
    - 3 = The report is somewhat fluent, but with some noticeable issues.\\
    - 4 = The report is mostly fluent, with only a few minor issues.\\
    - 5 = The report is fully fluent, with natural and smooth language.
    }

    \vspace{0.2cm}
    \textbf{Evaluation Steps:}
    \begin{enumerate}
        \item \texttt{Read the Report Carefully: Pay close attention to the language used, including grammar, phrasing, and overall readability.}
        \item \texttt{Identify Language Issues: Look for any grammatical errors, awkward sentences, or unnatural phrasing that may hinder the readability of the report.}
        \item \texttt{Assign a score for Fluency on a scale of 1 to 5, where 1 is the lowest and 5 is the highest based on the Evaluation Criteria.}
    \end{enumerate}
    
    \vspace{0.2cm}
    \textbf{Input:}\\
    \texttt{Financial report:}\\
    \{\{Summary\}\}

    \vspace{0.2cm}
    \textbf{Evaluation Form (Scores ONLY):}\\
    \vspace{0.2cm}
    \texttt{- Fluency (1-5):}
    \end{tcolorbox}

\subsection{Highlights}
\label{app:highlightprompt}

    \begin{tcolorbox}[title=Task: Source Segment Classification]
    \texttt{
    Please categorize the following text into three categories:}

\begin{enumerate}
        \item \texttt{Direct reference: Segments that directly mention numerical values or trends from the input time series data.}
        \item \texttt{Financial interpretation: Segments that infer or conclude based on the observed data without external knowledge.}
        \item \texttt{External knowledge: Segments that provide context or explanations using knowledge outside the observed time series data.}
    \end{enumerate}
    \texttt{Segments should be non-overlapping.}\\
    
    \texttt{Return the categorized segments in the following JSON format:}\\

     \texttt{\{'direct\_reference': ['segment1', 'segment2', ...], \\
        'financial\_interpretation': ['segment1', 'segment2', ...],  \\
         'external\_knowledge': ['segment1', 'segment2', ...]\}}\\
                \texttt{Text: \{\{sentence\}\}}
    \vspace{0.2cm}
    \end{tcolorbox}

\clearpage

\section{Uncertainty Lexicon}
\label{app:uncertainlex}
We used the following words as lexicon of words indicating uncertainty:

'unclear', 'unknown', 'doubtful', 'uncertain', 'unconfident', 'tentative', 'tentatively', 'unsettled', 'undecided', 'unresolved',
    'ambivalent', 'skeptical, 'questionable', 'questionably', 'unconvinced', 'might', 'maybe', 'possibly', 'could', 'may', 'could','potentially', 'conceivably', 'perhaps', 'perchance',
    'probably', 'likely', 'presumably', 'apparently', 'seem', 'appears',   'feasibly', 'reportedly', 'allegedly', 'purportedly', 'plausibly', 'plausible'.

\subsection{Evolution of hedging words over time}
\label{app:uncertainlexplots}

\begin{figure*}[h!]
    \centering
    \begin{subfigure}{0.32\textwidth}  
        \centering
        \includegraphics[width=\linewidth]{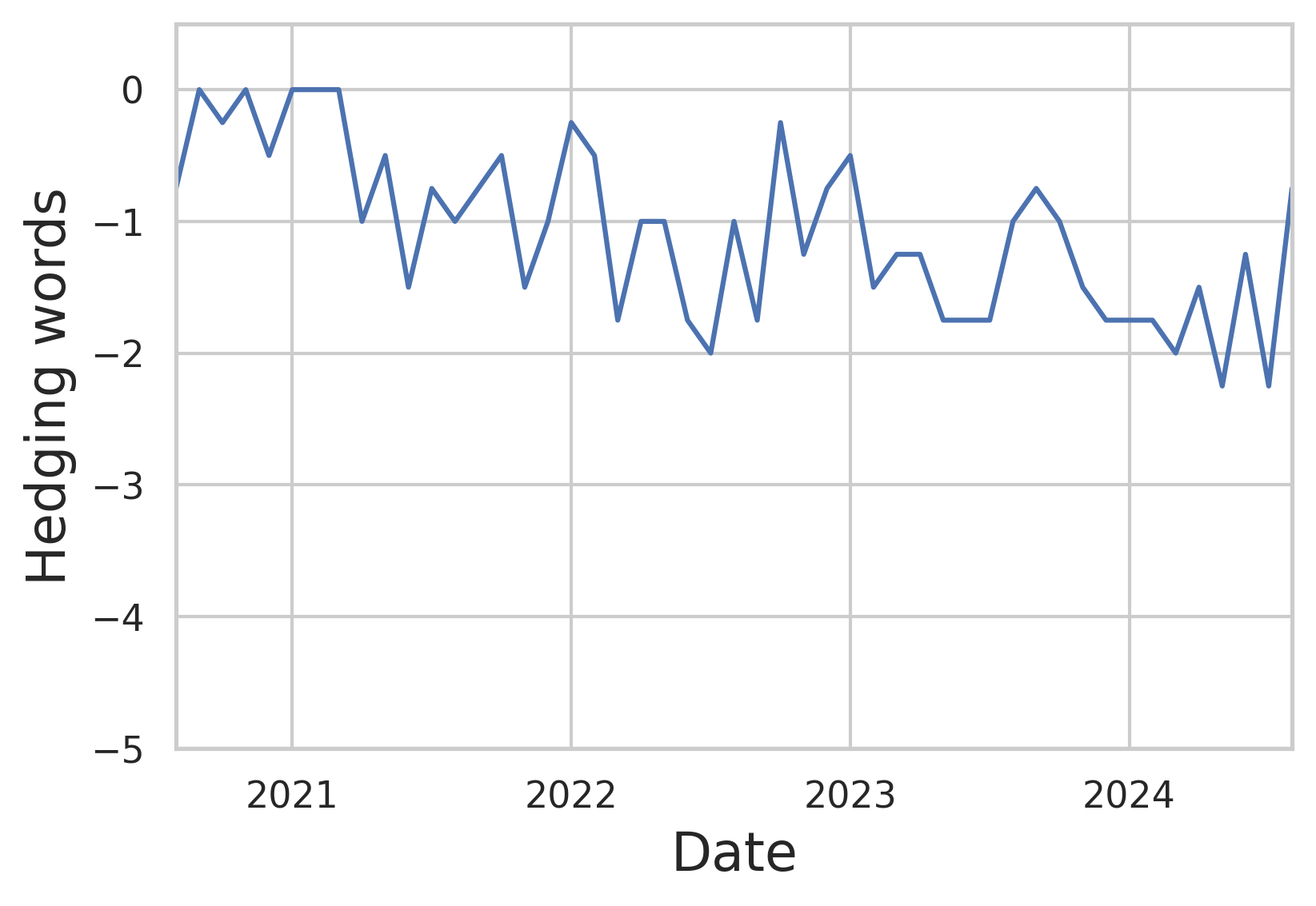}  
        \caption{GPT-4o}
    \end{subfigure}
    \begin{subfigure}{0.32\textwidth}  
        \centering
        \includegraphics[width=\linewidth]{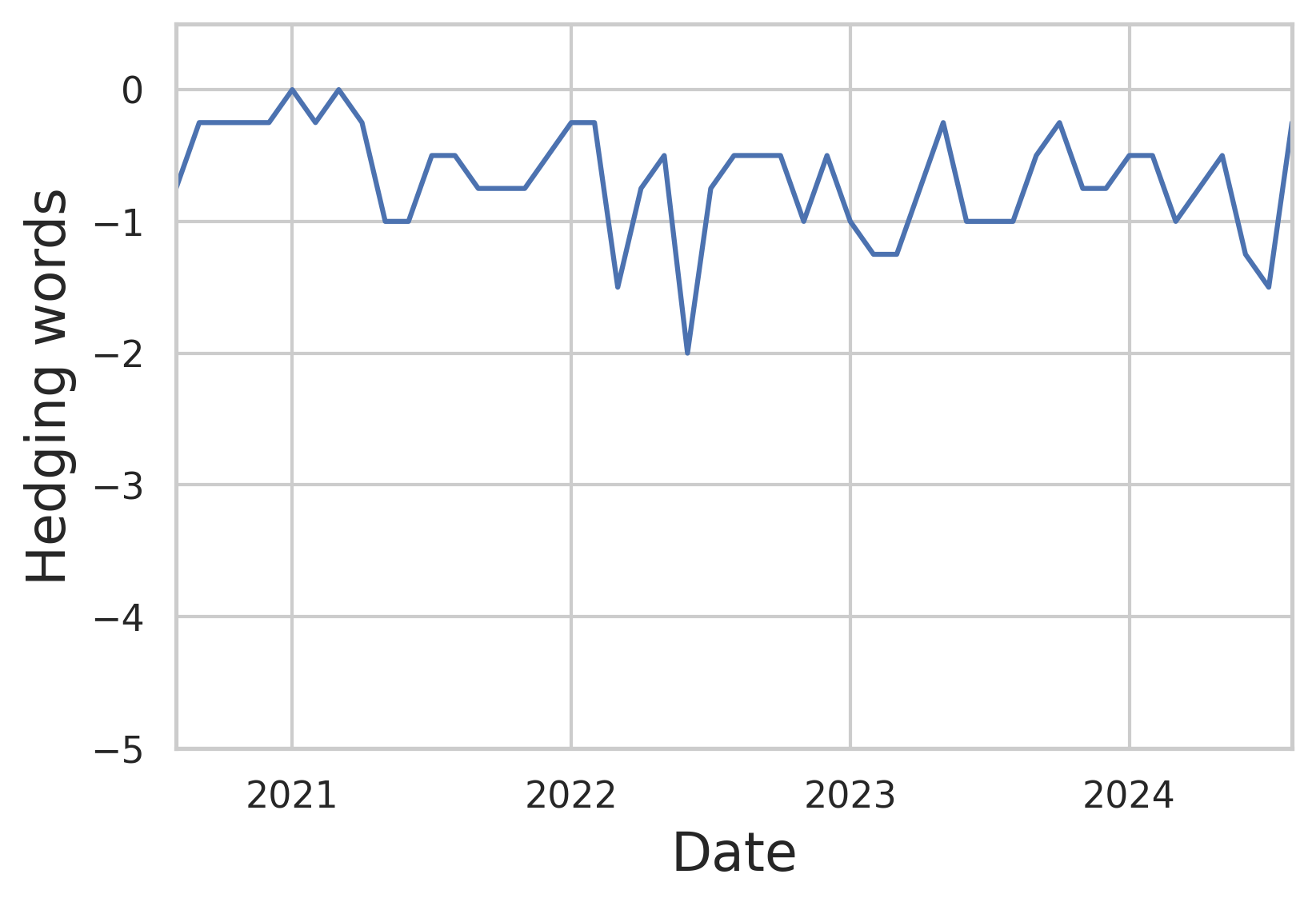}
        \caption{GPT-4o-mini}
        \label{fig:gpt-4o-mini_ti}
    \end{subfigure}
    \begin{subfigure}{0.32\textwidth}  
        \centering
        \includegraphics[width=\linewidth]{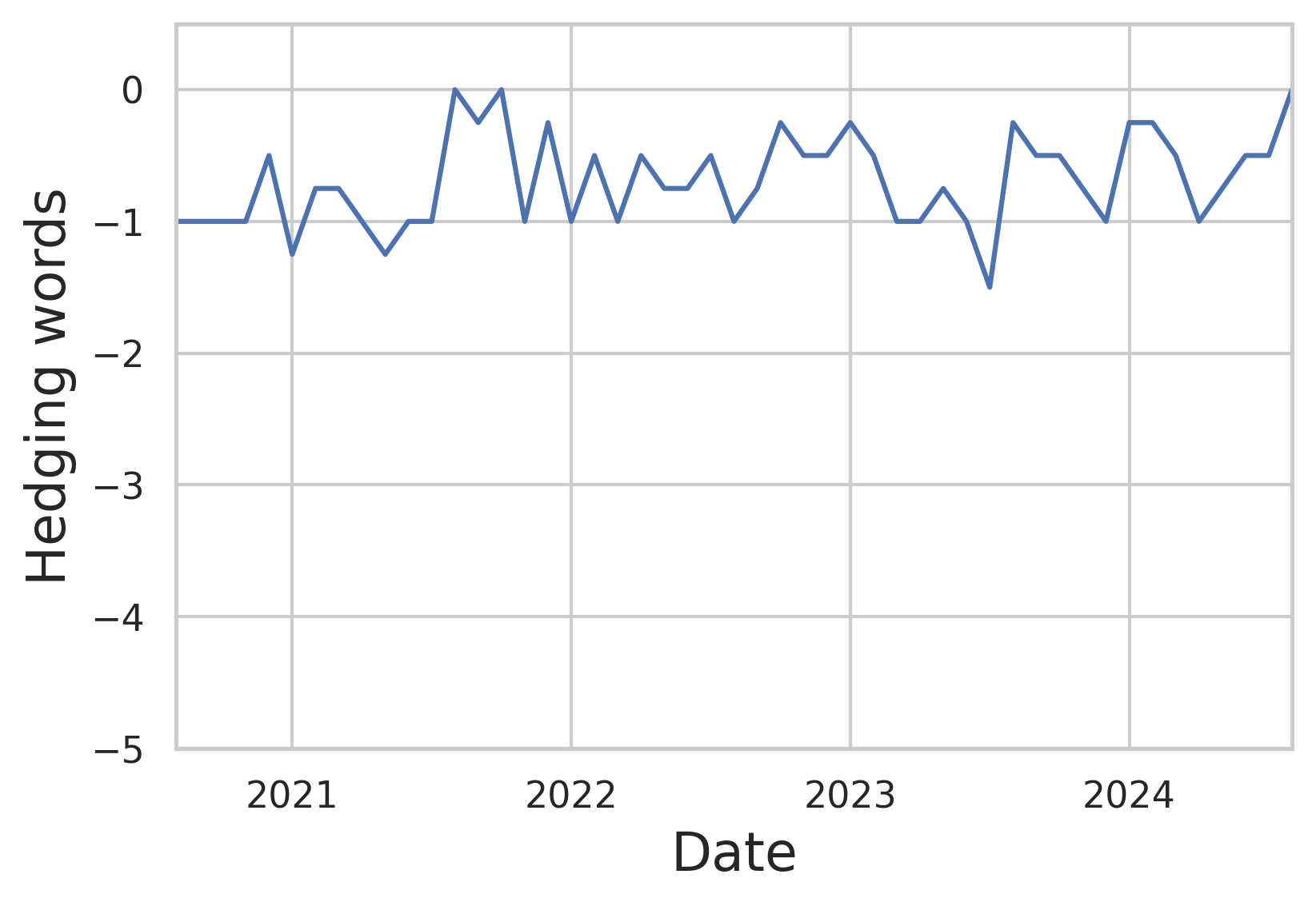}
        \caption{Gemini}
    \end{subfigure}\\
    \begin{subfigure}{0.32\textwidth}  
        \centering
        \includegraphics[width=\linewidth]{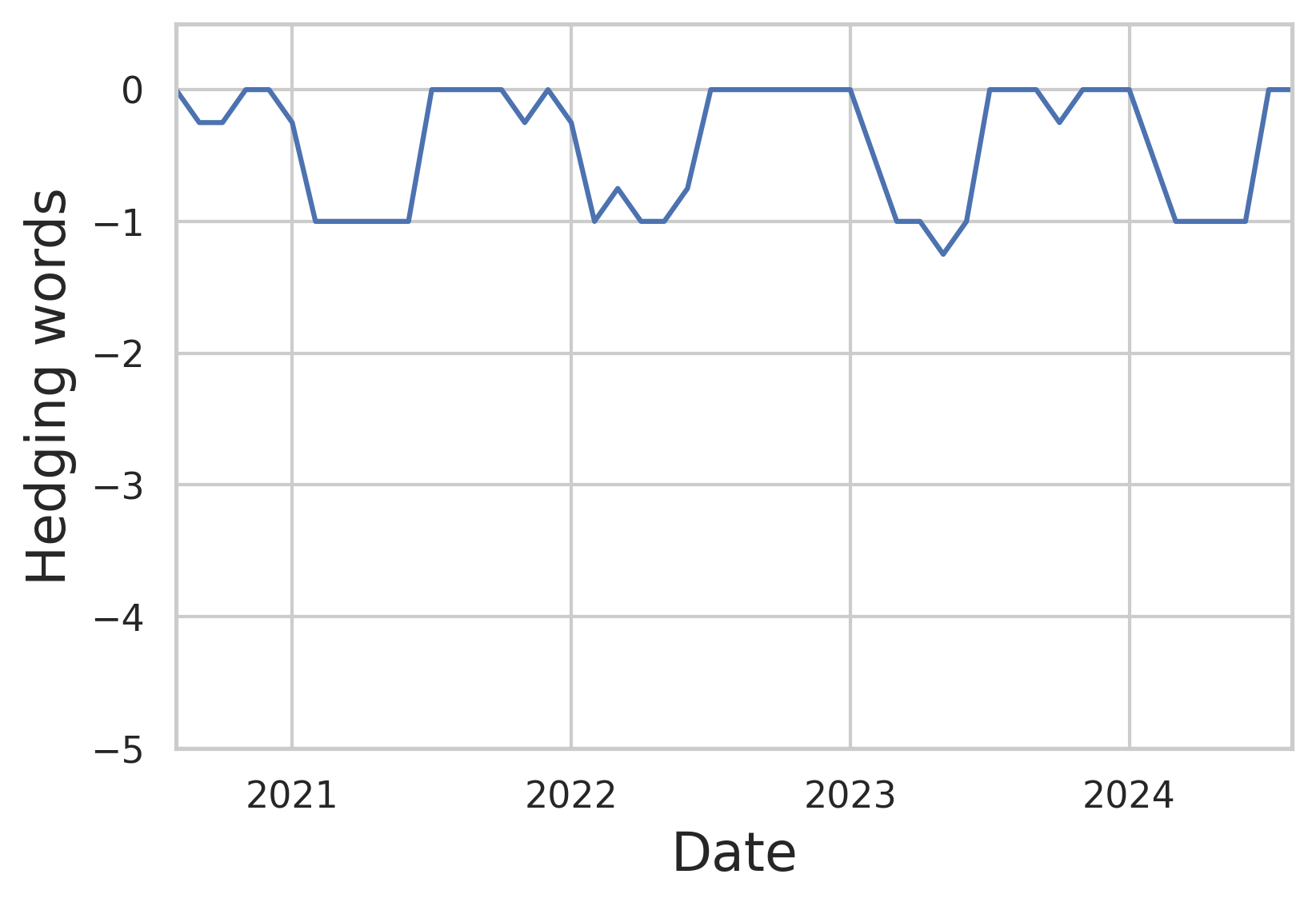}
        \caption{Phi-3}
    \end{subfigure}
        \begin{subfigure}{0.32\textwidth}  
        \centering
        \includegraphics[width=\linewidth]{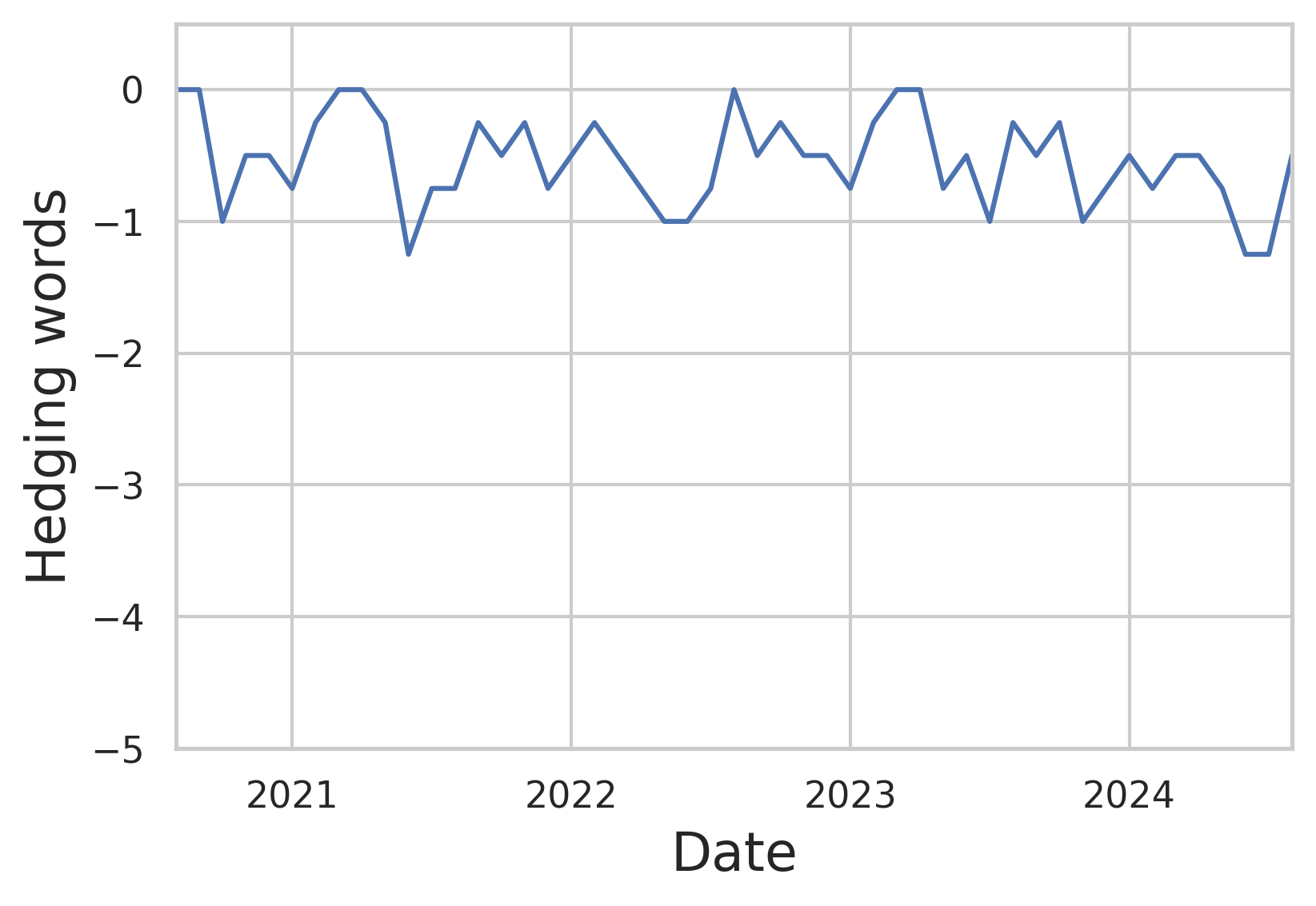}
        \caption{\llama}
    \end{subfigure}
    \caption{Evolution of hedging words over time for Short reports generated from real data.}
    \label{fig:hedge_short_real}
\end{figure*}

\begin{figure*}[h!]
    \centering
    \begin{subfigure}{0.24\textwidth}  
        \centering
        \includegraphics[width=\linewidth]{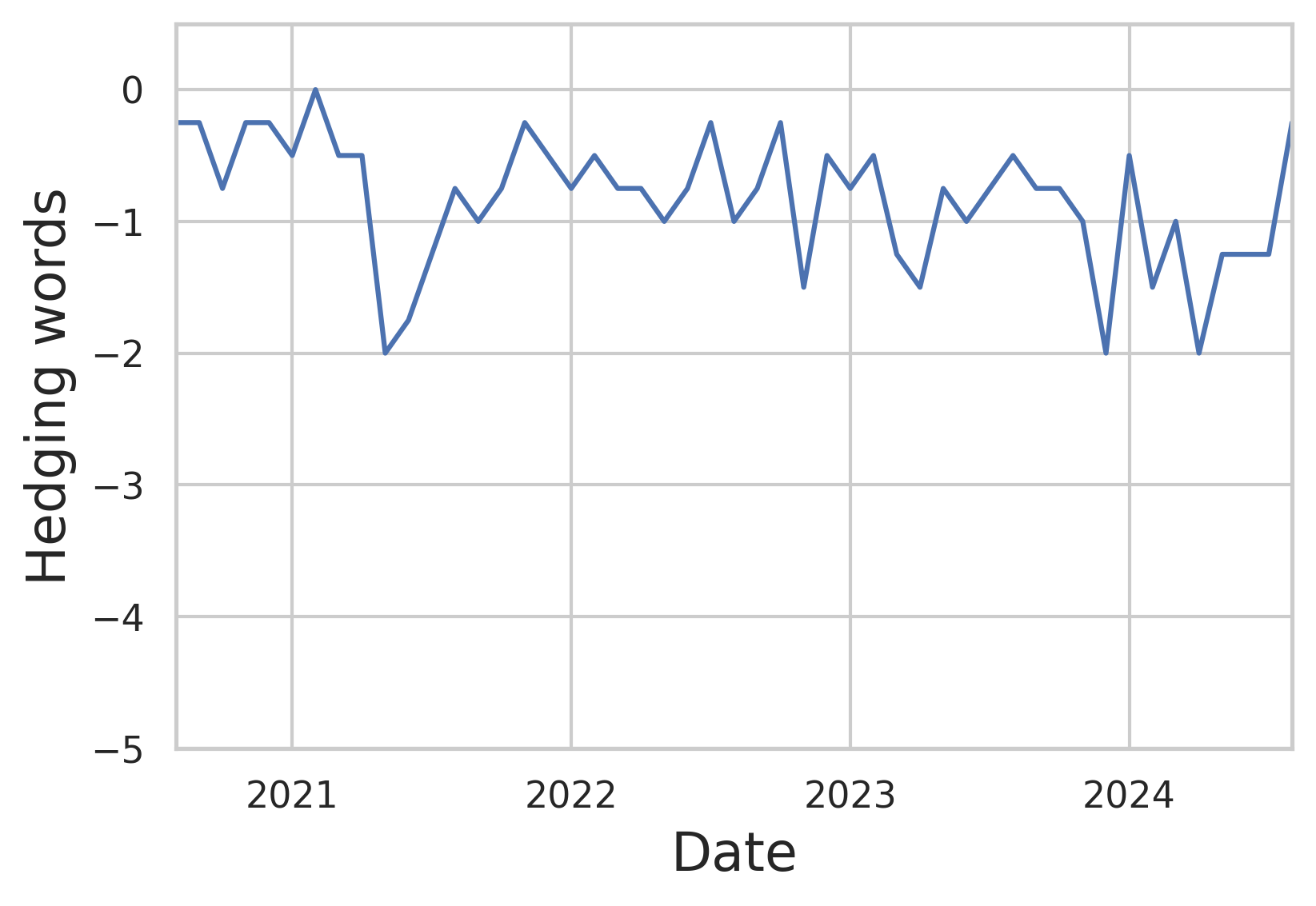}  
        \caption{GPT-4o}
    \end{subfigure}
    \begin{subfigure}{0.24\textwidth}  
        \centering
        \includegraphics[width=\linewidth]{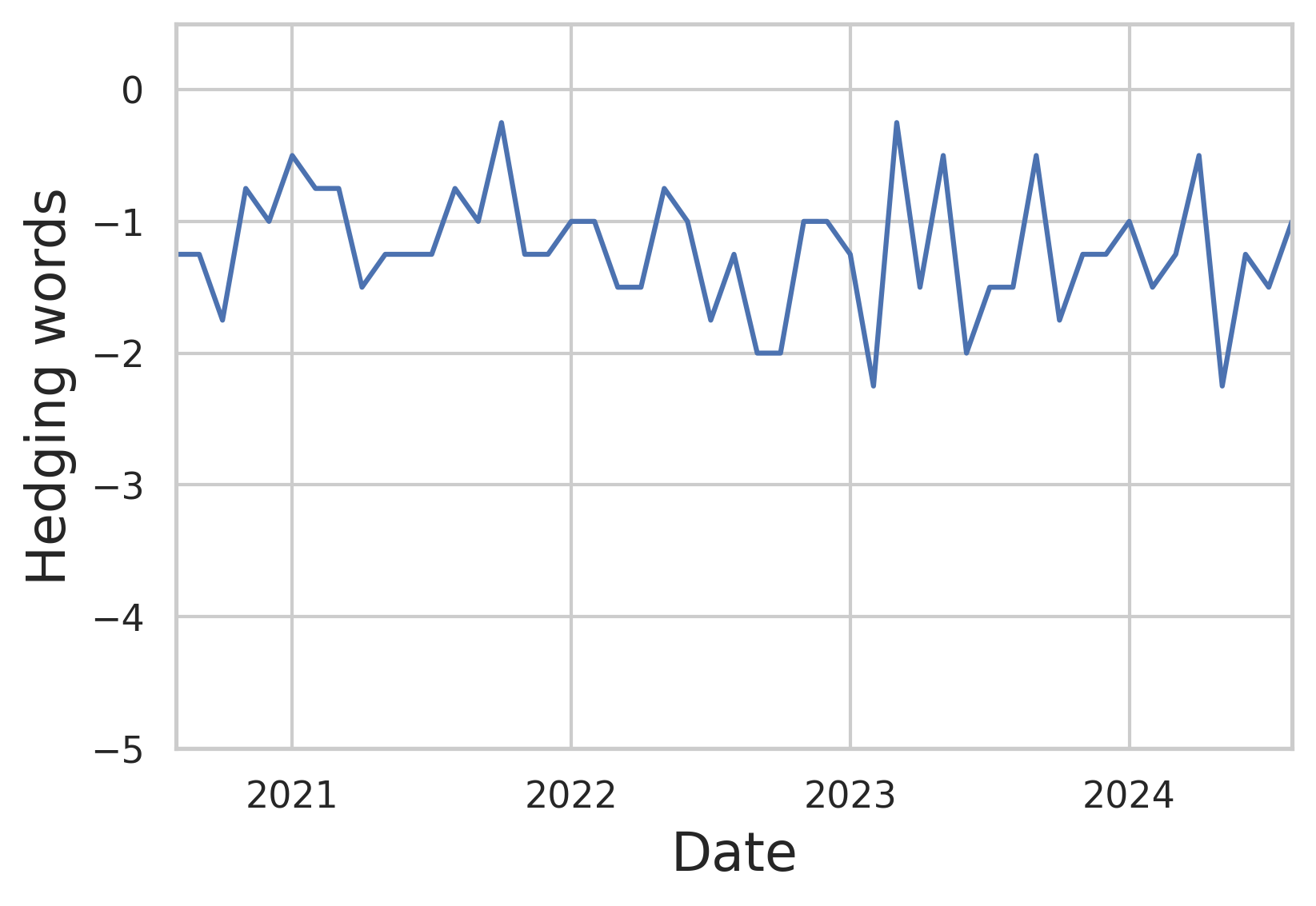}
        \caption{GPT-4o-mini}
    \end{subfigure}
    \begin{subfigure}{0.24\textwidth}  
        \centering
        \includegraphics[width=\linewidth]{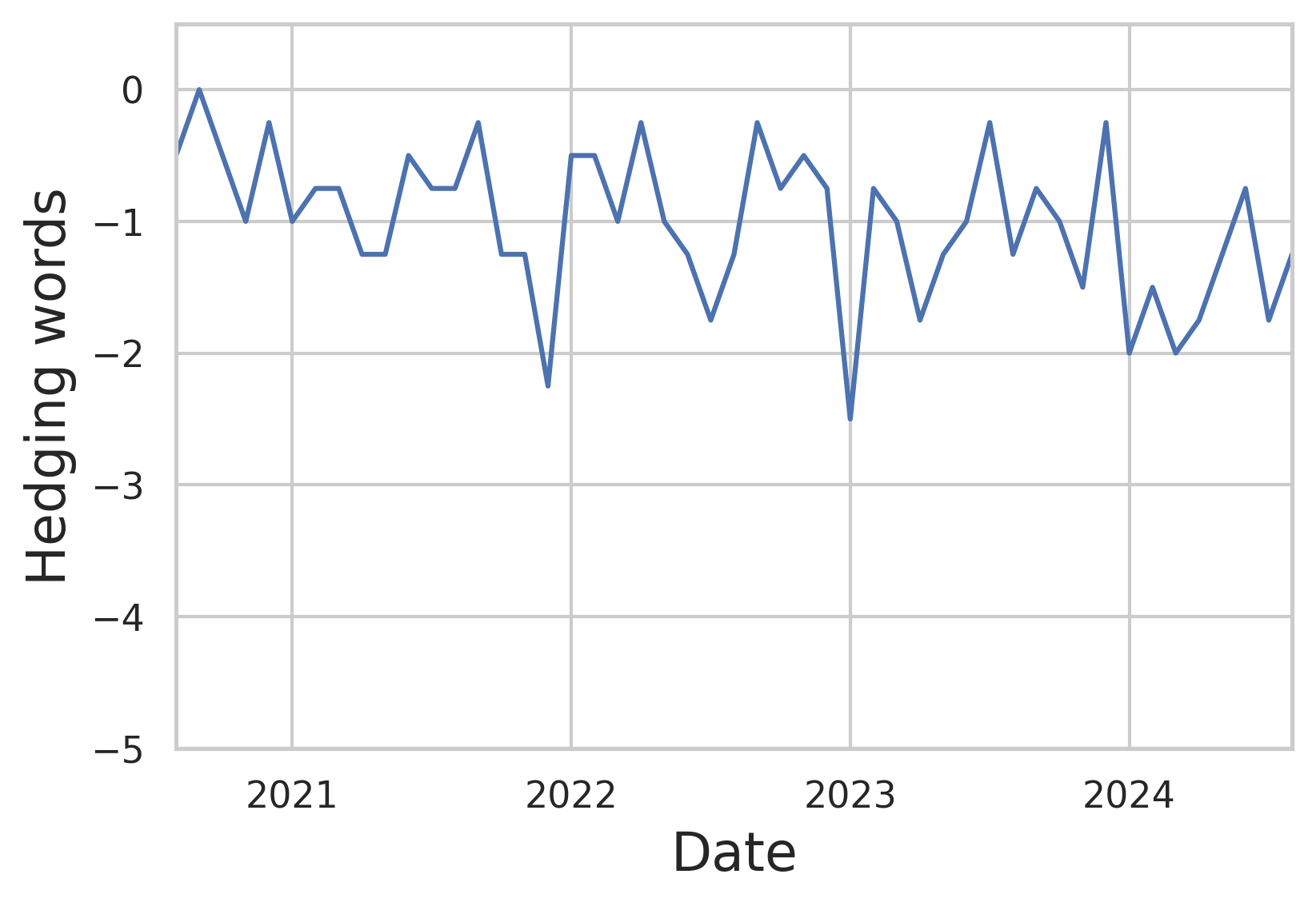}
        \caption{Gemini}
    \end{subfigure}
    \begin{subfigure}{0.24\textwidth}  
        \centering
        \includegraphics[width=\linewidth]{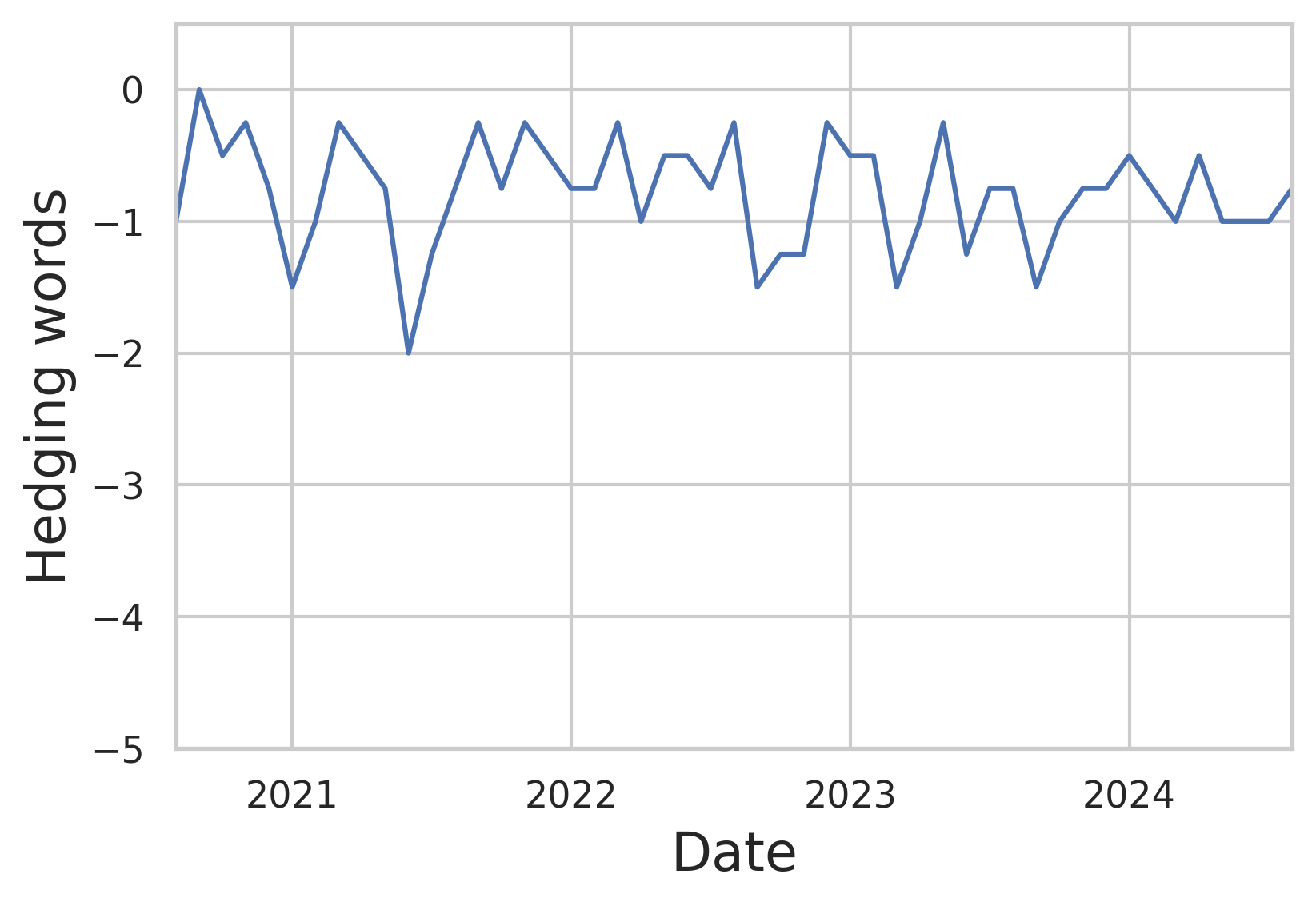}
        \caption{\llama}
    \end{subfigure}
    \caption{Evolution of hedging words over time for TI reports generated from real data.}
    \label{fig:hedge_ti_real}
\end{figure*}

\begin{figure*}[h!]
    \centering
    \begin{subfigure}{0.24\textwidth}  
        \centering
        \includegraphics[width=\linewidth]{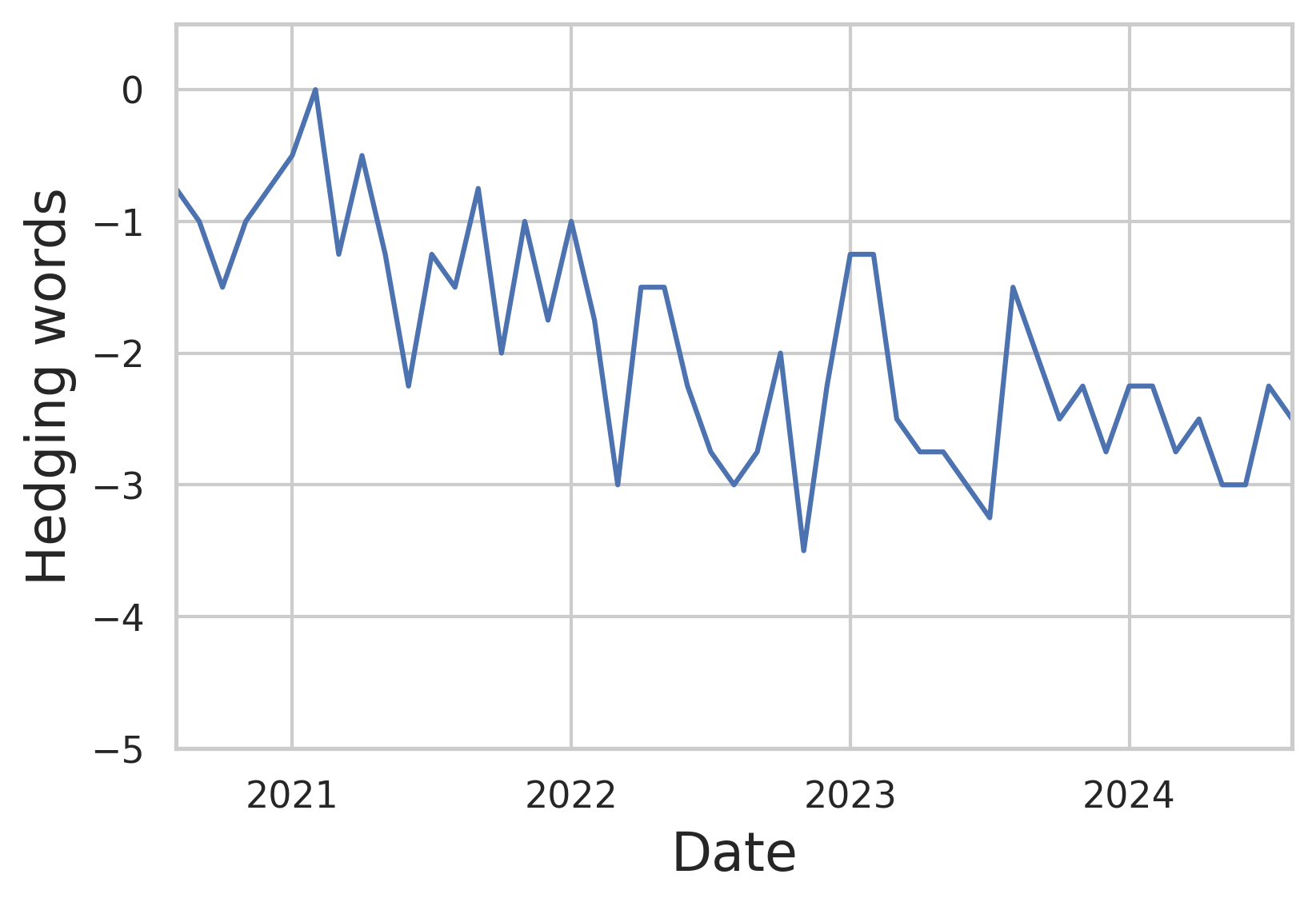}  
        \caption{GPT-4o}
    \end{subfigure}
    \begin{subfigure}{0.24\textwidth}  
        \centering
        \includegraphics[width=\linewidth]{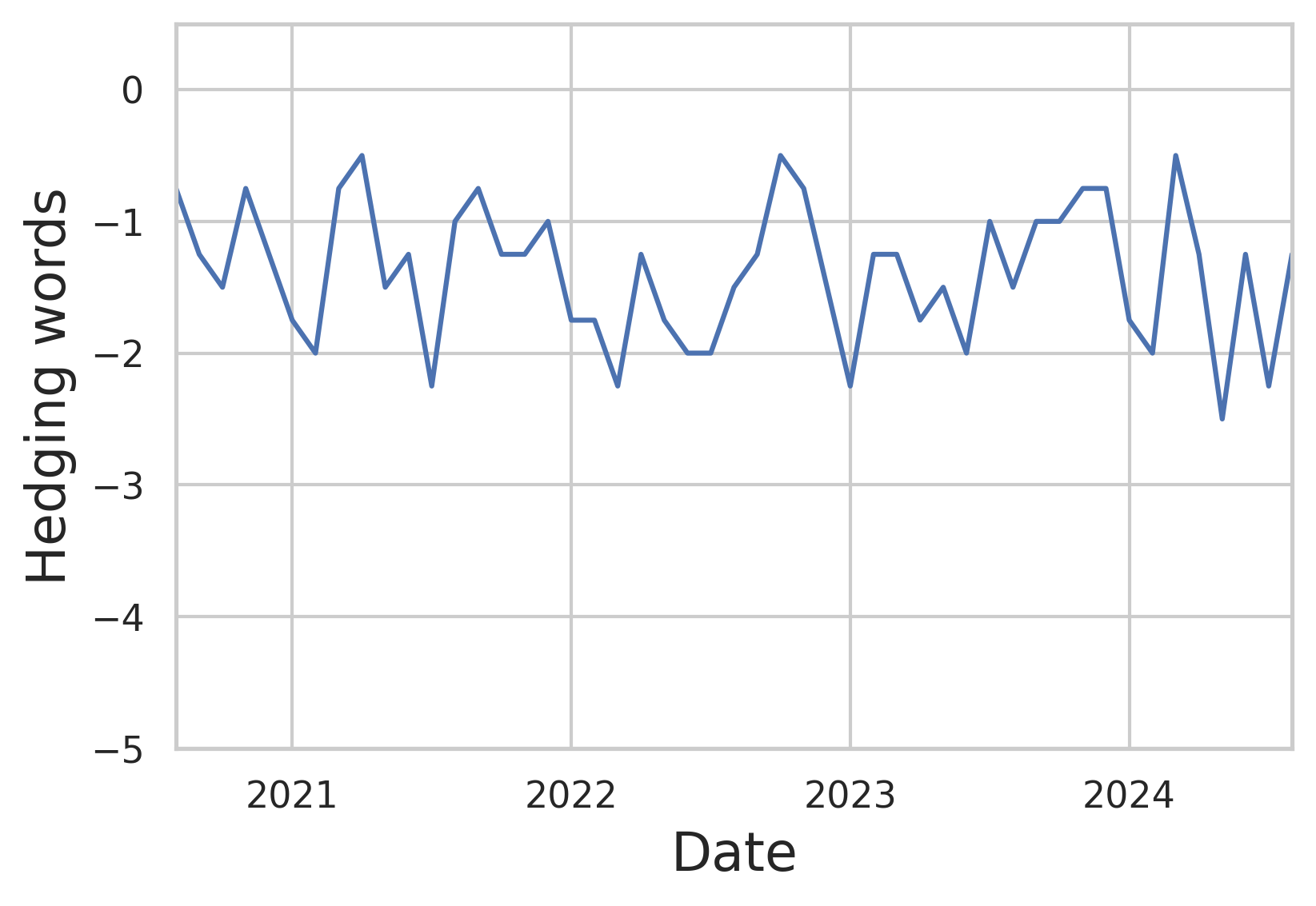}
        \caption{GPT-4o-mini}
    \end{subfigure}
    \begin{subfigure}{0.24\textwidth}  
        \centering
        \includegraphics[width=\linewidth]{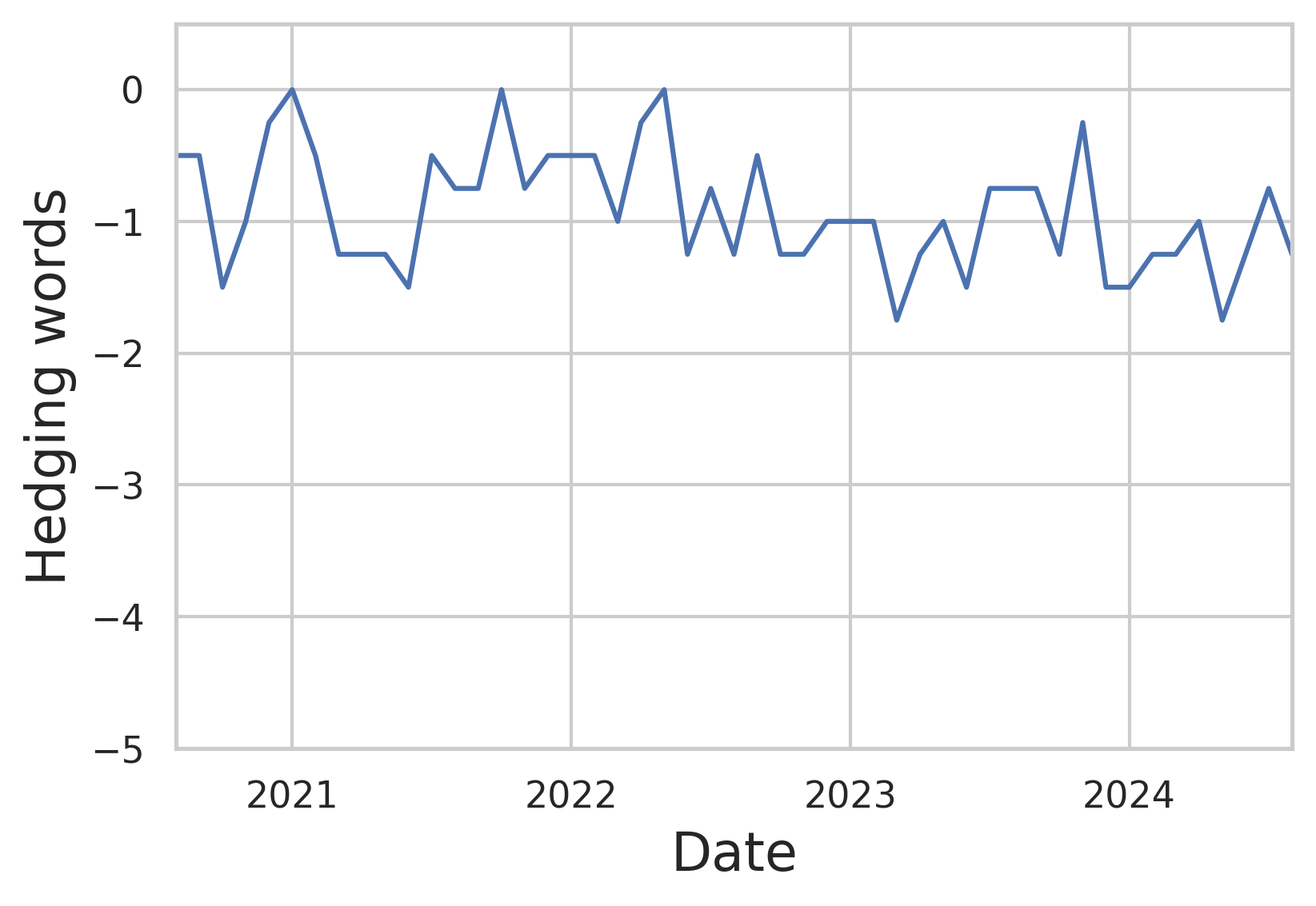}
        \caption{Gemini}
    \end{subfigure}
    \begin{subfigure}{0.24\textwidth}  
        \centering
        \includegraphics[width=\linewidth]{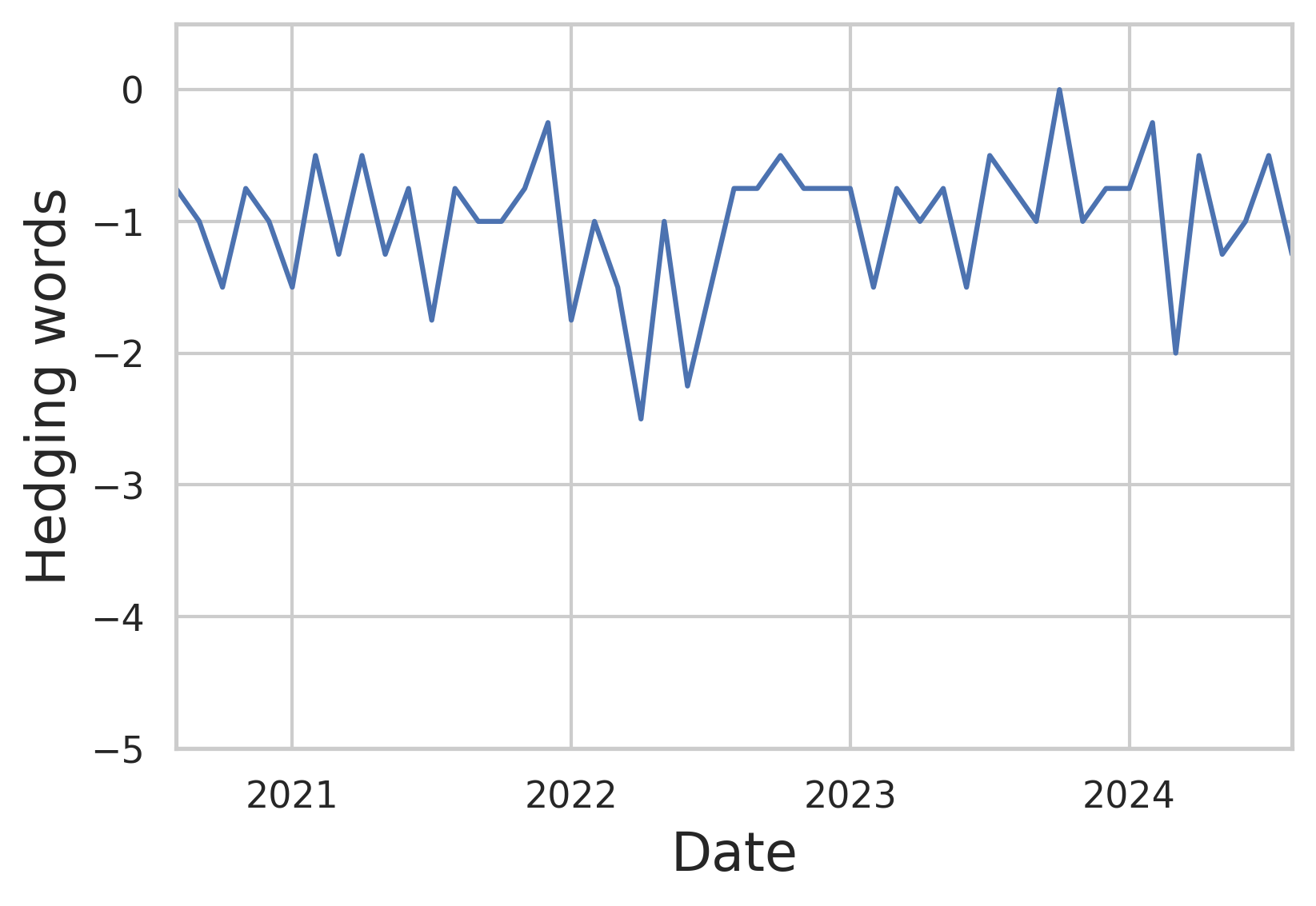}
        \caption{\llama}
    \end{subfigure}
    \caption{Evolution of hedging words over time for TI (plots) reports generated from real data.}
    \label{fig:hedge_ti_plots_real}
\end{figure*}

\begin{figure*}[h!]
    \centering
    \begin{subfigure}{0.32\textwidth}  
        \centering
        \includegraphics[width=\linewidth]{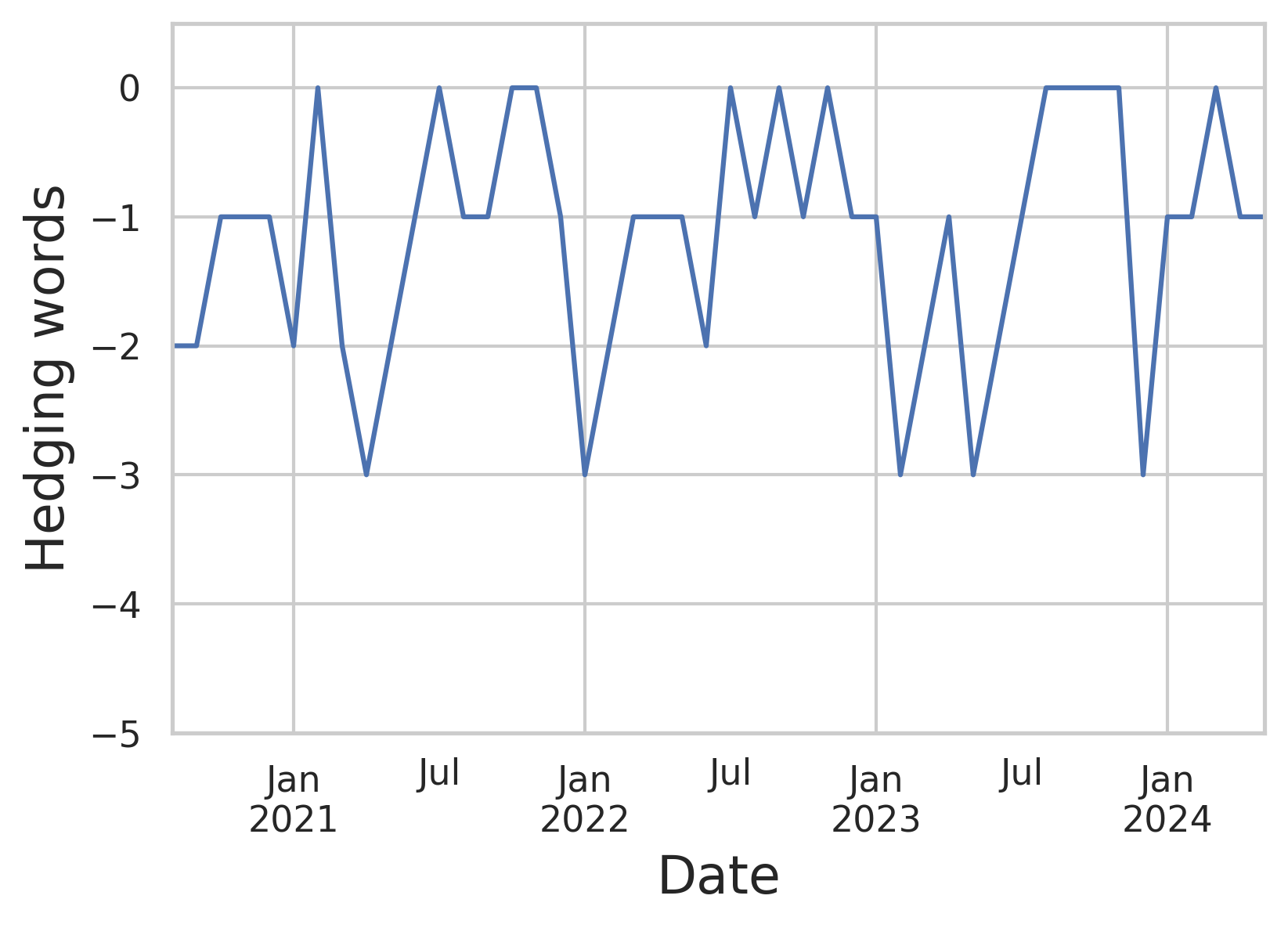}  
        \caption{GPT-4o}
    \end{subfigure}
    \begin{subfigure}{0.32\textwidth}  
        \centering
        \includegraphics[width=\linewidth]{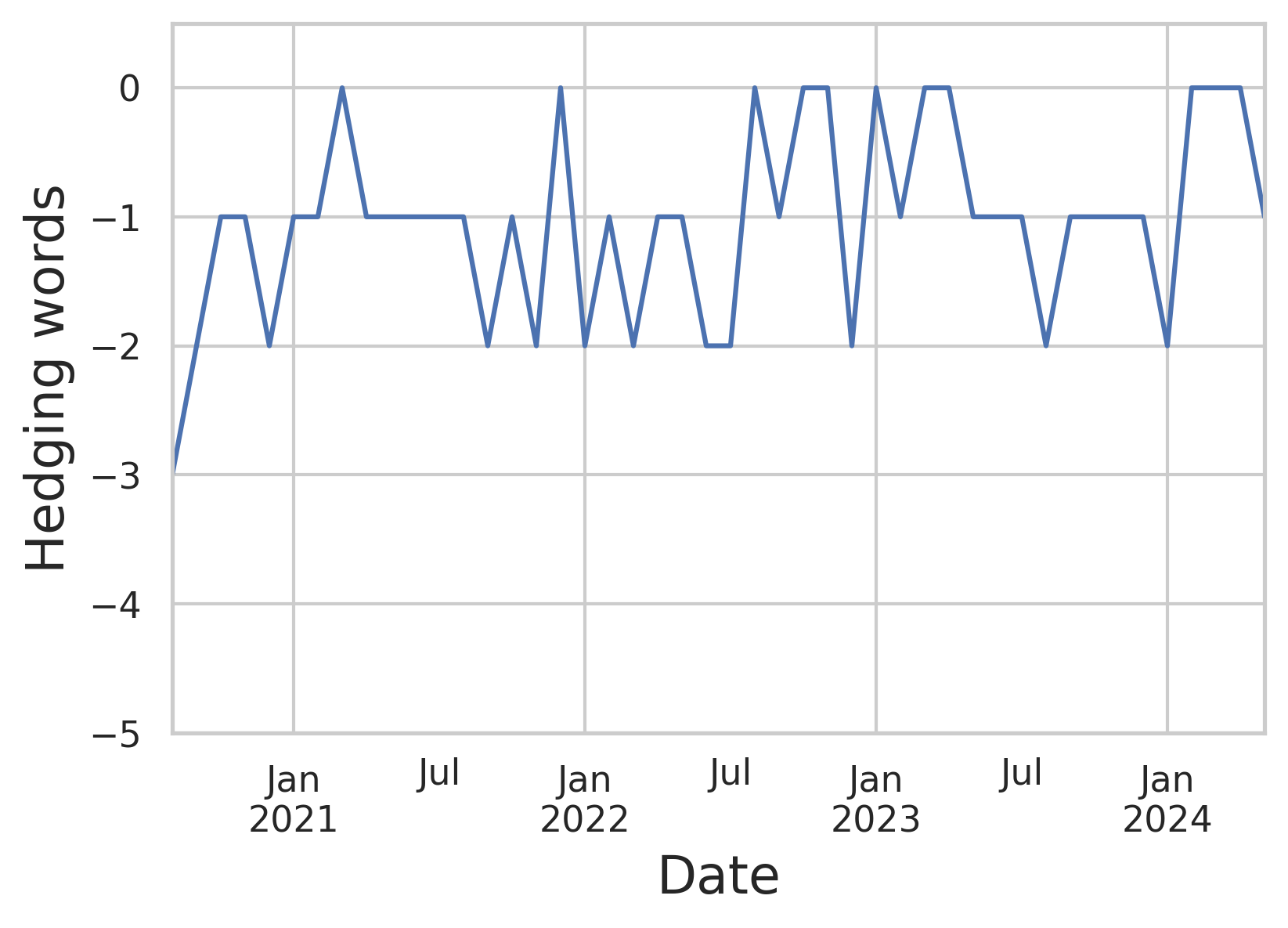}
        \caption{GPT-4o-mini}
    \end{subfigure}
    \begin{subfigure}{0.32\textwidth}  
        \centering
        \includegraphics[width=\linewidth]{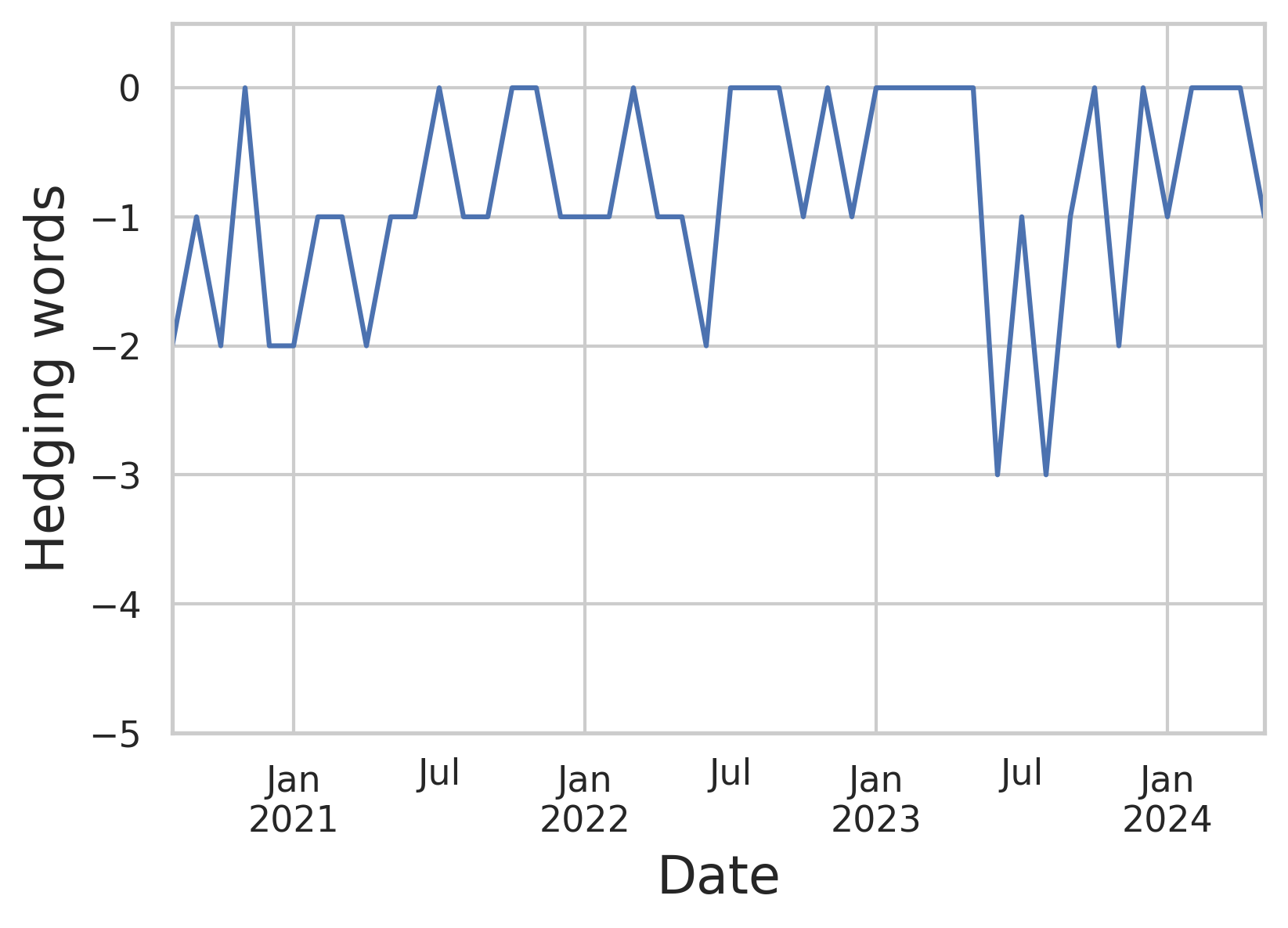}
        \caption{Gemini}
    \end{subfigure}\\
    \begin{subfigure}{0.32\textwidth}  
        \centering
        \includegraphics[width=\linewidth]{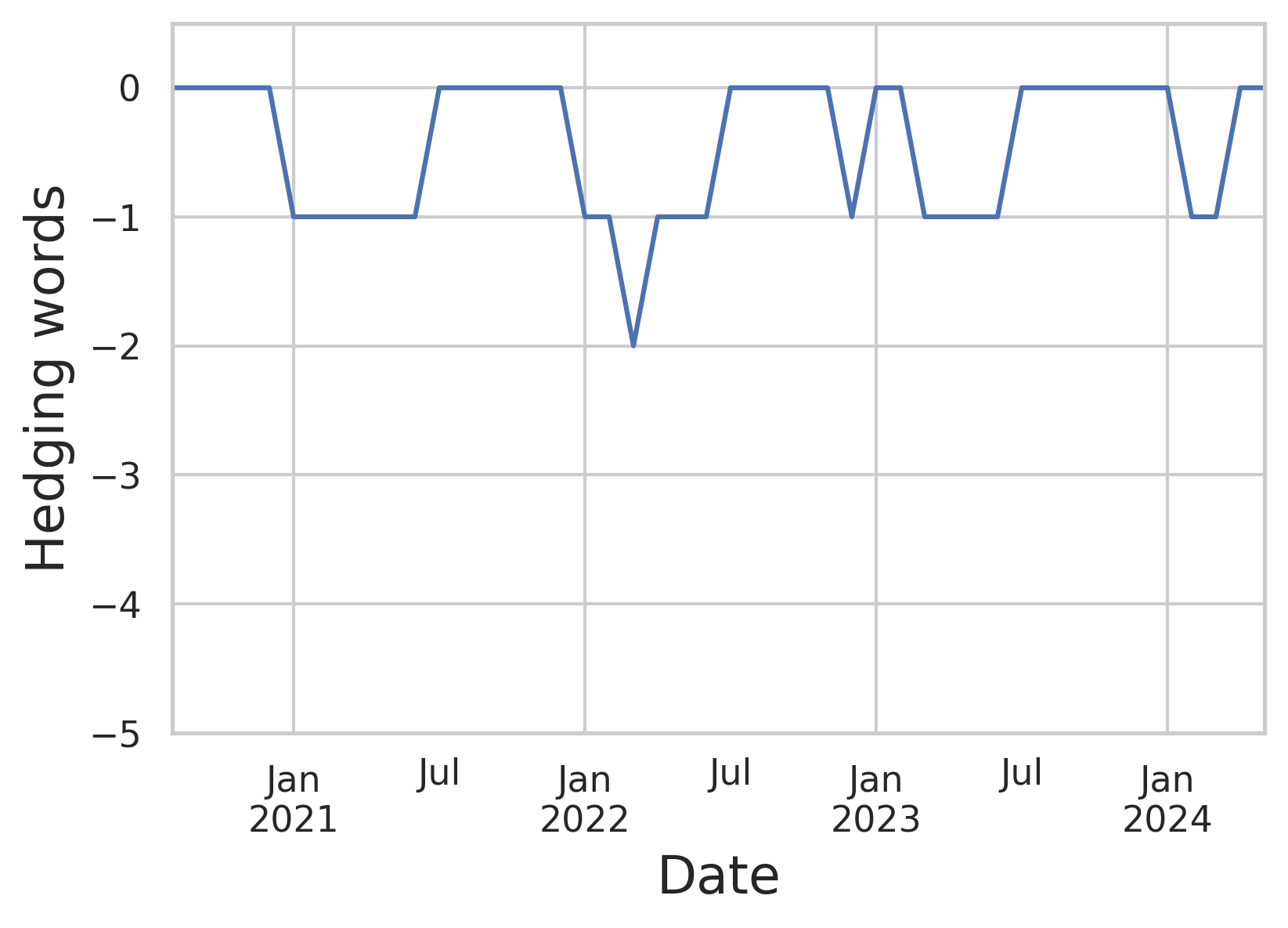}
        \caption{Phi-3}
    \end{subfigure}
        \begin{subfigure}{0.32\textwidth}  
        \centering
        \includegraphics[width=\linewidth]{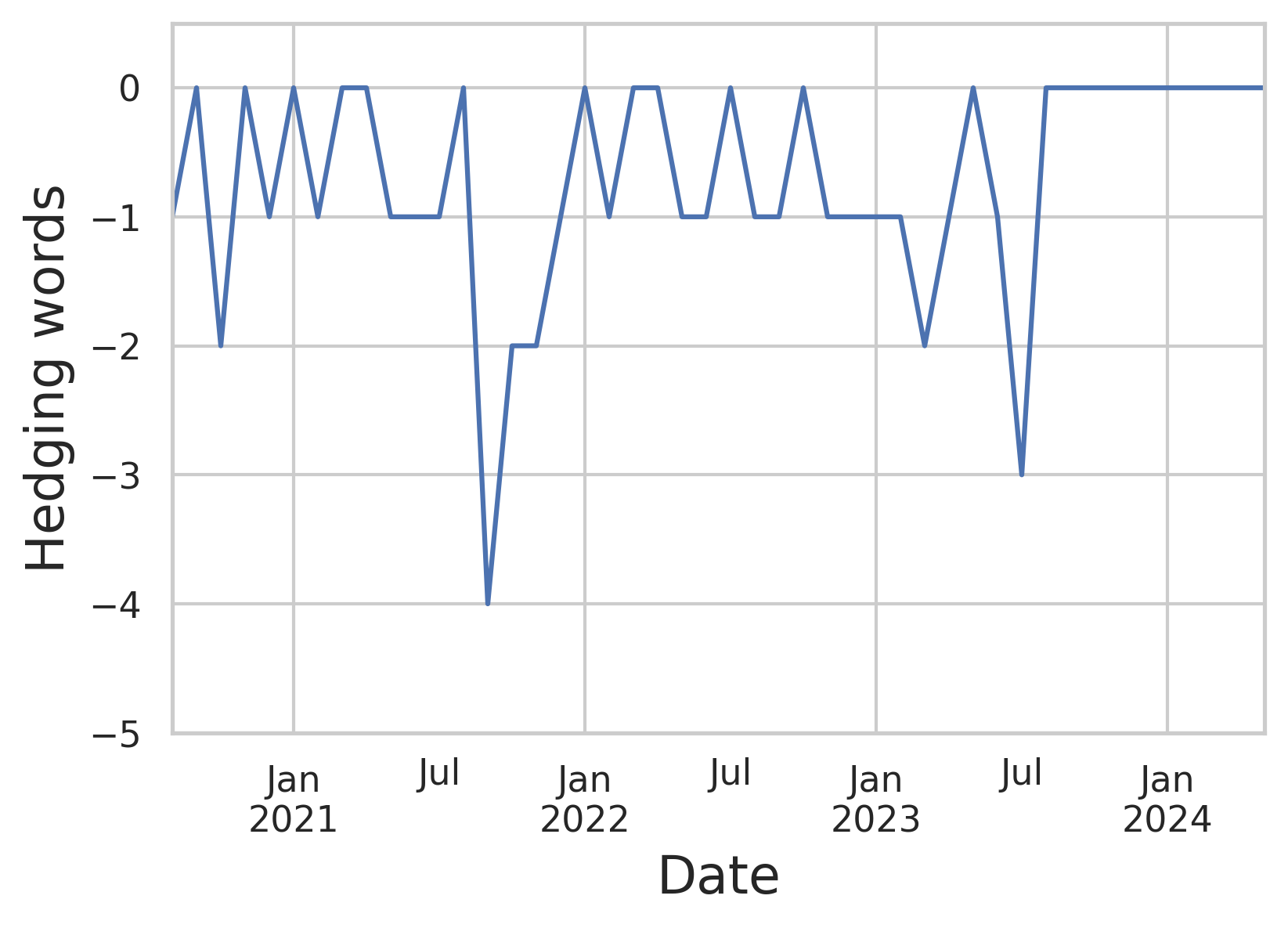}
        \caption{\llama}
    \end{subfigure}
    \caption{Evolution of hedging words over time for Short reports generated from synthetic data for the period 2019-2024.}
    \label{fig:hedge_short_syn}
\end{figure*}

\begin{figure*}[h!]
    \centering
    \begin{subfigure}{0.32\textwidth}  
        \centering
        \includegraphics[width=\linewidth]{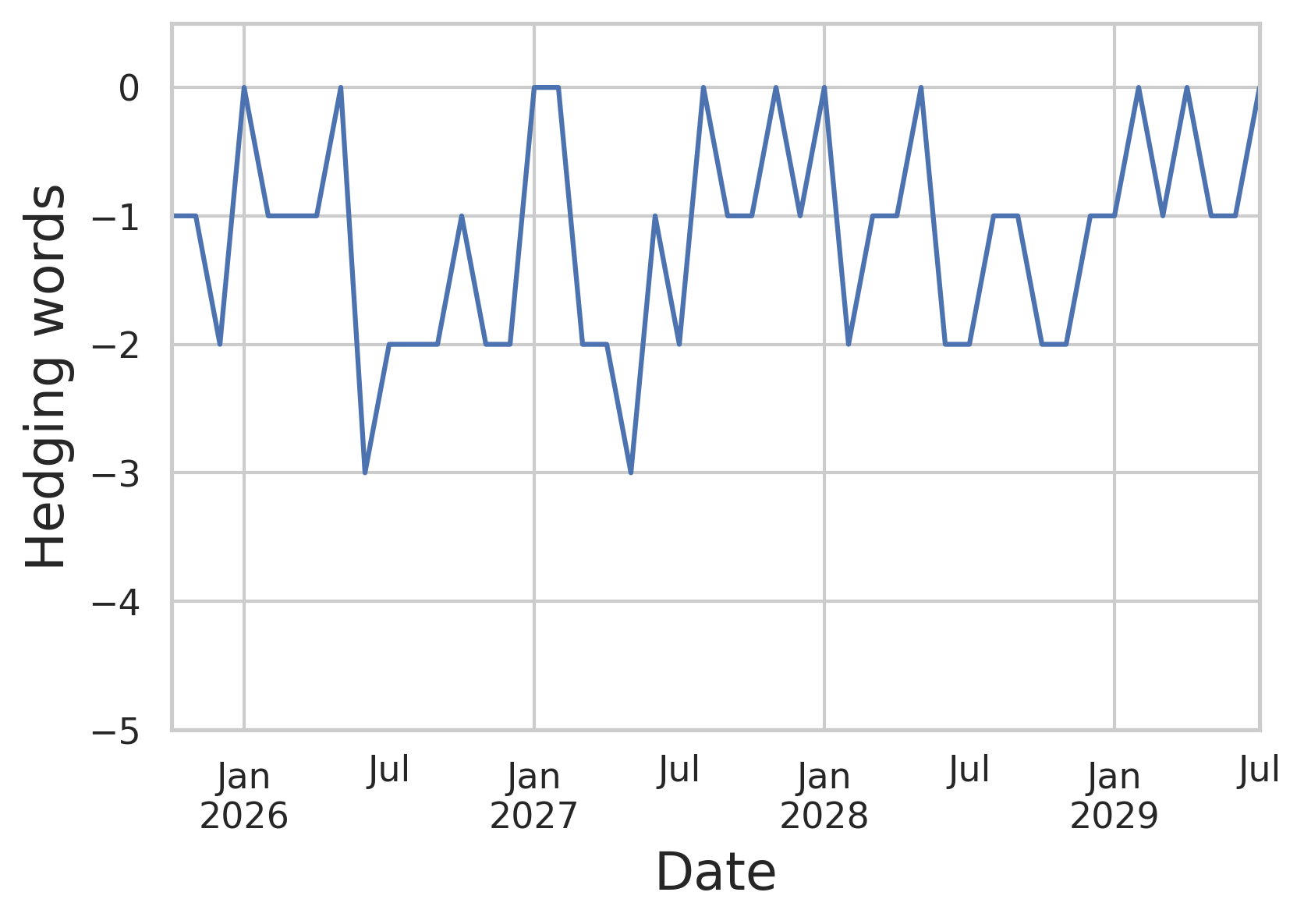}  
        \caption{GPT-4o}
    \end{subfigure}
    \begin{subfigure}{0.32\textwidth}  
        \centering
        \includegraphics[width=\linewidth]{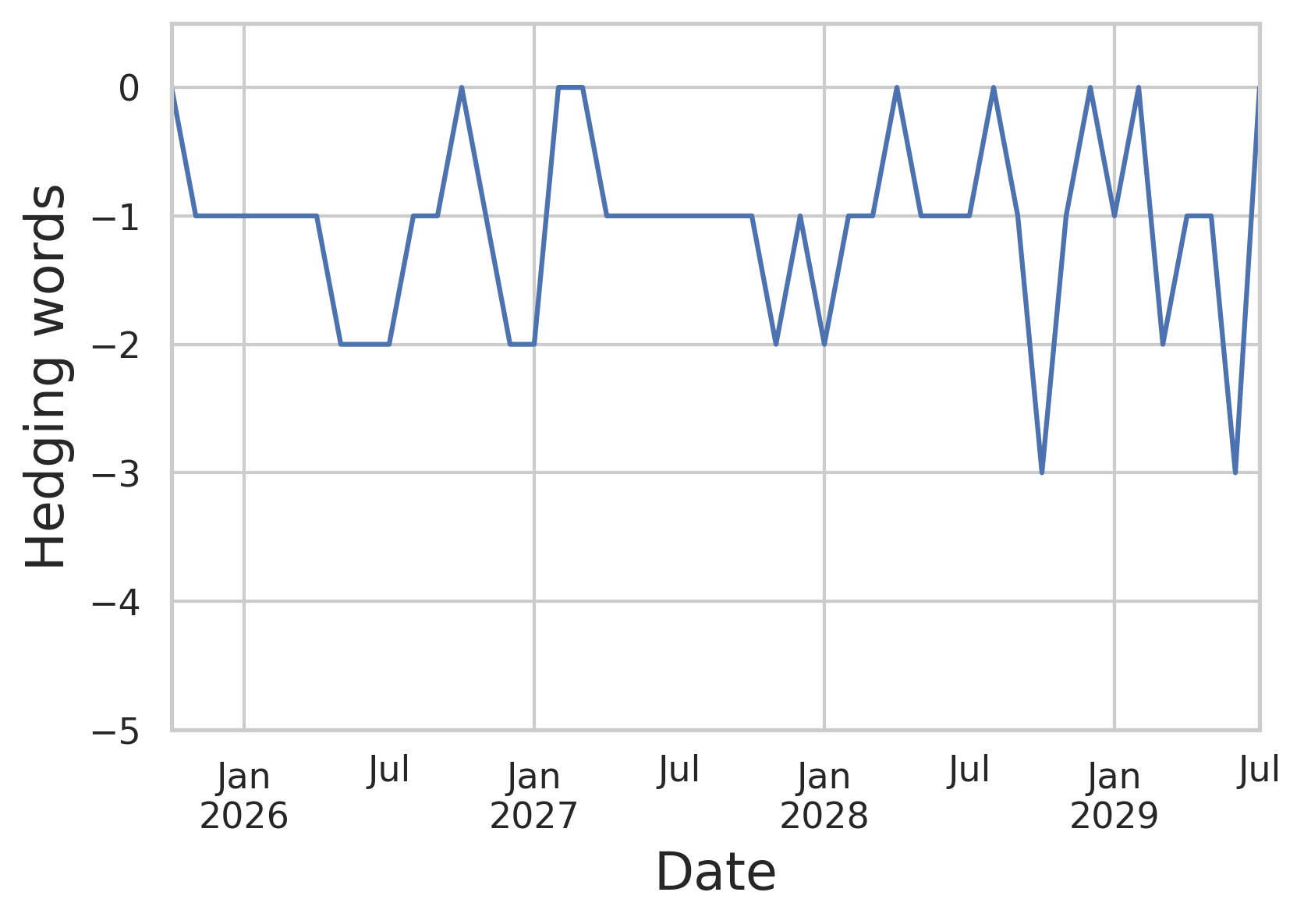}
        \caption{GPT-4o-mini}
    \end{subfigure}
    \begin{subfigure}{0.32\textwidth}  
        \centering
        \includegraphics[width=\linewidth]{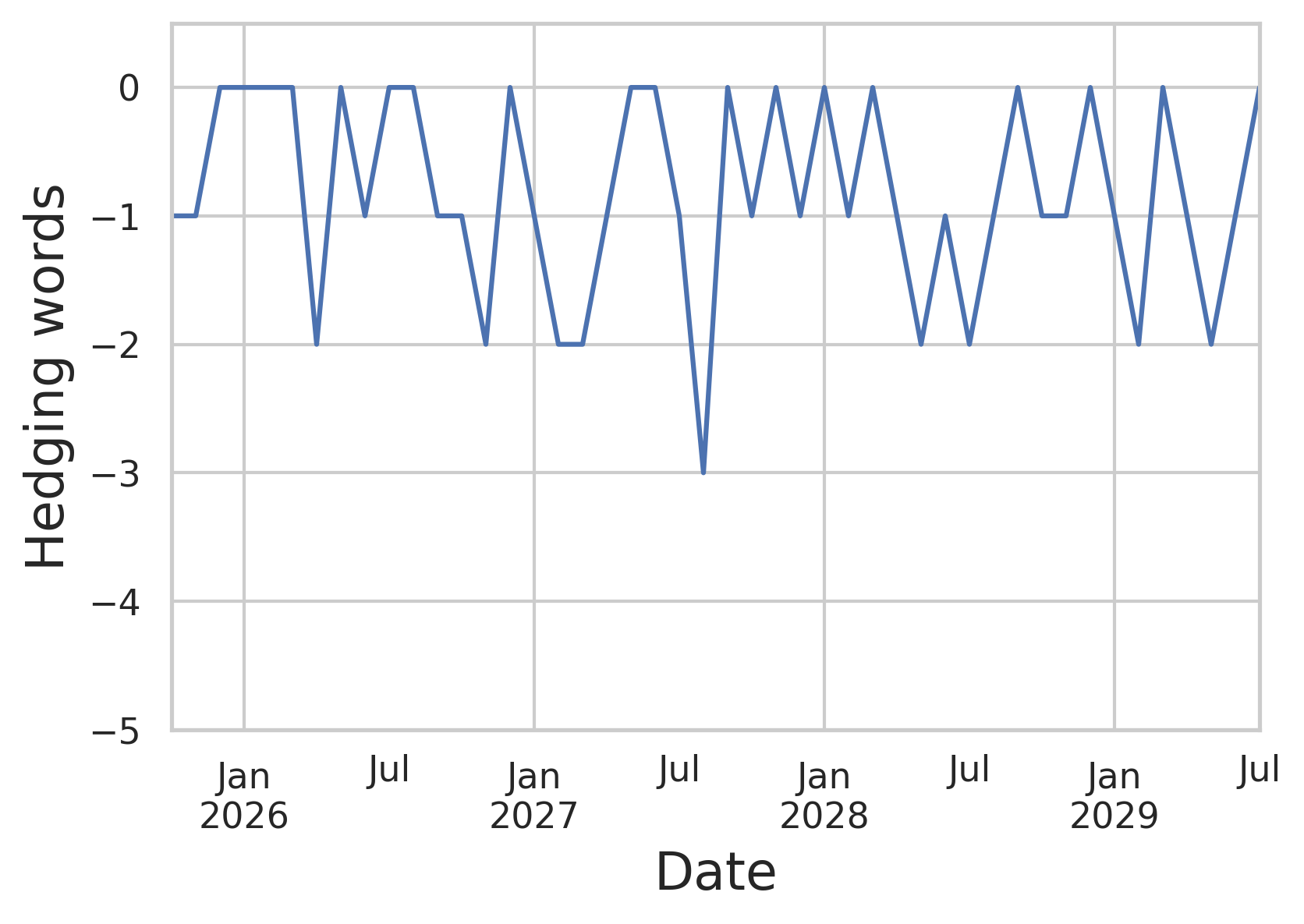}
        \caption{Gemini}
    \end{subfigure}\\
    \begin{subfigure}{0.32\textwidth}  
        \centering
        \includegraphics[width=\linewidth]{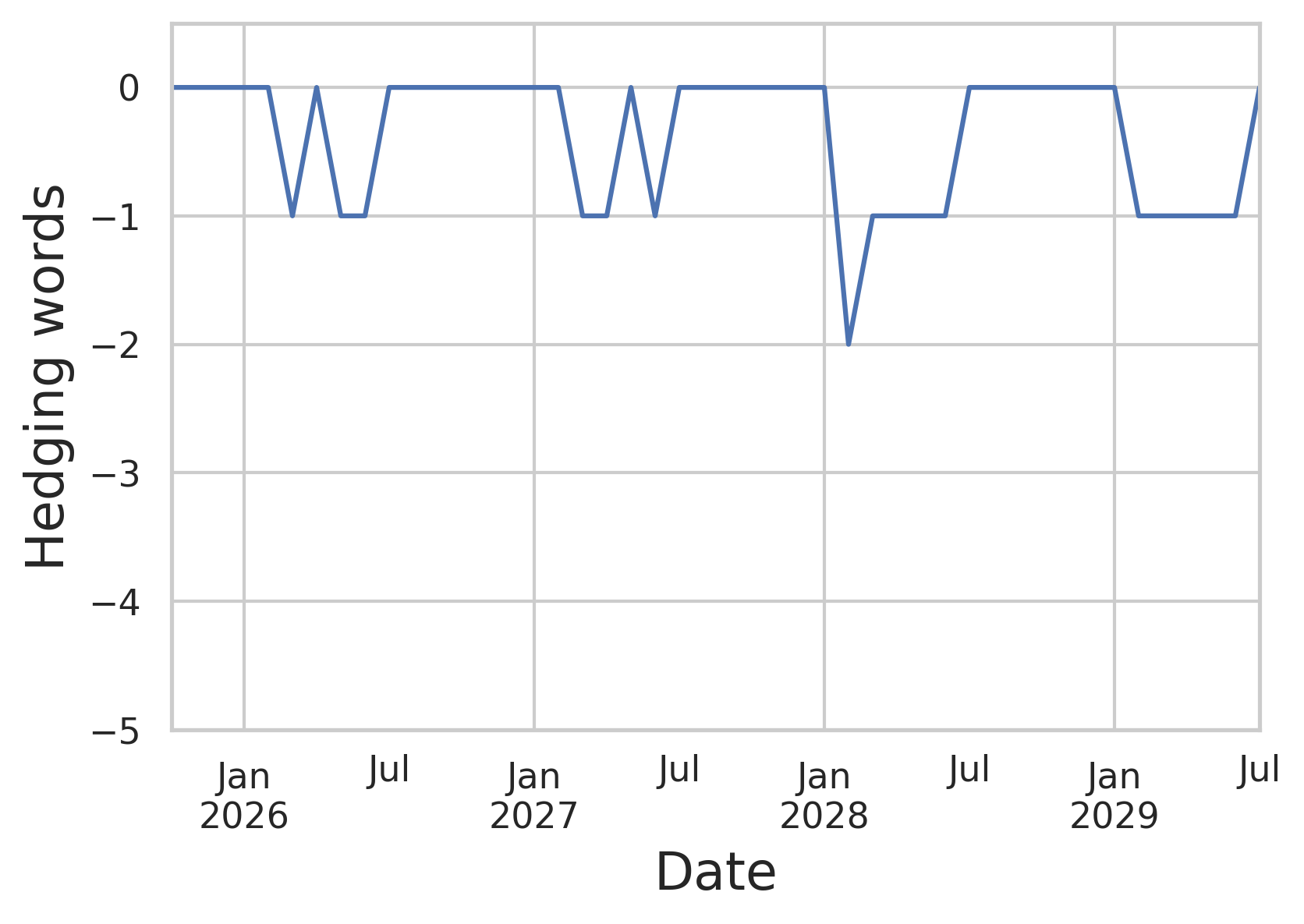}
        \caption{Phi-3}
    \end{subfigure}
        \begin{subfigure}{0.32\textwidth}  
        \centering
        \includegraphics[width=\linewidth]{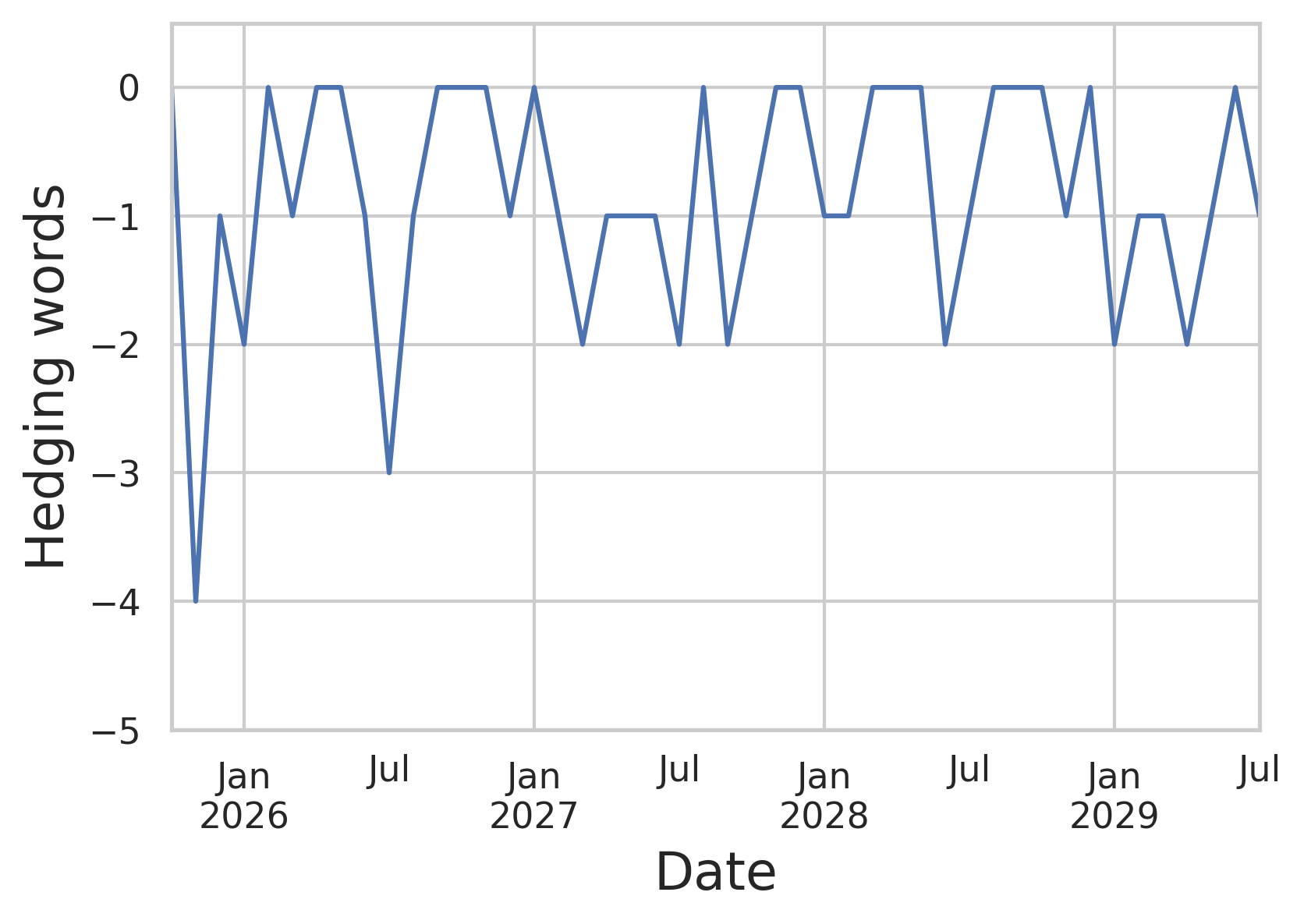}
        \caption{\llama}
    \end{subfigure}
    \caption{Evolution of hedging words over time for Short reports generated from synthetic data for the period 2024-2029.}
    \label{fig:hedge_short_syn2}
\end{figure*}

\begin{figure*}[h!]
    \centering
    \begin{subfigure}{0.24\textwidth}  
        \centering
        \includegraphics[width=\linewidth]{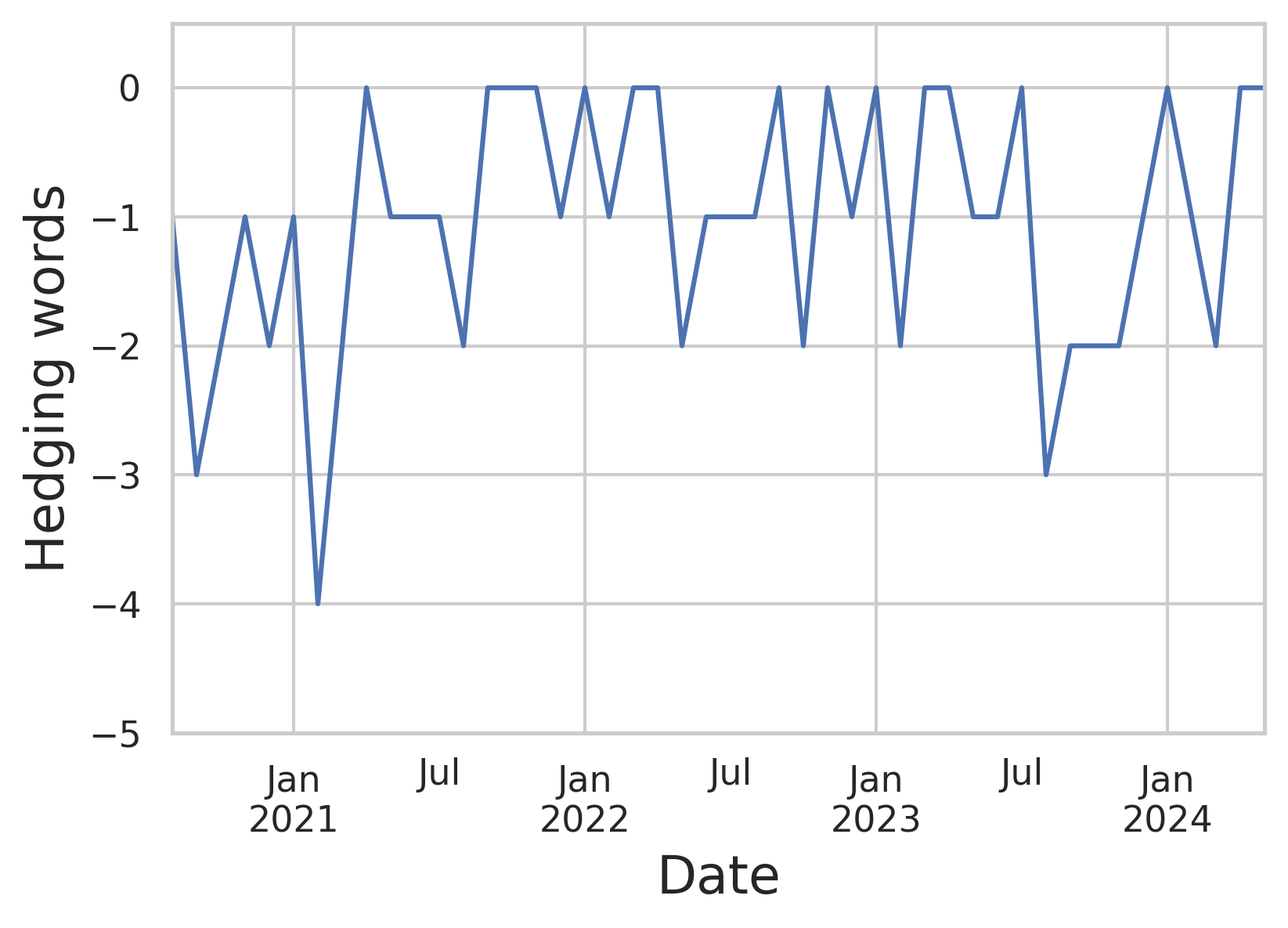}  
        \caption{GPT-4o}
    \end{subfigure}
    \begin{subfigure}{0.24\textwidth}  
        \centering
        \includegraphics[width=\linewidth]{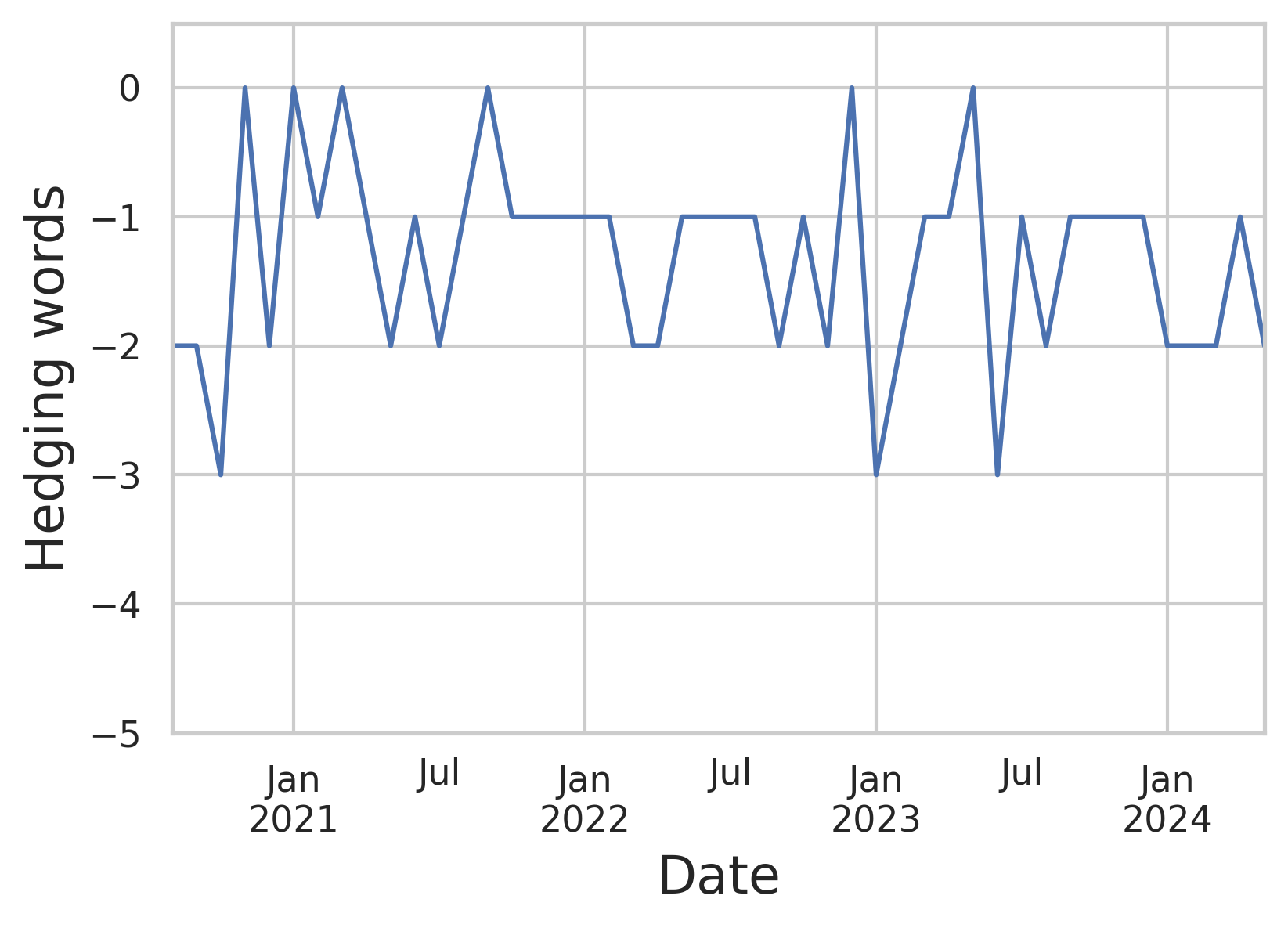}
        \caption{GPT-4o-mini}
    \end{subfigure}
    \begin{subfigure}{0.24\textwidth}  
        \centering
        \includegraphics[width=\linewidth]{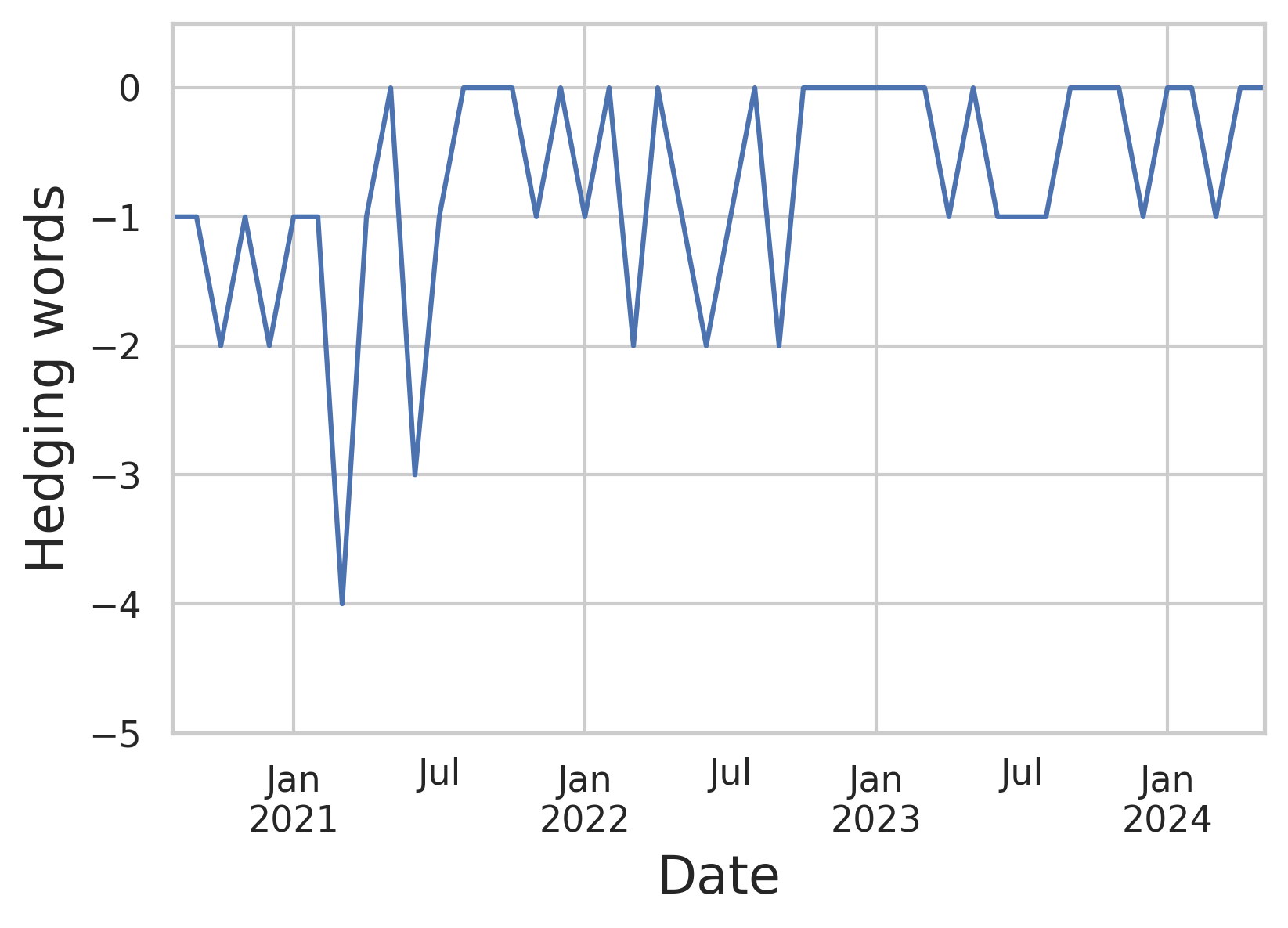}
        \caption{Gemini}
    \end{subfigure}
    \begin{subfigure}{0.24\textwidth}  
        \centering
        \includegraphics[width=\linewidth]{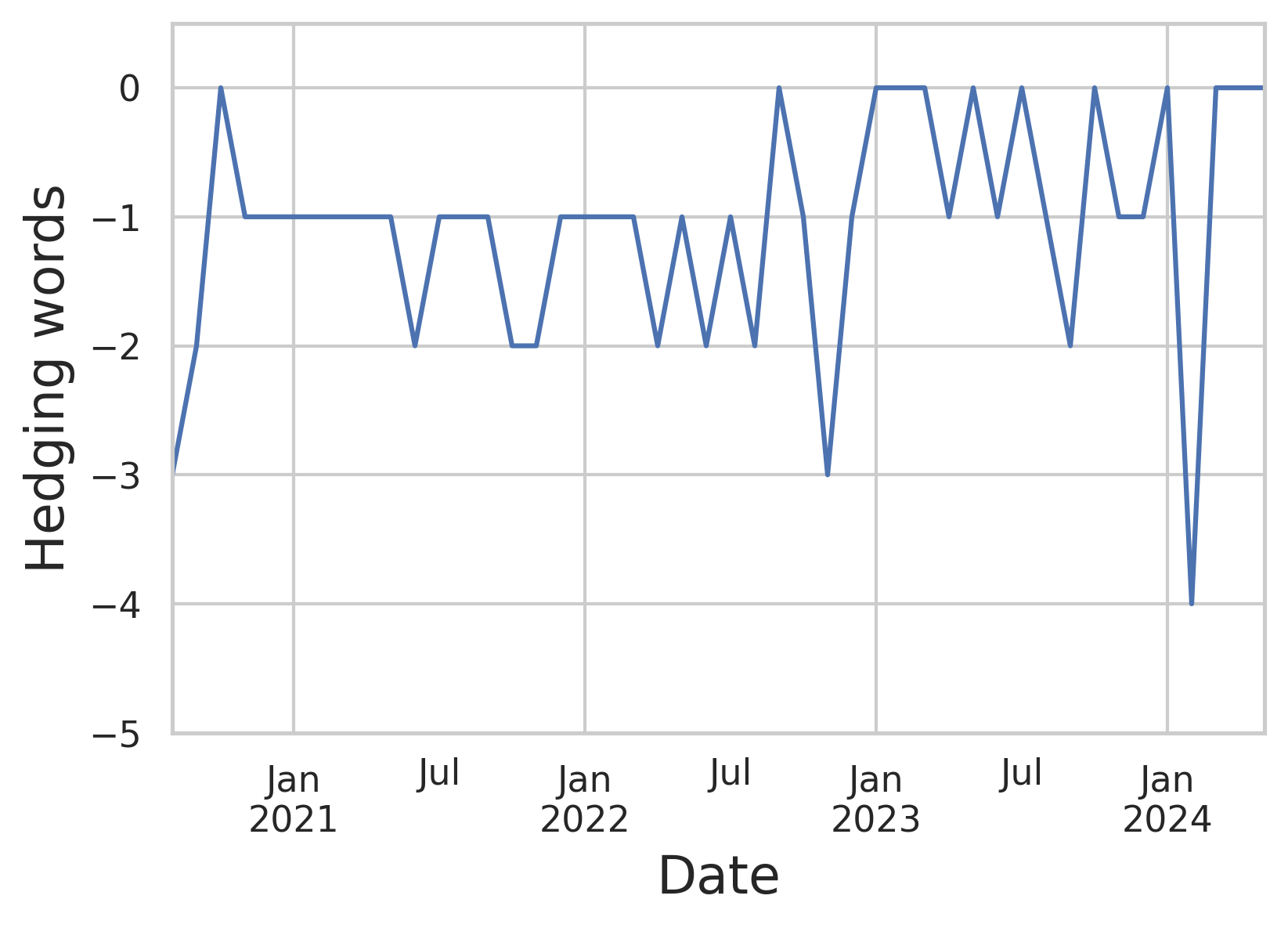}
        \caption{\llama}
    \end{subfigure}
    \caption{Evolution of hedging words over time for TI reports generated from from synthetic data for the period 2019-2024.}
    \label{fig:hedge_ti_syn}
\end{figure*}

\begin{figure*}[h!]
    \centering
    \begin{subfigure}{0.24\textwidth}  
        \centering
        \includegraphics[width=\linewidth]{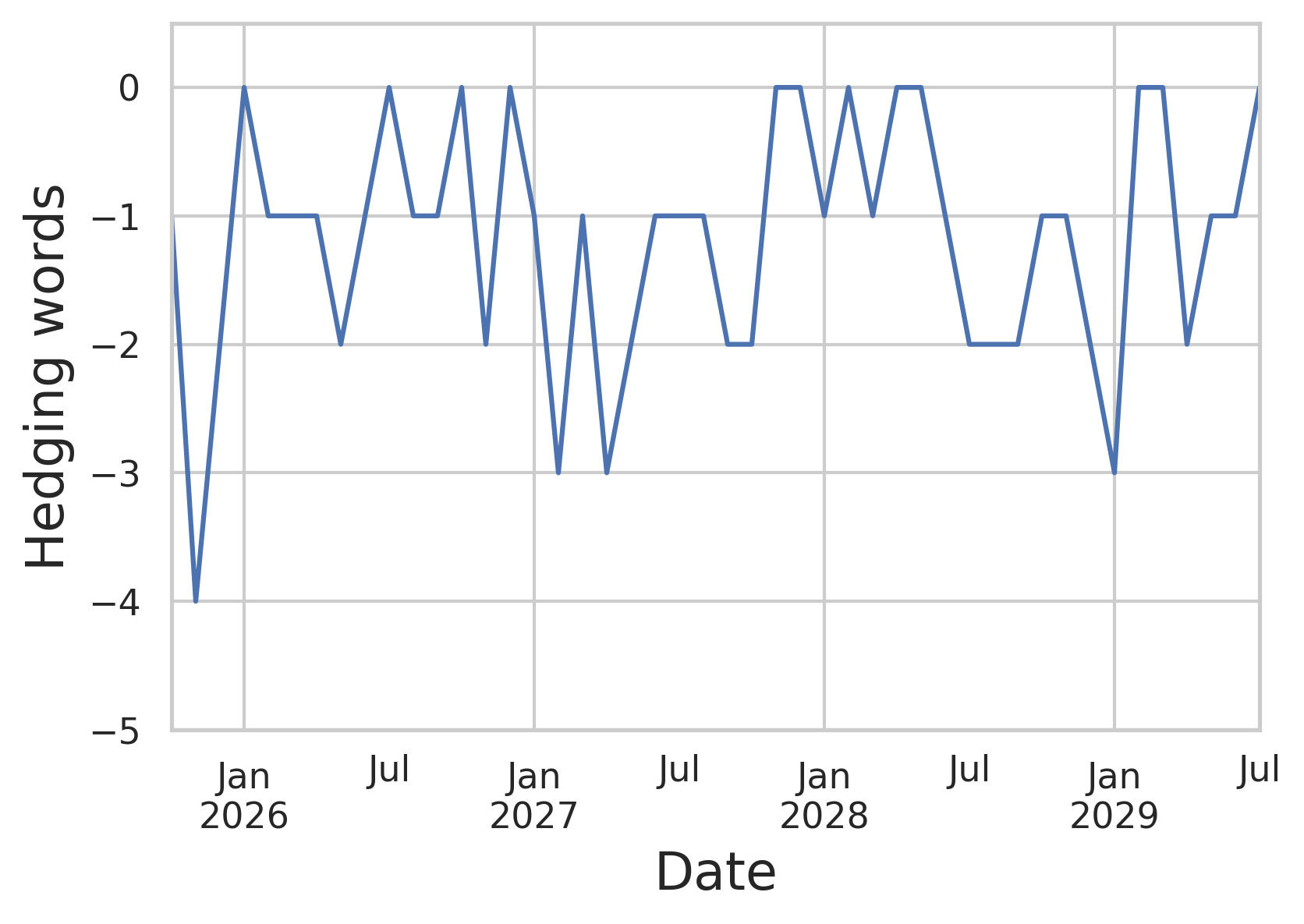}  
        \caption{GPT-4o}
    \end{subfigure}
    \begin{subfigure}{0.24\textwidth}  
        \centering
        \includegraphics[width=\linewidth]{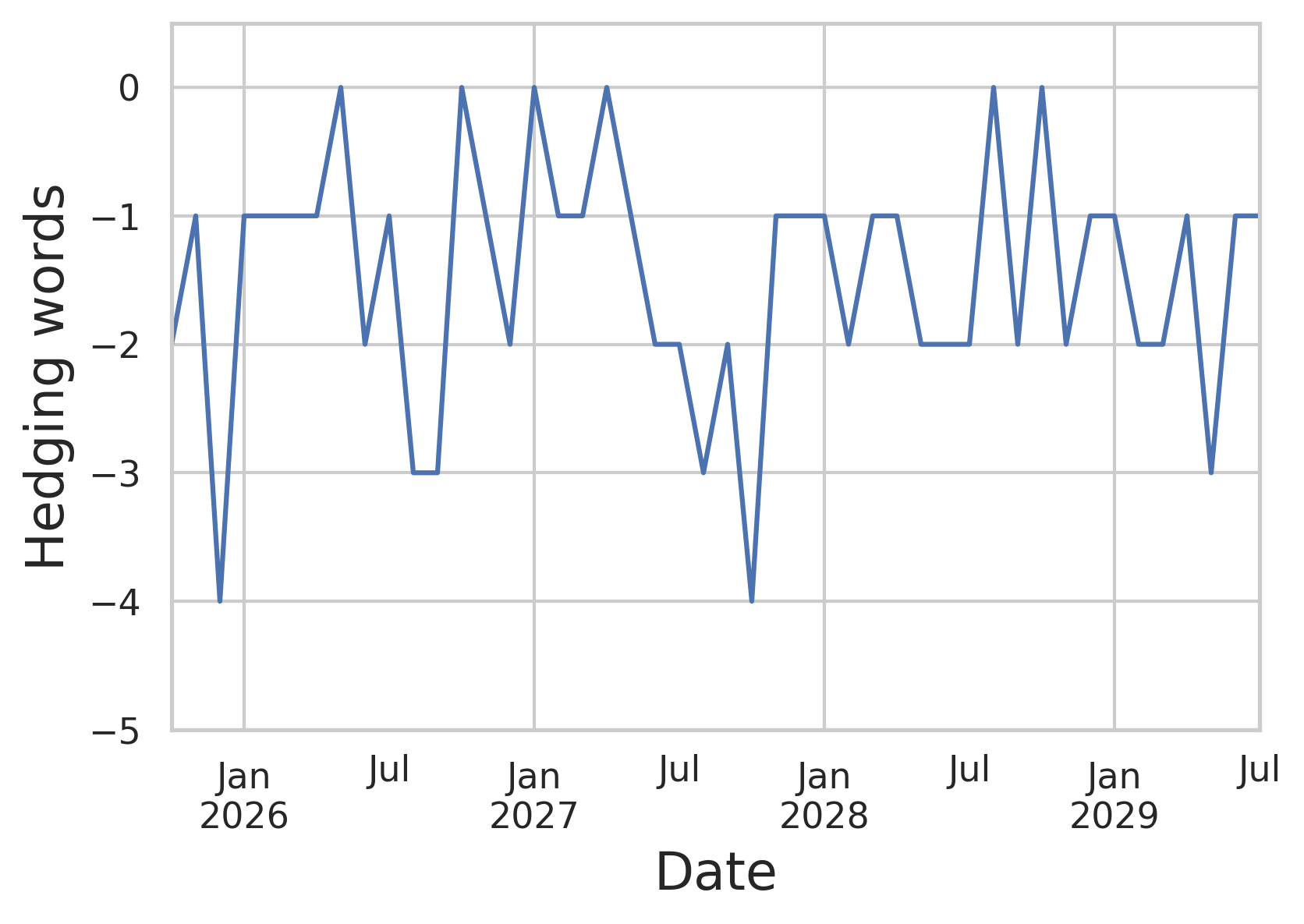}
        \caption{GPT-4o-mini}
    \end{subfigure}
    \begin{subfigure}{0.24\textwidth}  
        \centering
        \includegraphics[width=\linewidth]{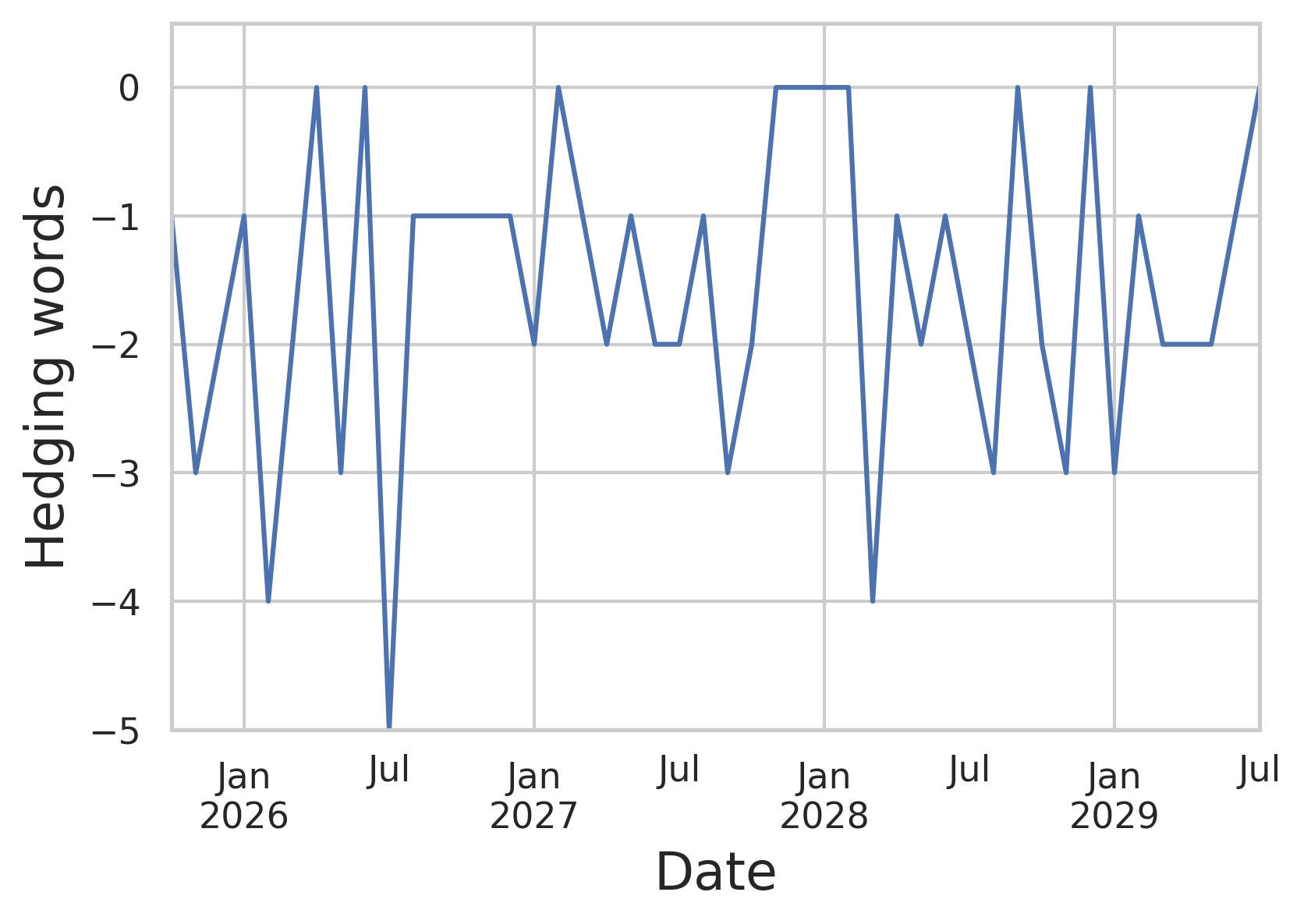}
        \caption{Gemini}
    \end{subfigure}
    \begin{subfigure}{0.24\textwidth}  
        \centering
        \includegraphics[width=\linewidth]{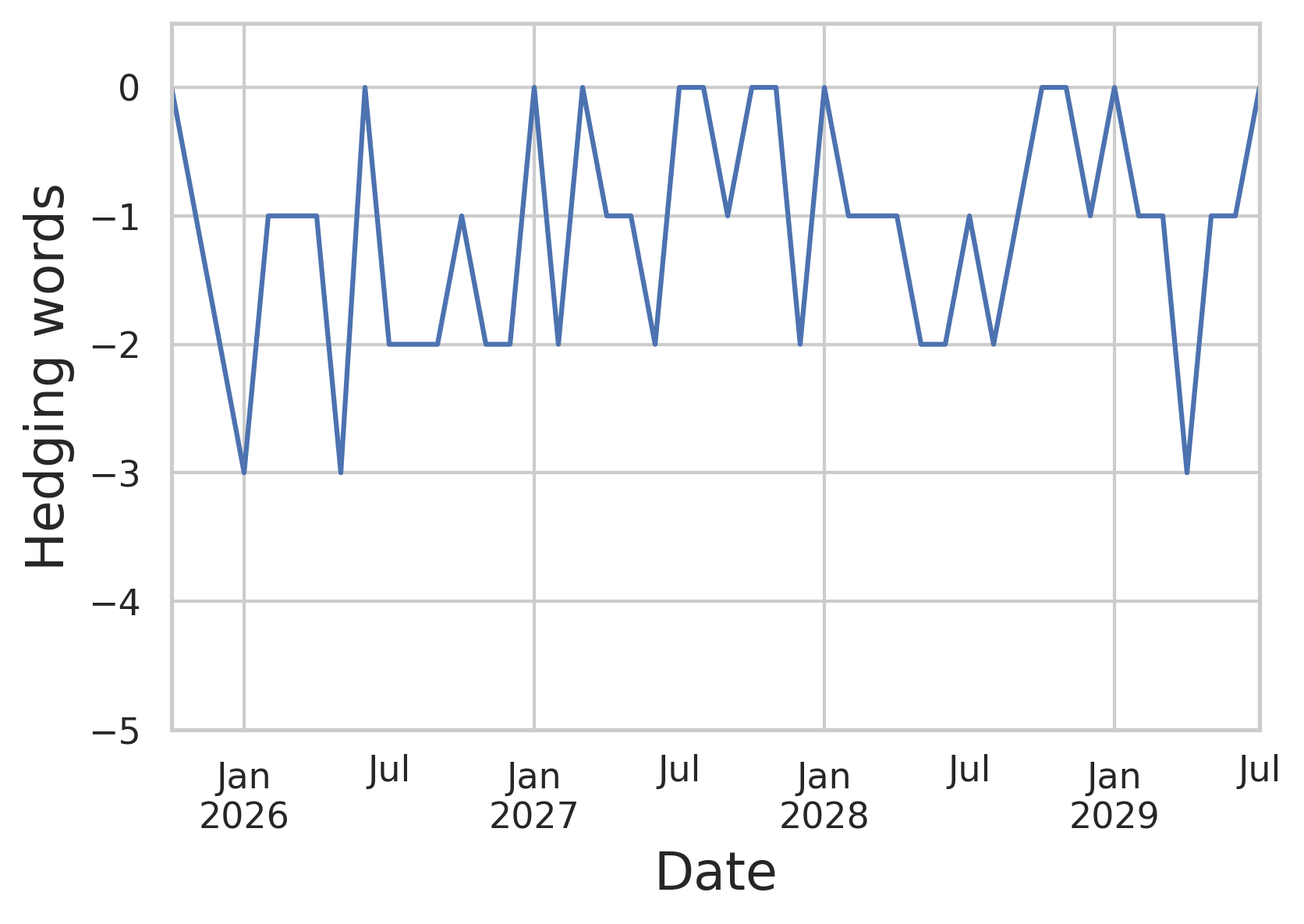}
        \caption{\llama}
    \end{subfigure}
    \caption{Evolution of hedging words over time for TI reports generated from from synthetic data for the period 2024-2029.}
    \label{fig:hedge_ti_syn2}
\end{figure*}


\begin{figure*}[h!]
    \centering
    \begin{subfigure}{0.24\textwidth}  
        \centering
        \includegraphics[width=\linewidth]{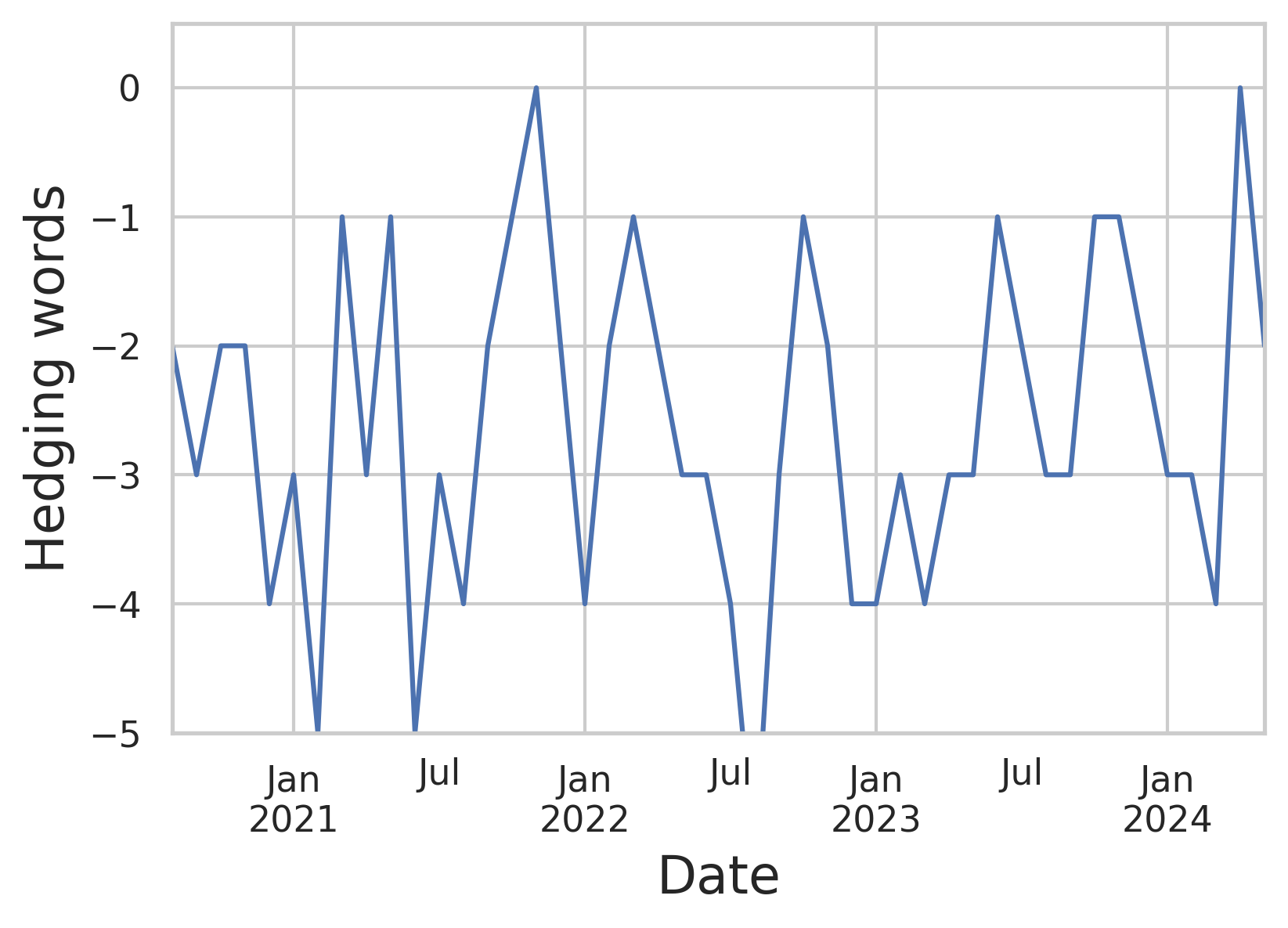}  
        \caption{GPT-4o}
    \end{subfigure}
    \begin{subfigure}{0.24\textwidth}  
        \centering
        \includegraphics[width=\linewidth]{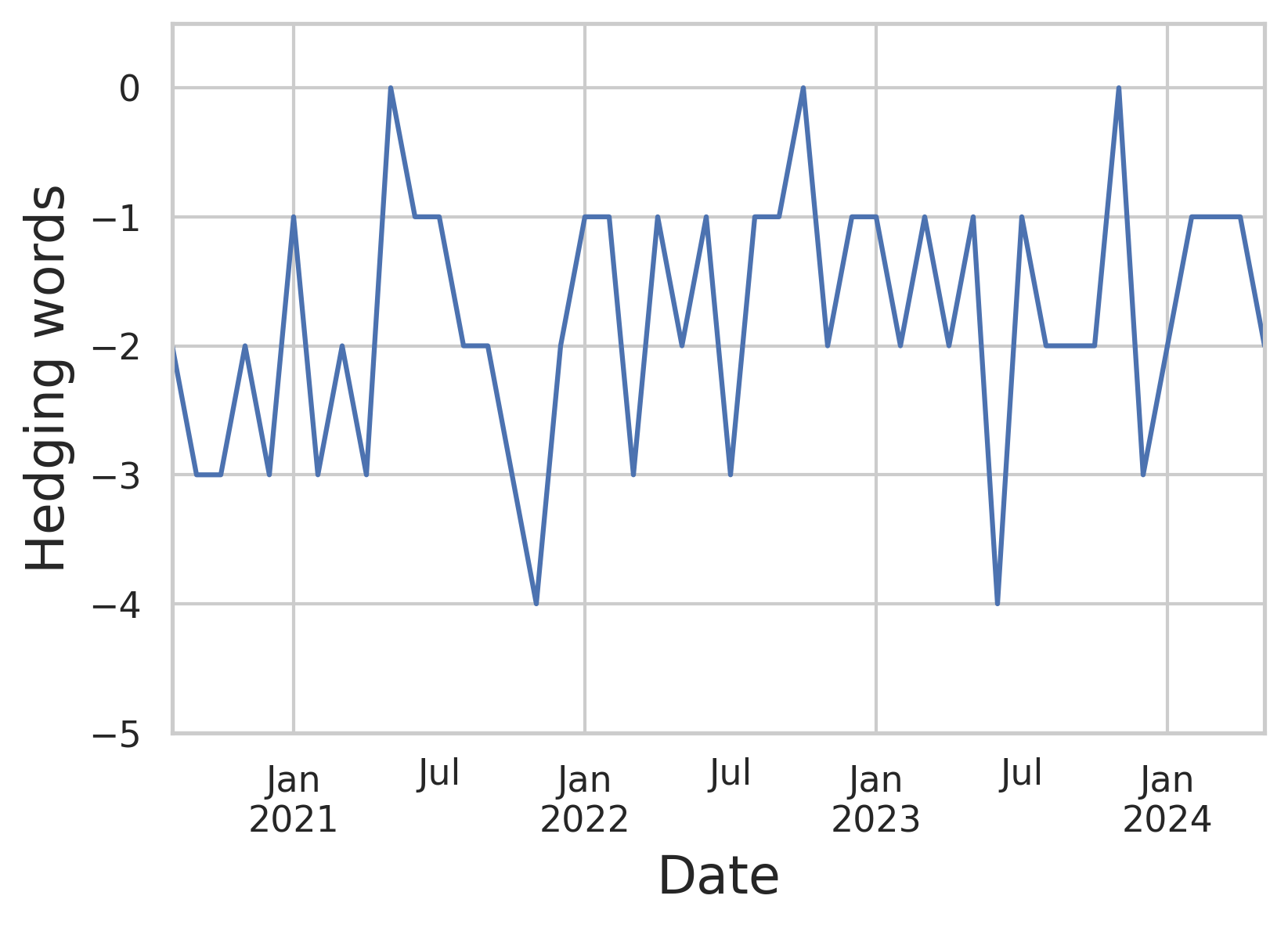}
        \caption{GPT-4o-mini}
    \end{subfigure}
    \begin{subfigure}{0.24\textwidth}  
        \centering
        \includegraphics[width=\linewidth]{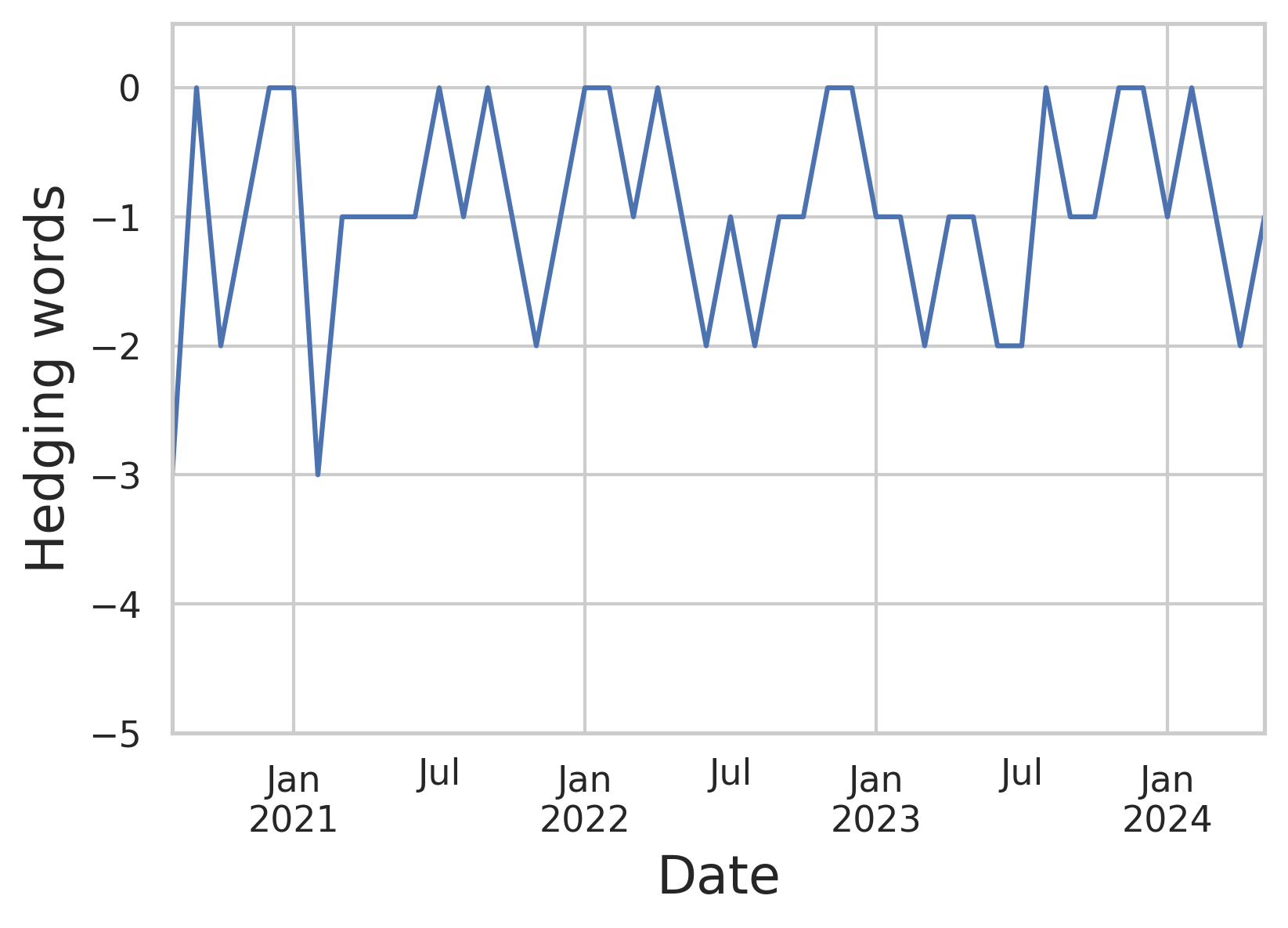}
        \caption{Gemini}
    \end{subfigure}
    \begin{subfigure}{0.24\textwidth}  
        \centering
        \includegraphics[width=\linewidth]{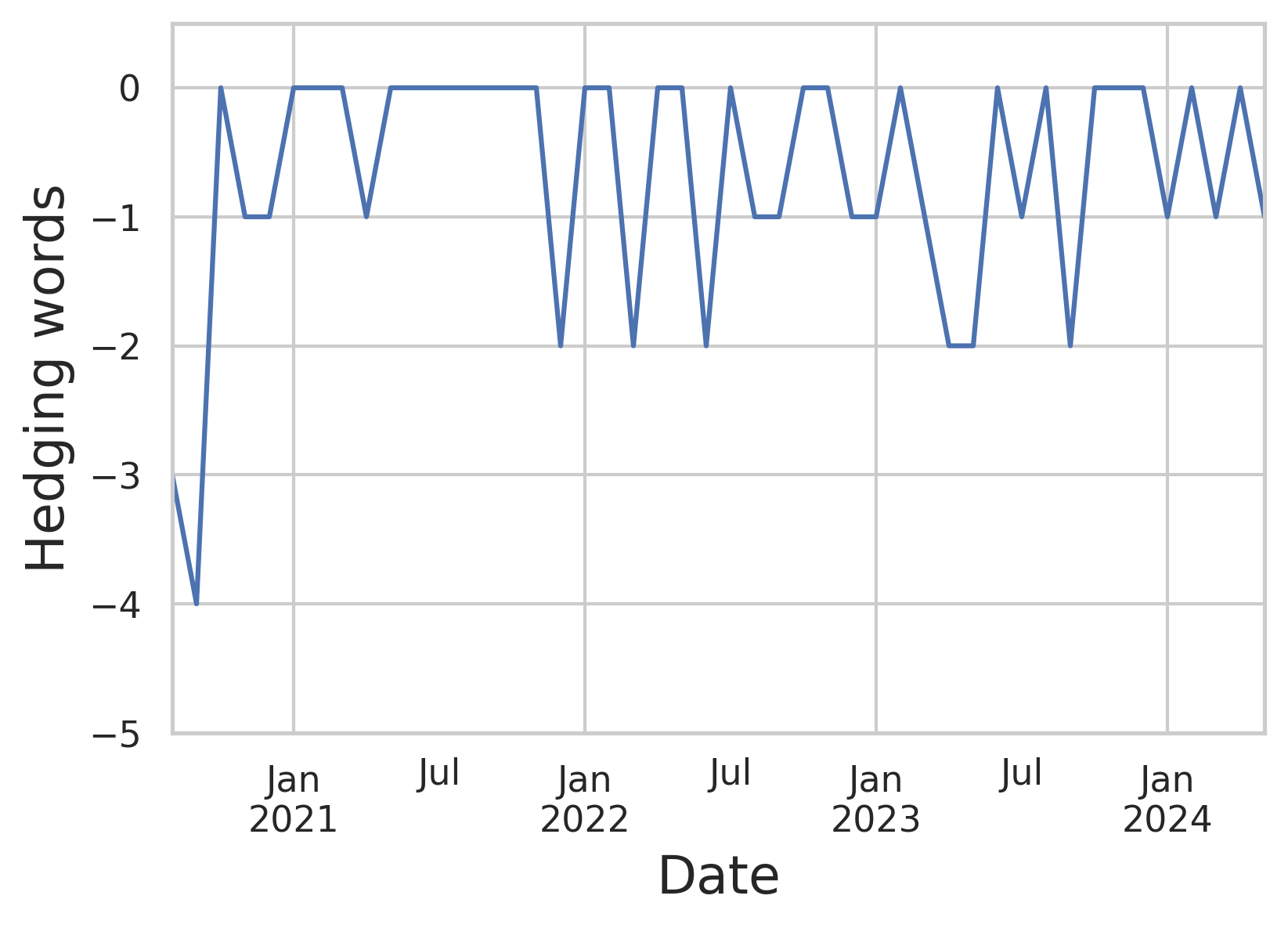}
        \caption{\llama}
    \end{subfigure}
    \caption{Evolution of hedging words over time for TI (plots) reports generated from from synthetic data for the period 2019-2024.}
    \label{fig:hedge_ti_plots_syn}
\end{figure*}

\begin{figure*}[h!]
    \centering
    \begin{subfigure}{0.24\textwidth}  
        \centering
        \includegraphics[width=\linewidth]{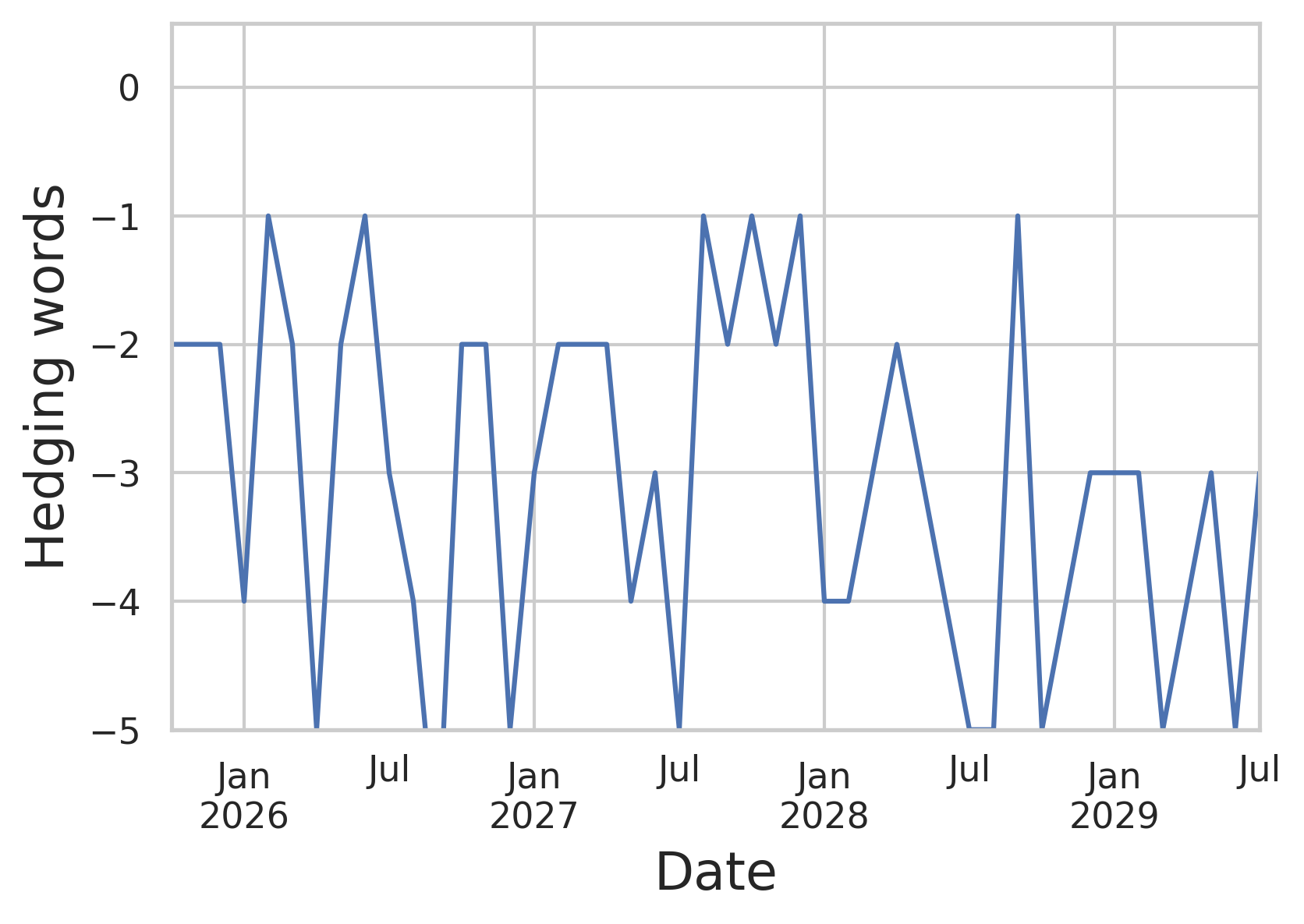}  
        \caption{GPT-4o}
        \label{fig:gpt-4o_ti}
    \end{subfigure}
    \begin{subfigure}{0.24\textwidth}  
        \centering
        \includegraphics[width=\linewidth]{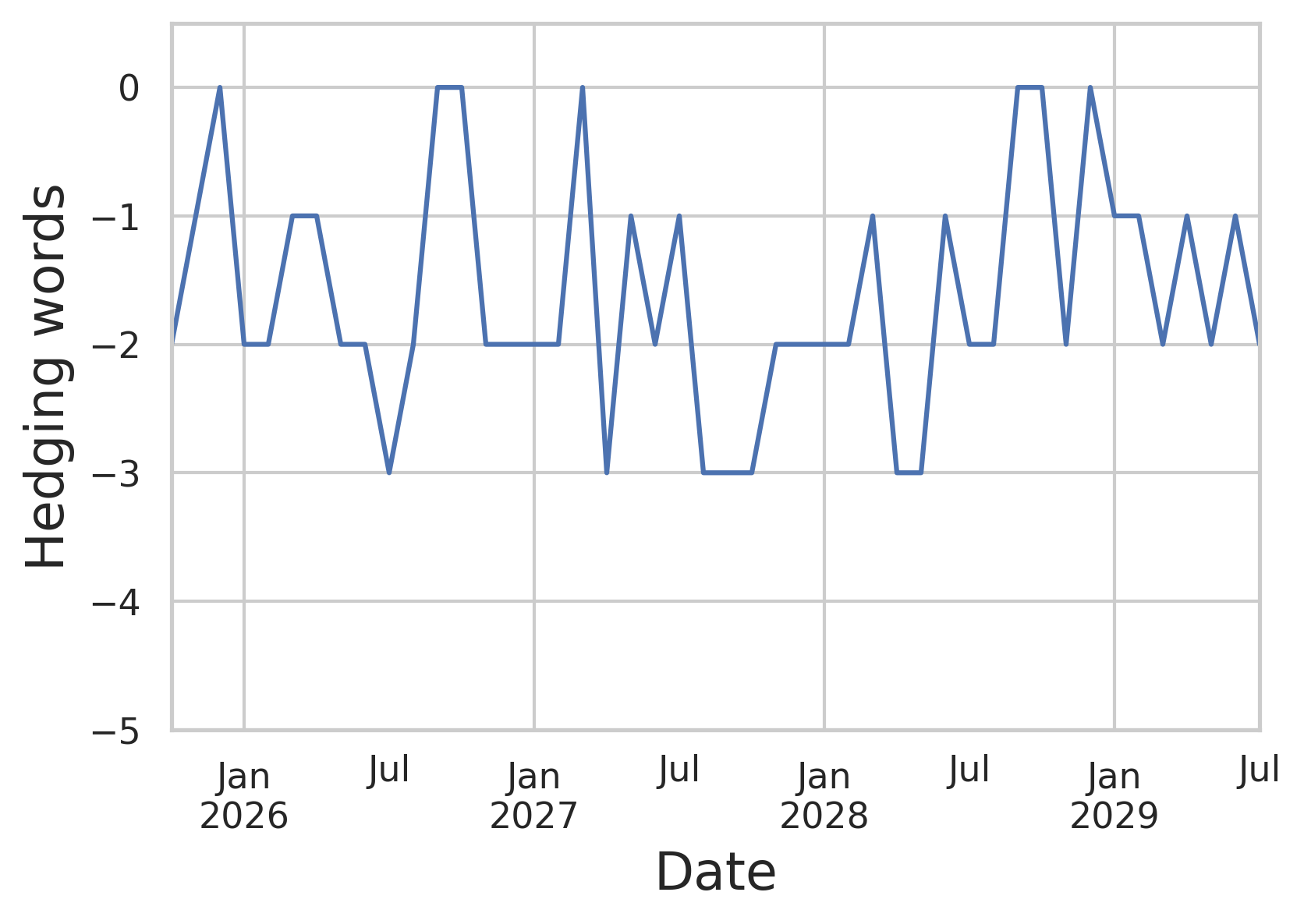}
        \caption{GPT-4o-mini}
    \end{subfigure}
    \begin{subfigure}{0.24\textwidth}  
        \centering
        \includegraphics[width=\linewidth]{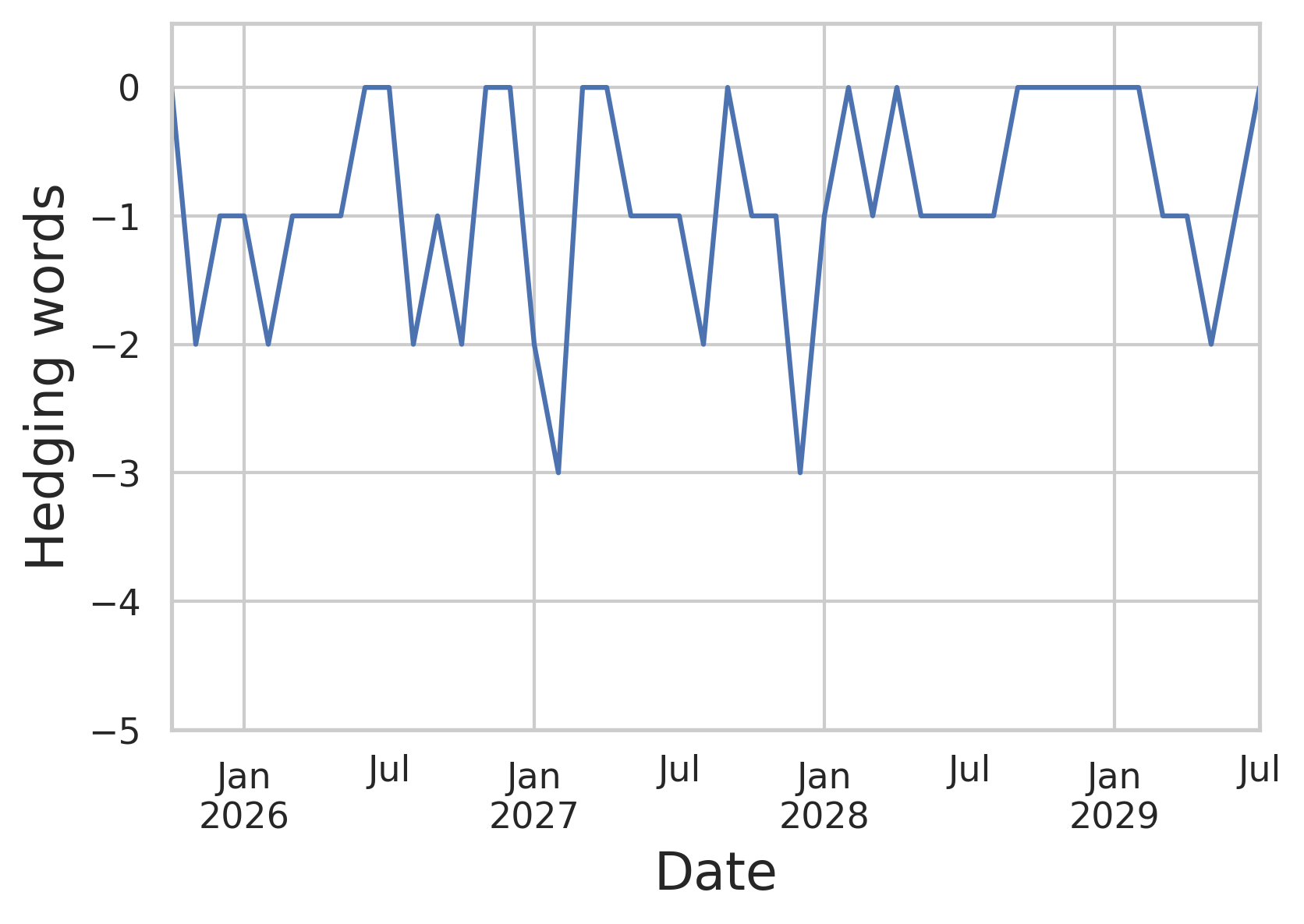}
        \caption{Gemini}
    \end{subfigure}
    \begin{subfigure}{0.24\textwidth}  
        \centering
        \includegraphics[width=\linewidth]{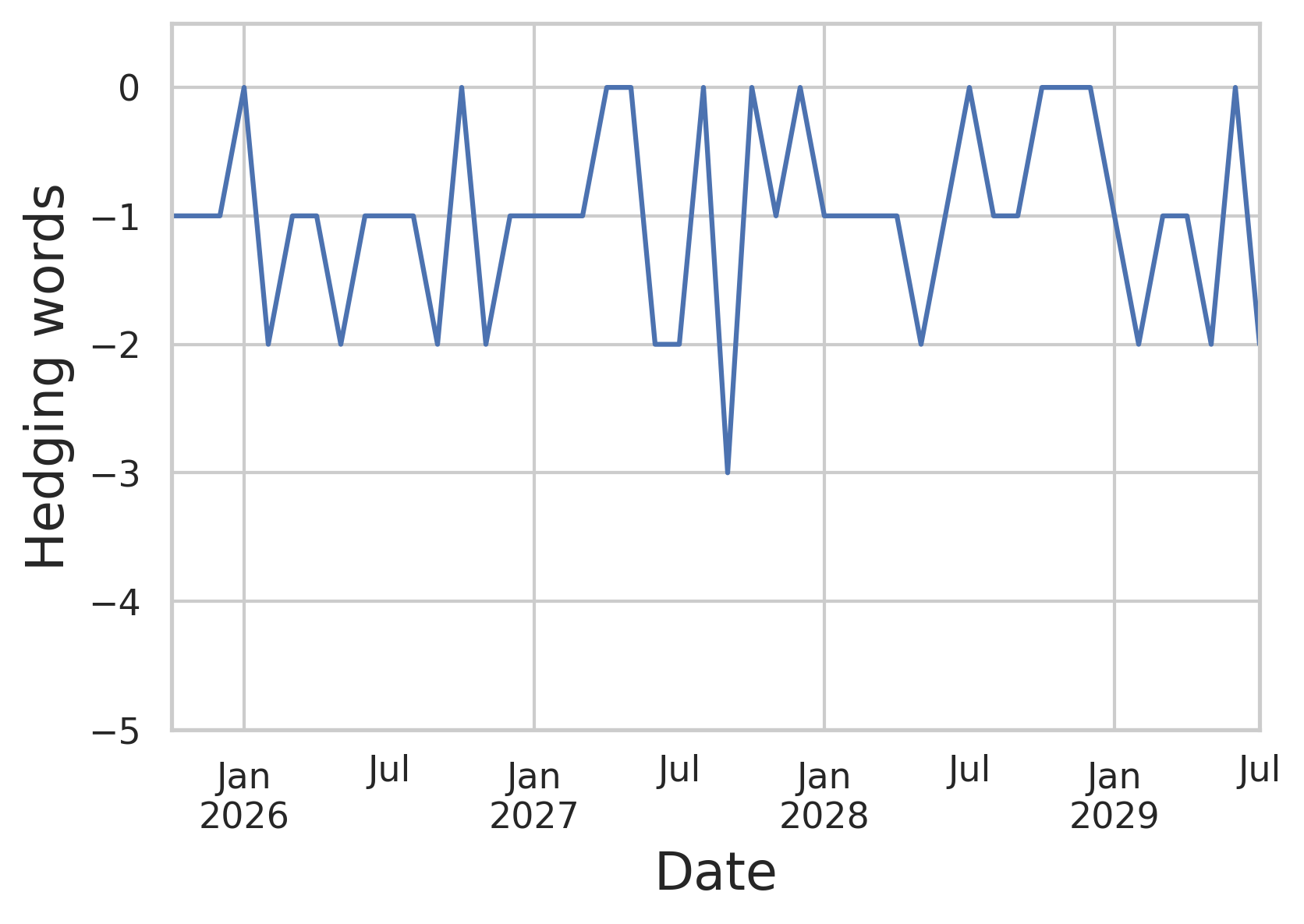}
        \caption{\llama}
    \end{subfigure}
    \caption{Evolution of hedging words over time for TI (plots) reports generated from from synthetic data for the period 2024-2029.}
    \label{fig:hedge_ti_plots_syn2}
\end{figure*}

\clearpage

\section{Linguistic Analysis}
\label{app:linguist}
Table \ref{apptab:linguist} shows the full table for linguistic analysis of all types of reports generated. 

\begin{table*}[h]
    \centering
\resizebox{0.99\textwidth}{!}{
\begin{tabular}{llcccccccc|ccccc}
\toprule
                           \textbf{Report type} &        \textbf{Model} &  \textbf{Rep. len} &  \textbf{Sent. len} &  \textbf{TTR (w)} &  \textbf{TTR (a)} &  \textbf{Polarity} &  \textbf{Subjectivity} &  \textbf{Terms} &  \textbf{Readability} &  \textbf{MA} &  \textbf{RSI} &  \textbf{MACD} &  \textbf{BB} &  \textbf{Retracemt.} \\
\midrule
                                 Short &       \gpt&        226.6 &          27.6 &         0.79 &         0.07 &      0.11 &          0.42 &   0.03 &         33.4 &               0.00 &        0.00 &         0.00 &       0.00 &              0.00 \\
                                  Short &  \gptm  &        204.7 &          25.8 &         0.80 &         0.06 &      0.11 &          0.42 &   0.03 &         35.8 &               0.01 &        0.00 &         0.00 &       0.00 &              0.00 \\
                                  Short &       \gemini &        150.1 &          24.7 &         0.82 &         0.07 &      0.13 &          0.47 &   0.04 &         38.4 &               0.01 &        0.00 &         0.00 &       0.00 &              0.00 \\
                                 Short &         Phi-3 &        199.0 &          23.7 &         0.53 &         0.02 &      0.24 &          0.52 &   0.01 &         62.5 &               0.00 &        0.00 &         0.00 &       0.00 &              0.00 \\
                                  Short &        \llama &        216.5 &          19.5 &         0.73 &         0.10 &      0.13 &          0.43 &   0.02 &         47.2 &               0.02 &        0.00 &         0.00 &       0.00 &              0.00 \\ \hline
                  TI &       \gpt&        354.2 &          27.2 &         0.73 &         0.06 &      0.09 &          0.41 &   0.05 &         28.7 &               0.87 &        1.00 &         1.00 &       0.57 &              0.00 \\
                   TI &  \gptm  &        328.7 &          27.0 &         0.76 &         0.06 &      0.07 &          0.42 &   0.06 &         26.6 &               0.99 &        1.00 &         0.96 &       0.20 &              0.00 \\
                   TI &       \gemini &        291.7 &          22.8 &         0.70 &         0.05 &      0.09 &          0.44 &   0.06 &         35.6 &               0.93 &        1.00 &         1.00 &       0.52 &              0.00 \\
                   TI &        \llama &        452.0 &          11.3 &         0.60 &         0.07 &      0.10 &          0.42 &   0.04 &         42.0 &               0.91 &        0.94 &         0.87 &       0.21 &              0.00 \\ \hline
             TI (plots) &       \gpt&        334.1 &          27.8 &         0.75 &         0.06 &      0.09 &          0.41 &   0.04 &         28.8 &               1.00 &        1.00 &         0.02 &       0.00 &              0.00 \\
             TI (plots) &  \gptm  &        320.4 &          26.3 &         0.76 &         0.06 &      0.08 &          0.41 &   0.04 &         29.2 &               1.00 &        1.00 &         0.01 &       0.00 &              0.00 \\
            TI (plots) &       \gemini &        275.7 &          23.5 &         0.71 &         0.05 &      0.10 &          0.44 &   0.07 &         39.6 &               0.94 &        1.00 &         0.94 &       0.84 &              0.21 \\
             TI (plots) &        \llama &        293.4 &          14.8 &         0.71 &         0.12 &      0.08 &          0.41 &   0.04 &         45.2 &               0.52 &        0.61 &         0.45 &       0.58 &              0.28 \\ \hline \hline
                            (S) Short &       \gpt &        232.5 &          27.4 &         0.77 &         0.10 &      0.11 &          0.41 &   0.02 &         38.3 &               0.00 &        0.00 &         0.00 &       0.00 &              0.00 \\
                            (S) Short &  \gptm  &        217.5 &          23.8 &         0.78 &         0.10 &      0.10 &          0.38 &   0.03 &         41.0 &               0.00 &        0.00 &         0.00 &       0.00 &              0.00 \\
                            (S) Short &       \gemini &        158.3 &          24.5 &         0.80 &         0.10 &      0.13 &          0.48 &   0.03 &         39.8 &               0.01 &        0.00 &         0.00 &       0.00 &              0.00 \\
                            (S) Short &         Phi-3 &        206.8 &          24.0 &         0.48 &         0.03 &      0.24 &          0.55 &   0.01 &         64.2 &               0.00 &        0.00 &         0.00 &       0.00 &              0.00 \\
                            (S) Short &        \llama &        218.2 &          17.4 &         0.71 &         0.13 &      0.13 &          0.40 &   0.02 &         48.0 &               0.02 &        0.01 &         0.00 &       0.01 &              0.00 \\ \hline
            (S) TI &       \gpt &        354.3 &          28.2 &         0.73 &         0.09 &      0.09 &          0.40 &   0.05 &         32.3 &               0.89 &        1.00 &         1.00 &       0.54 &              0.00 \\
             (S) TI &  \gptm  &        334.6 &          25.5 &         0.74 &         0.09 &      0.08 &          0.40 &   0.06 &         30.9 &               0.98 &        1.00 &         0.95 &       0.33 &              0.00 \\
            (S) TI &       \gemini &        302.0 &          22.6 &         0.69 &         0.08 &      0.08 &          0.42 &   0.06 &         35.8 &               0.80 &        1.00 &         1.00 &       0.58 &              0.00 \\
             (S) TI &        \llama &        493.8 &          10.1 &         0.57 &         0.10 &      0.10 &          0.41 &   0.03 &         41.2 &               0.91 &        0.95 &         0.84 &       0.14 &              0.00 \\ \hline
  (S) TI (plots) &       \gpt &        348.5 &          27.9 &         0.74 &         0.10 &      0.09 &          0.42 &   0.03 &         33.3 &               1.00 &        1.00 &         0.02 &       0.00 &              0.00 \\
 (S) TI (plots) &  \gptm  &        326.2 &          24.5 &         0.74 &         0.09 &      0.08 &          0.39 &   0.03 &         33.9 &               0.99 &        1.00 &         0.01 &       0.00 &              0.00 \\
  (S) TI (plots) &       \gemini &        287.1 &          23.8 &         0.68 &         0.07 &      0.10 &          0.44 &   0.06 &         38.5 &               0.77 &        1.00 &         0.87 &       0.79 &              0.17 \\
  (S) TI (plots) &        \llama &        281.4 &          13.9 &         0.70 &         0.16 &      0.08 &          0.39 &   0.03 &         43.3 &               0.55 &        0.67 &         0.38 &       0.39 &              0.43 \\
\bottomrule
\end{tabular}
   }
   \caption{Linguistic analysis of all report types: short, technical indicator (TI), technical indicator reports generated using time series and plots (TI (plots)), real and synthetic (S). Table presents average report lengths (Rep. len), sentence lengths (Sent. len), Type-Token Ratio within each report (TTR (w)) and across reports (TTR (a)), sentiment polarity, report subjectivity, proportion of financial terms in the report, Flesch reading ease score (Readability), proportion of reports mentioning each of the technical indicators (MA, RSI, MACD, BB and Retracement).}
   \label{apptab:linguist}
\end{table*}

\clearpage

\section{Segments Analysis}
\label{app:seganalysis}
Tables \ref{apptab:segmanalysis} and \ref{apptab:segmanalysis-synth} present the average length and the proportion of the presence of each type of highlight (DR, FI, EK) across different types of reports (Short, TI, and TI with plots) for each model.

\begin{table}[h!]
    \centering
\resizebox{\columnwidth}{!}{
\begin{tabular}{llcccccc}
\toprule
\textbf{Report type} & \textbf{Model} & \textbf{DR (avg len)} & \textbf{FI (avg len)} & \textbf{EK (avg len)} & \textbf{DR (prop)} & \textbf{FI (prop)} & \textbf{EK (prop)} \\ 
\midrule
\multirow{4}{*}{Short} 
 & \gpt& 12.34 & 11.60 & 8.28 & 0.44 & 0.46 & 0.11 \\
 & \gptm  & 12.55 & 11.96 & 7.62 & 0.44 & 0.47 & 0.10 \\
 & \gemini & 11.46 & 11.40 & 9.98 & 0.47 & 0.46 & 0.07 \\
 & Phi3 & 13.89 & 9.53 & 6.78 & 0.70 & 0.29 & 0.00 \\
 & \llama & 12.93 & 12.66 & 10.23 & 0.50 & 0.43 & 0.08 \\
 \midrule
\multirow{3}{*}{TI}          
 & \gpt & 10.97 & 11.23 & 7.42 & 0.44 & 0.46 & 0.10 \\
 & \gptm  & 11.72 & 13.12 & 8.78 & 0.39 & 0.51 & 0.10 \\
 & \gemini & 11.18 & 11.58 & 9.40 & 0.39 & 0.53 & 0.08 \\
 & \llama & 13.12 & 13.10 & 11.11 & 0.40 & 0.49 & 0.11 \\
 \midrule
\multirow{3}{*}{TI with plots} 
 & \gpt& 12.34 & 14.10 & 9.31 & 0.37 & 0.52 & 0.11 \\
 & \gptm  & 12.71 & 14.10 & 9.58 & 0.37 & 0.51 & 0.12 \\
 & \gemini & 11.08 & 11.61 & 8.92 & 0.39 & 0.53 & 0.08 \\
 & \llama & 13.75 & 14.95 & 12.83 & 0.43 & 0.46 & 0.11 \\
\bottomrule
\end{tabular}
}
\caption{Segment analysis of real indexes}
\label{apptab:segmanalysis}
\end{table}

\begin{table}[h!]
    \centering
\resizebox{\columnwidth}{!}{
\begin{tabular}{llcccccc}
\toprule
\textbf{Report type} & \textbf{Model} & \textbf{DR (avg len)} & \textbf{FI (avg len)} & \textbf{EK (avg len)} & \textbf{DR (prop)} & \textbf{FI (prop)} & \textbf{EK (prop)} \\ 
\midrule
\multirow{3}{*}{Short} 
 & \gpt& 13.15 & 11.59 & 7.05 & 0.48 & 0.47 & 0.05 \\
 & \gptm  & 13.33 & 11.16 & 7.67 & 0.47 & 0.47 & 0.06 \\
 & \gemini & 12.56 & 12.02 & 15.53 & 0.48 & 0.47 & 0.04 \\
 & Phi3   & 16.39 & 9.46  &  0    &  0.76 &     0.24 & 0.00 \\
 & \llama  & 13.29 & 12.51 & 10.29 & 0.51 & 0.44 & 0.05 \\
 \midrule
\multirow{3}{*}{TI}          
 & \gpt& 11.26 & 11.48 & 6.93 & 0.45 & 0.47 & 0.07 \\
 & \gptm  & 12.28 & 13.46 & 8.72 & 0.40 & 0.52 & 0.08 \\
 & \gemini & 12.64 & 12.24 & 7.86 & 0.40 & 0.55 & 0.05 \\
 & \llama &  14.05 &       13.57 &       10.23 &     0.43 &     0.48 &     0.08 \\
 \midrule
\multirow{3}{*}{TI with plots} 
 & \gpt& 12.18 & 13.49 & 9.08 & 0.40 & 0.51 & 0.09 \\
 & \gptm  & 13.54 & 13.97 & 10.78 & 0.40 & 0.51 & 0.09 \\
 & \gemini & 12.02 & 11.74 & 9.85 & 0.41 & 0.53 & 0.05 \\
 & \llama &       13.74 &       14.81 &       12.77 &     0.43 &     0.47 &     0.10 \\
\bottomrule
\end{tabular}
}
\caption{Segment analysis of synthetic indexes}
\label{apptab:segmanalysis-synth}
\end{table}

\subsection{Proportion of segment identification over time}

\begin{figure*}[h!]
    \centering
    \begin{subfigure}{0.32\textwidth}  
        \centering
        \includegraphics[width=\linewidth]{span_class_smooth_gpt-4o_short.png}  
        \caption{GPT-4o}
    \end{subfigure}
    \begin{subfigure}{0.32\textwidth}  
        \centering
        \includegraphics[width=\linewidth]{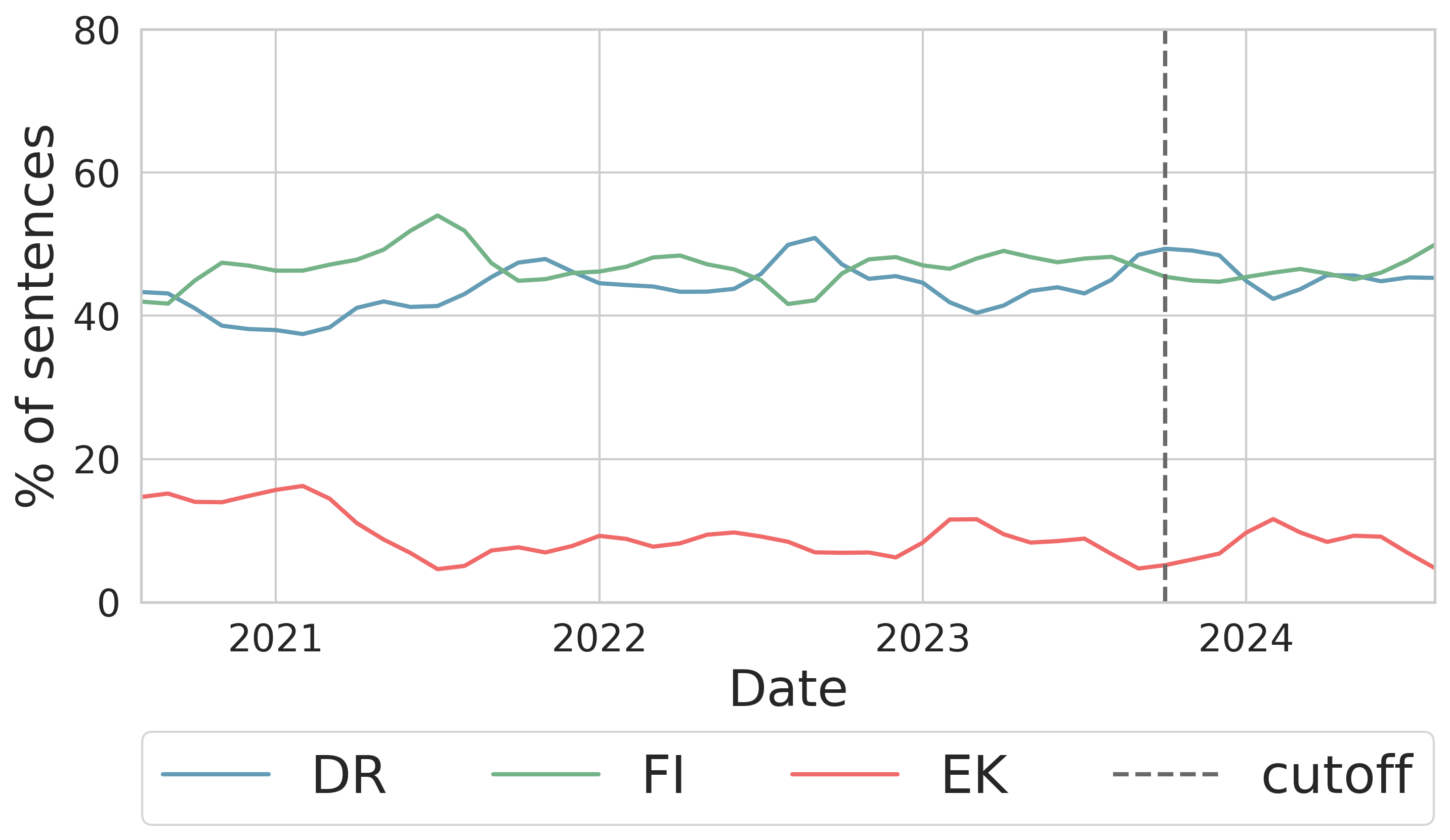}
        \caption{GPT-4o-mini}
    \end{subfigure}
    \begin{subfigure}{0.32\textwidth}  
        \centering
        \includegraphics[width=\linewidth]{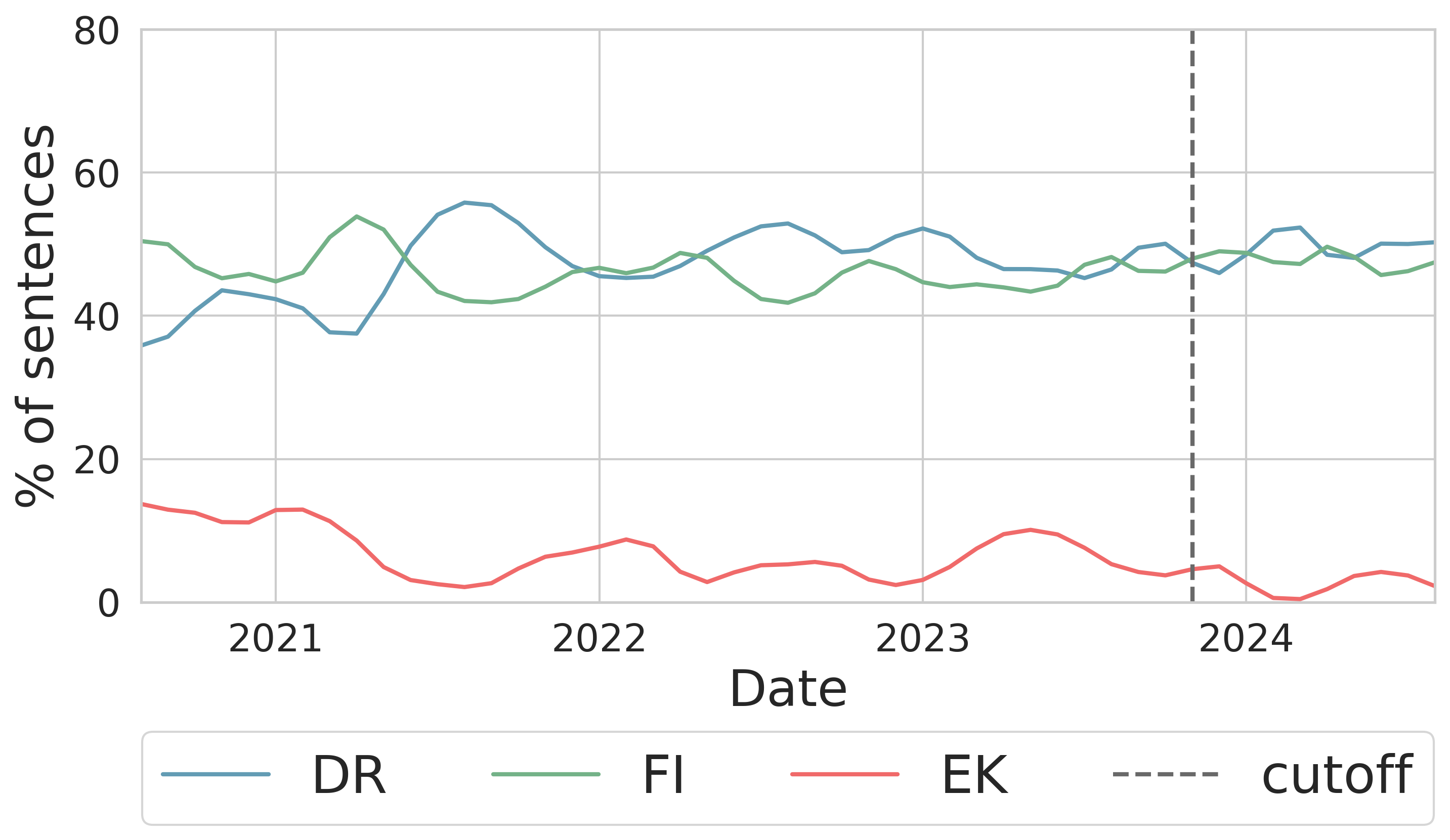}
        \caption{Gemini}
    \end{subfigure}\\
    \begin{subfigure}{0.32\textwidth}  
        \centering
        \includegraphics[width=\linewidth]{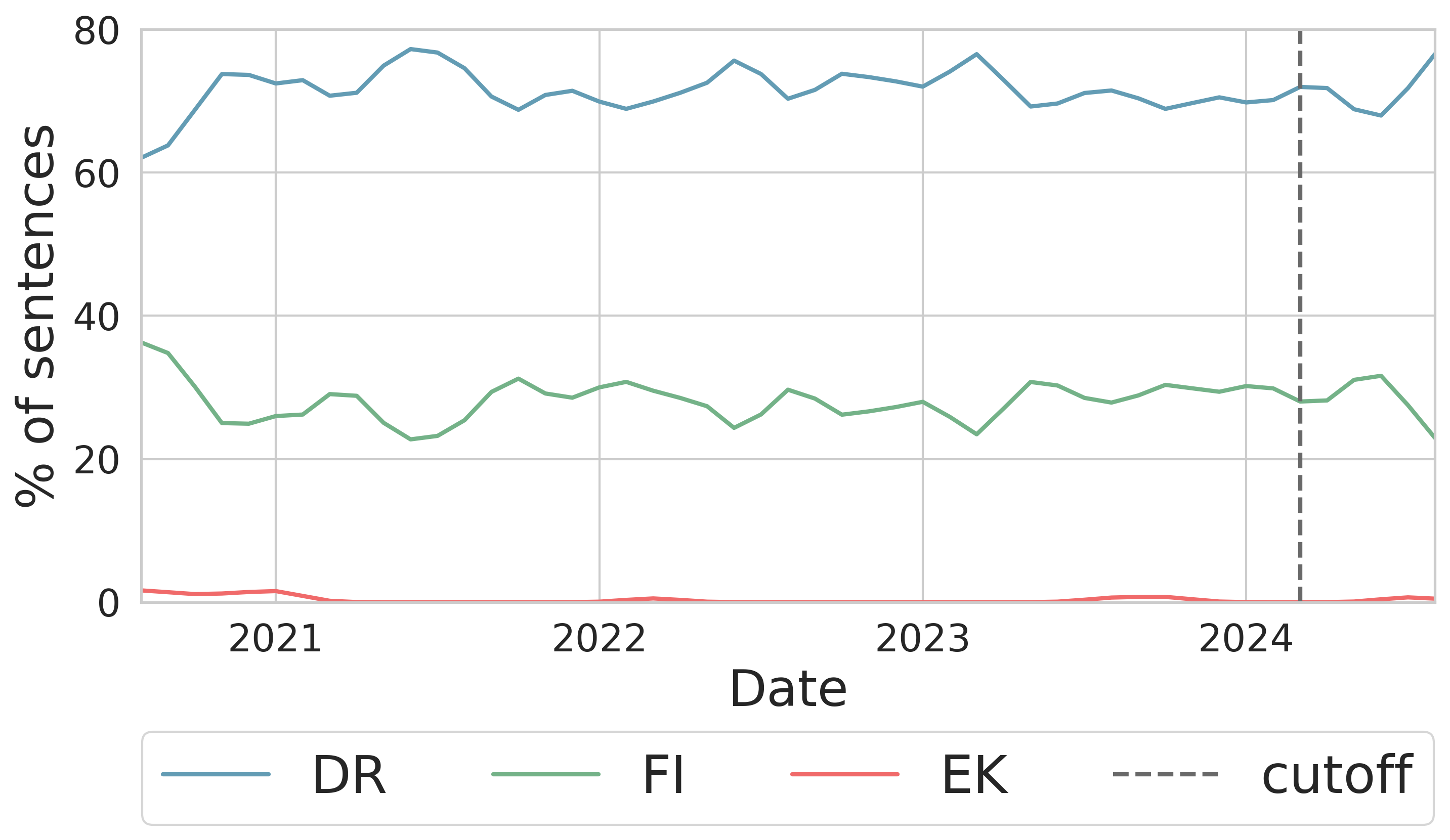}
        \caption{Phi-3}
    \end{subfigure}
    \begin{subfigure}{0.32\textwidth}  
        \centering
        \includegraphics[width=\linewidth]{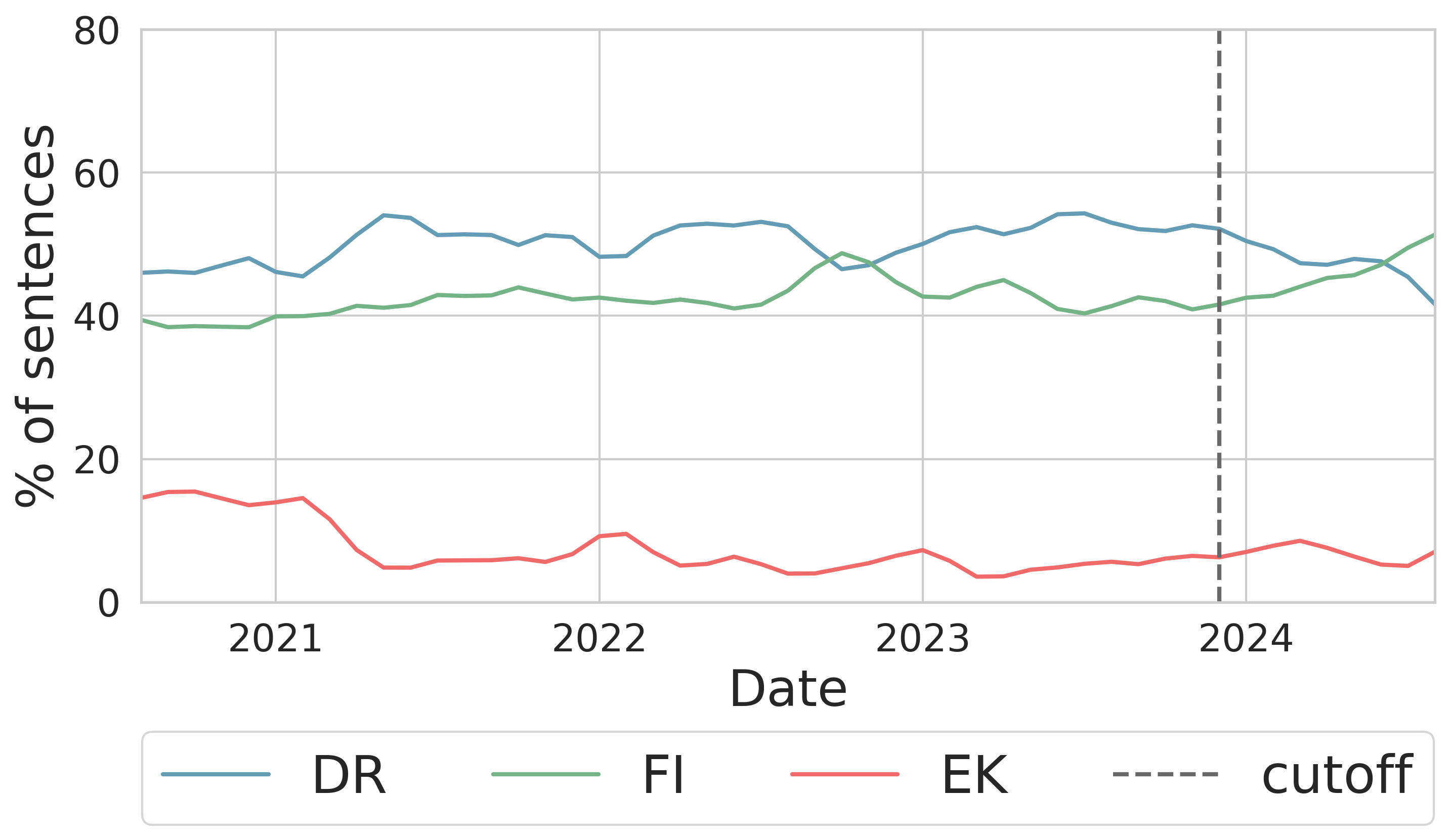}
        \caption{\llama}
    \end{subfigure}
    \caption{Short Reports highlighting segments over time for different models.}
    \label{fig:span-time-plot-short-real}
\end{figure*}

\begin{figure*}[h!]
    \centering
    \begin{subfigure}{0.24\textwidth}  
        \centering
        \includegraphics[width=\linewidth]{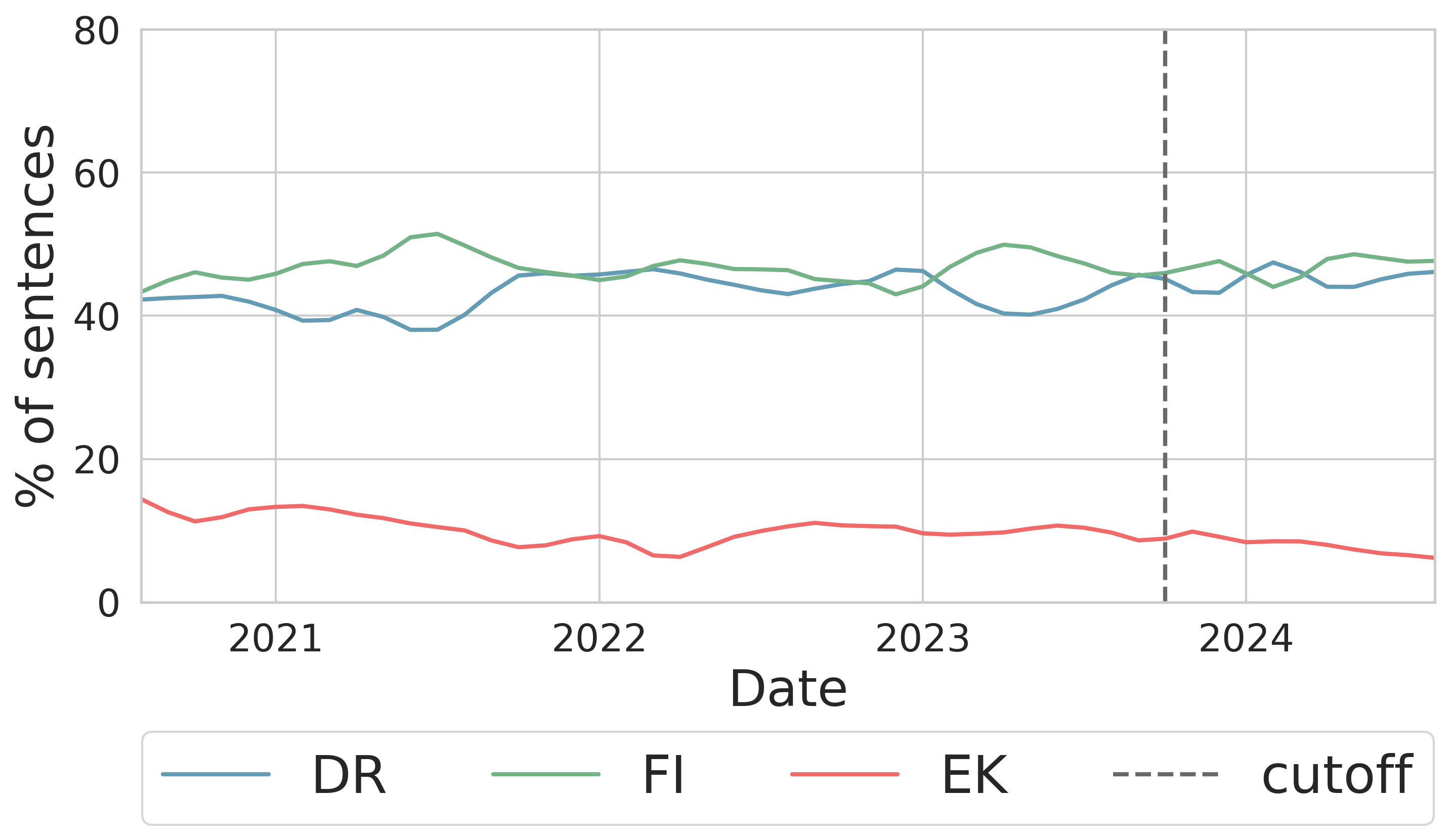}  
        \caption{GPT-4o}
    \end{subfigure}
    \begin{subfigure}{0.24\textwidth}  
        \centering
        \includegraphics[width=\linewidth]{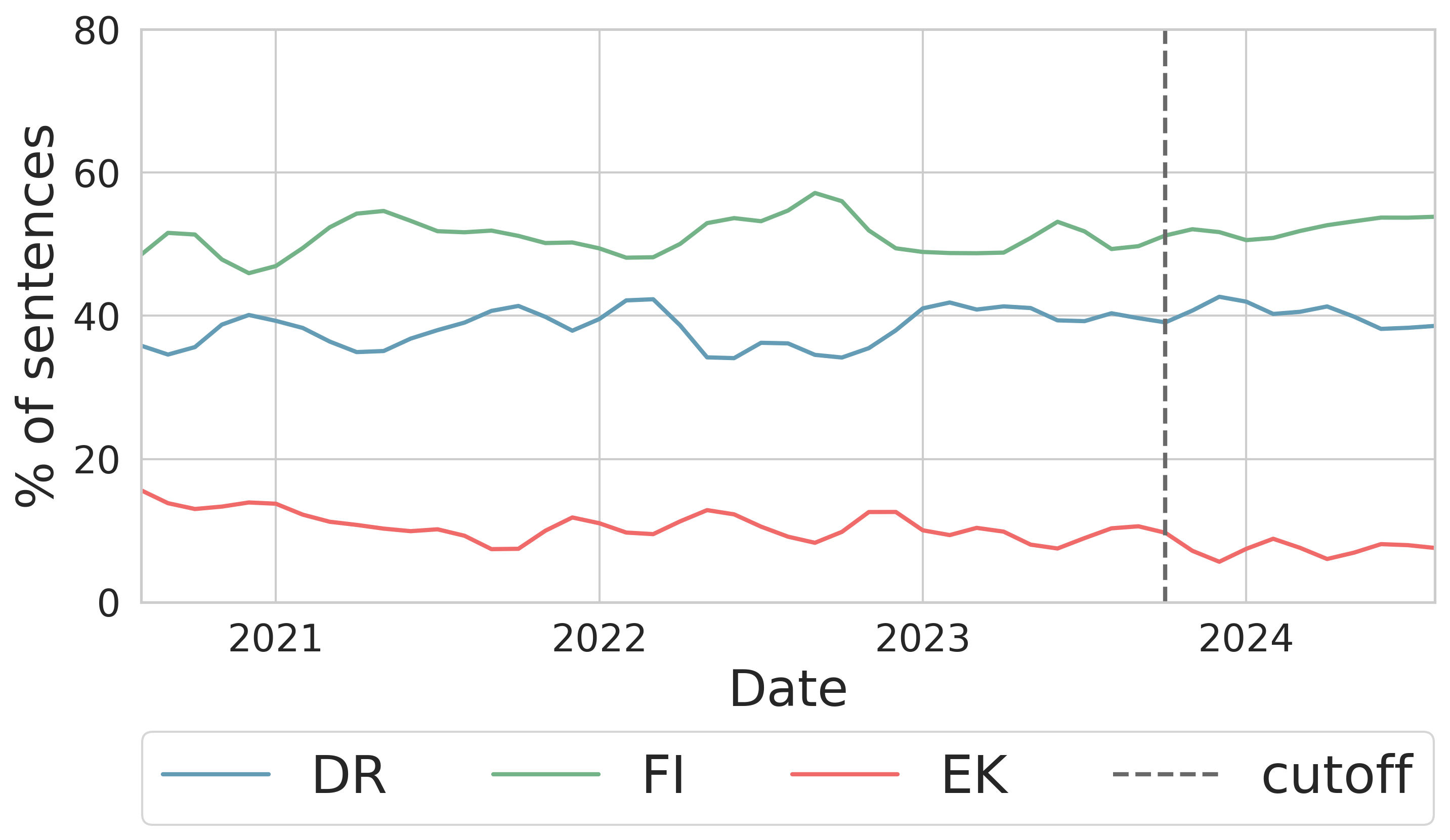}
        \caption{GPT-4o-mini}
    \end{subfigure}
    \begin{subfigure}{0.24\textwidth}  
        \centering
        \includegraphics[width=\linewidth]{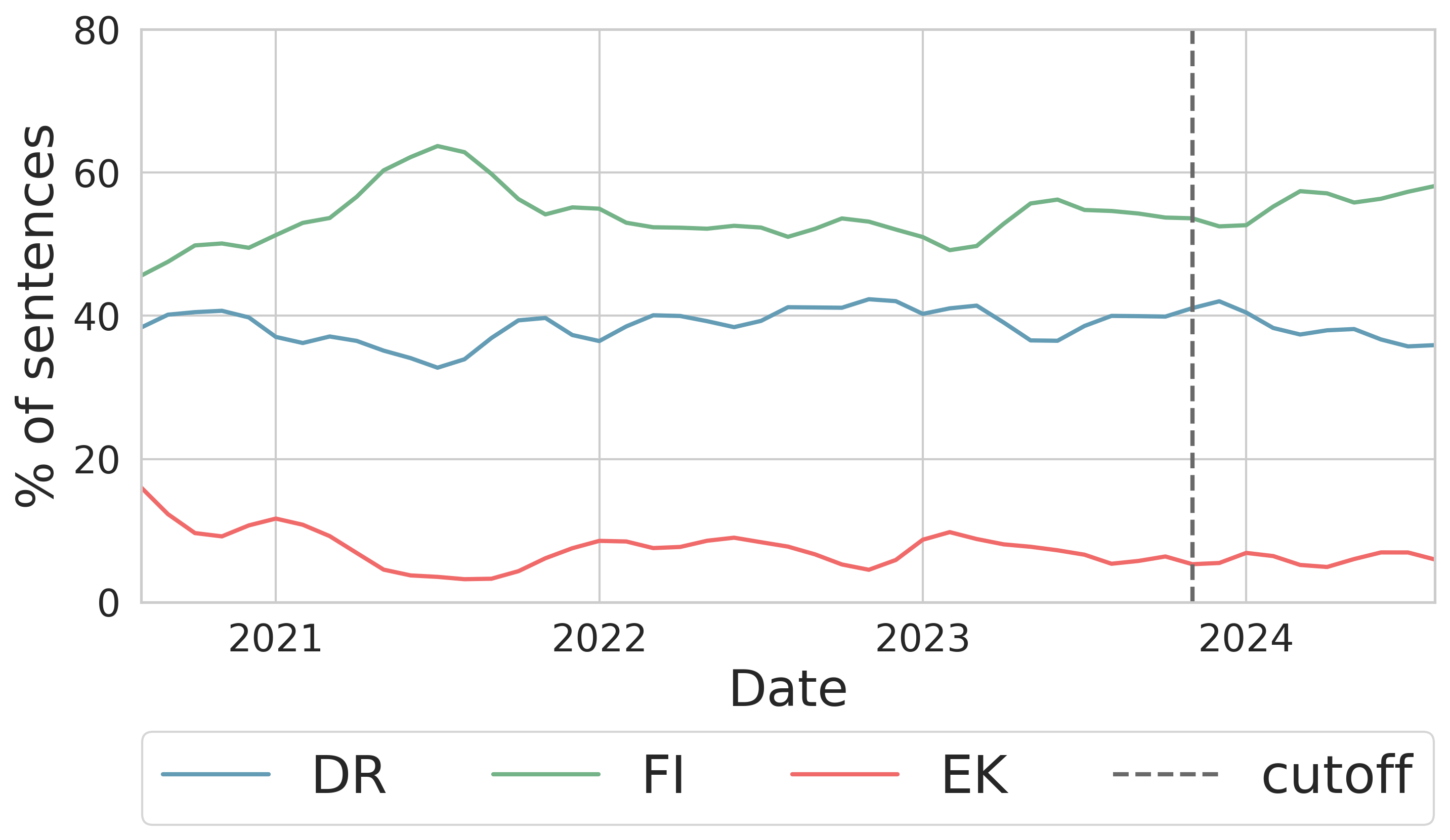}
        \caption{Gemini}
    \end{subfigure}
    \begin{subfigure}{0.24\textwidth}  
        \centering
        \includegraphics[width=\linewidth]{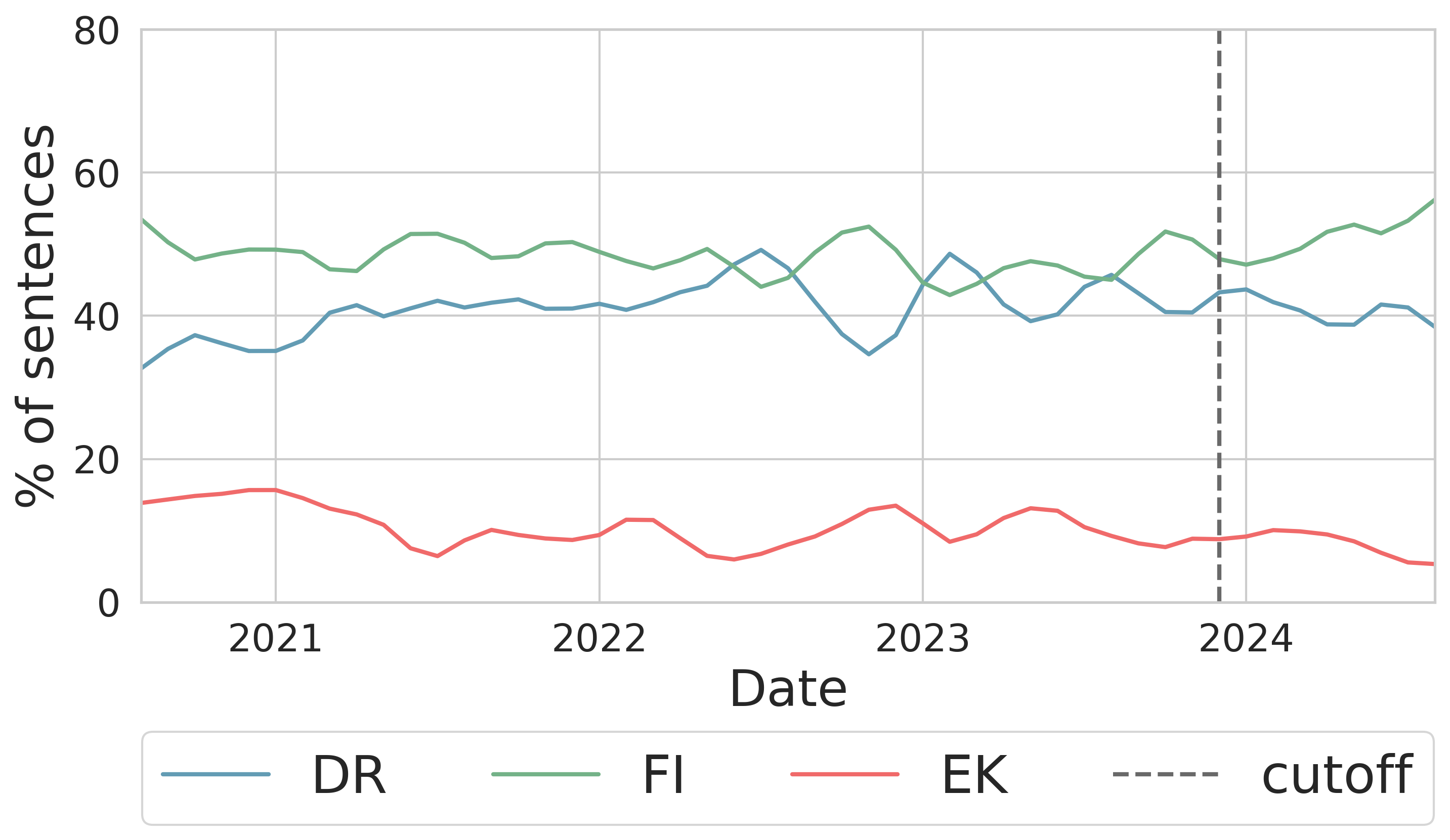}
        \caption{\llama}
    \end{subfigure}
    \caption{Technical Indicator Reports highlighting segments over time for different models.}
    \label{fig:span-time-plot-ti-real}
\end{figure*}

\begin{figure*}[h!]
    \centering
    \begin{subfigure}{0.24\textwidth}  
        \centering
        \includegraphics[width=\linewidth]{span_class_smooth_gpt-4o_technical_indicators_with_plots.png}  
        \caption{GPT-4o}
        \label{fig:gpt-4o_ti_plots}
    \end{subfigure}
    \begin{subfigure}{0.24\textwidth}  
        \centering
        \includegraphics[width=\linewidth]{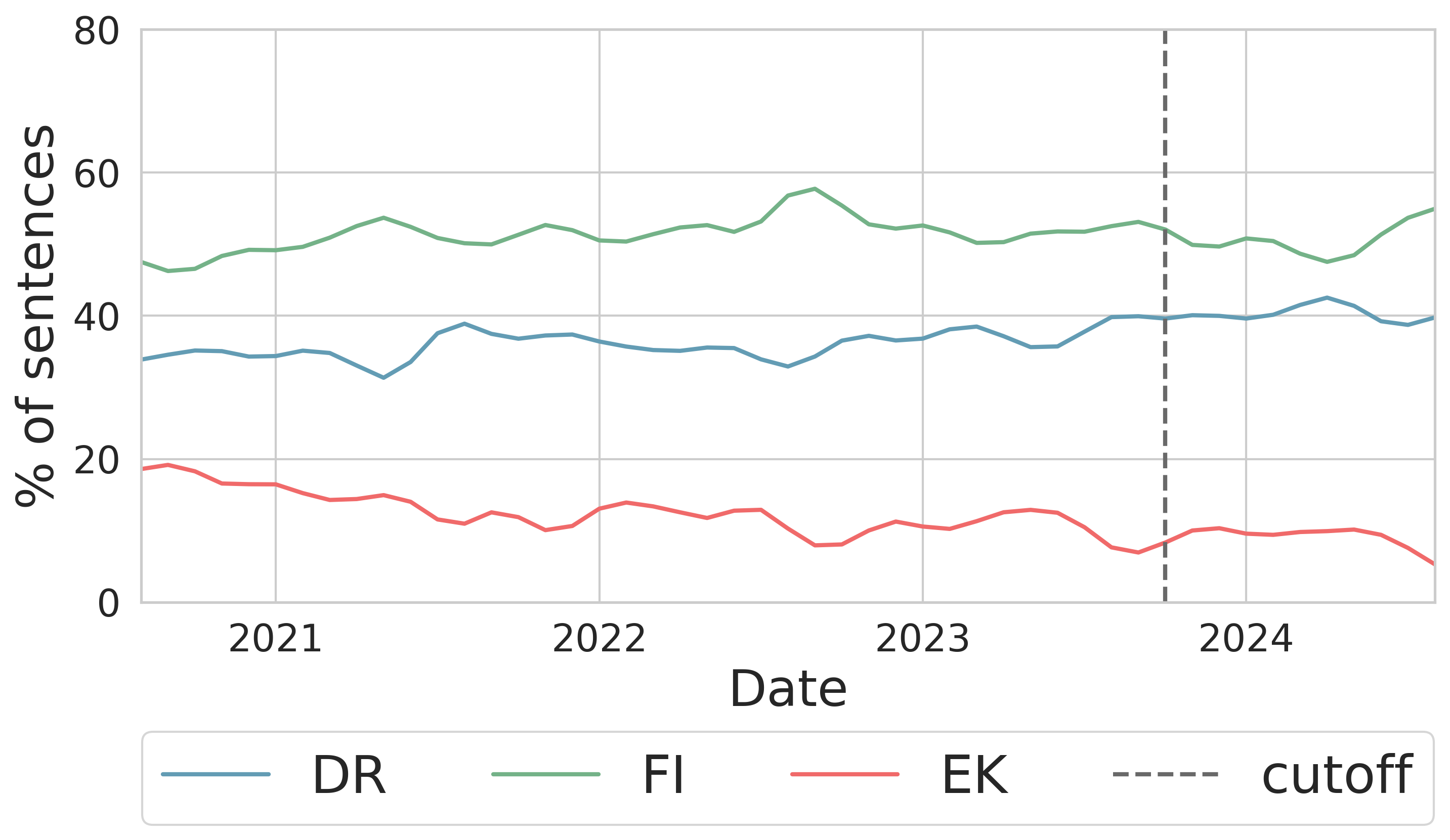}
        \caption{GPT-4o-mini}
        \label{fig:gpt-4o-mini_ti_plots}
    \end{subfigure}
    \begin{subfigure}{0.24\textwidth}  
        \centering
        \includegraphics[width=\linewidth]{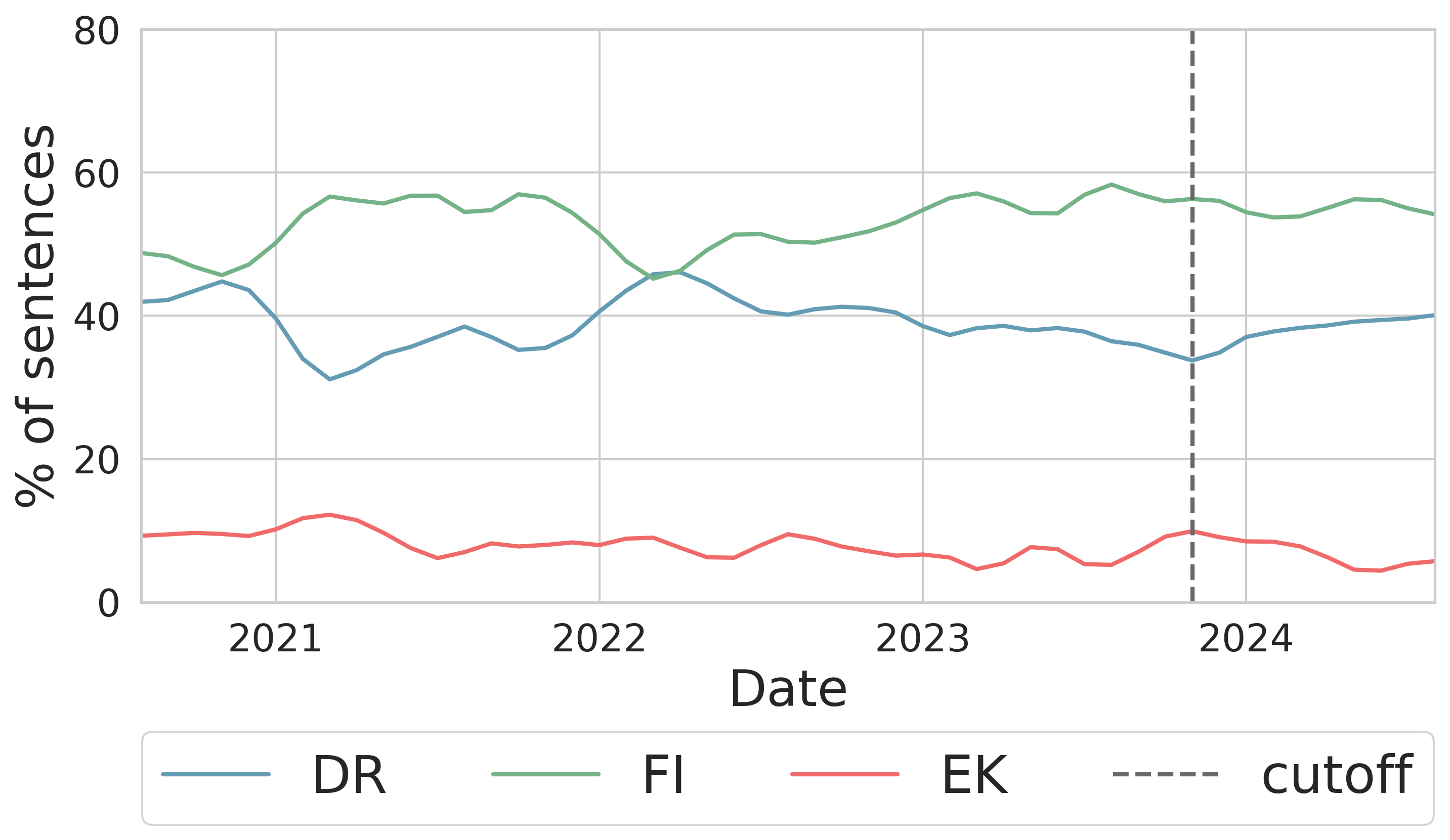}
        \caption{Gemini}
    \end{subfigure}
    \begin{subfigure}{0.24\textwidth}  
        \centering
        \includegraphics[width=\linewidth]{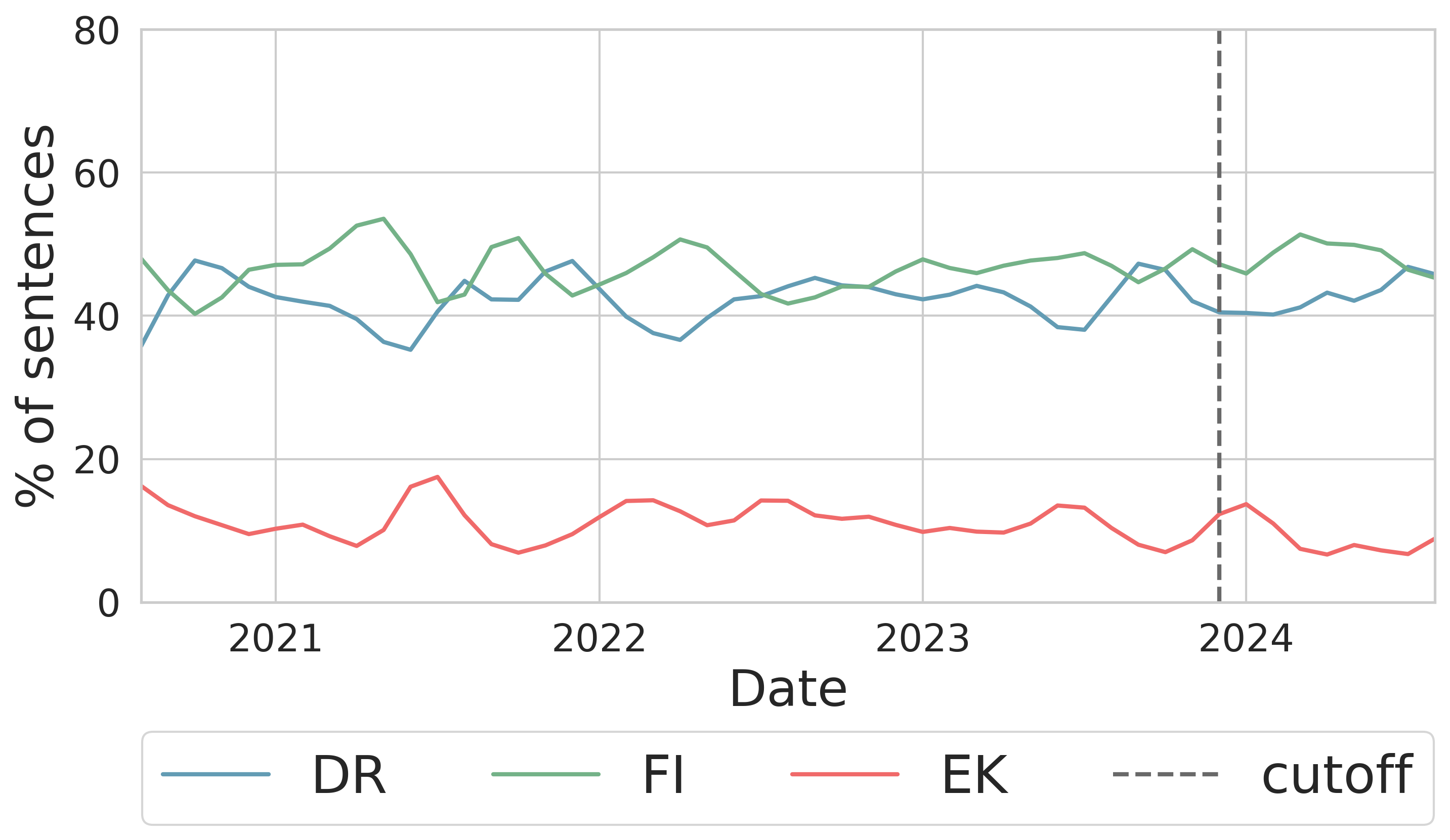}
        \caption{Gemini}
    \end{subfigure}
    \caption{Technical Indicator Reports (with plots) highlighting segments over time for different models.}
    \label{fig:span-time-plot-ti-plot-real}
\end{figure*}

Following the analysis shown in Section sec:span-over-time, Figures \ref{fig:span-time-plot-short-real}, \ref{fig:span-time-plot-ti-real} and \ref{fig:span-time-plot-ti-plot-real} show the evolution of segment identification highlight over time for all models and types of reports.

\begin{figure*}[h!]
    \centering
    \begin{subfigure}{0.32\textwidth}  
        \centering
        \includegraphics[width=\linewidth]{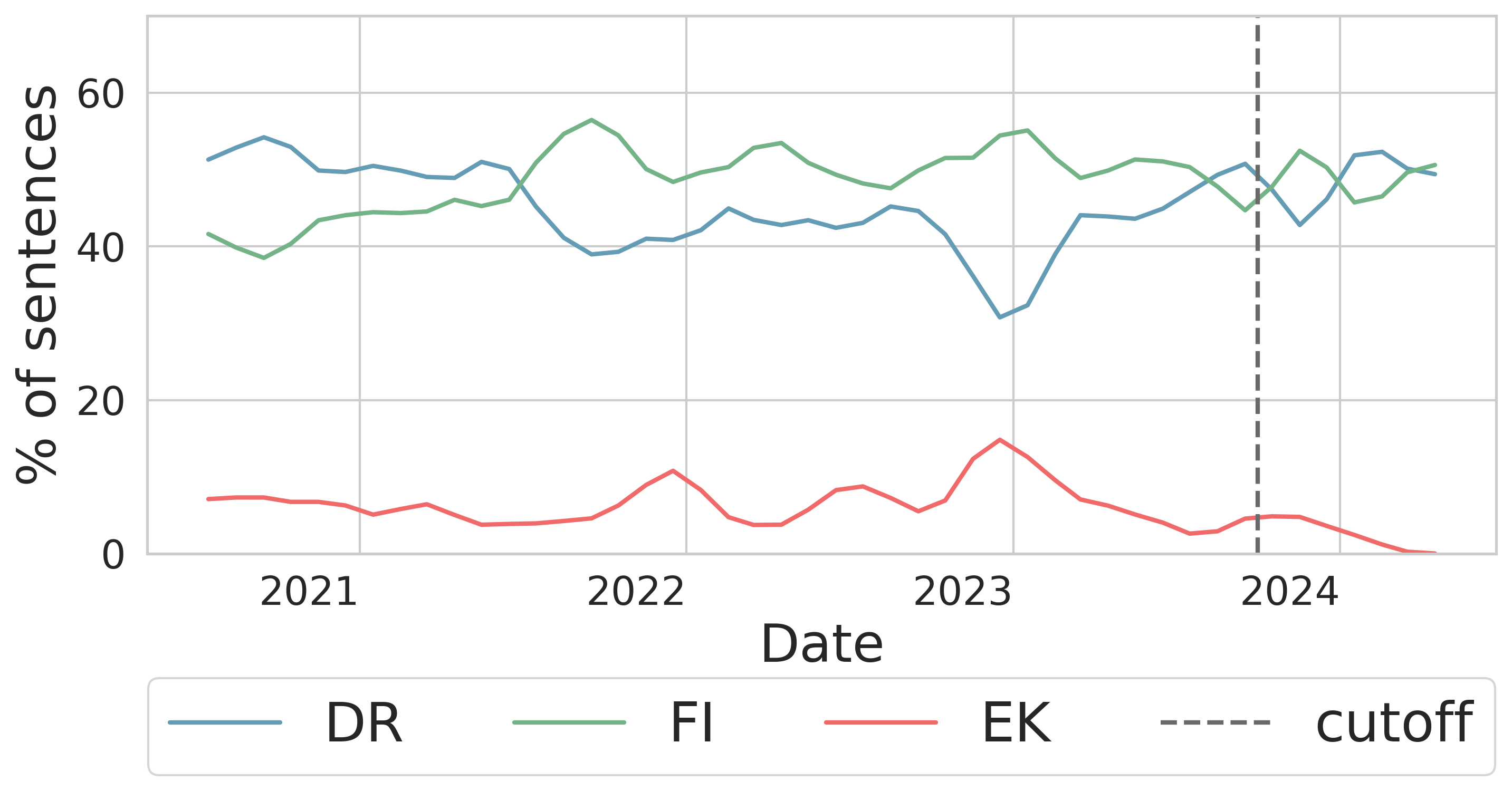}  
        \caption{GPT-4o}
    \end{subfigure}
    \begin{subfigure}{0.32\textwidth}  
        \centering
        \includegraphics[width=\linewidth]{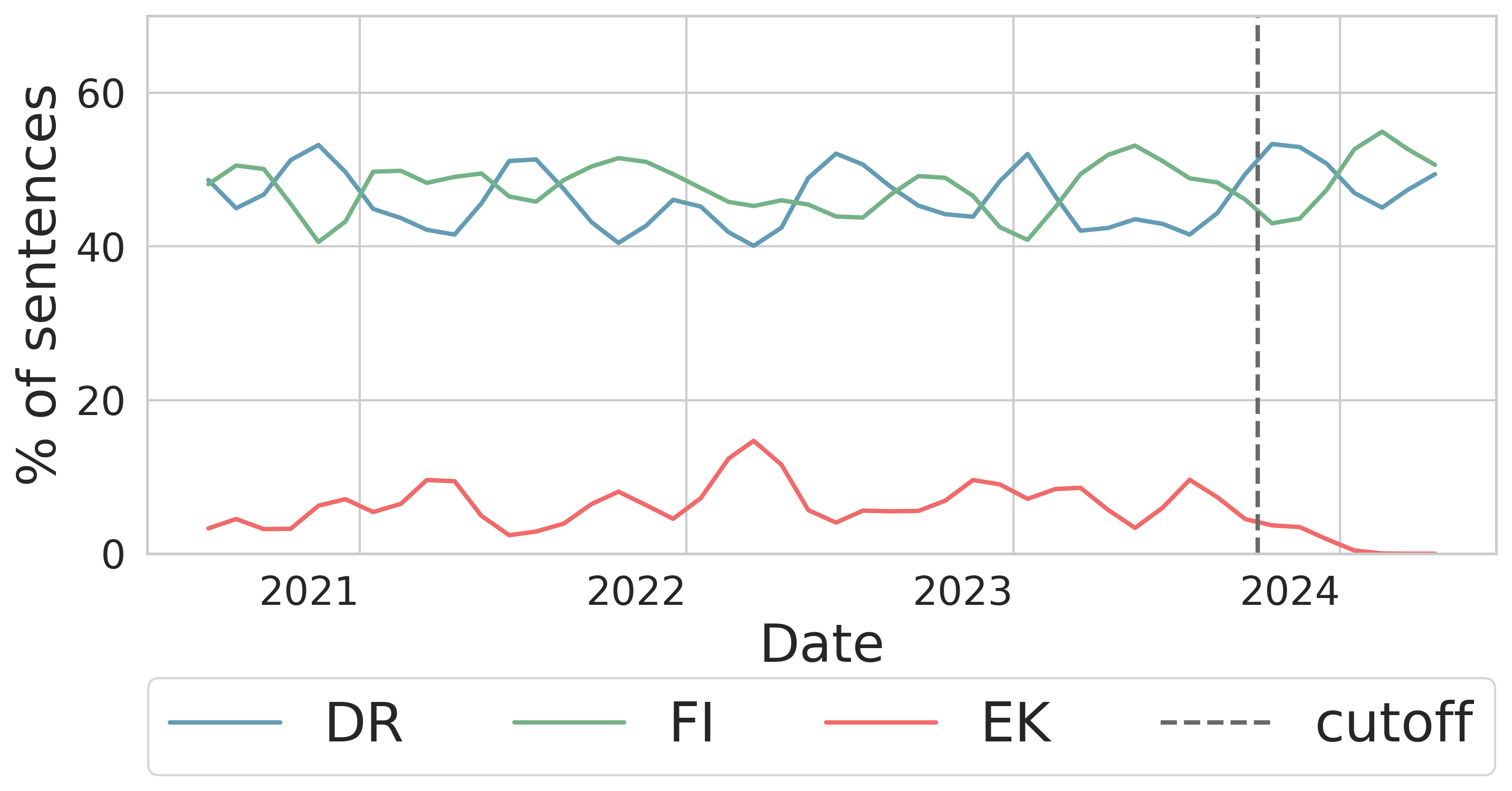}
        \caption{GPT-4o-mini}
    \end{subfigure}
    \begin{subfigure}{0.32\textwidth}  
        \centering
        \includegraphics[width=\linewidth]{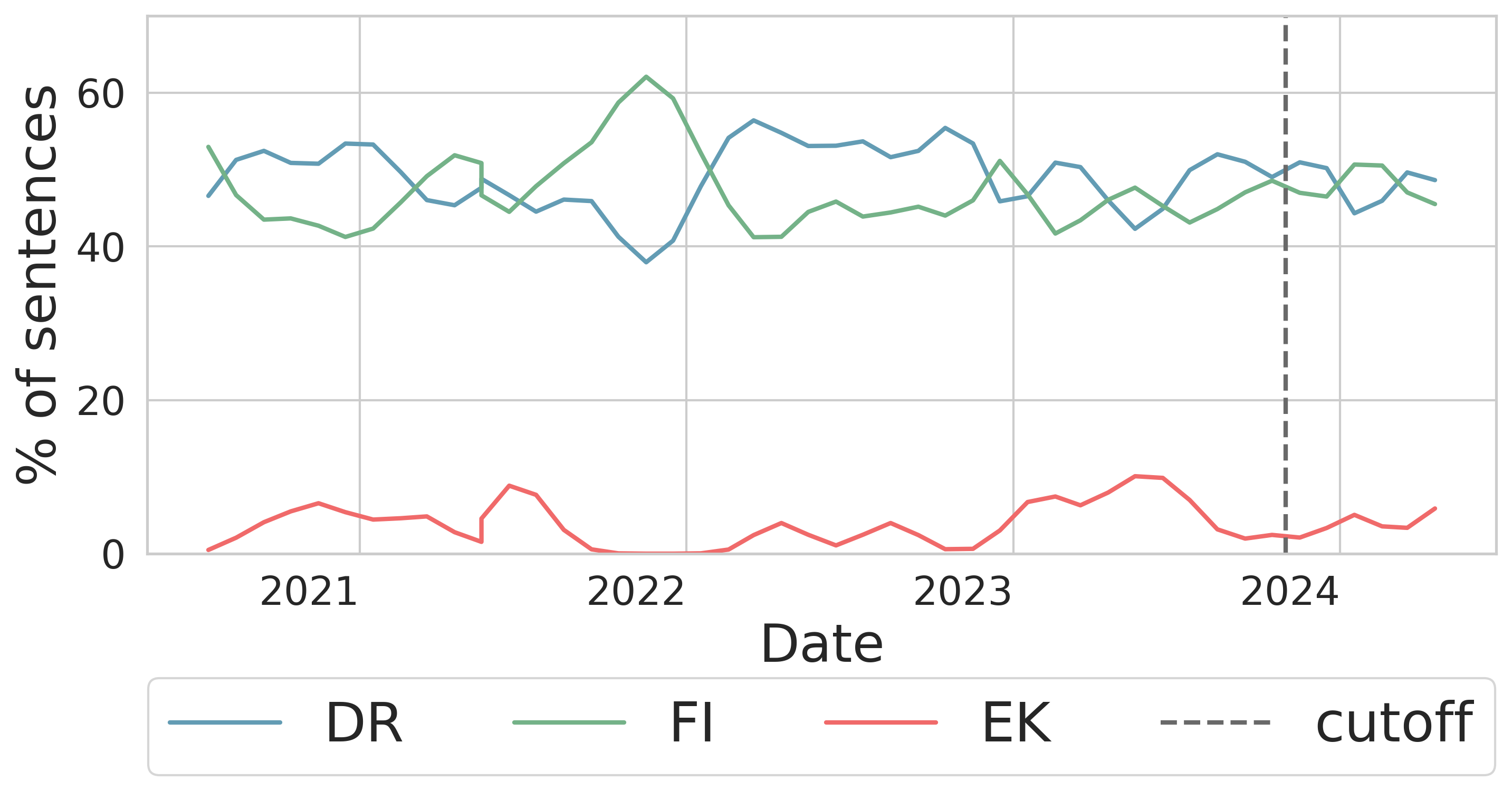}
        \caption{Gemini}
    \end{subfigure}\\
    \begin{subfigure}{0.32\textwidth}  
        \centering
        \includegraphics[width=\linewidth]{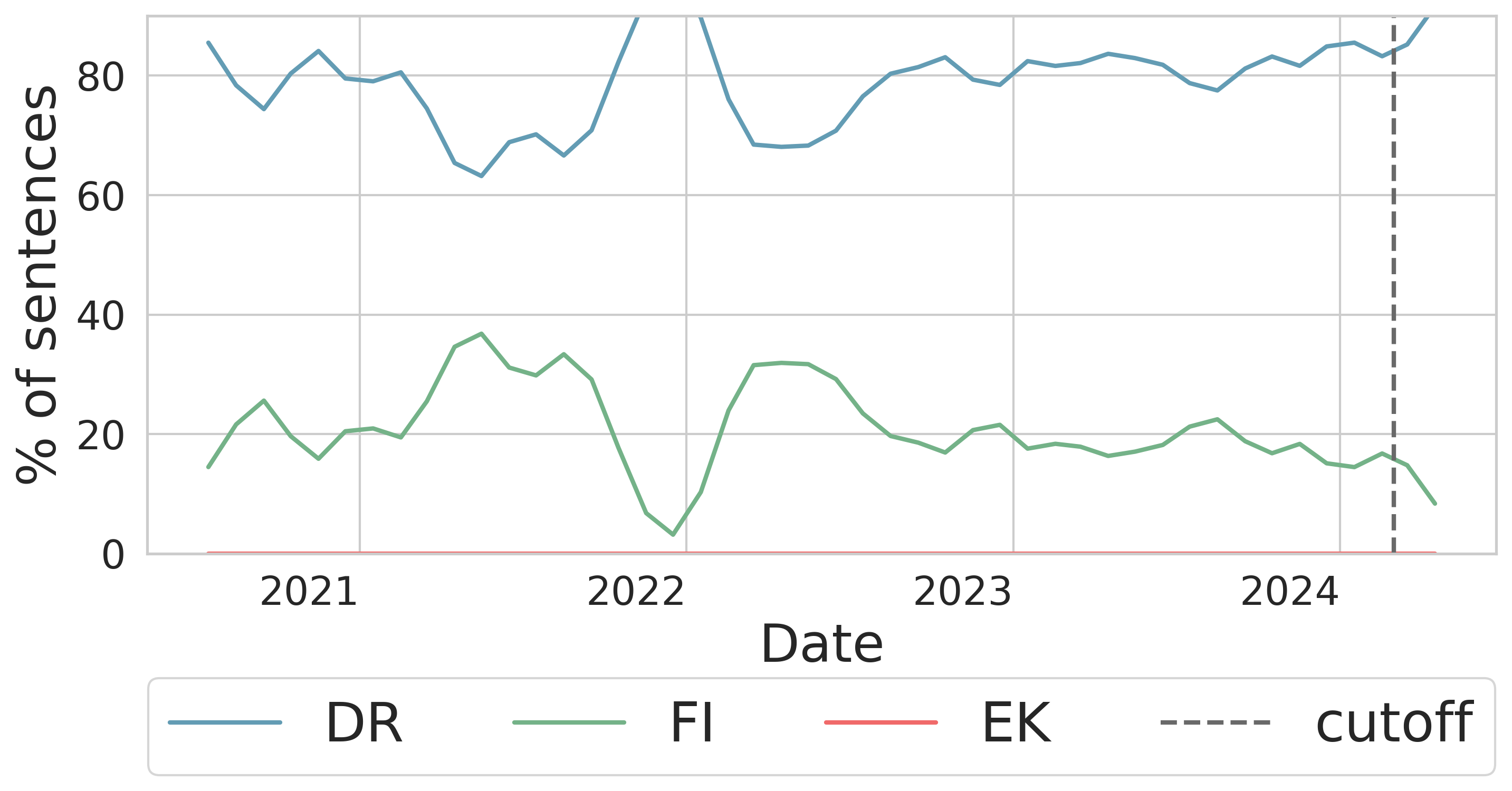}
        \caption{Phi-3}
    \end{subfigure}
    \begin{subfigure}{0.32\textwidth}  
        \centering
        \includegraphics[width=\linewidth]{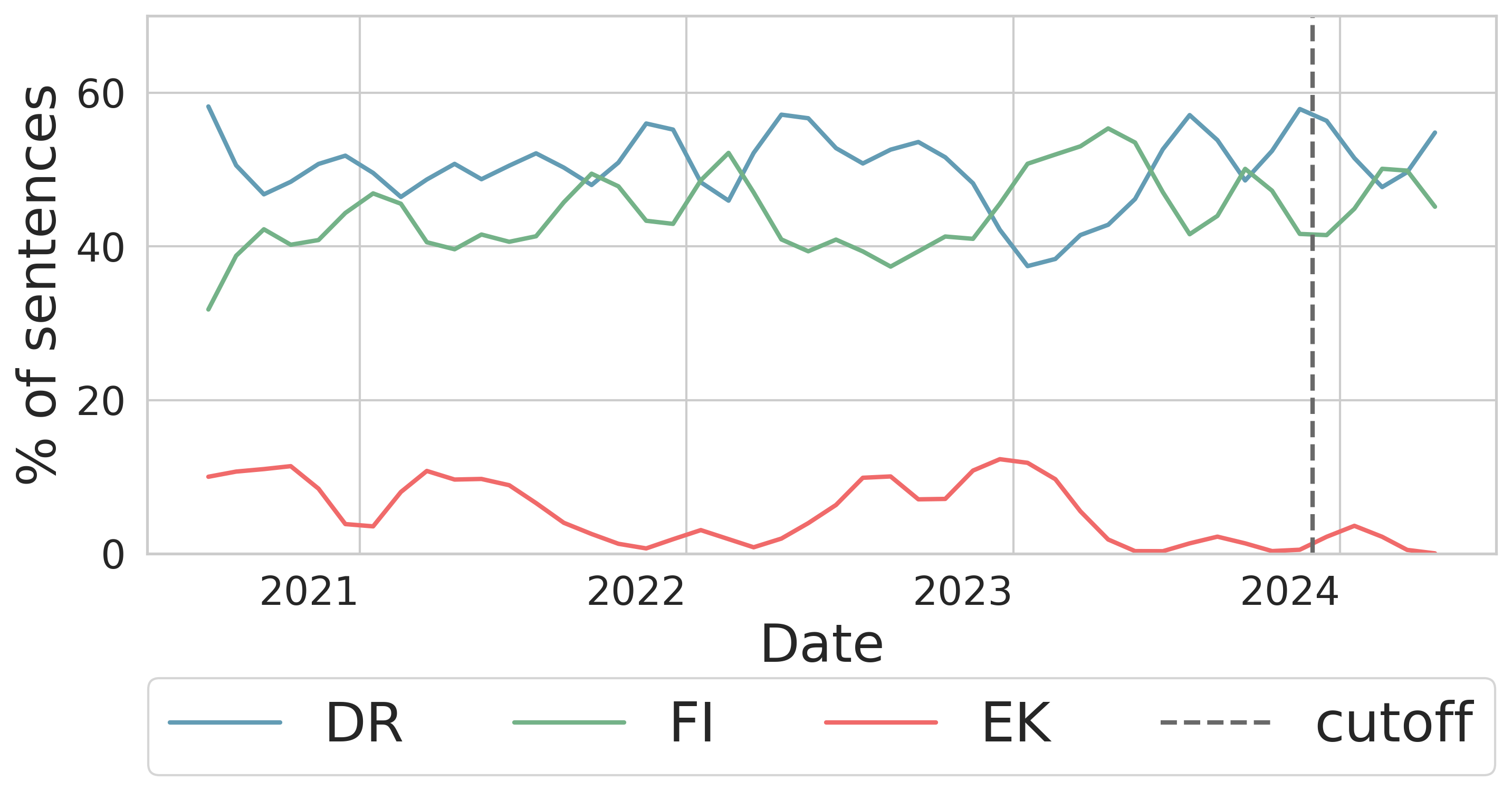}
        \caption{\llama}
        \label{fig:llama_short}
    \end{subfigure}
    \caption{Short Reports from synthetic time series (2019-2024) highlighting segments over time for different models.}
    \label{fig:span-time-plot-short-synt1}
\end{figure*}

\begin{figure*}[h!]
    \centering
    \begin{subfigure}{0.32\textwidth}  
        \centering
        \includegraphics[width=\linewidth]{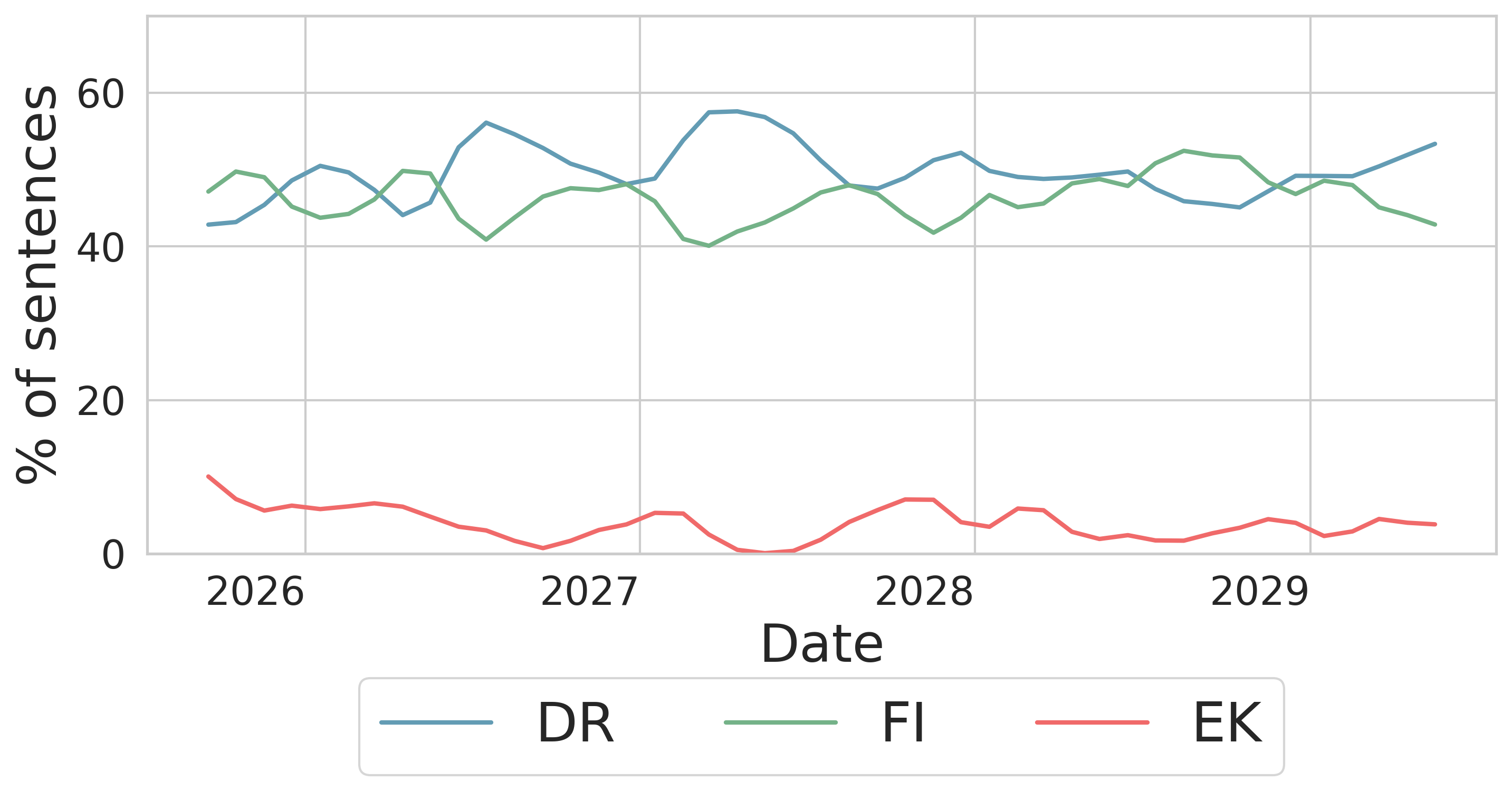}  
        \caption{GPT-4o}
    \end{subfigure}
    \begin{subfigure}{0.32\textwidth}  
        \centering
        \includegraphics[width=\linewidth]{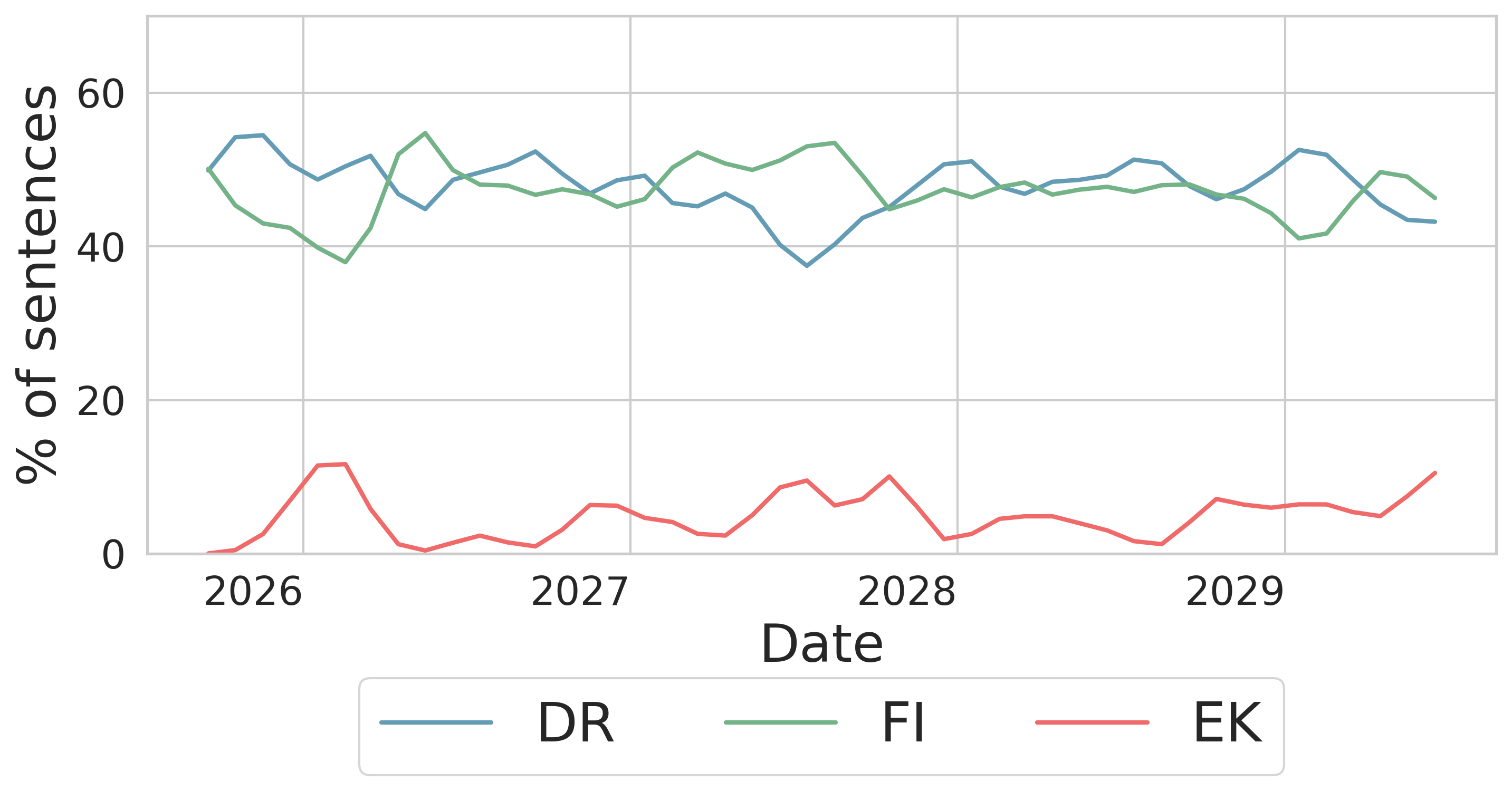}
        \caption{GPT-4o-mini}
    \end{subfigure}
    \begin{subfigure}{0.32\textwidth}  
        \centering
        \includegraphics[width=\linewidth]{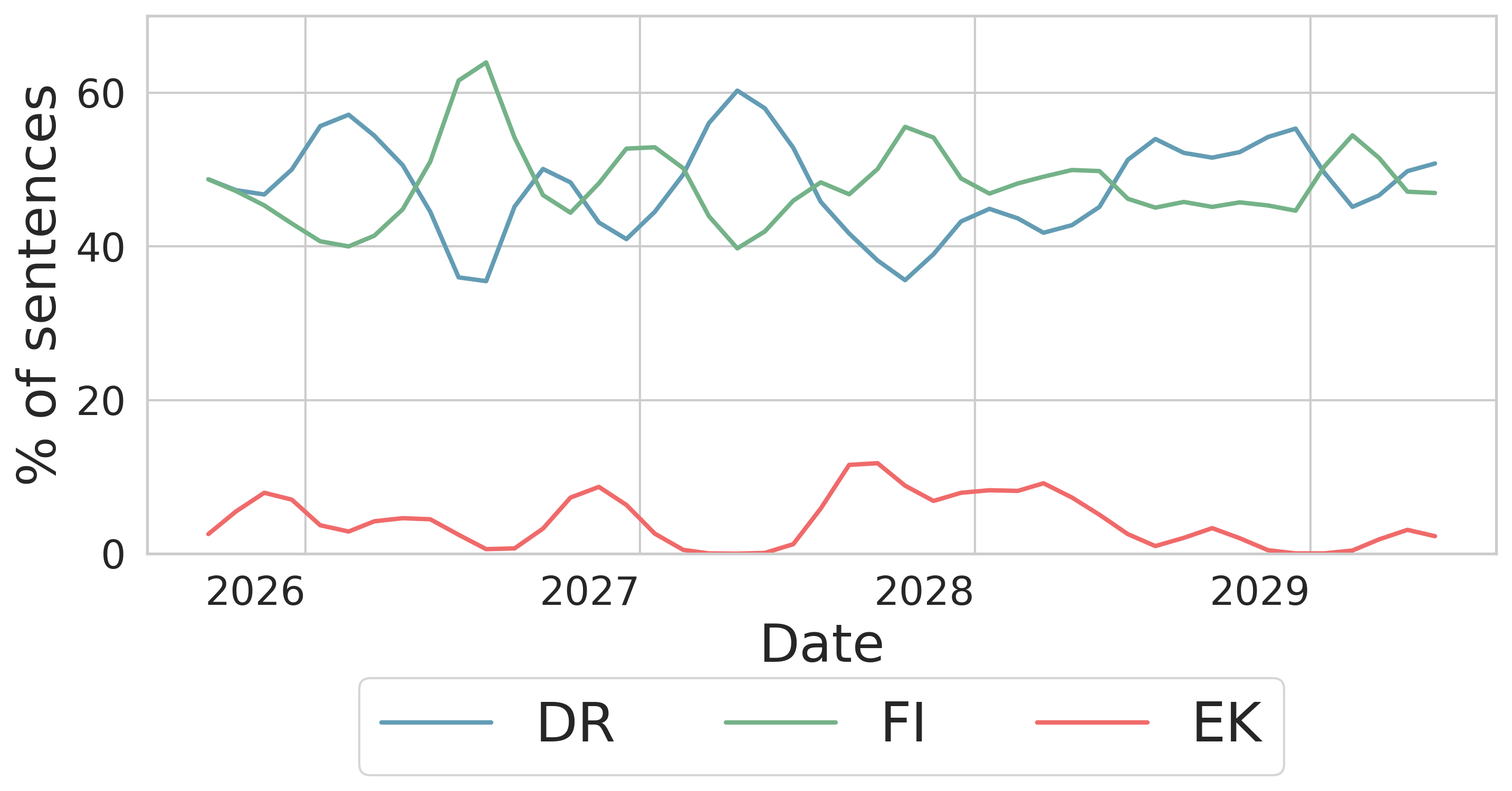}
        \caption{Gemini}
    \end{subfigure}\\
    \begin{subfigure}{0.32\textwidth}  
        \centering
        \includegraphics[width=\linewidth]{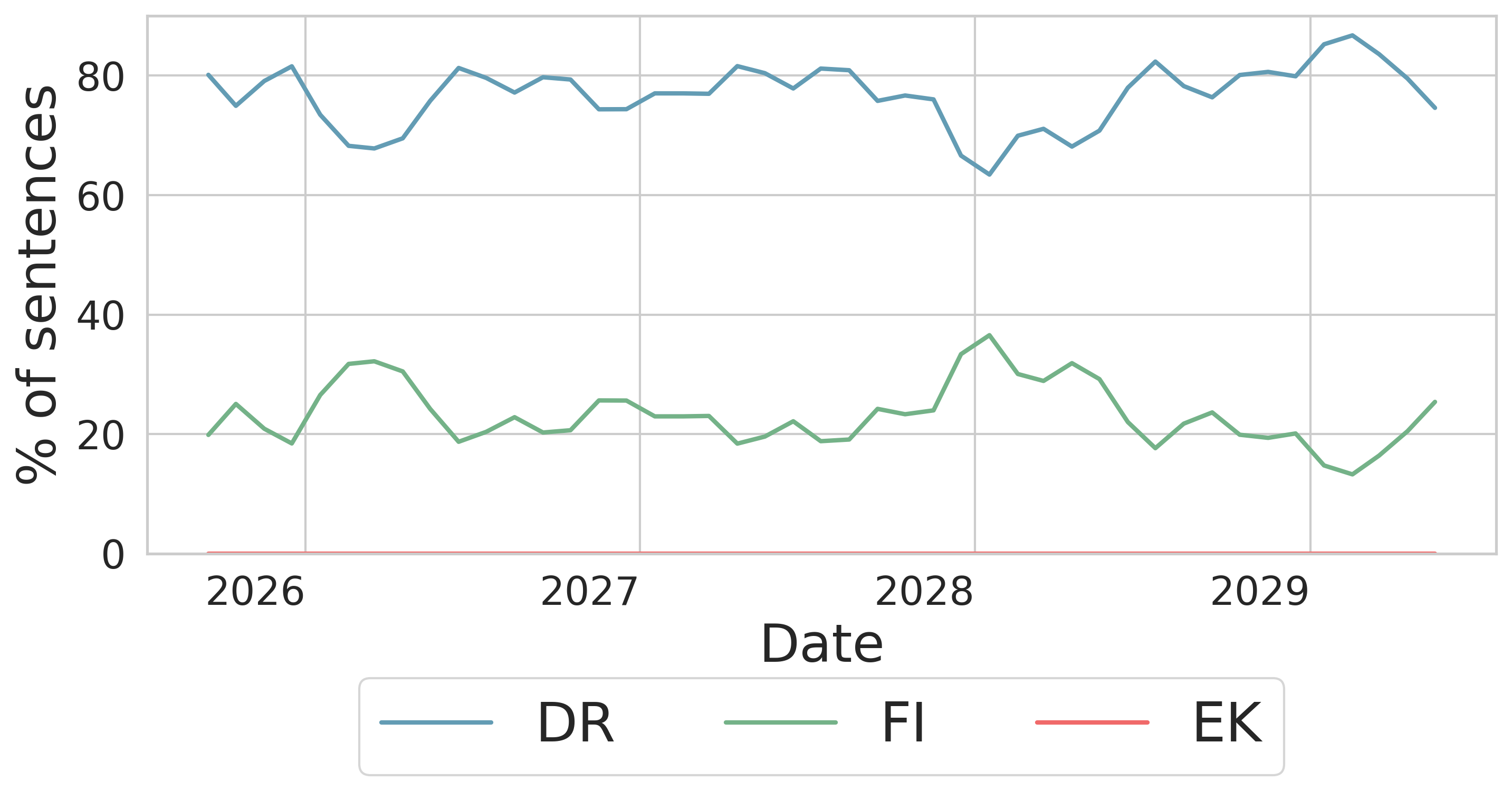}
        \caption{Phi-3}
    \end{subfigure}
    \begin{subfigure}{0.32\textwidth}  
        \centering
        \includegraphics[width=\linewidth]{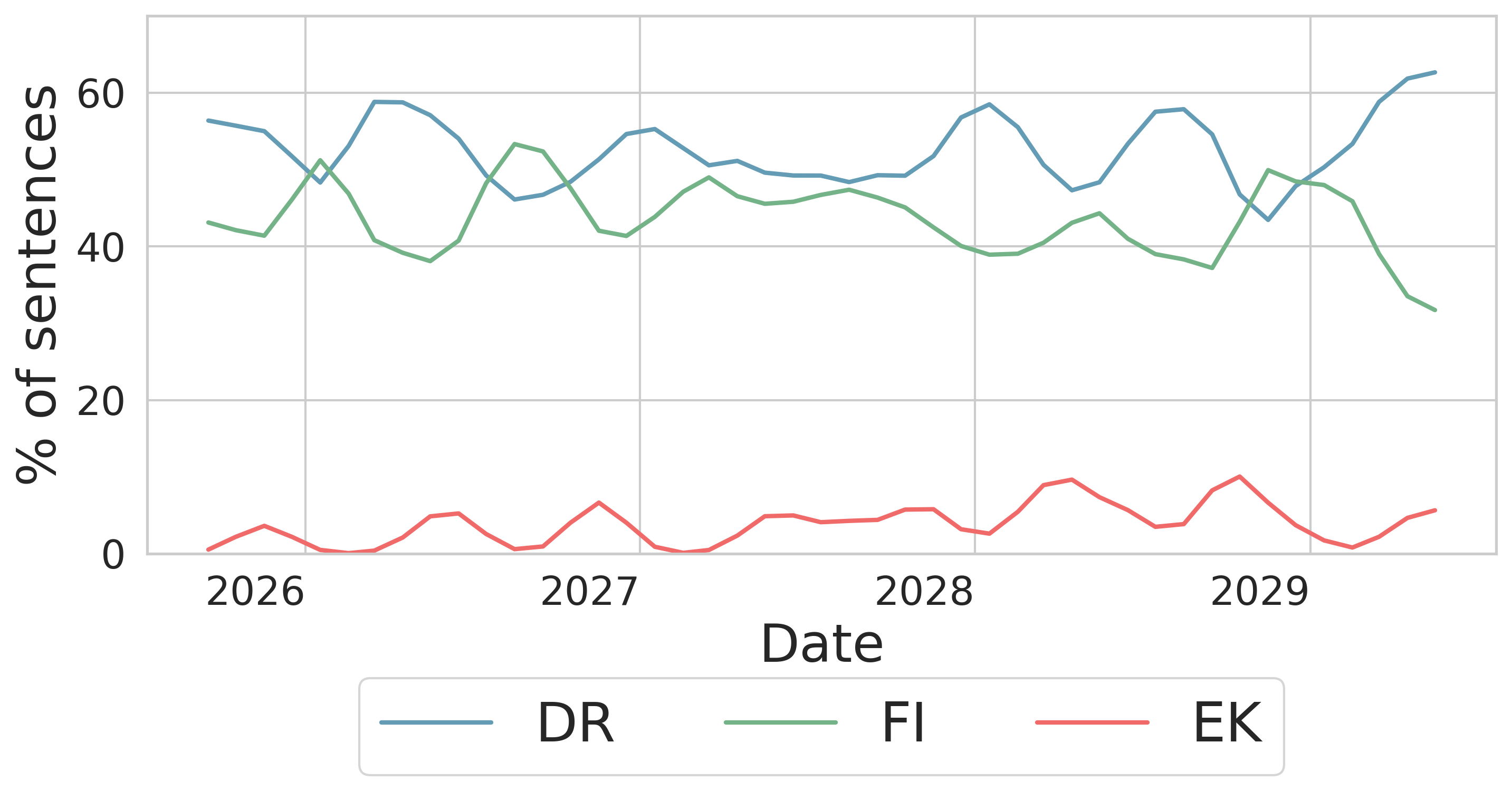}
        \caption{\llama}
    \end{subfigure}
    \caption{Short Reports from synthetic time series (2024-2029) highlighting segments over time for different models.}
    \label{fig:span-time-plot-short-synt2}
\end{figure*}

\begin{figure*}[h!]
    \centering
    \begin{subfigure}{0.24\textwidth}  
        \centering
        \includegraphics[width=\linewidth]{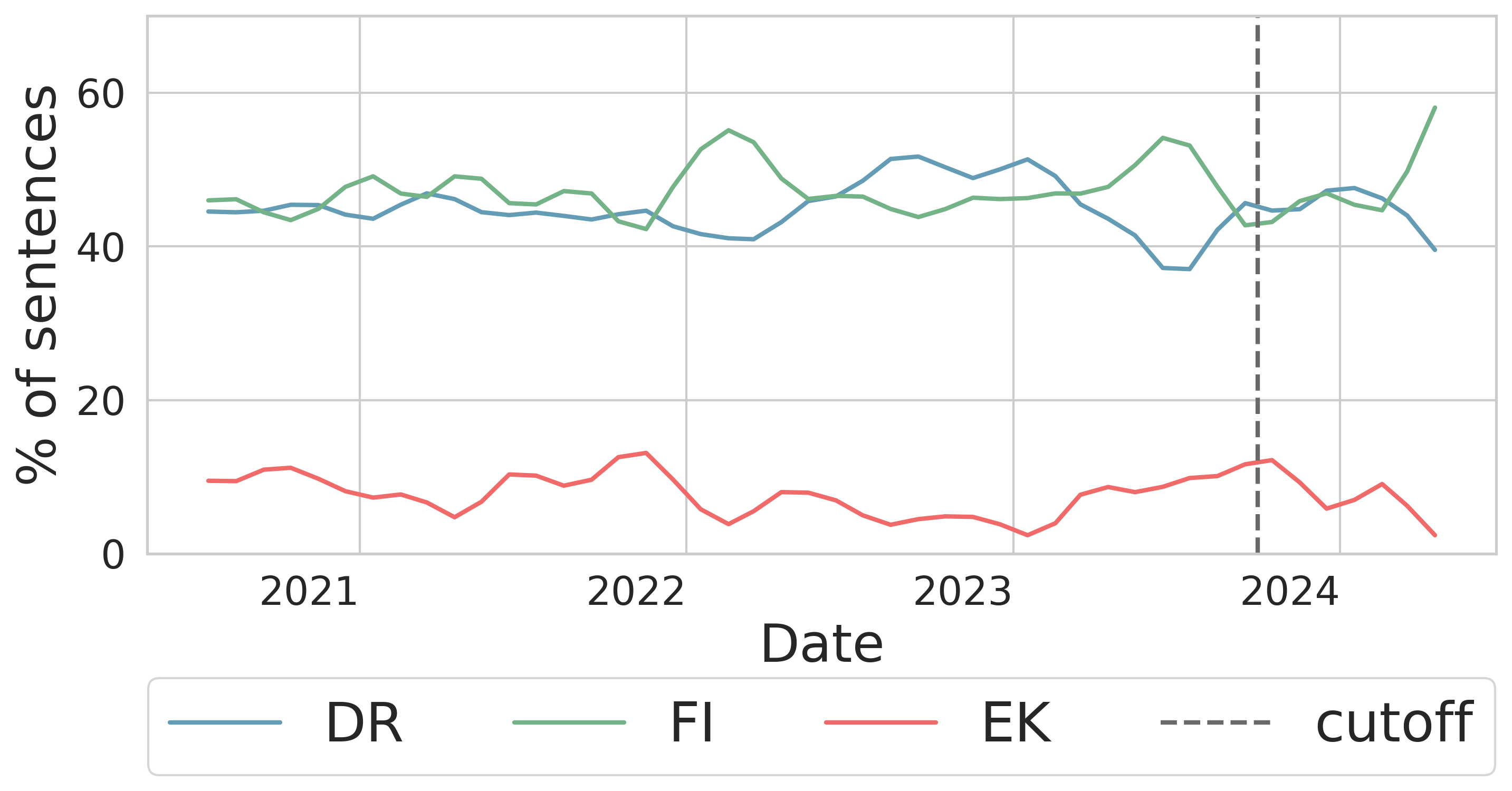}  
        \caption{GPT-4o}
    \end{subfigure}
    \begin{subfigure}{0.24\textwidth}  
        \centering
        \includegraphics[width=\linewidth]{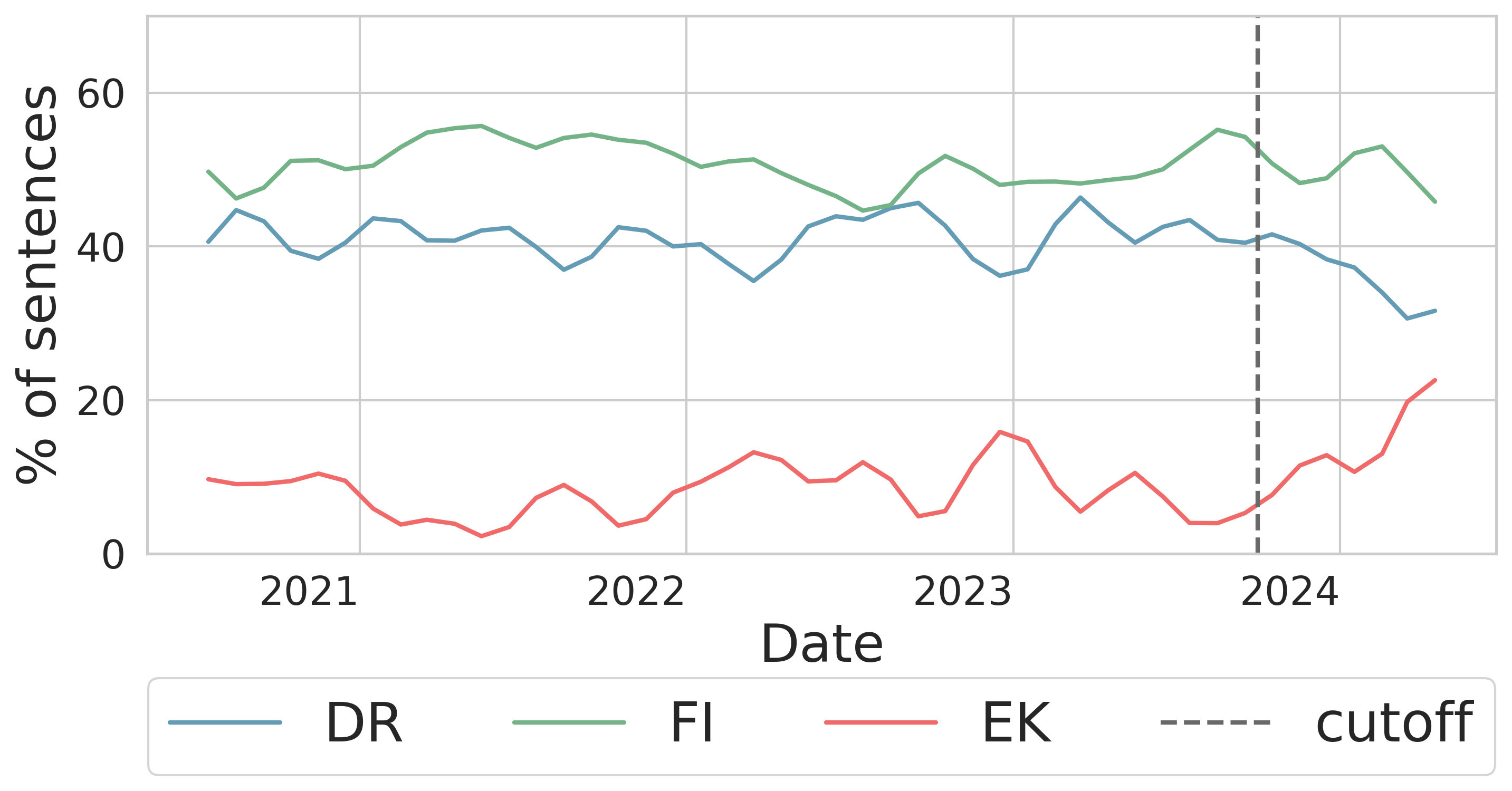}
        \caption{GPT-4o-mini}
    \end{subfigure}
    \begin{subfigure}{0.24\textwidth}  
        \centering
        \includegraphics[width=\linewidth]{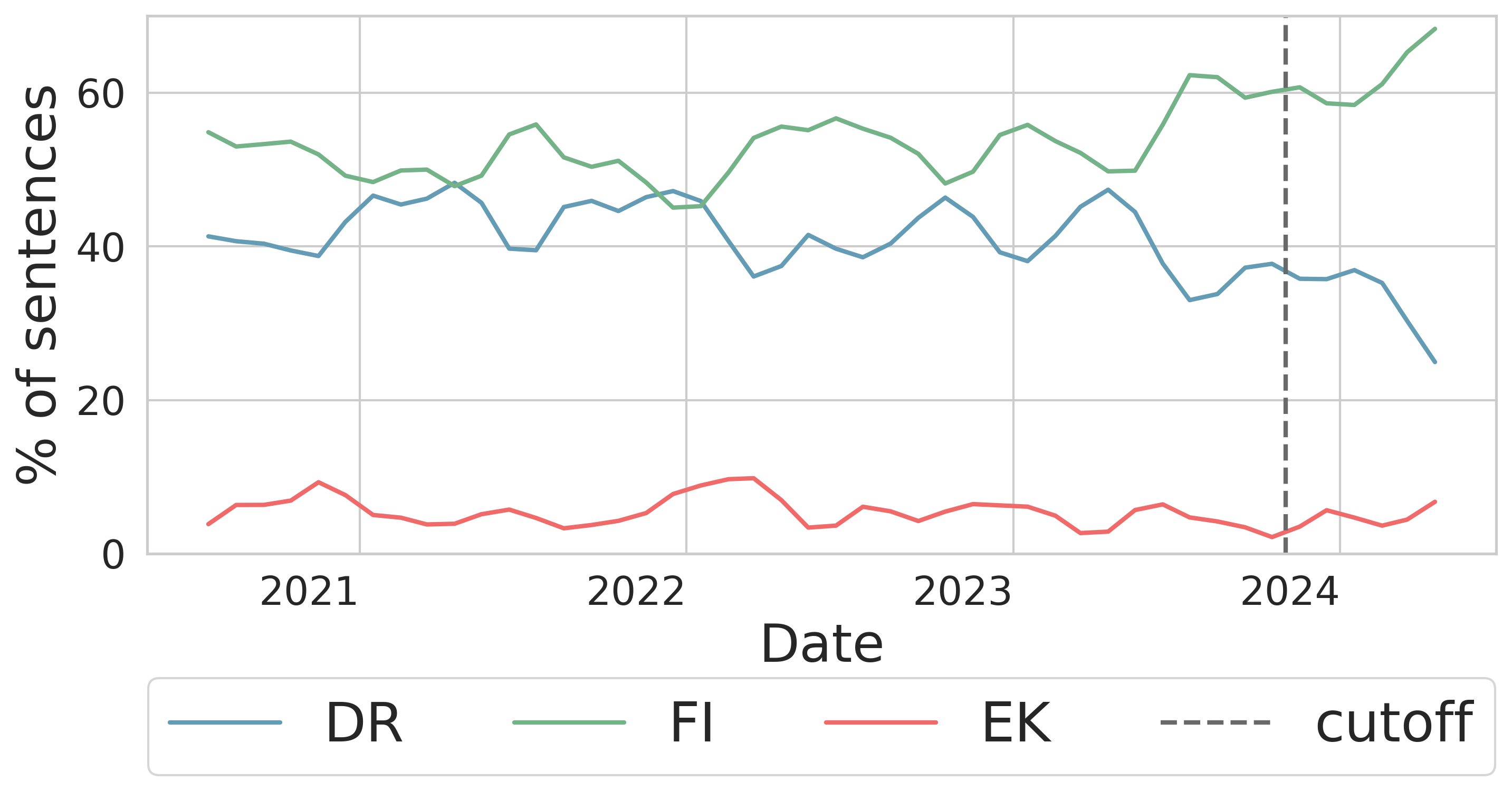}
        \caption{Gemini}
    \end{subfigure}
    \begin{subfigure}{0.24\textwidth}  
        \centering
        \includegraphics[width=\linewidth]{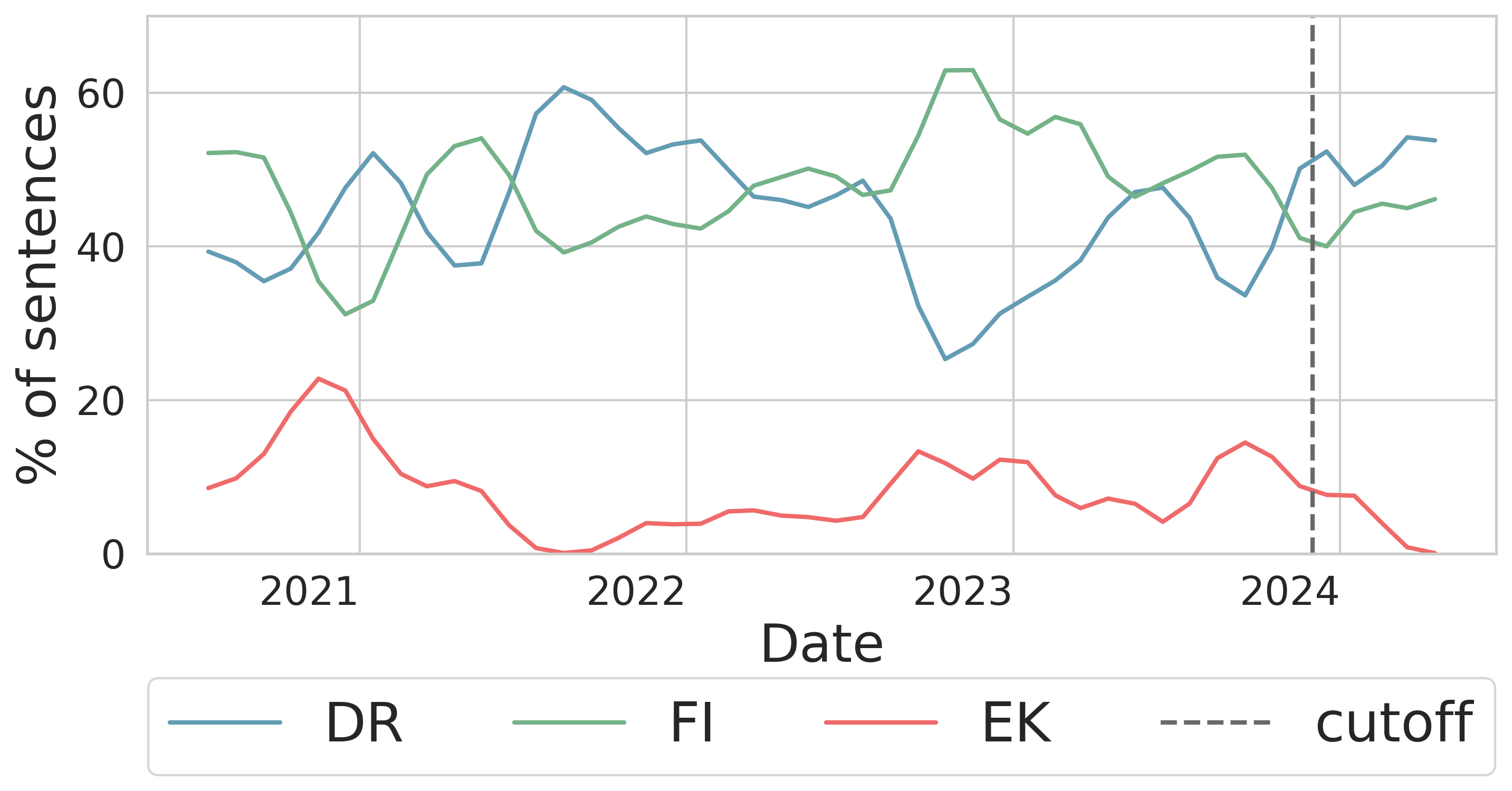}
        \caption{\llama}
    \end{subfigure}
    \caption{Technical Indicator Reports from synthetic time series (2019-2024) highlighting segments over time for different models.}
    \label{fig:span-time-plot-ti-synt1}
\end{figure*}

\begin{figure*}[h!]
    \centering
    \begin{subfigure}{0.24\textwidth}  
        \centering
        \includegraphics[width=\linewidth]{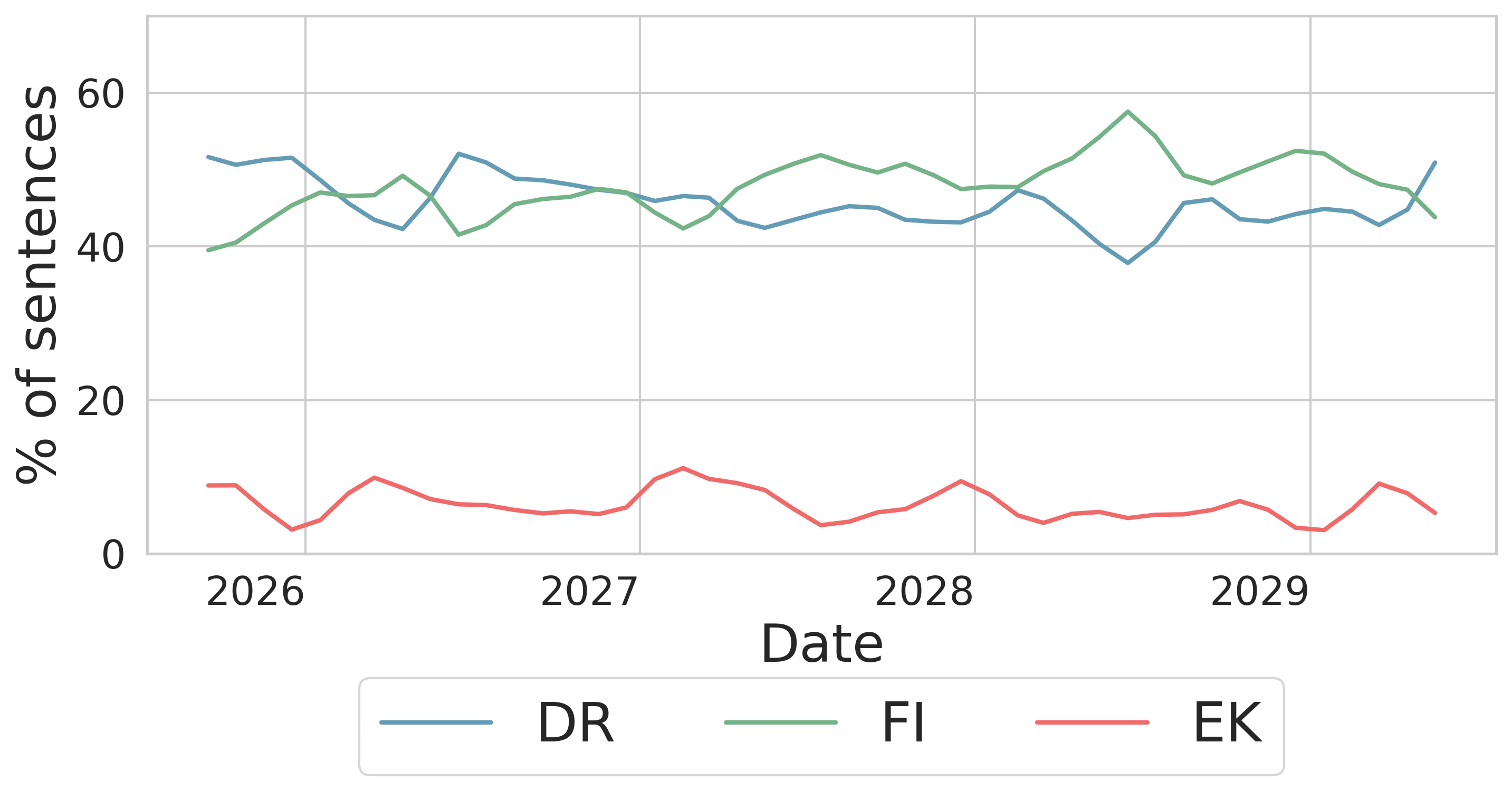}  
        \caption{GPT-4o}
    \end{subfigure}
    \begin{subfigure}{0.24\textwidth}  
        \centering
        \includegraphics[width=\linewidth]{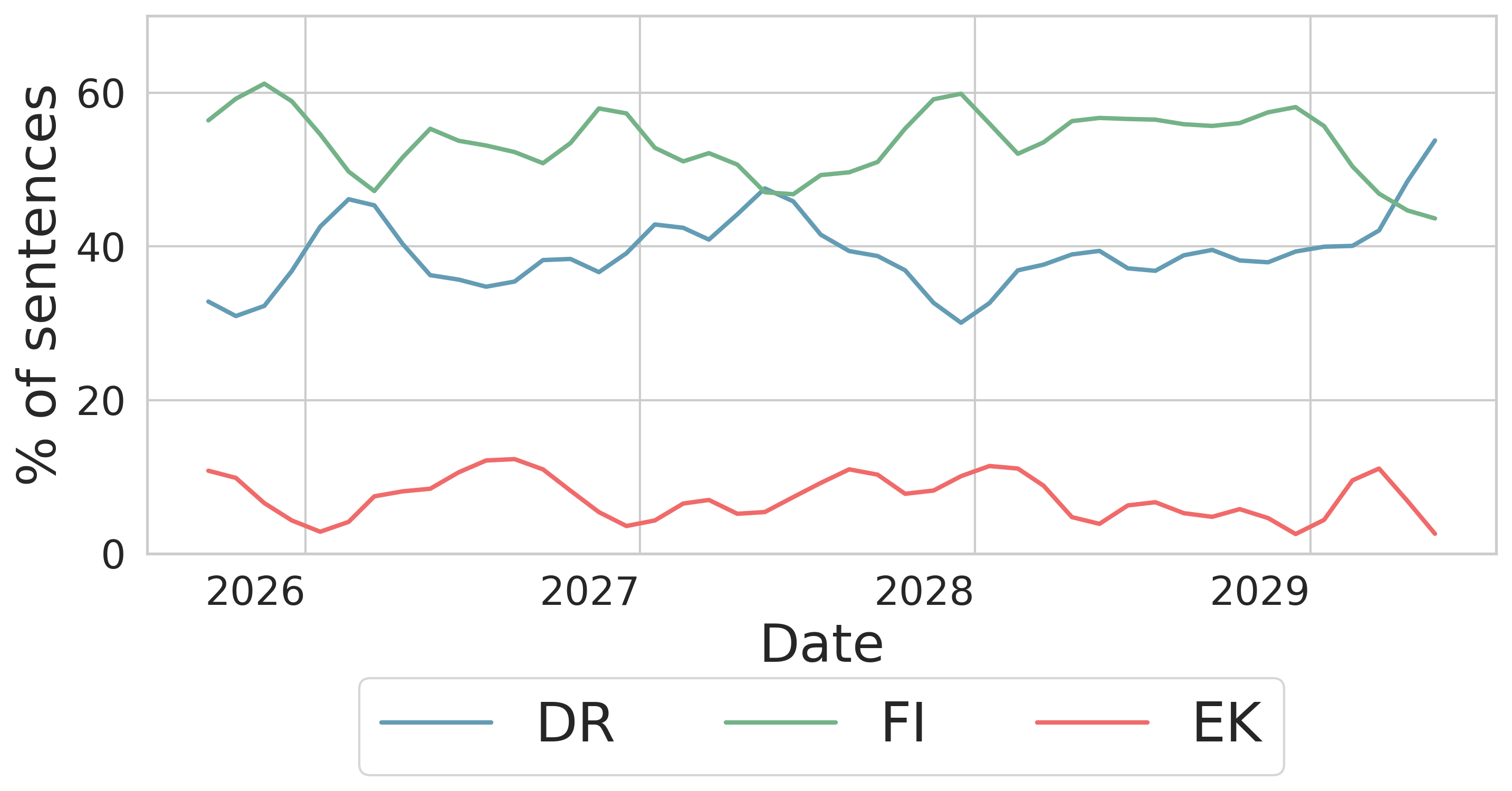}
        \caption{GPT-4o-mini}
    \end{subfigure}
    \begin{subfigure}{0.24\textwidth}  
        \centering
        \includegraphics[width=\linewidth]{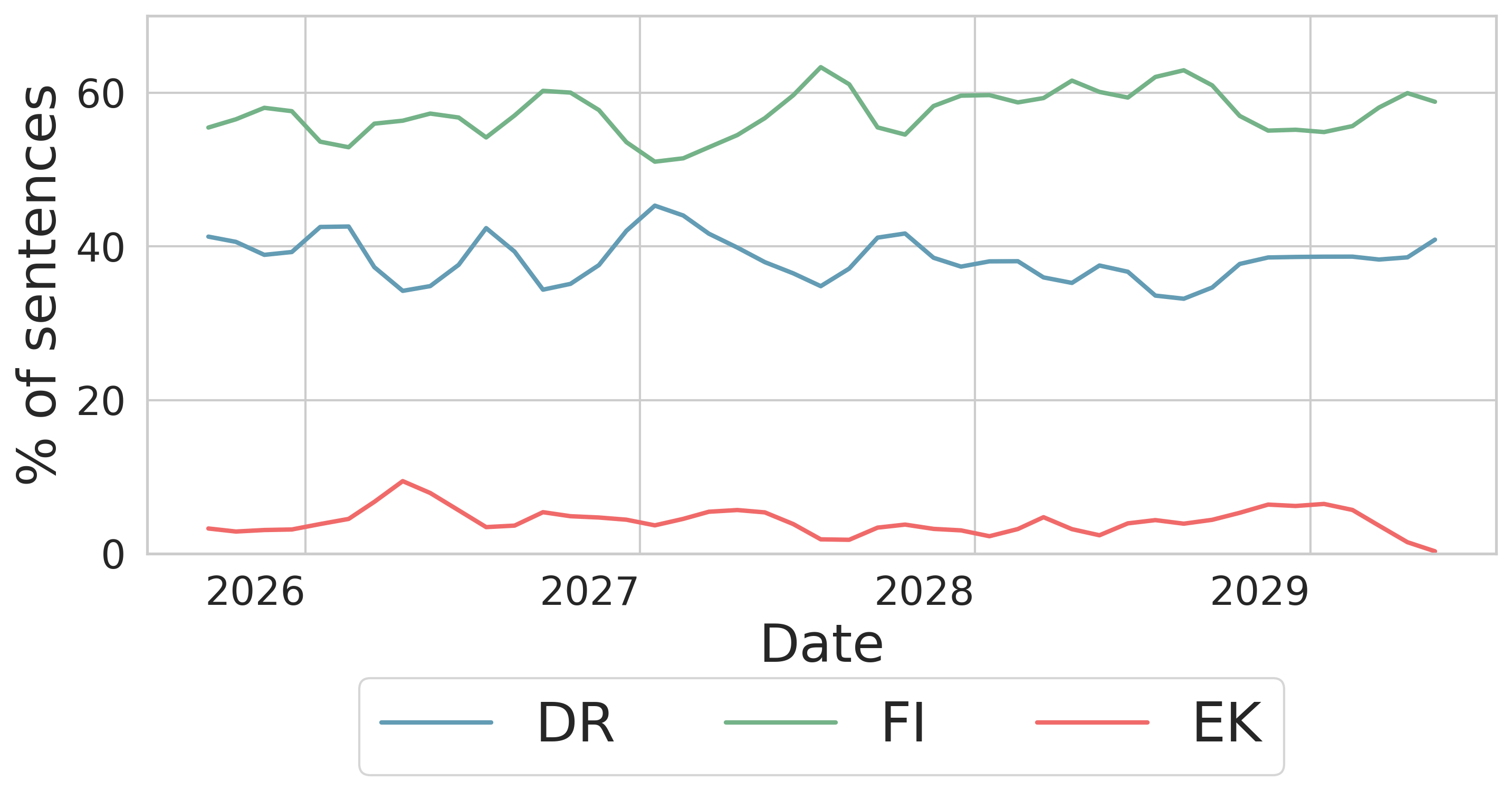}
        \caption{Gemini}
    \end{subfigure}
    \begin{subfigure}{0.24\textwidth}  
        \centering
        \includegraphics[width=\linewidth]{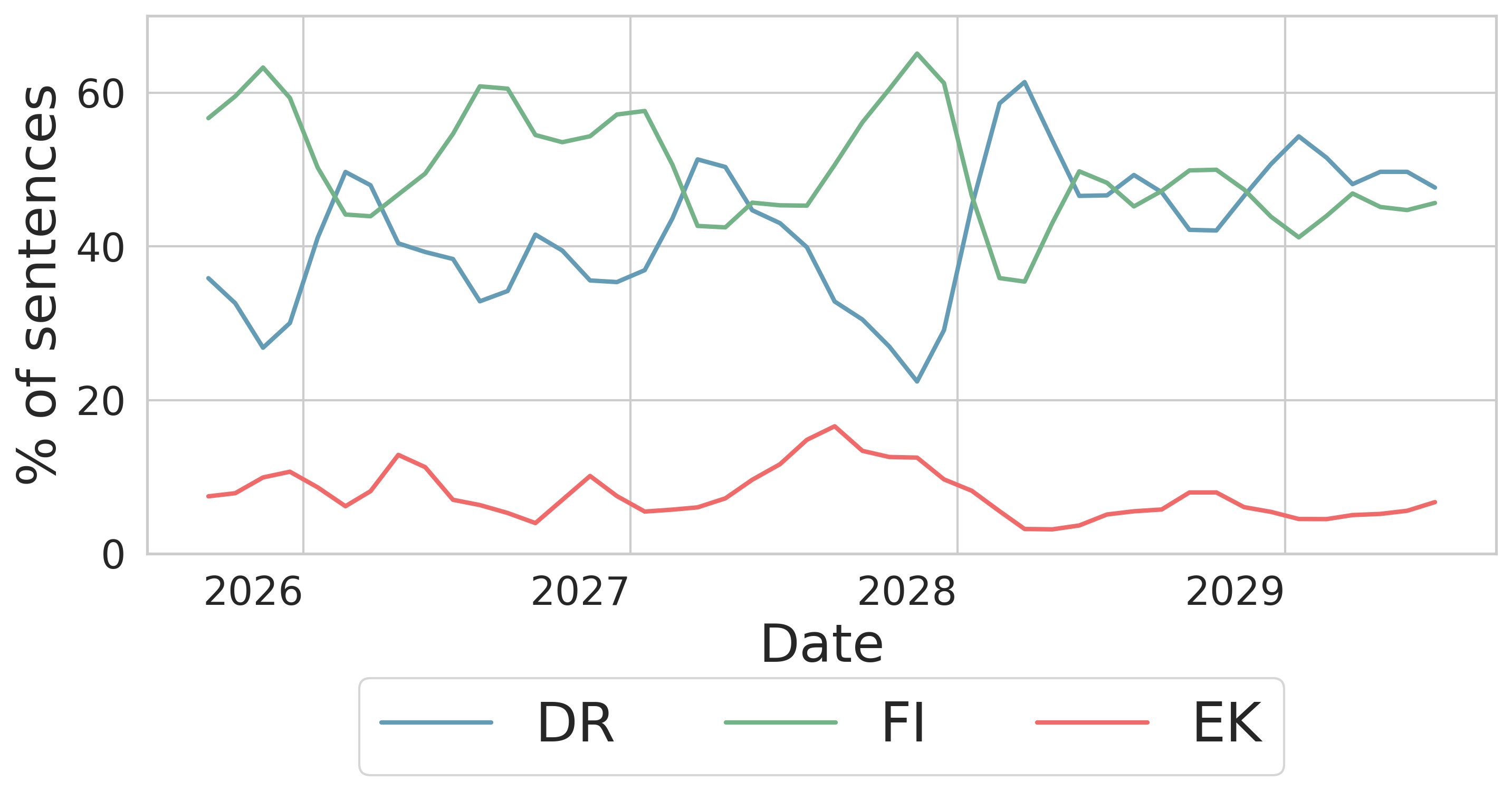}
        \caption{\llama}
    \end{subfigure}
    \caption{Technical Indicator Reports from synthetic time series (2024-2029) highlighting segments over time for different models.}
    \label{fig:span-time-plot-ti-synt2}
\end{figure*}


\begin{figure*}[h!]
    \centering
    \begin{subfigure}{0.24\textwidth}  
        \centering
        \includegraphics[width=\linewidth]{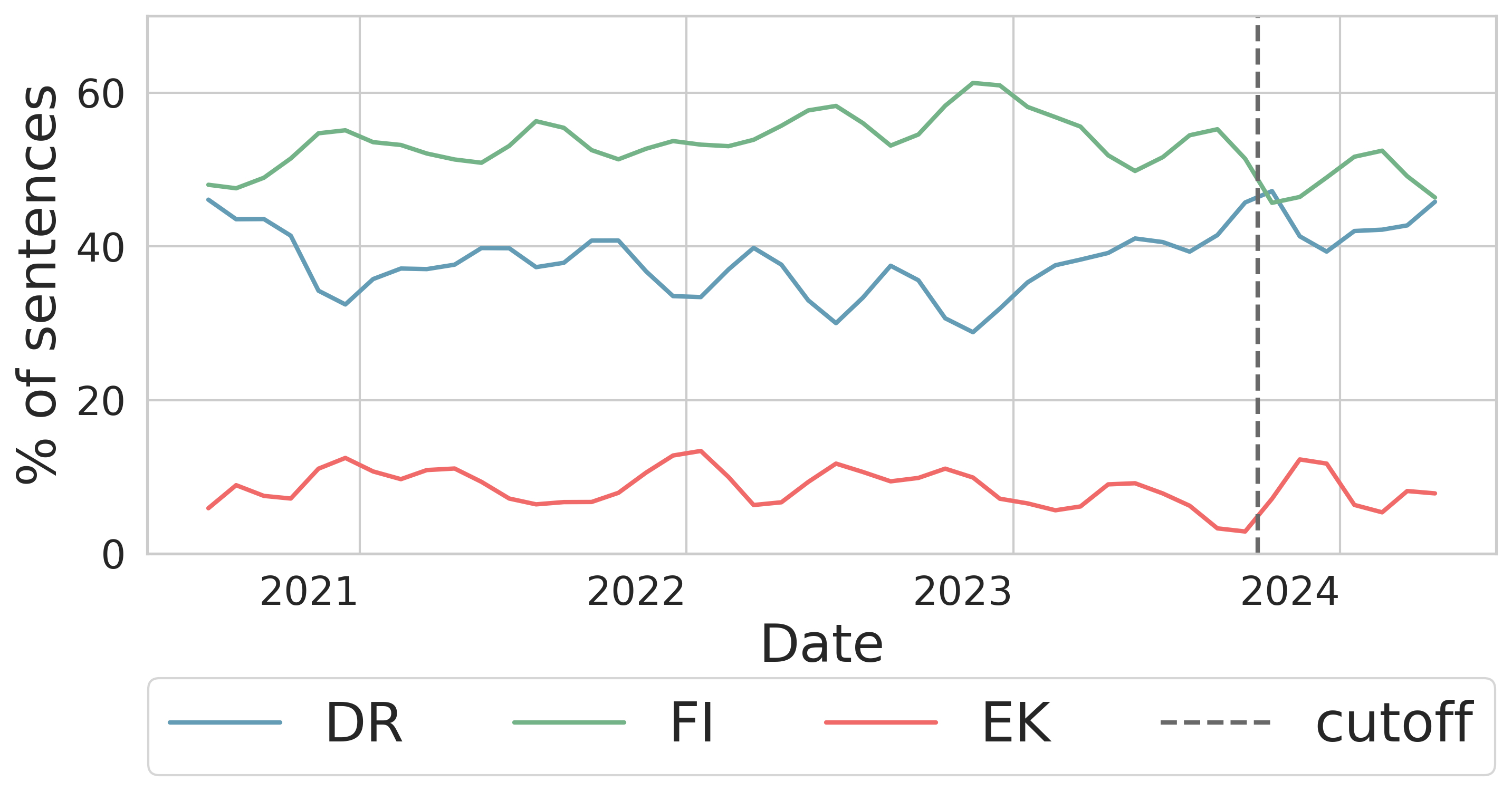}  
        \caption{GPT-4o}
    \end{subfigure}
    \begin{subfigure}{0.24\textwidth}  
        \centering
        \includegraphics[width=\linewidth]{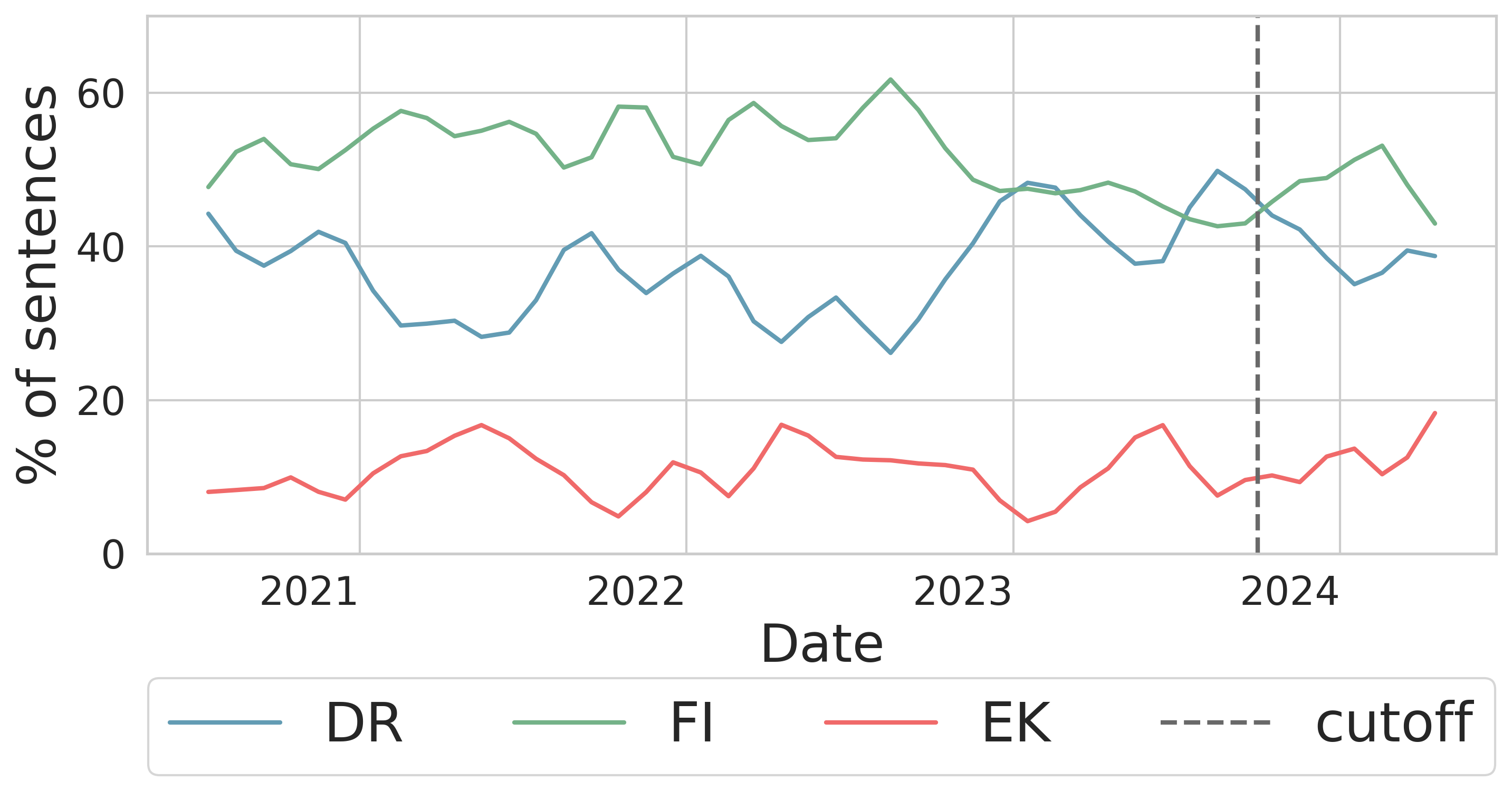}
        \caption{GPT-4o-mini}
    \end{subfigure}
    \begin{subfigure}{0.24\textwidth}  
        \centering
        \includegraphics[width=\linewidth]{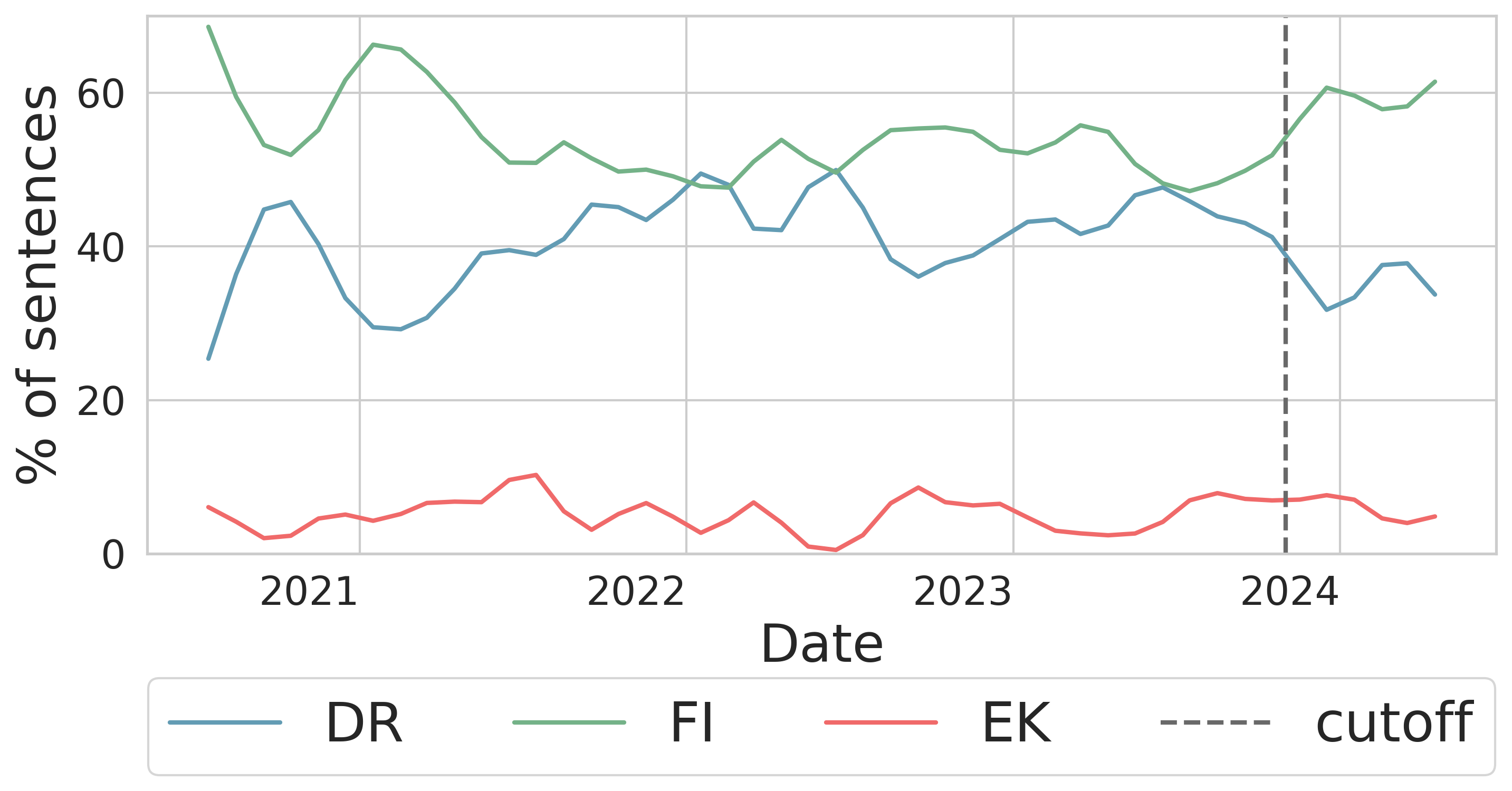}
        \caption{Gemini}
    \end{subfigure}
    \begin{subfigure}{0.24\textwidth}  
        \centering
        \includegraphics[width=\linewidth]{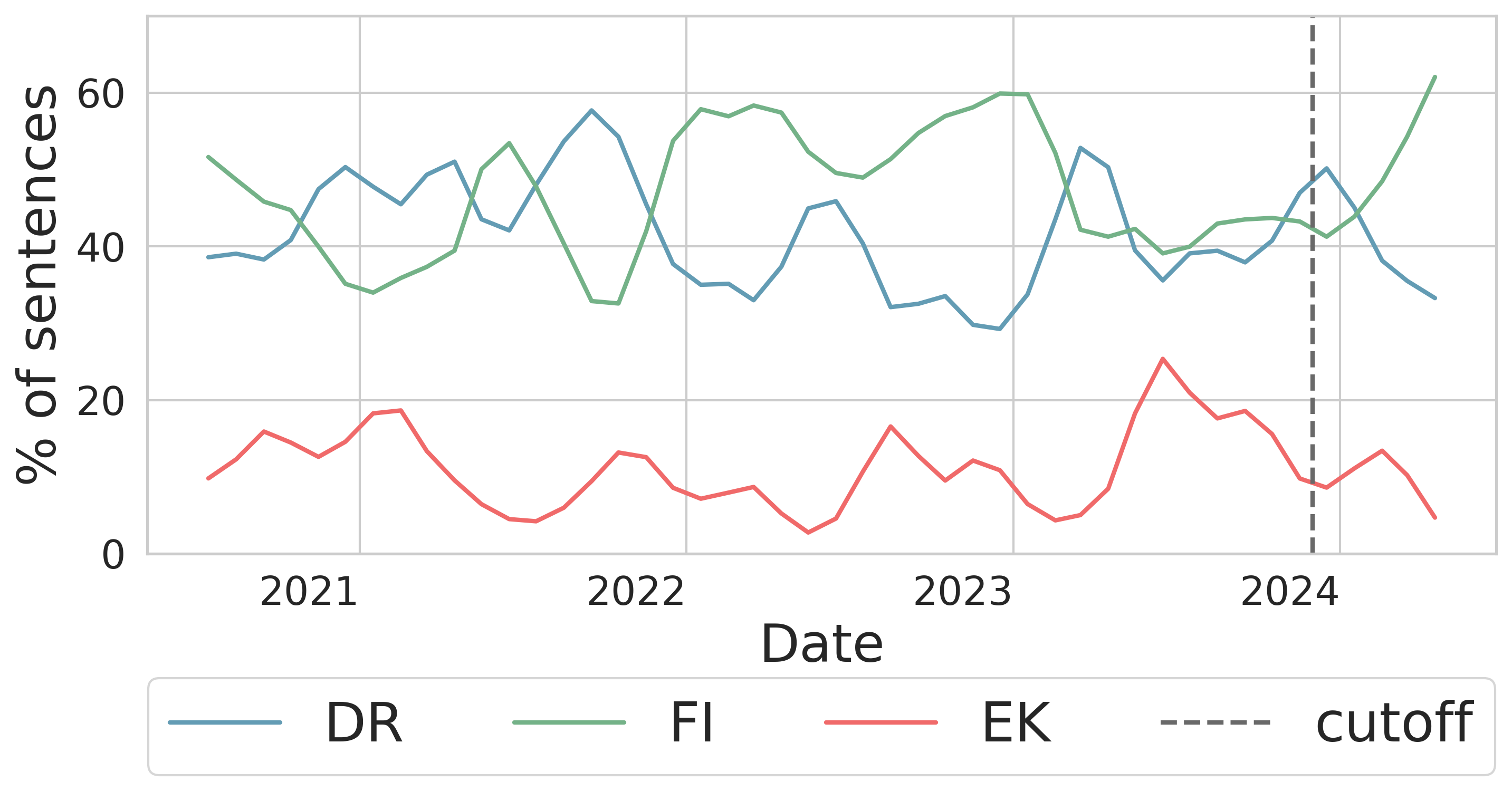}
        \caption{\llama}
    \end{subfigure}
    \caption{Technical Indicator (plots) Reports from synthetic time series (2019-2024) highlighting segments over time for different models.}
    \label{fig:span-time-plot-ti-plots-synt1}
\end{figure*}

\begin{figure*}[h!]
    \centering
    \begin{subfigure}{0.24\textwidth}  
        \centering
        \includegraphics[width=\linewidth]{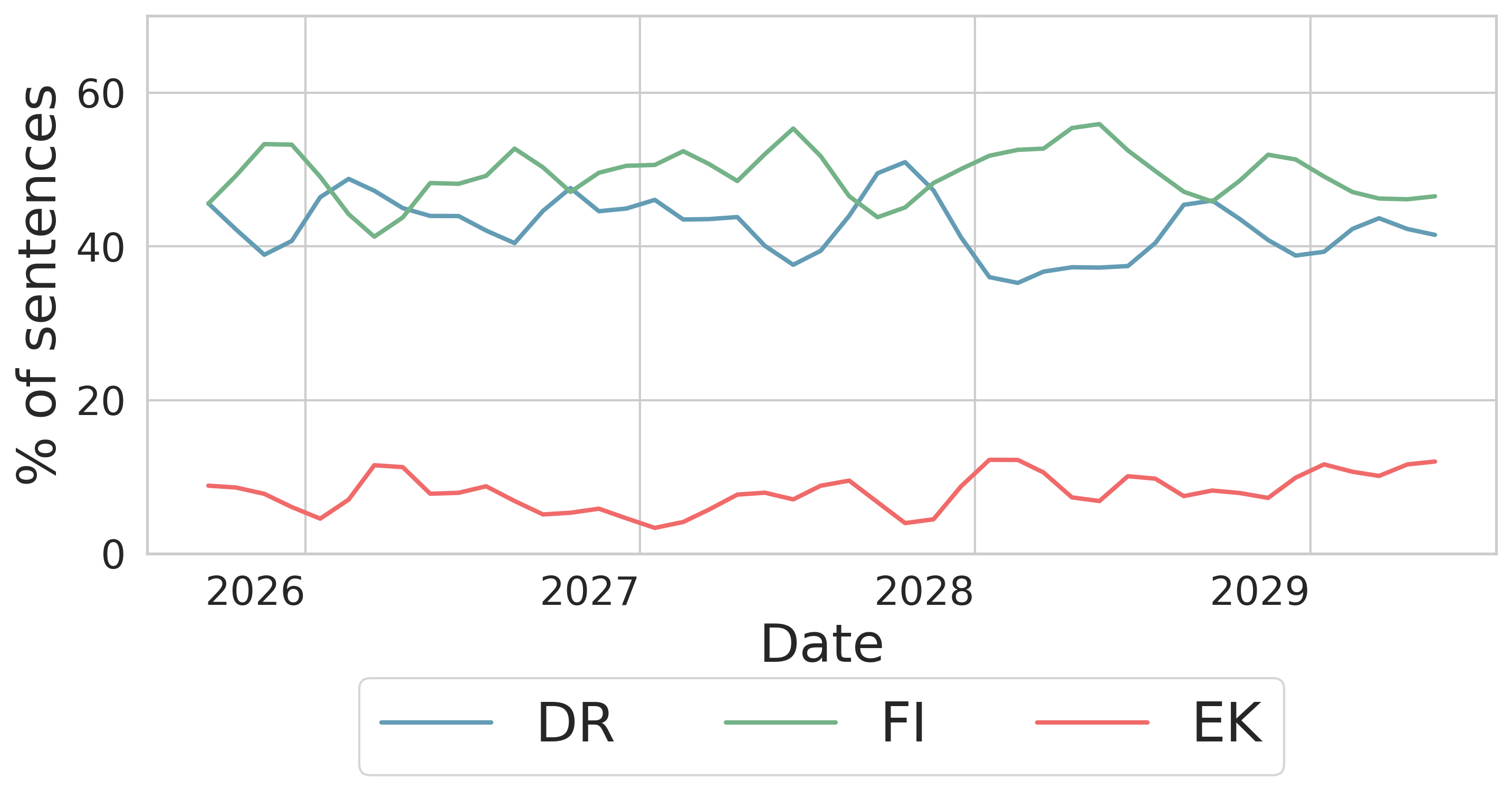}  
        \caption{GPT-4o}
    \end{subfigure}
    \begin{subfigure}{0.24\textwidth}  
        \centering
        \includegraphics[width=\linewidth]{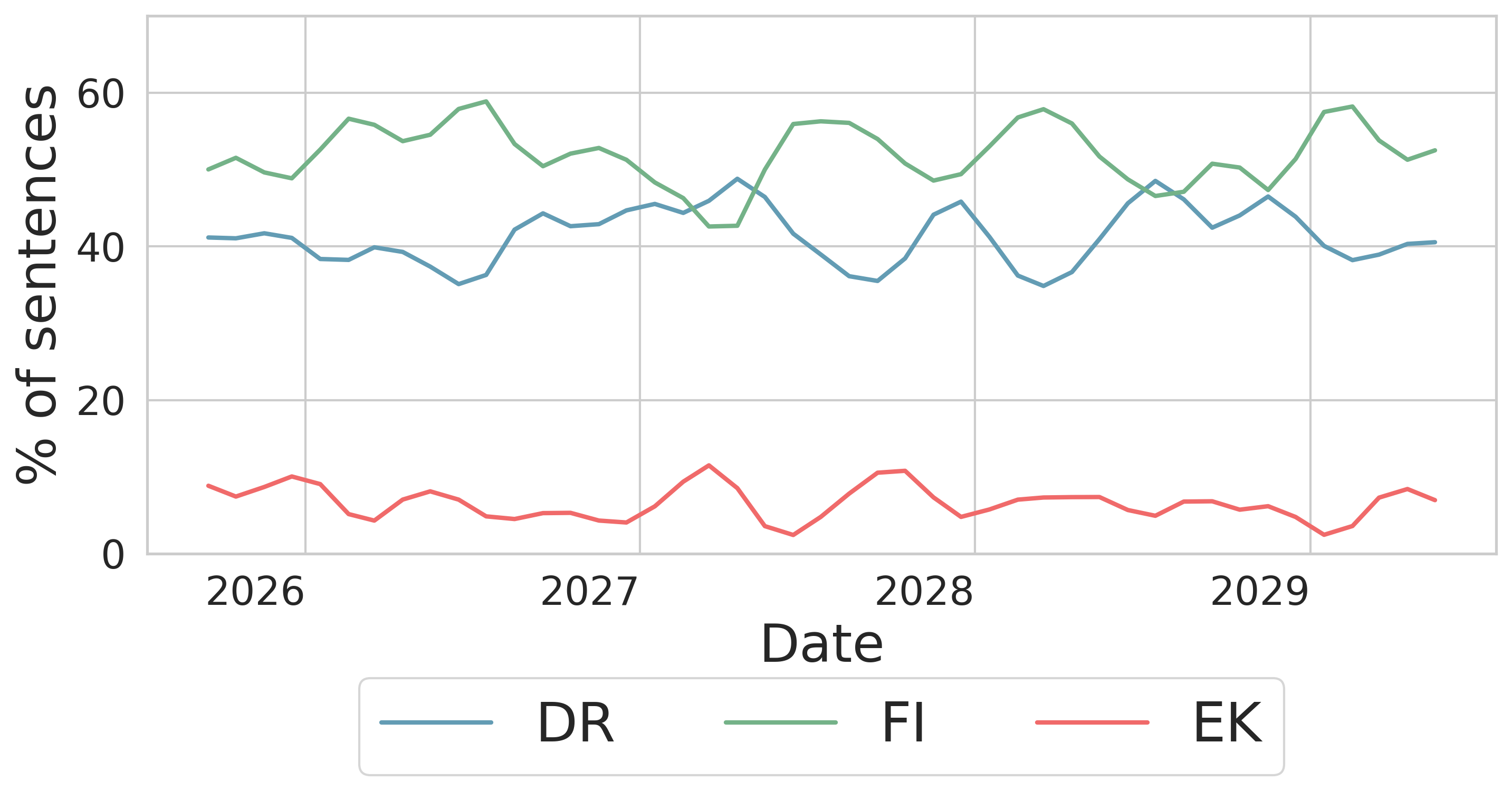}
        \caption{GPT-4o-mini}
    \end{subfigure}
    \begin{subfigure}{0.24\textwidth}  
        \centering
        \includegraphics[width=\linewidth]{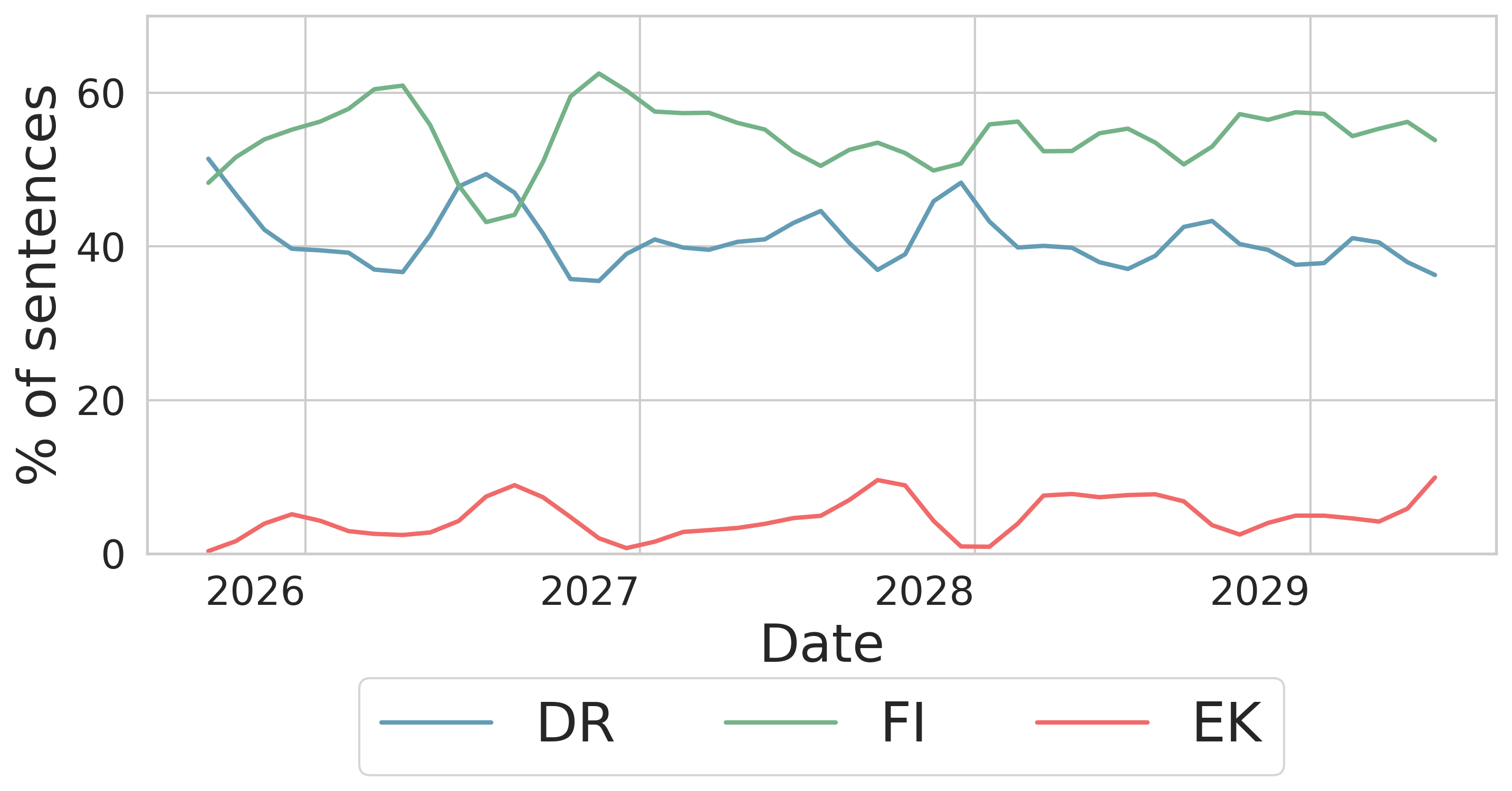}
        \caption{Gemini}
    \end{subfigure}
    \begin{subfigure}{0.24\textwidth}  
        \centering
        \includegraphics[width=\linewidth]{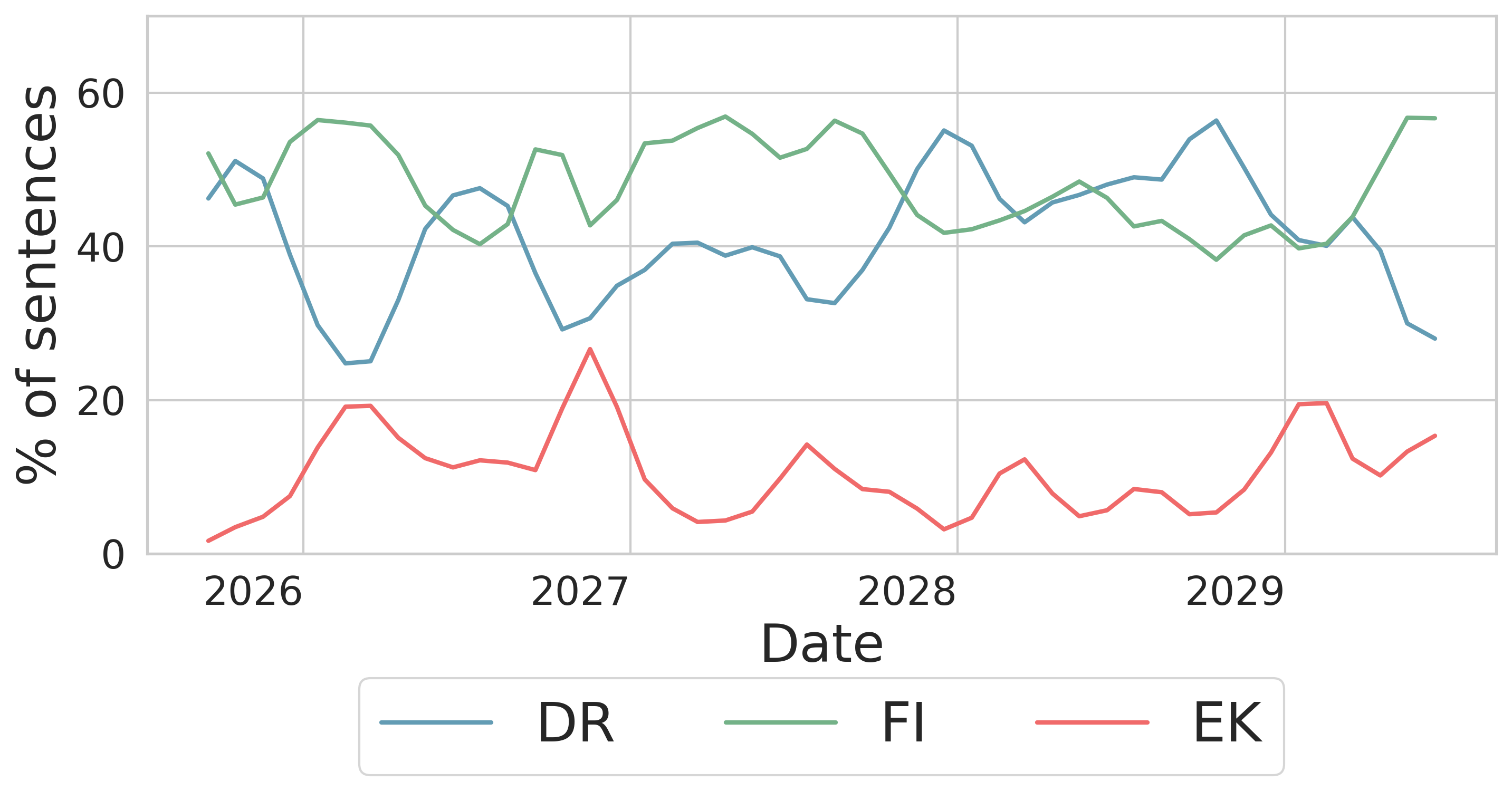}
        \caption{\llama}
    \end{subfigure}
    \caption{Technical Indicator (plots) Reports from synthetic time series (2024-2029) highlighting segments over time for different models.}
    \label{fig:span-time-plot-ti-plots-synt2}
\end{figure*}

\clearpage

\section{Additional analysis G-Eval scores}
\label{sec:app:geval-time}
\begin{figure*}[h!]
    \centering
        \begin{subfigure}{0.32\textwidth}  
        \centering
        \includegraphics[width=\linewidth]{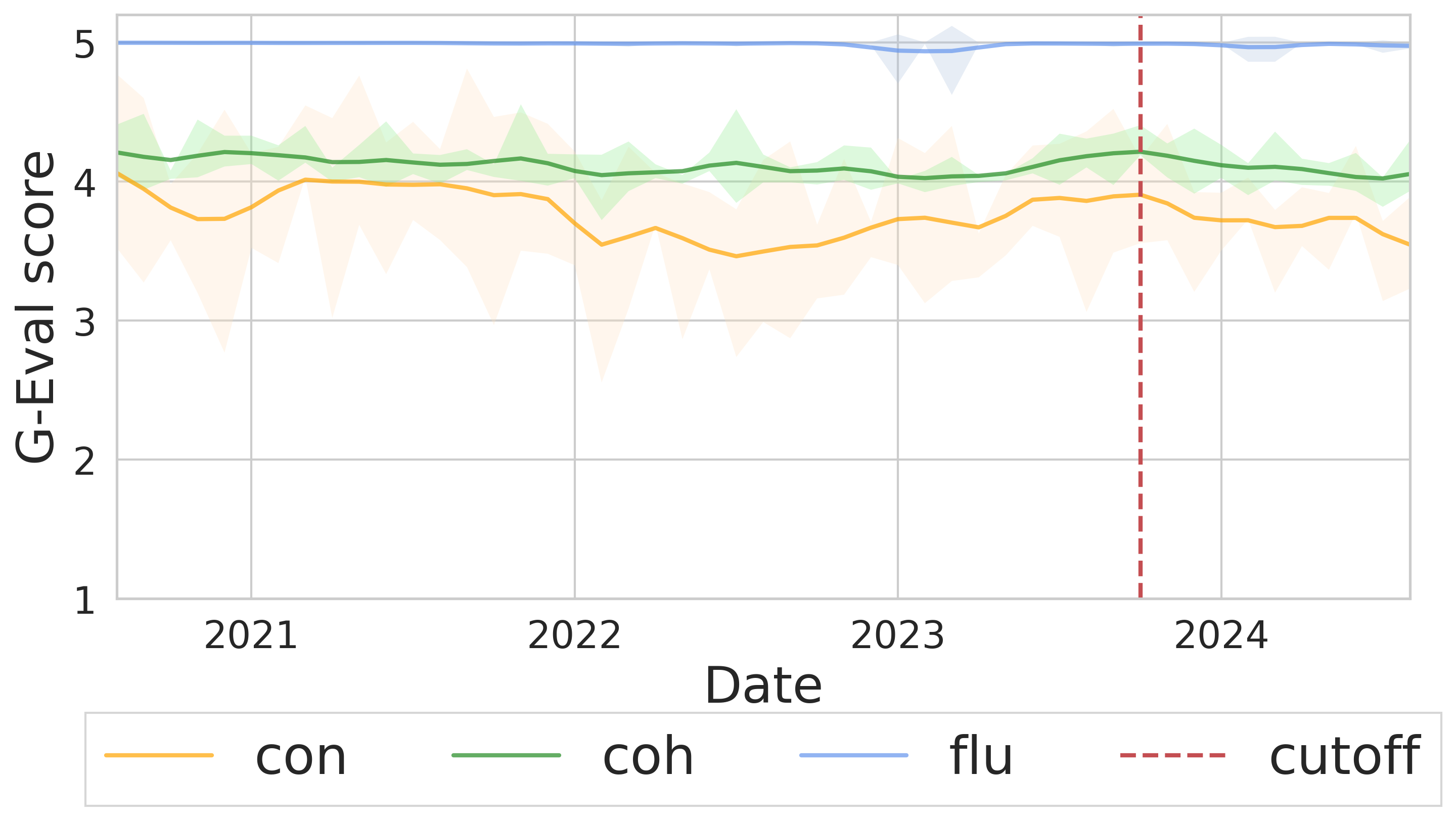}
        \caption{GPT-4o -- real}
    \end{subfigure}
    \begin{subfigure}{0.32\textwidth}  
        \centering
        \includegraphics[width=\linewidth]{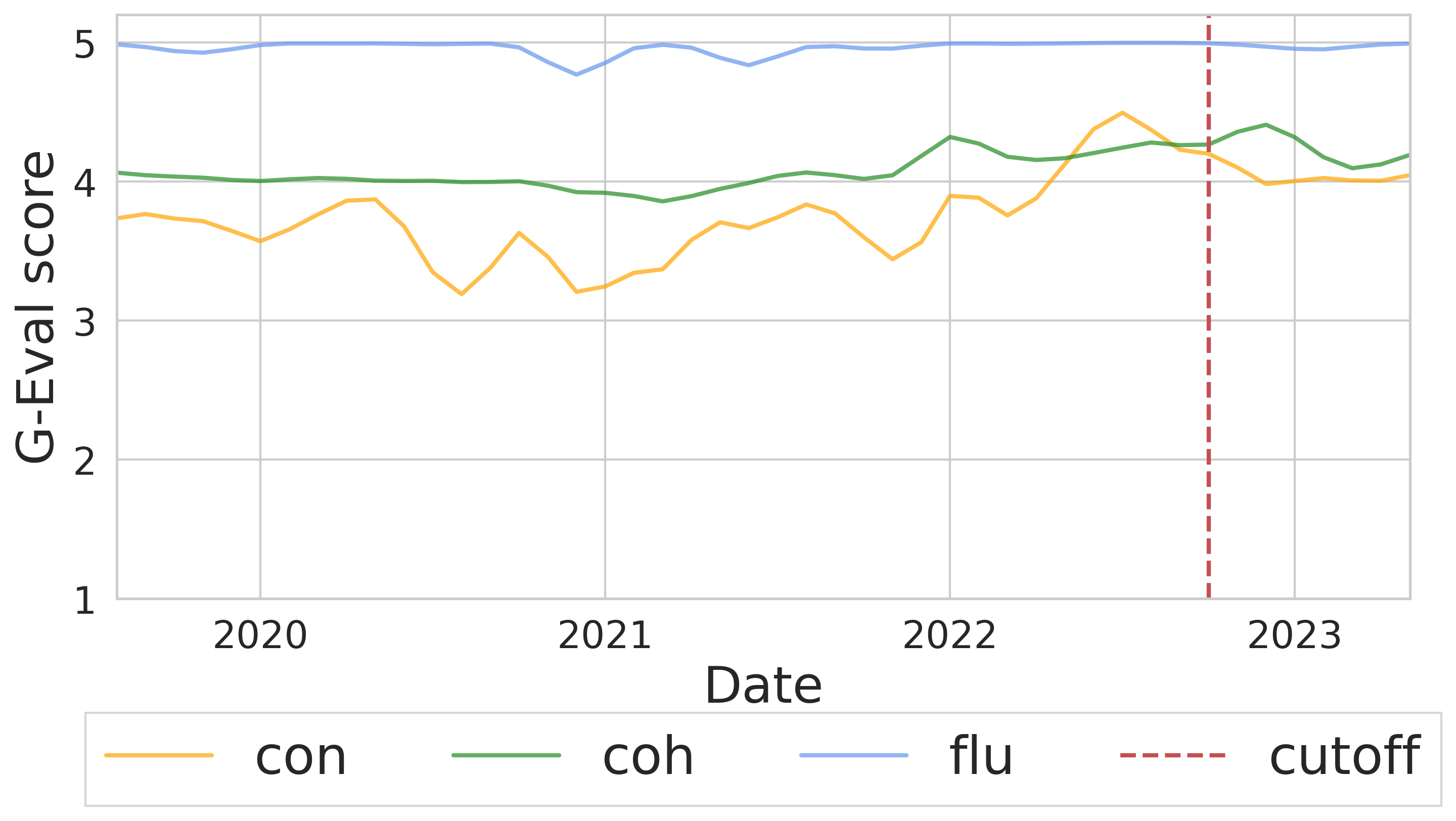}
        \caption{\gpt -- Synt (2019-2024)}
    \end{subfigure}
    \begin{subfigure}{0.32\textwidth}  
        \centering
        \includegraphics[width=\linewidth]{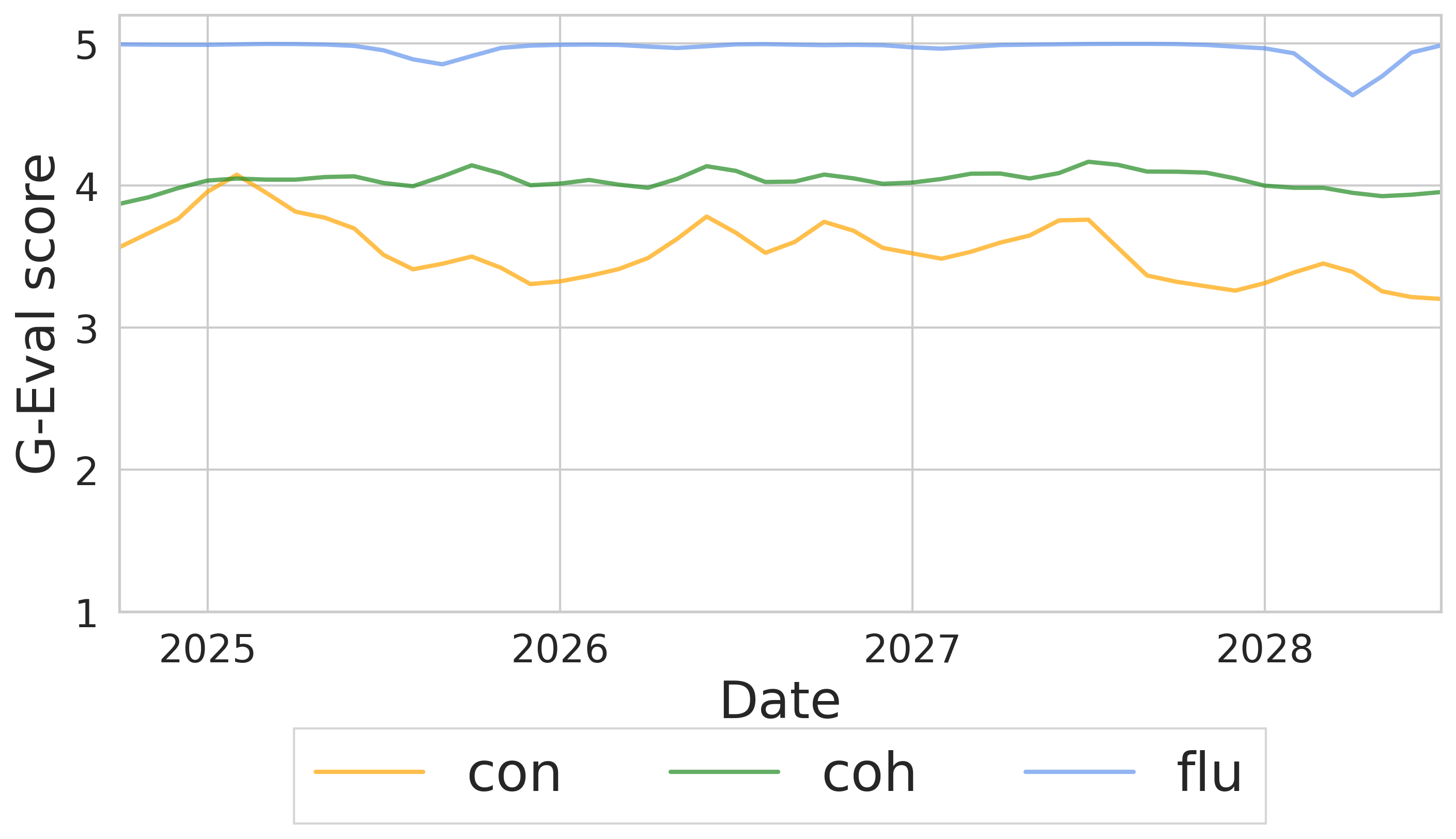}
        \caption{\gpt -- Synt (2024-2029)}
    \end{subfigure}\\
            \begin{subfigure}{0.32\textwidth}  
        \centering
        \includegraphics[width=\linewidth]{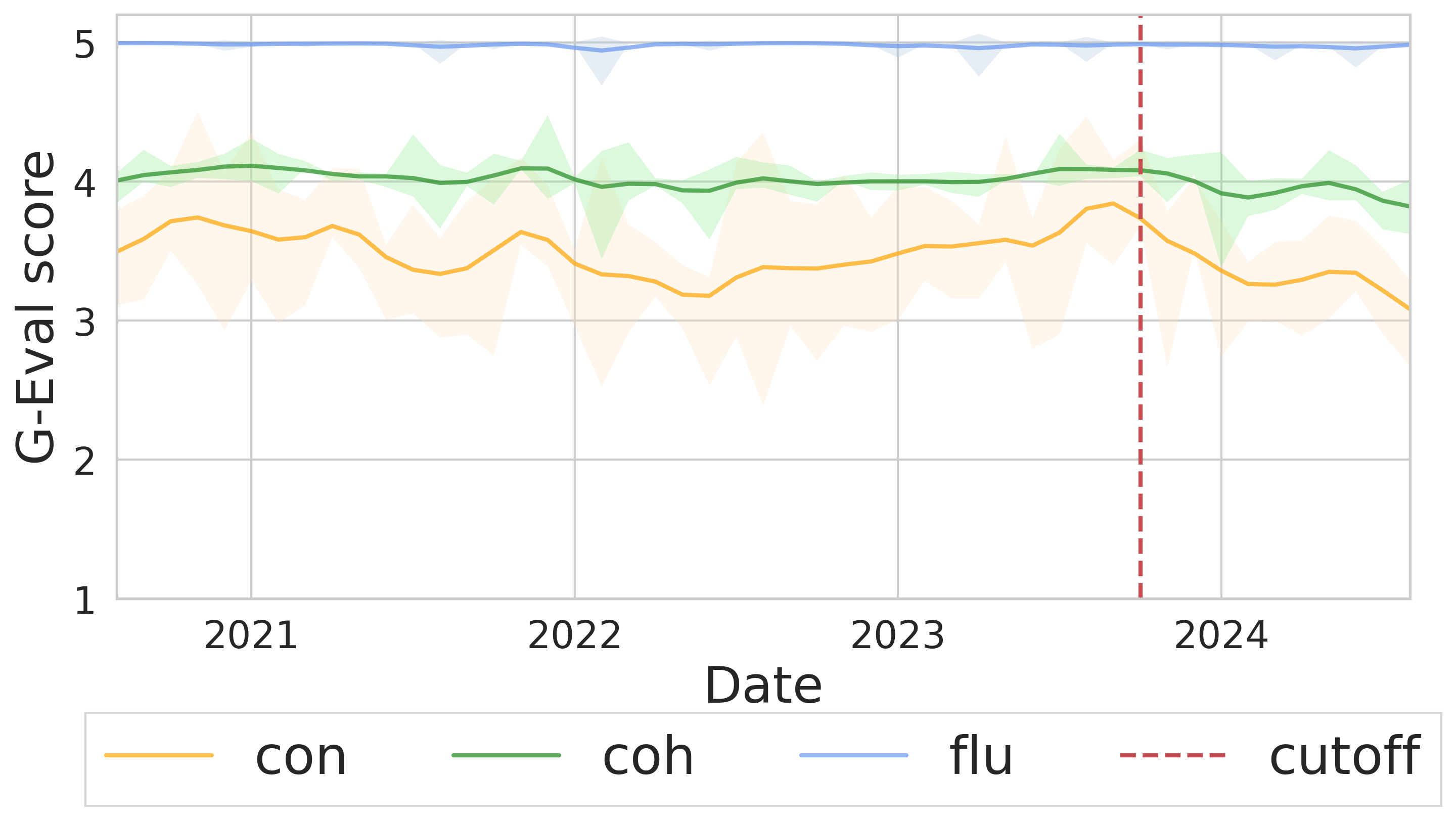}
        \caption{GPT-4o-mini -- real}
    \end{subfigure}
    \begin{subfigure}{0.32\textwidth}  
        \centering
        \includegraphics[width=\linewidth]{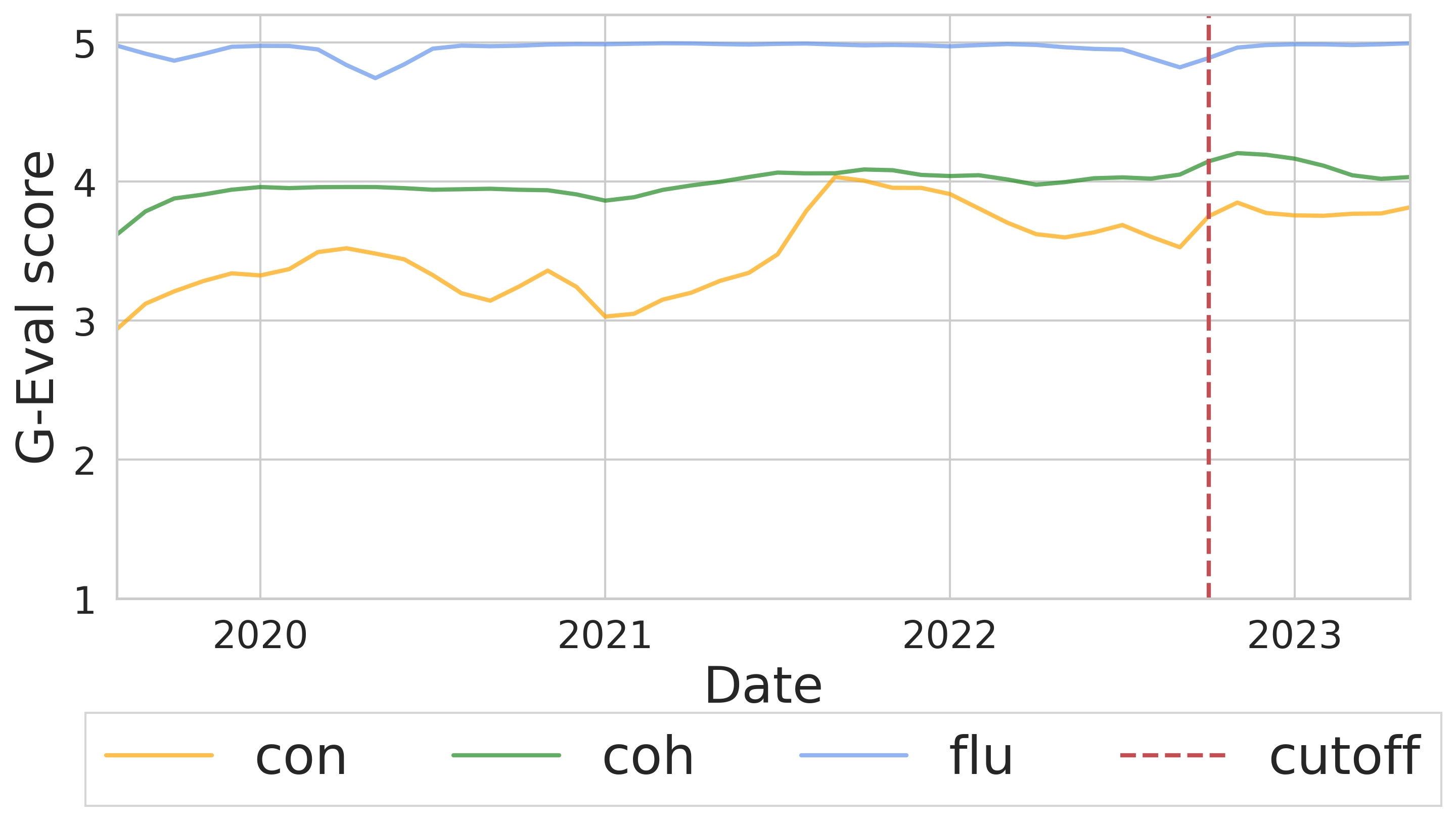}
        \caption{\gptm -- Synt (2019-2024)}
    \end{subfigure}
    \begin{subfigure}{0.32\textwidth}  
        \centering
        \includegraphics[width=\linewidth]{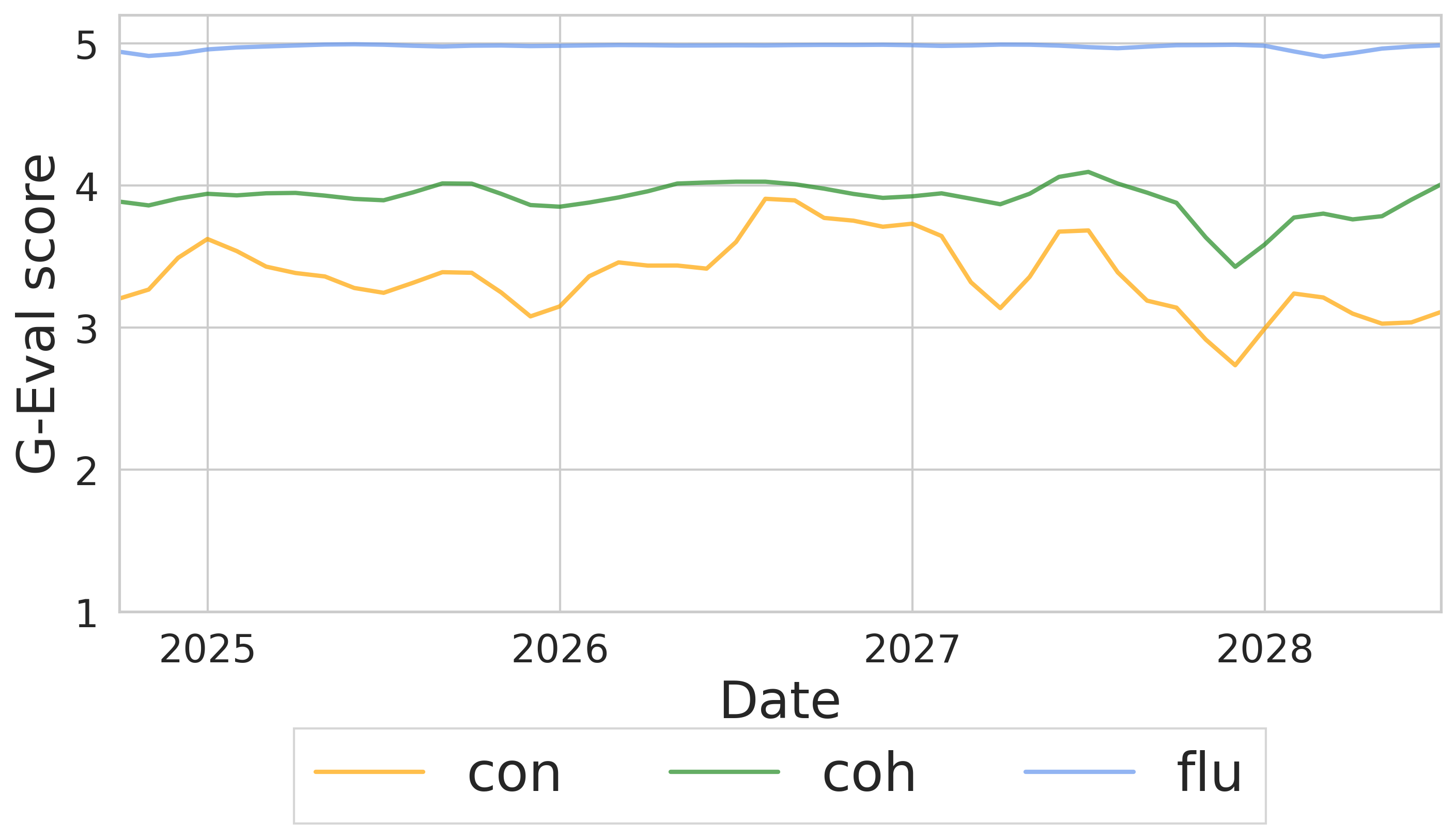}
        \caption{\gptm -- Synt (2024-2029)}
    \end{subfigure}\\
    \begin{subfigure}{0.32\textwidth}  
        \centering
        \includegraphics[width=\linewidth]{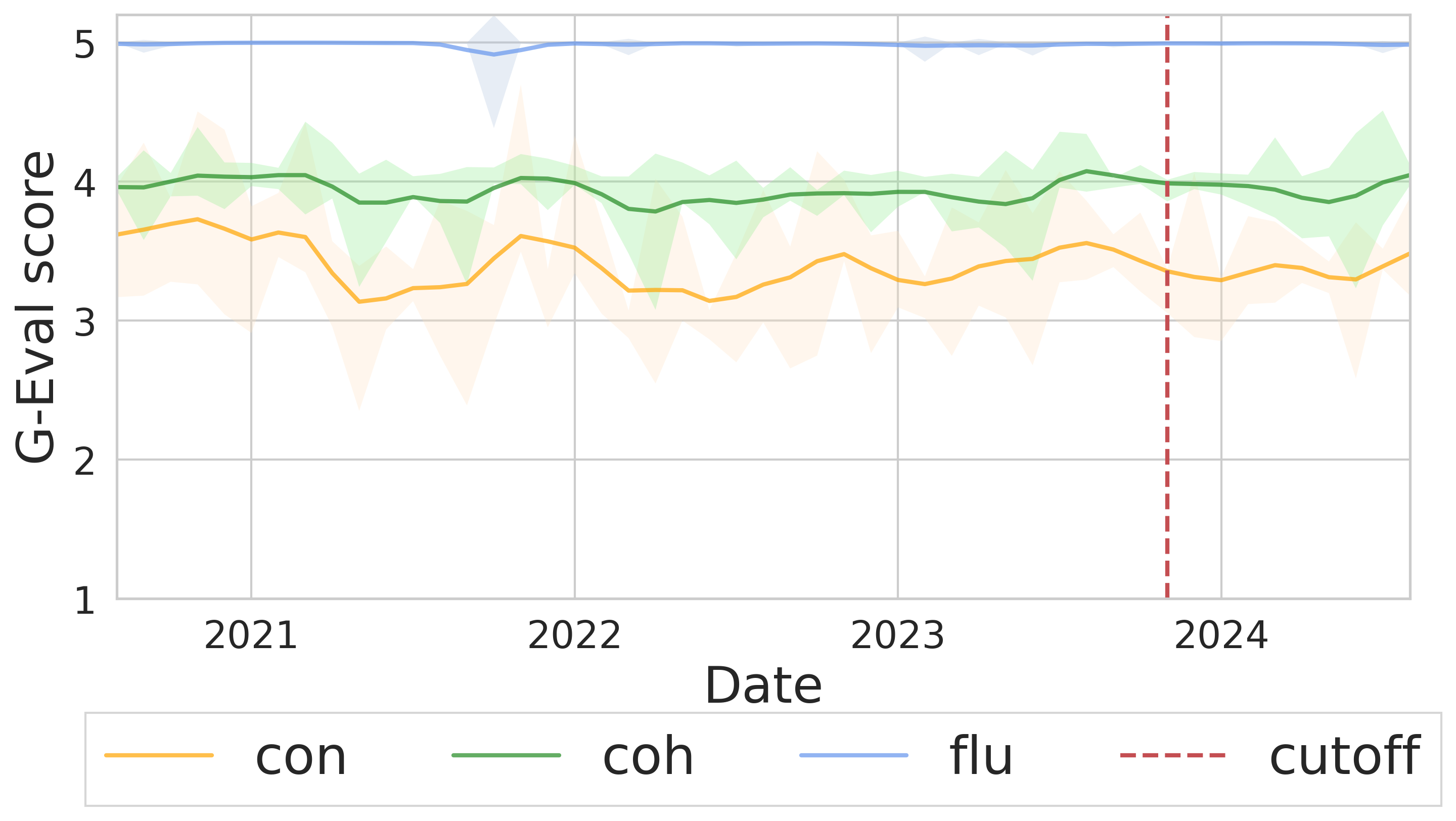}
        \caption{Gemini -- real}
    \end{subfigure}
    \begin{subfigure}{0.32\textwidth}  
        \centering
        \includegraphics[width=\linewidth]{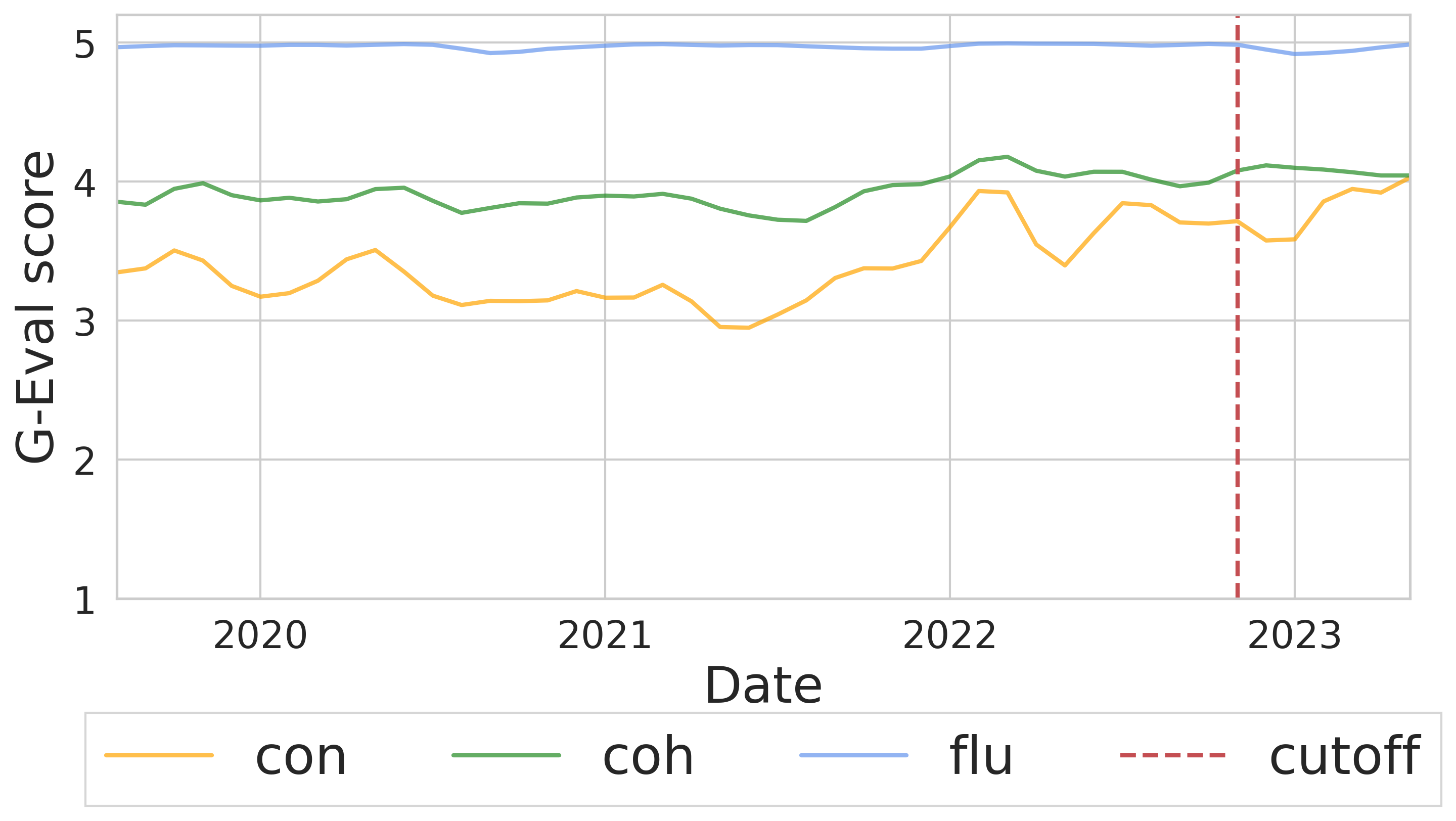}
        \caption{Gemini -- Synt (2019-2024)}
    \end{subfigure}
    \begin{subfigure}{0.32\textwidth}  
        \centering
        \includegraphics[width=\linewidth]{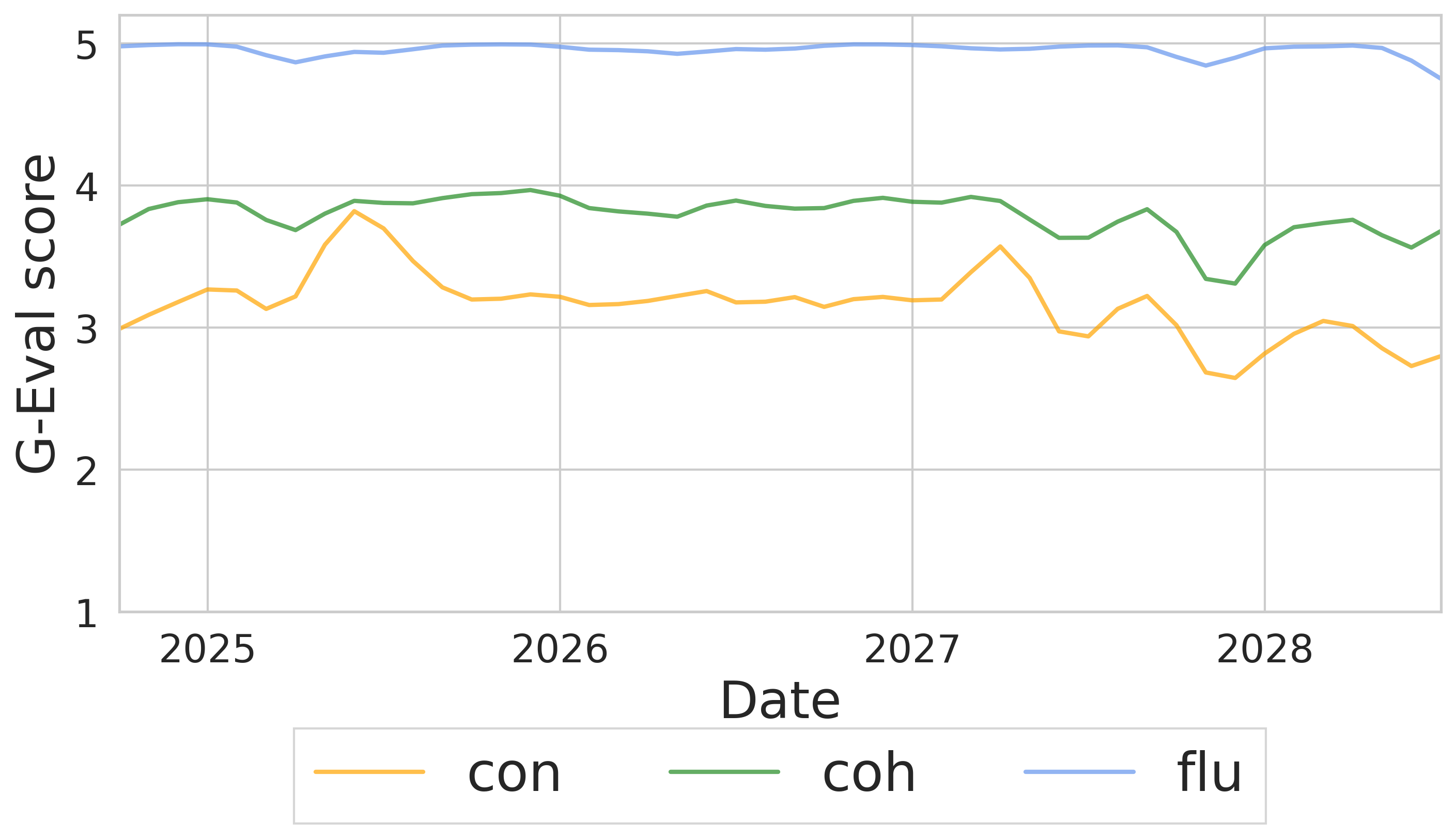}
        \caption{Gemini -- Synt (2024-2029)}
    \end{subfigure}\\
    
        \begin{subfigure}{0.32\textwidth}  
        \centering
        \includegraphics[width=\linewidth]{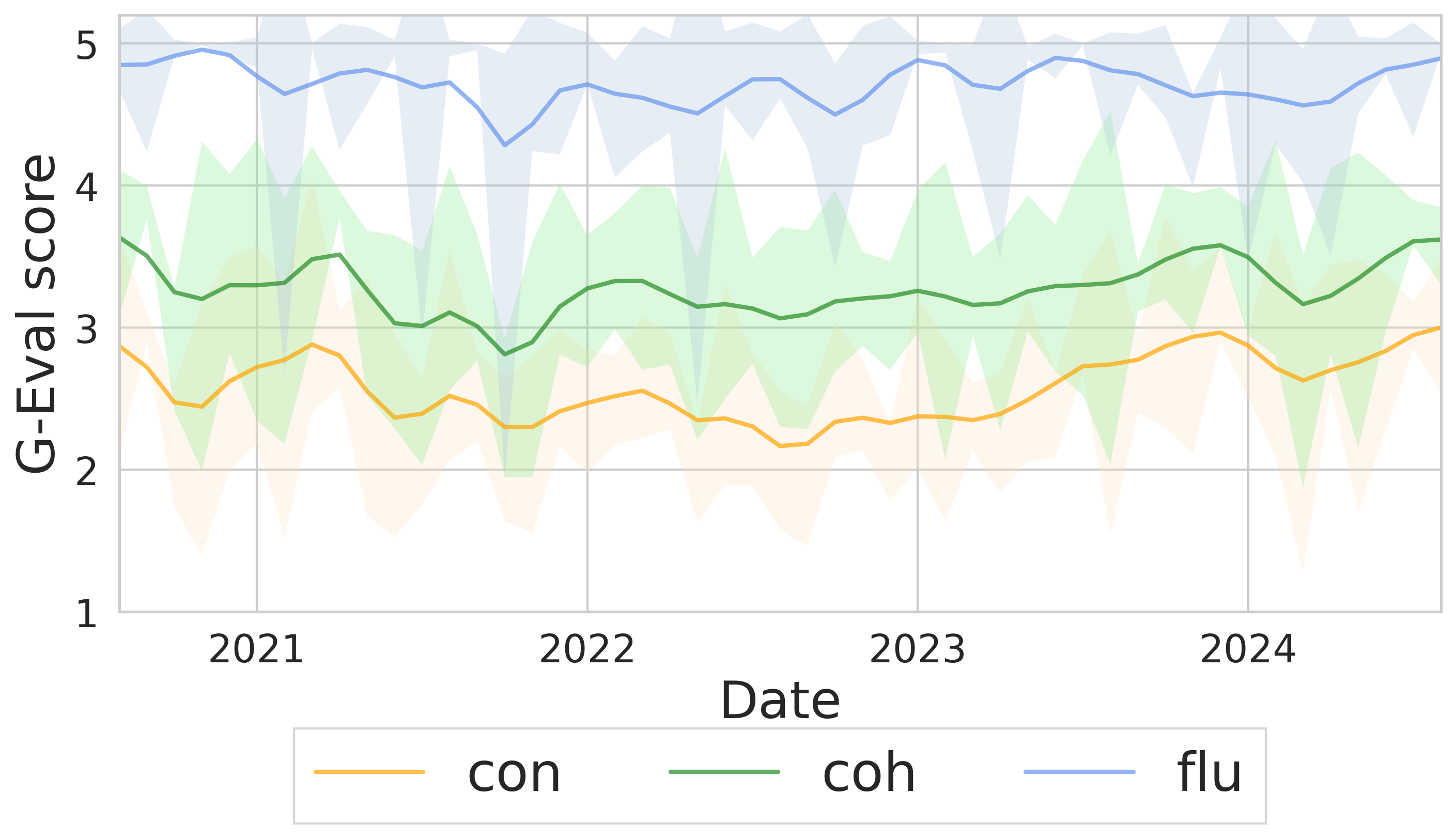}
        \caption{\llama -- real}
    \end{subfigure}
    \begin{subfigure}{0.32\textwidth}  
        \centering
        \includegraphics[width=\linewidth]{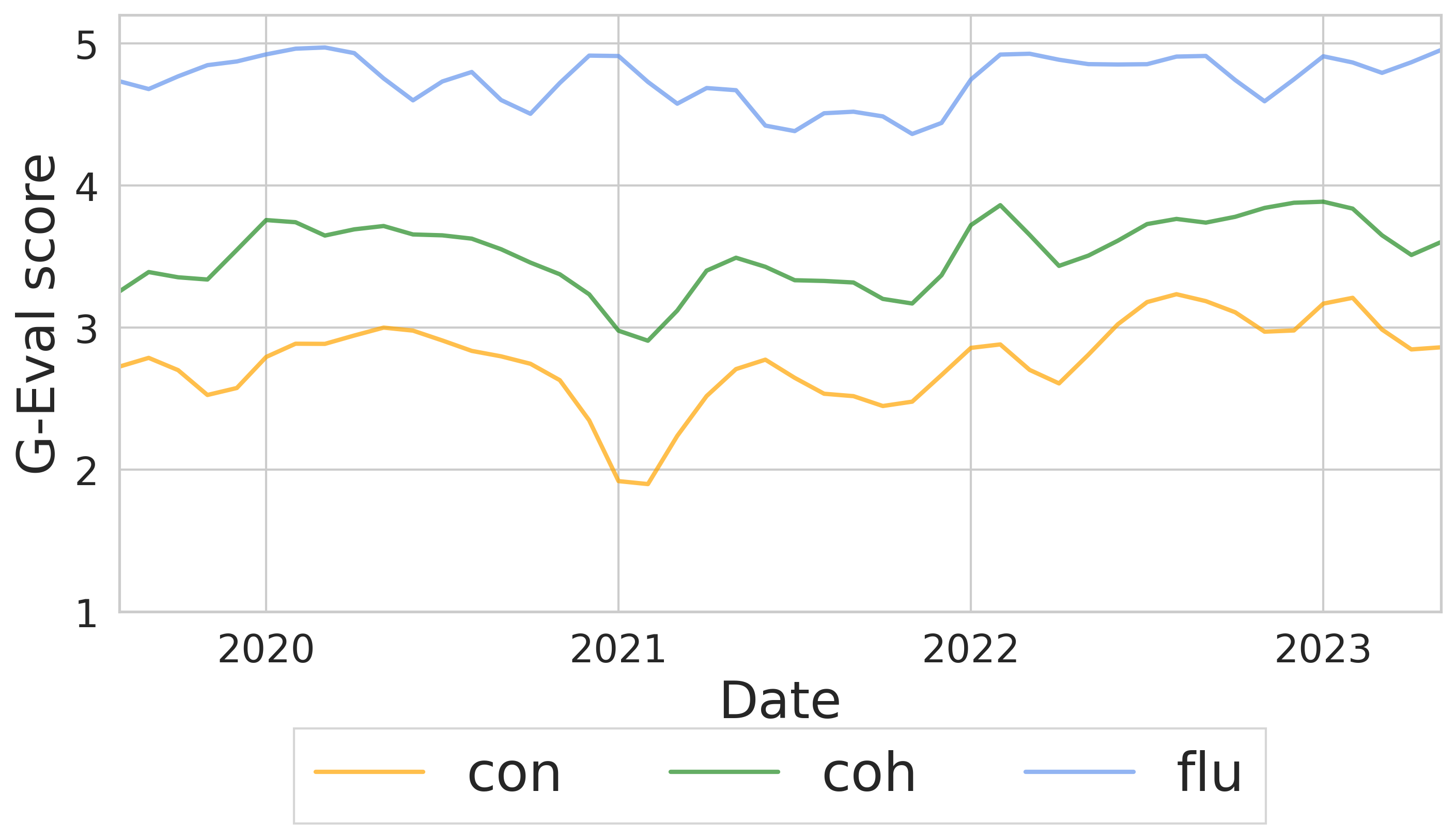}
        \caption{\llama -- Synt (2019-2024)}
    \end{subfigure}
    \begin{subfigure}{0.32\textwidth}  
        \centering
        \includegraphics[width=\linewidth]{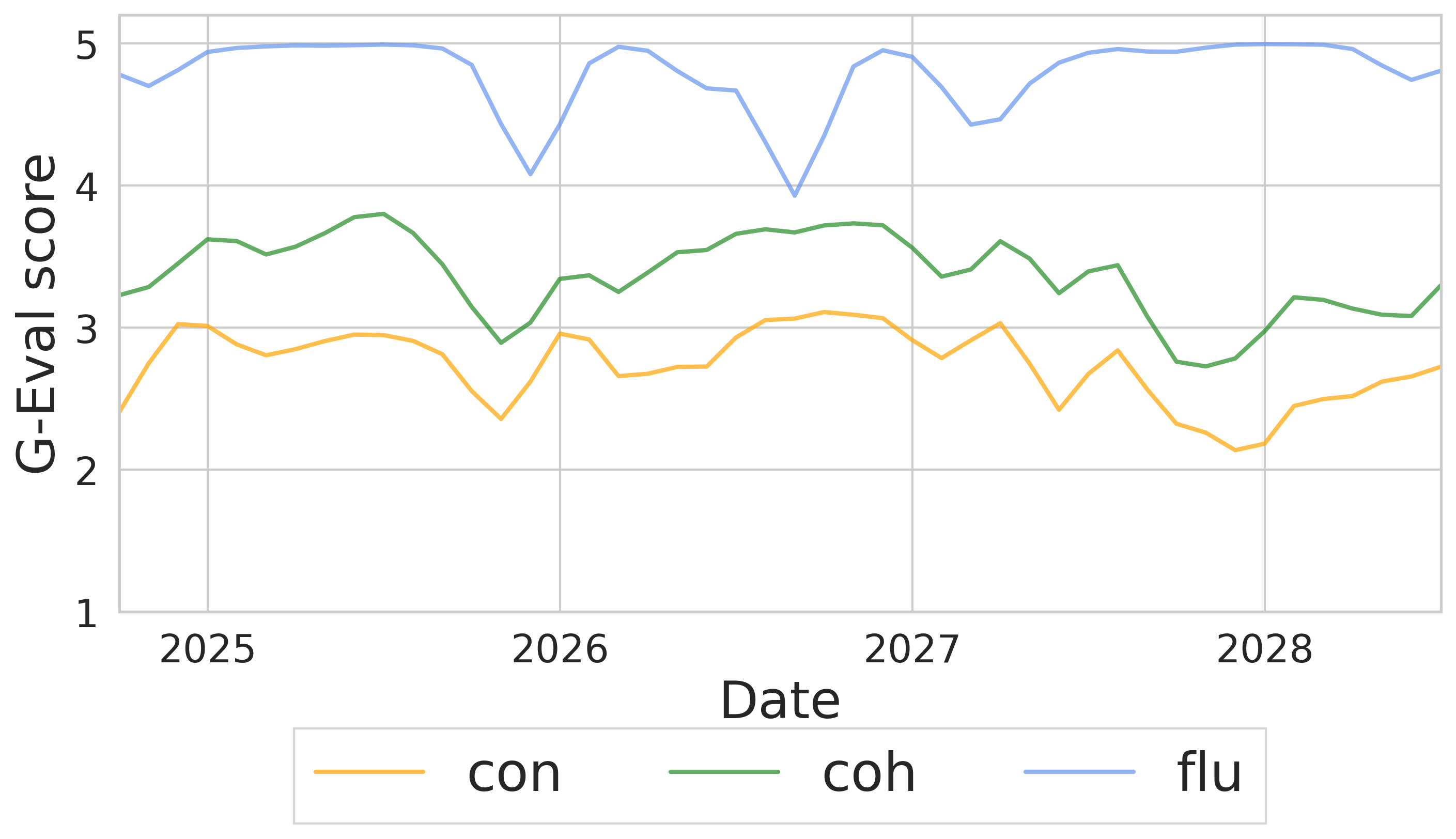}
        \caption{\llama -- Synt (2024-2029)}
    \end{subfigure}\\
    
        \begin{subfigure}{0.32\textwidth}  
        \centering
        \includegraphics[width=\linewidth]{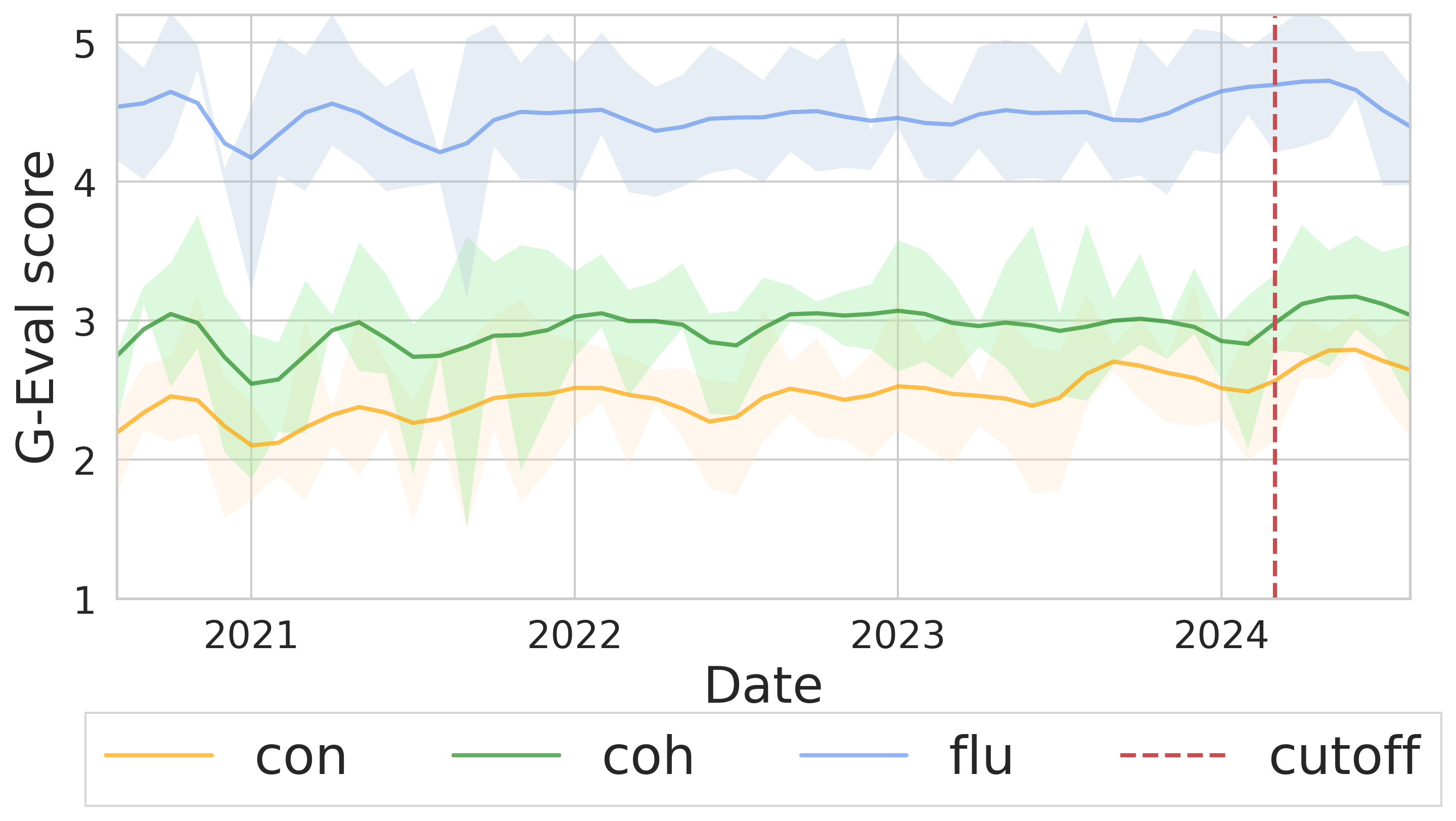}
        \caption{Phi-3 -- real}
    \end{subfigure}
    \begin{subfigure}{0.32\textwidth}  
        \centering
        \includegraphics[width=\linewidth]{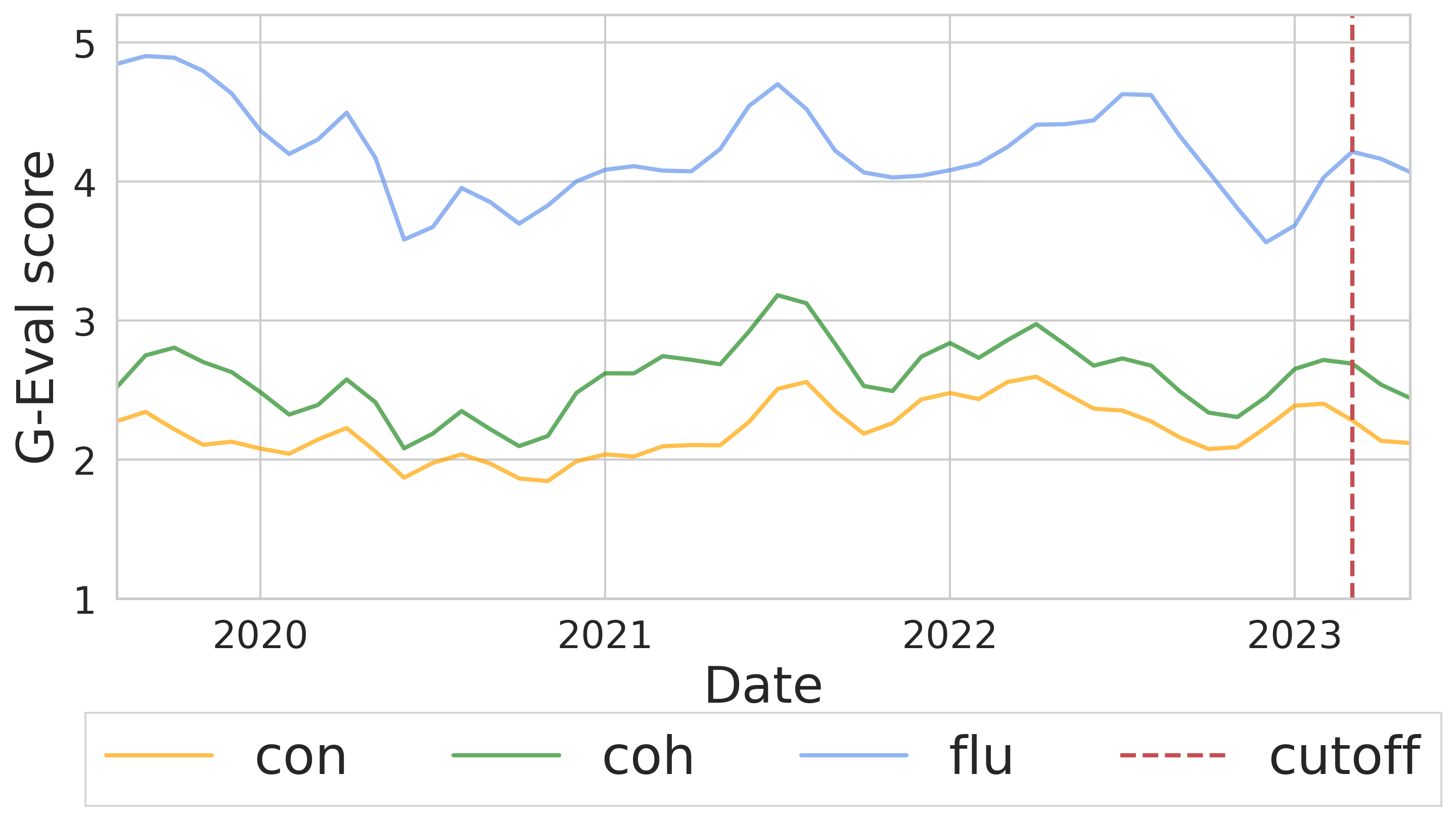}
        \caption{Phi-3 -- Synt (2019-2024)}
    \end{subfigure}
    \begin{subfigure}{0.32\textwidth}  
        \centering
        \includegraphics[width=\linewidth]{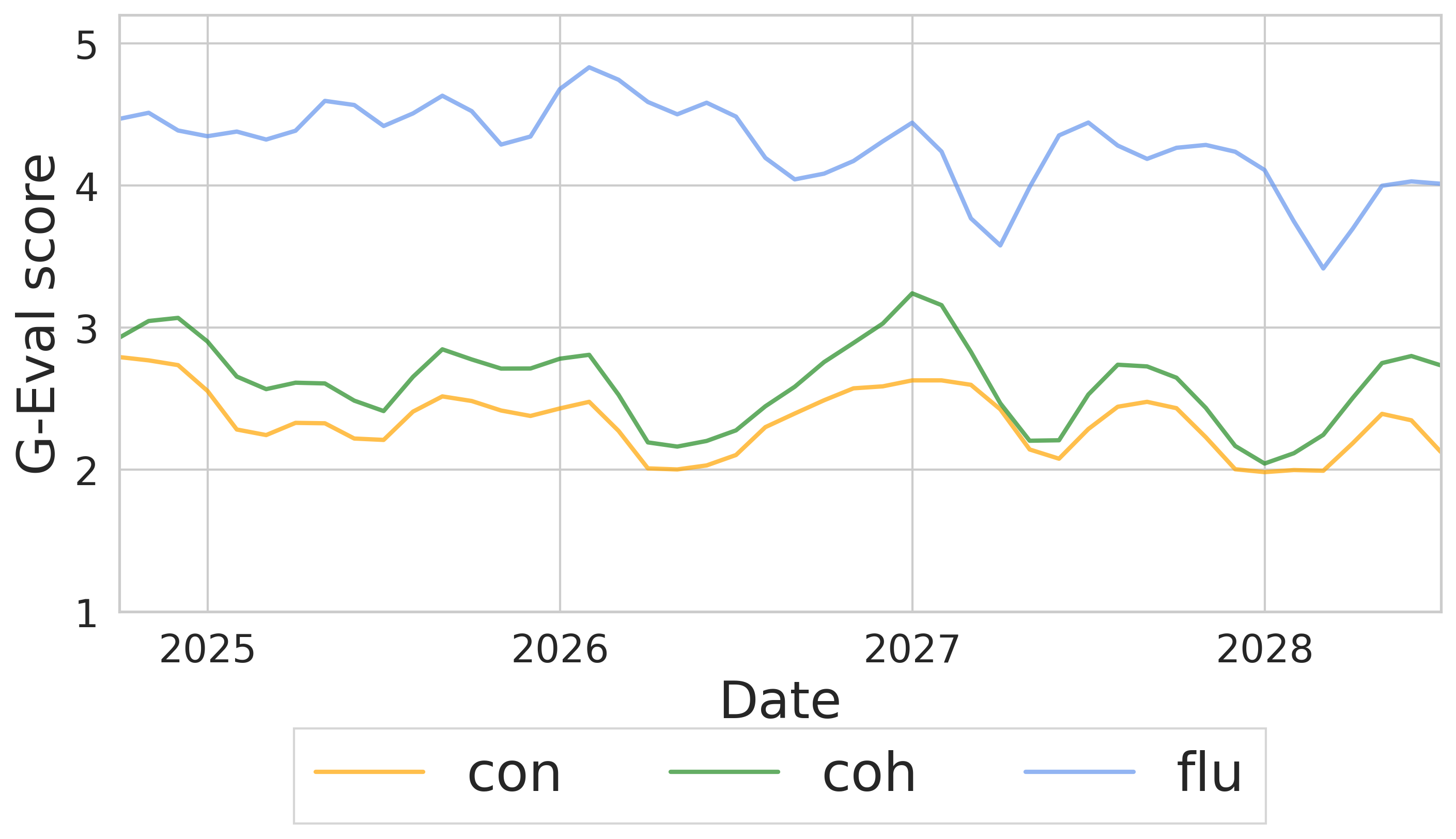}
        \caption{Phi-3 -- Synt (2024-2029)}
    \end{subfigure}\\
    \caption{Short Reports G-EVAl scores over time.}
    \label{fig:geval-time-short}
\end{figure*}

\begin{figure*}[h!]
    \centering
        \begin{subfigure}{0.32\textwidth}  
        \centering
        \includegraphics[width=\linewidth]{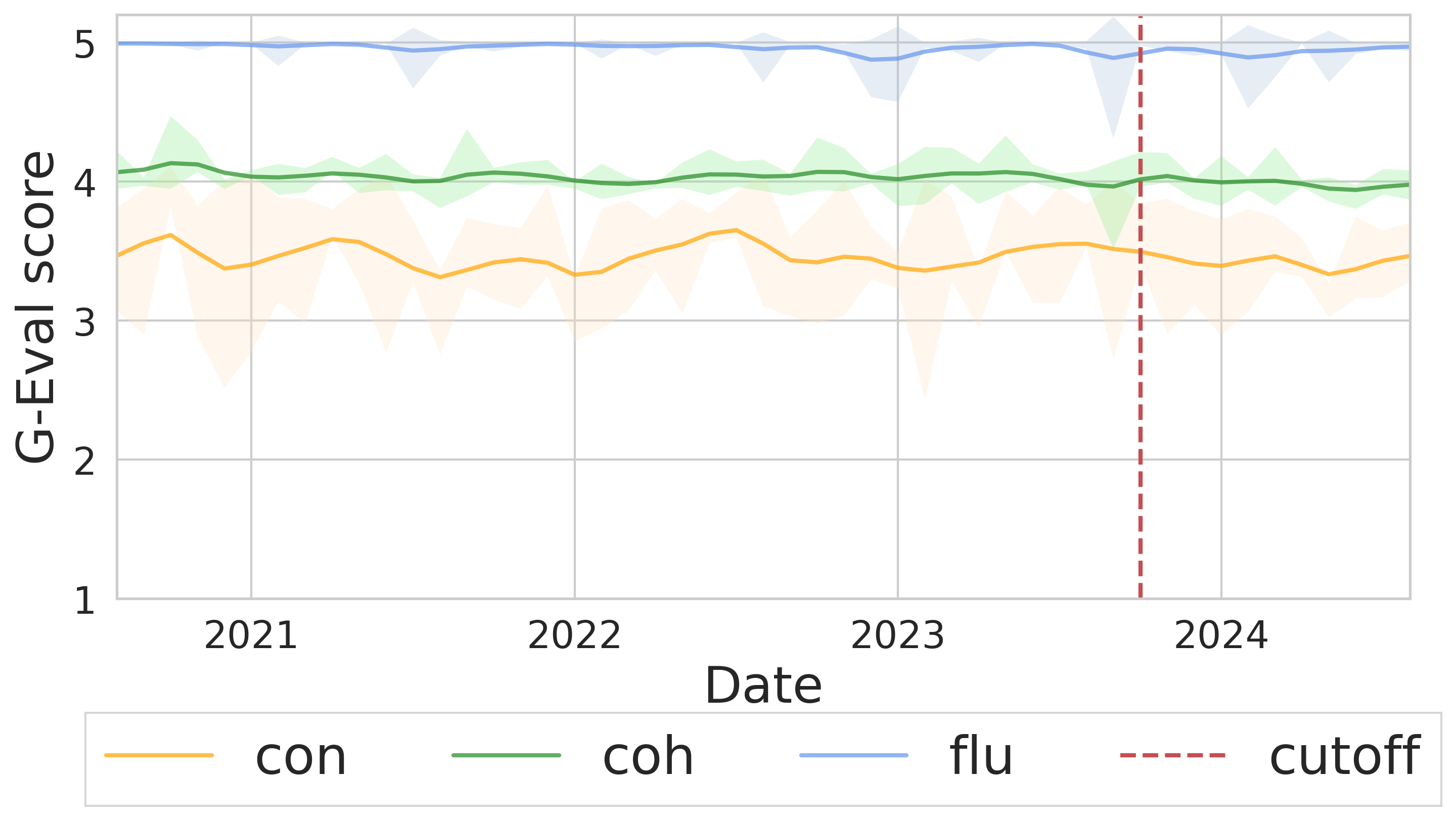}
        \caption{GPT-4o -- real}
    \end{subfigure}
    \begin{subfigure}{0.32\textwidth}  
        \centering
        \includegraphics[width=\linewidth]{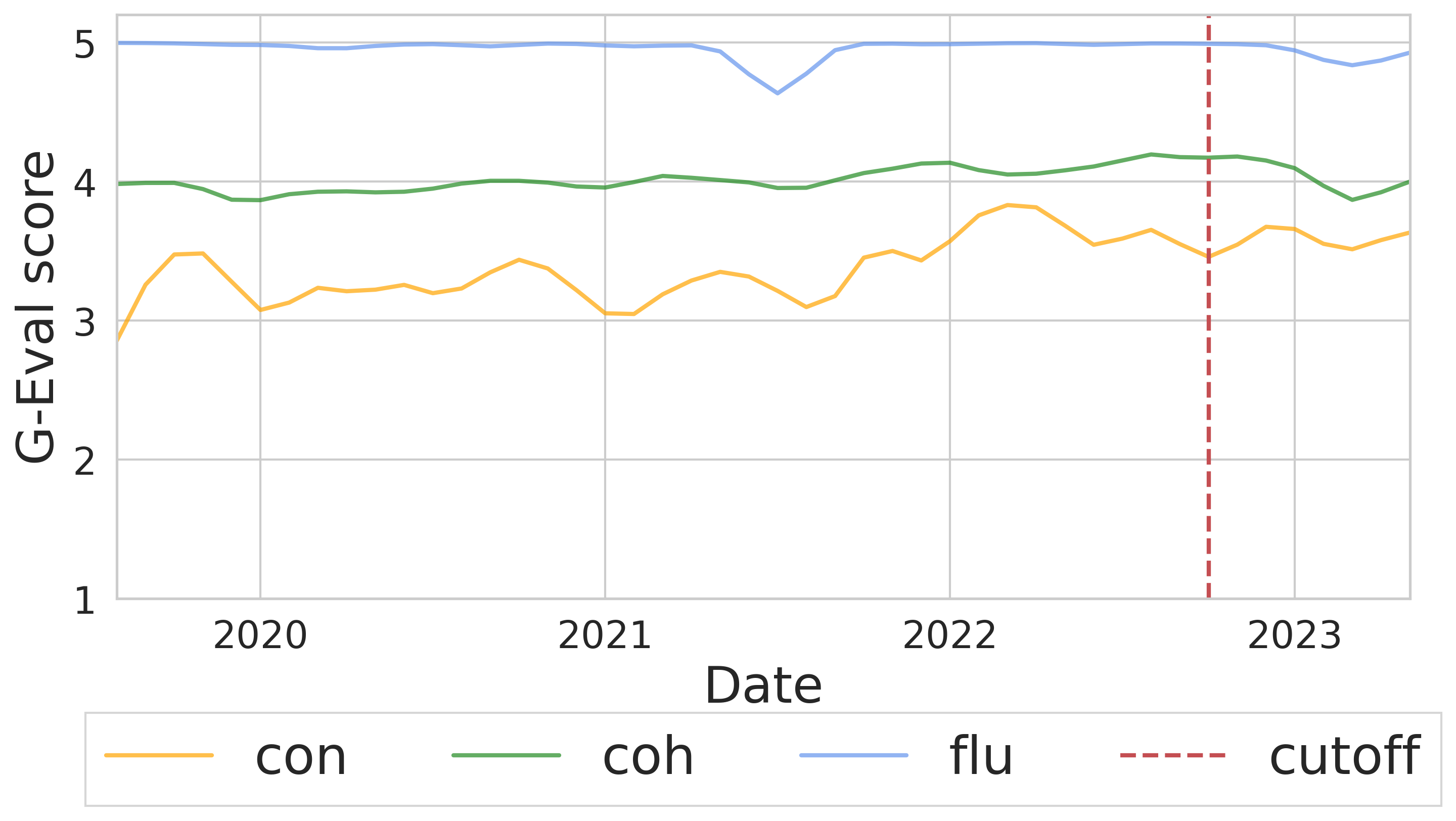}
        \caption{\gpt -- Synt (2019-2024)}
    \end{subfigure}
    \begin{subfigure}{0.32\textwidth}  
        \centering
        \includegraphics[width=\linewidth]{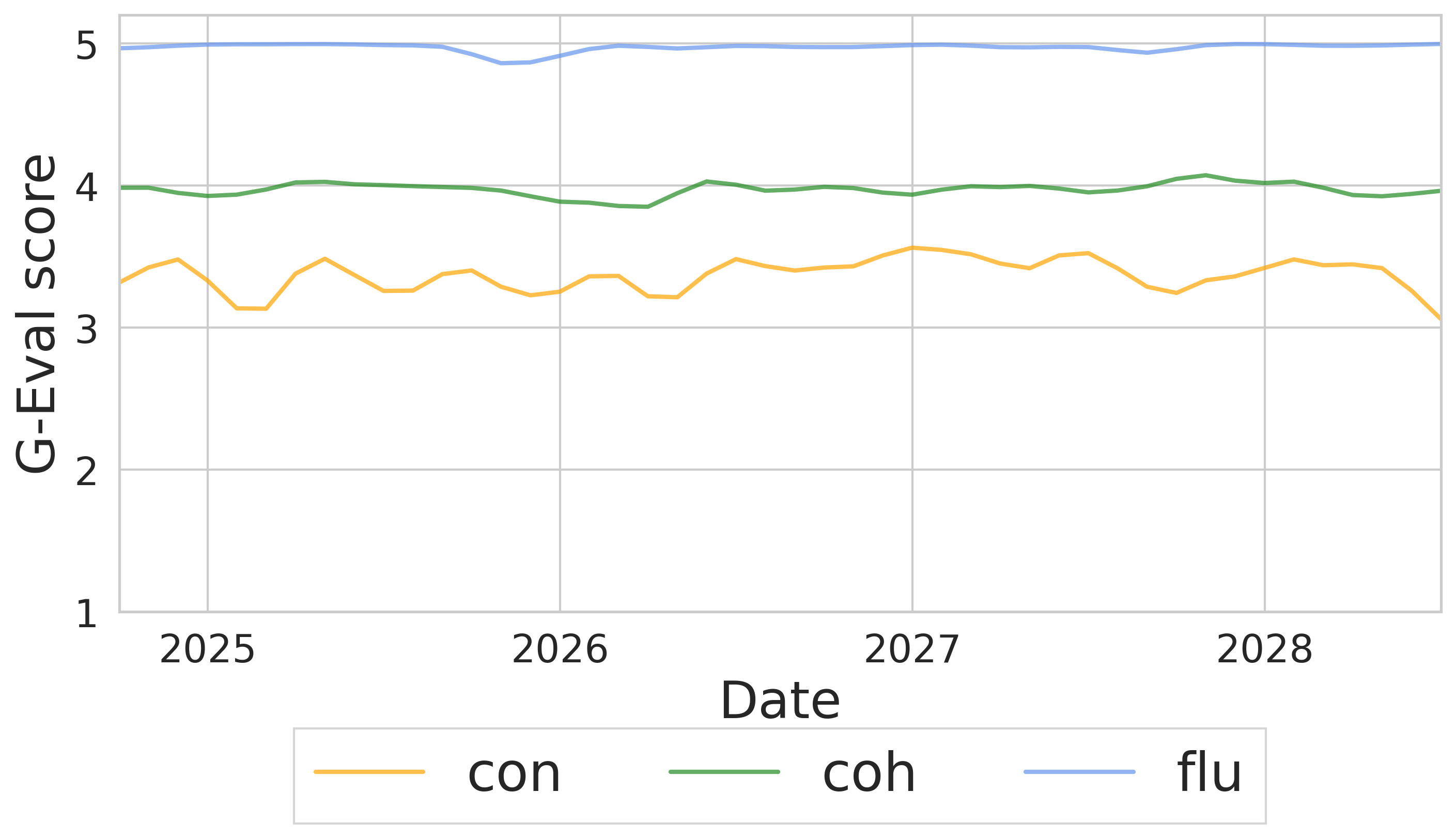}
        \caption{\gpt -- Synt (2024-2029)}
    \end{subfigure}\\
            \begin{subfigure}{0.32\textwidth}  
        \centering
        \includegraphics[width=\linewidth]{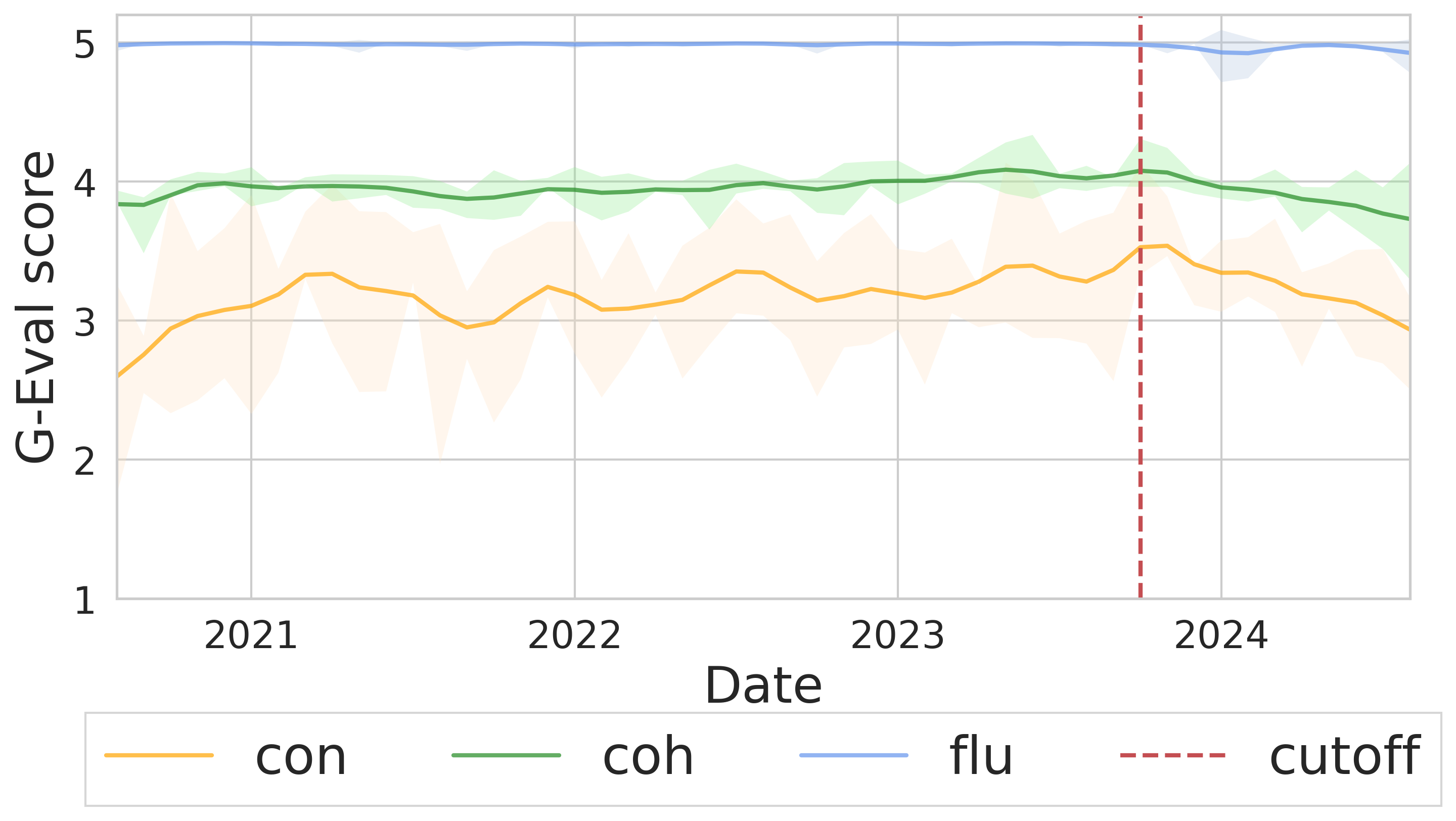}
        \caption{GPT-4o-mini -- real}
    \end{subfigure}
    \begin{subfigure}{0.32\textwidth}  
        \centering
        \includegraphics[width=\linewidth]{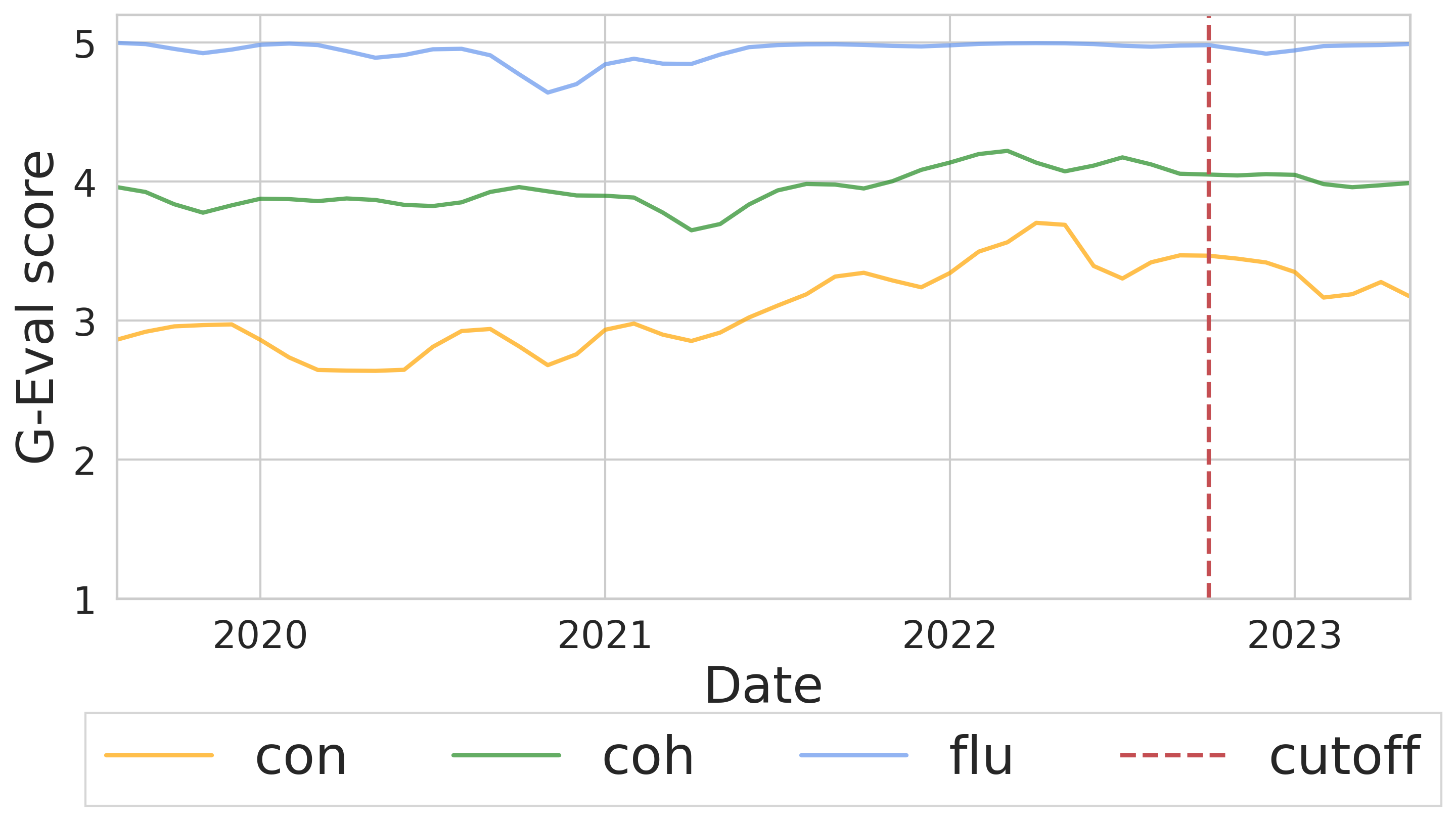}
        \caption{\gptm -- Synt (2019-2024)}
    \end{subfigure}
    \begin{subfigure}{0.32\textwidth}  
        \centering
        \includegraphics[width=\linewidth]{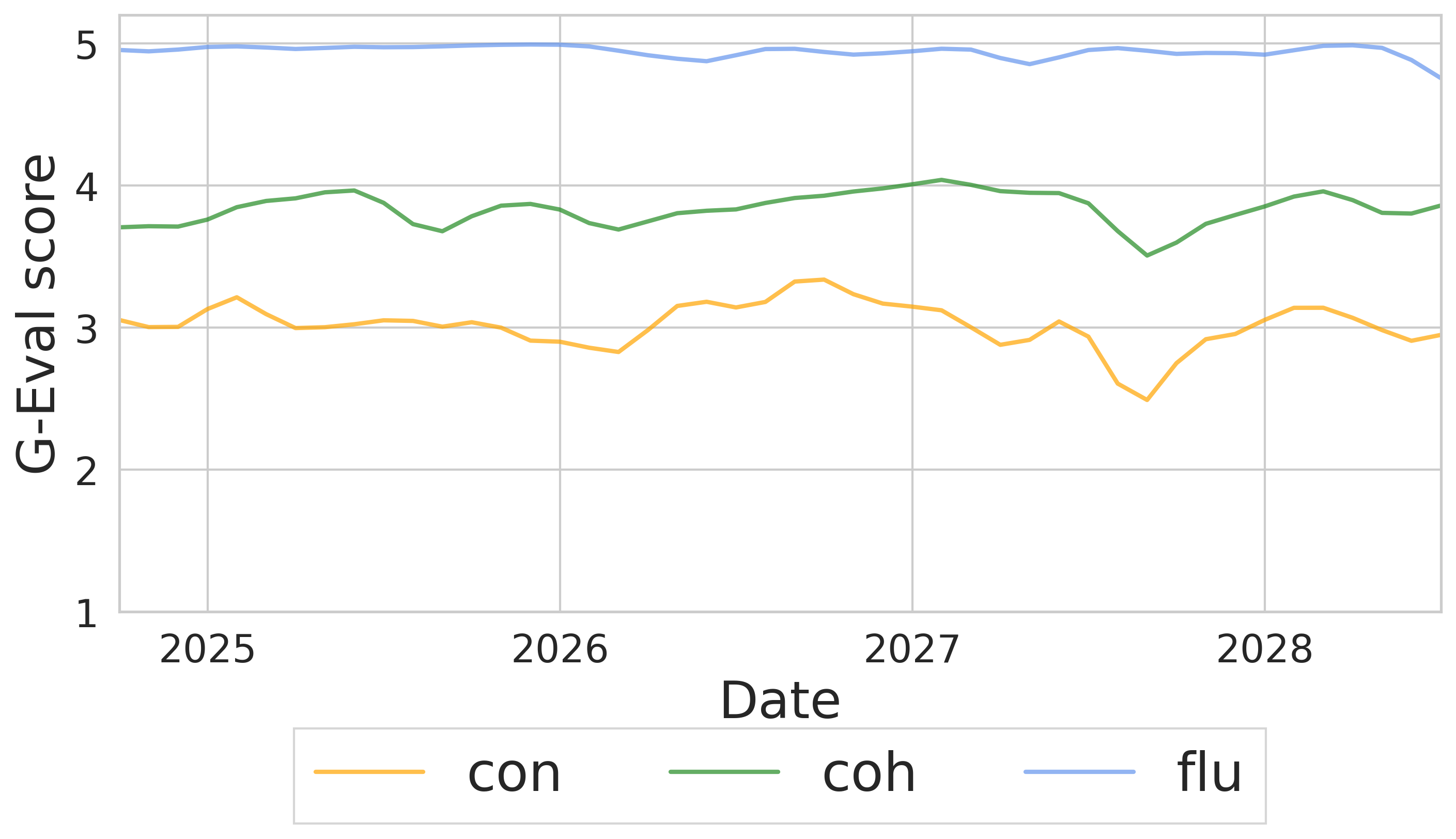}
        \caption{\gptm -- Synt (2024-2029)}
    \end{subfigure}\\
    \begin{subfigure}{0.32\textwidth}  
        \centering
        \includegraphics[width=\linewidth]{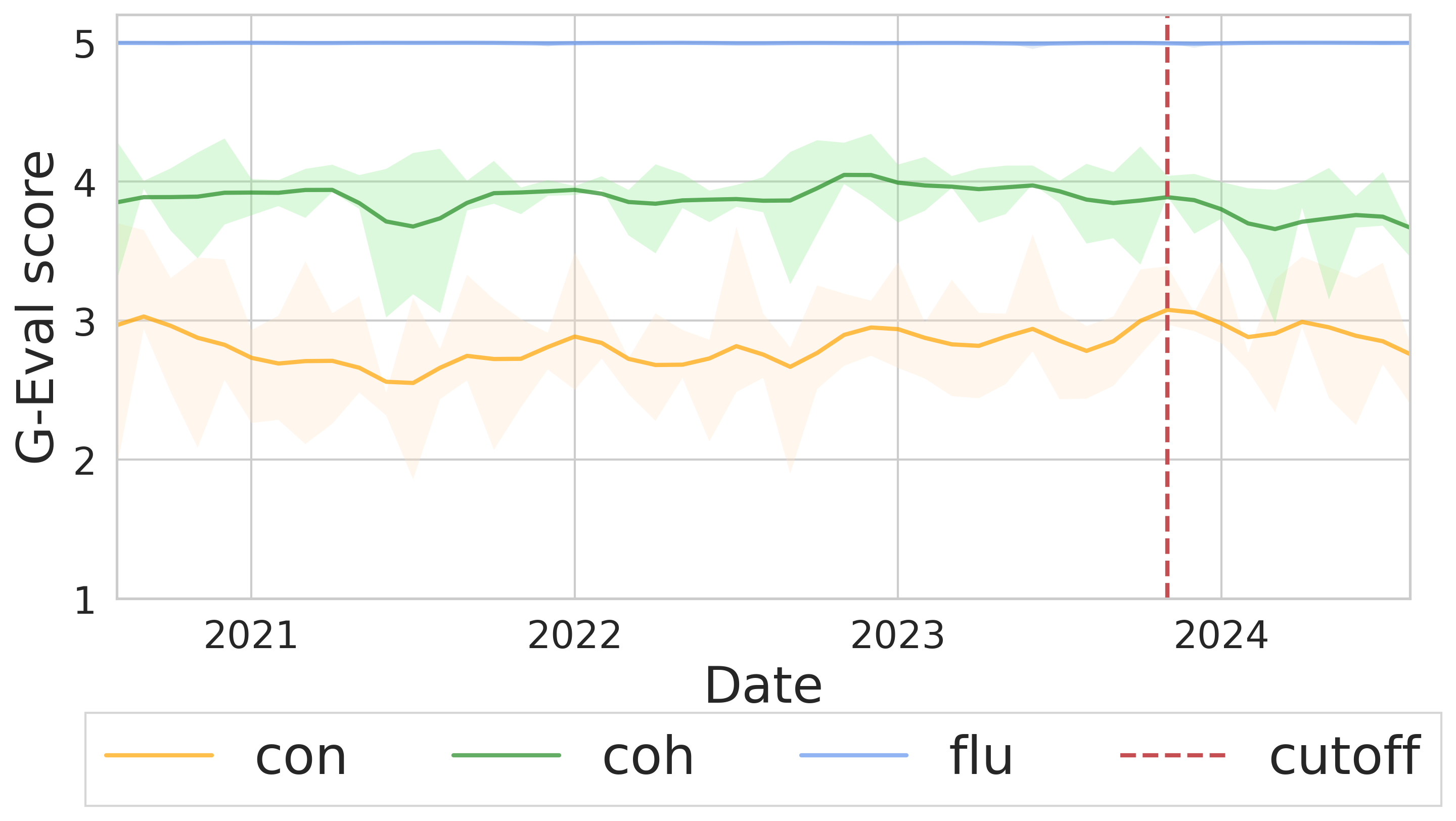}
        \caption{Gemini -- real}
    \end{subfigure}
    \begin{subfigure}{0.32\textwidth}  
        \centering
        \includegraphics[width=\linewidth]{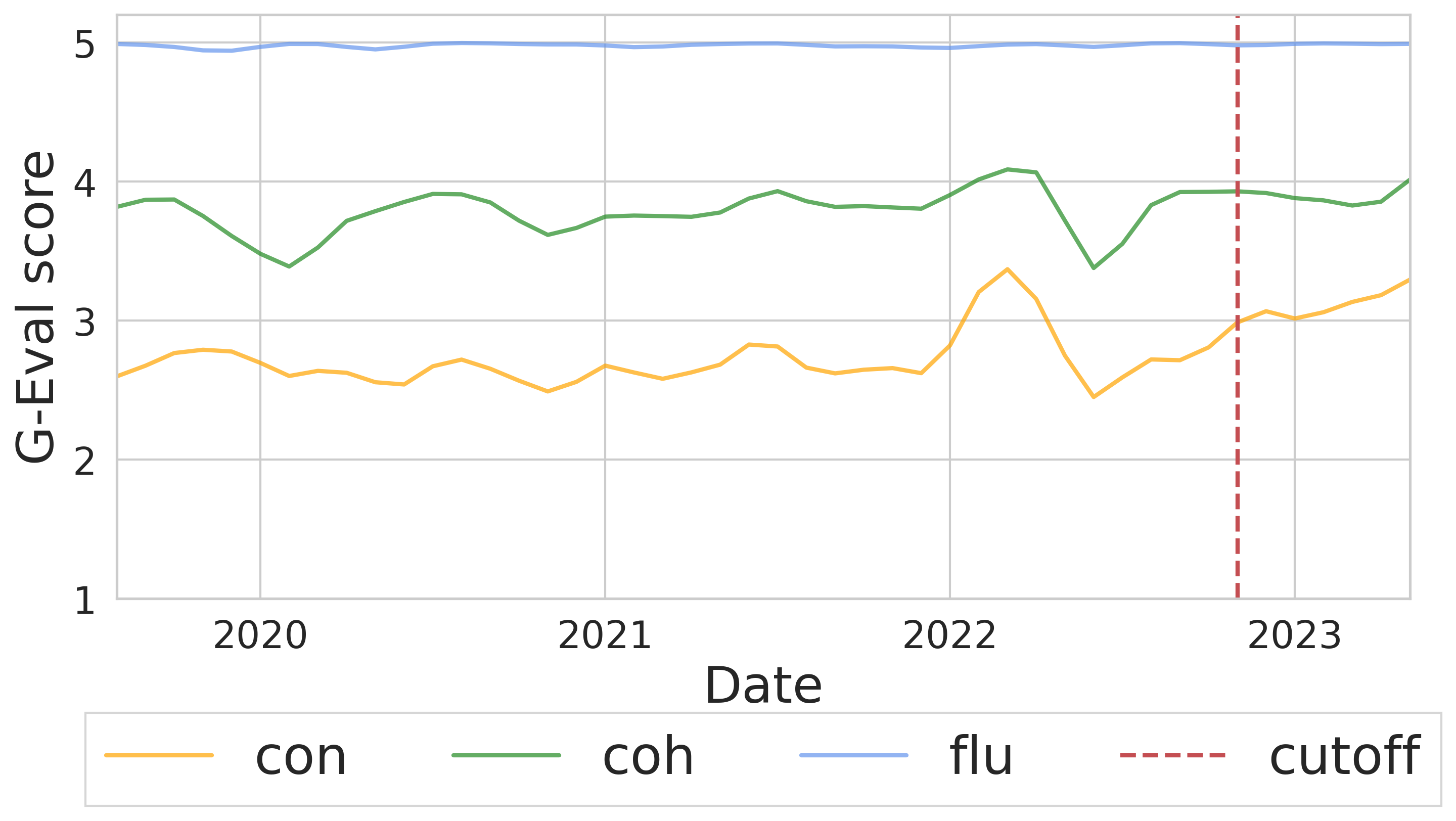}
        \caption{Gemini -- Synt (2019-2024)}
    \end{subfigure}
    \begin{subfigure}{0.32\textwidth}  
        \centering
        \includegraphics[width=\linewidth]{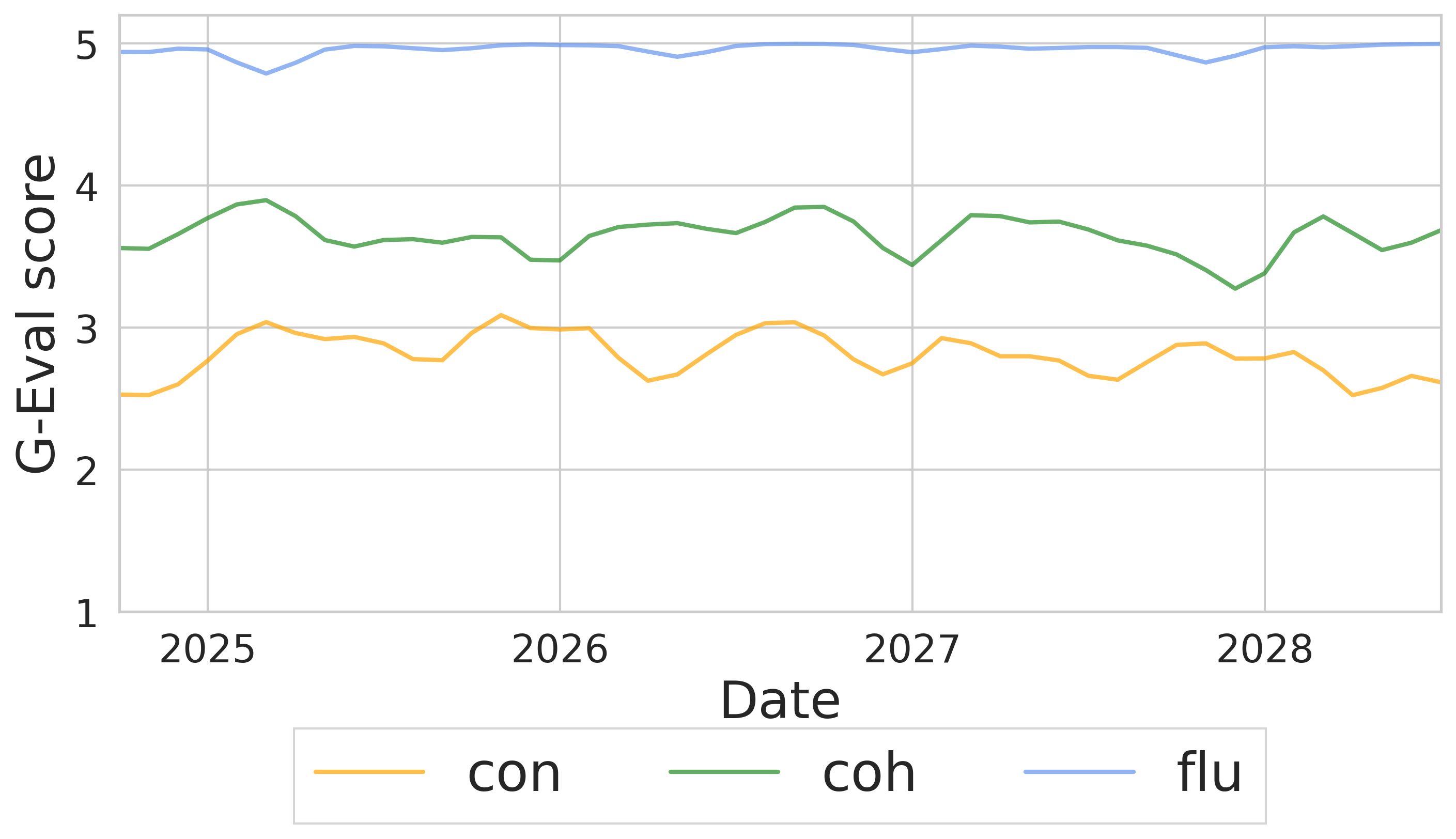}
        \caption{Gemini -- Synt (2024-2029)}
    \end{subfigure}\\
    
        \begin{subfigure}{0.32\textwidth}  
        \centering
        \includegraphics[width=\linewidth]{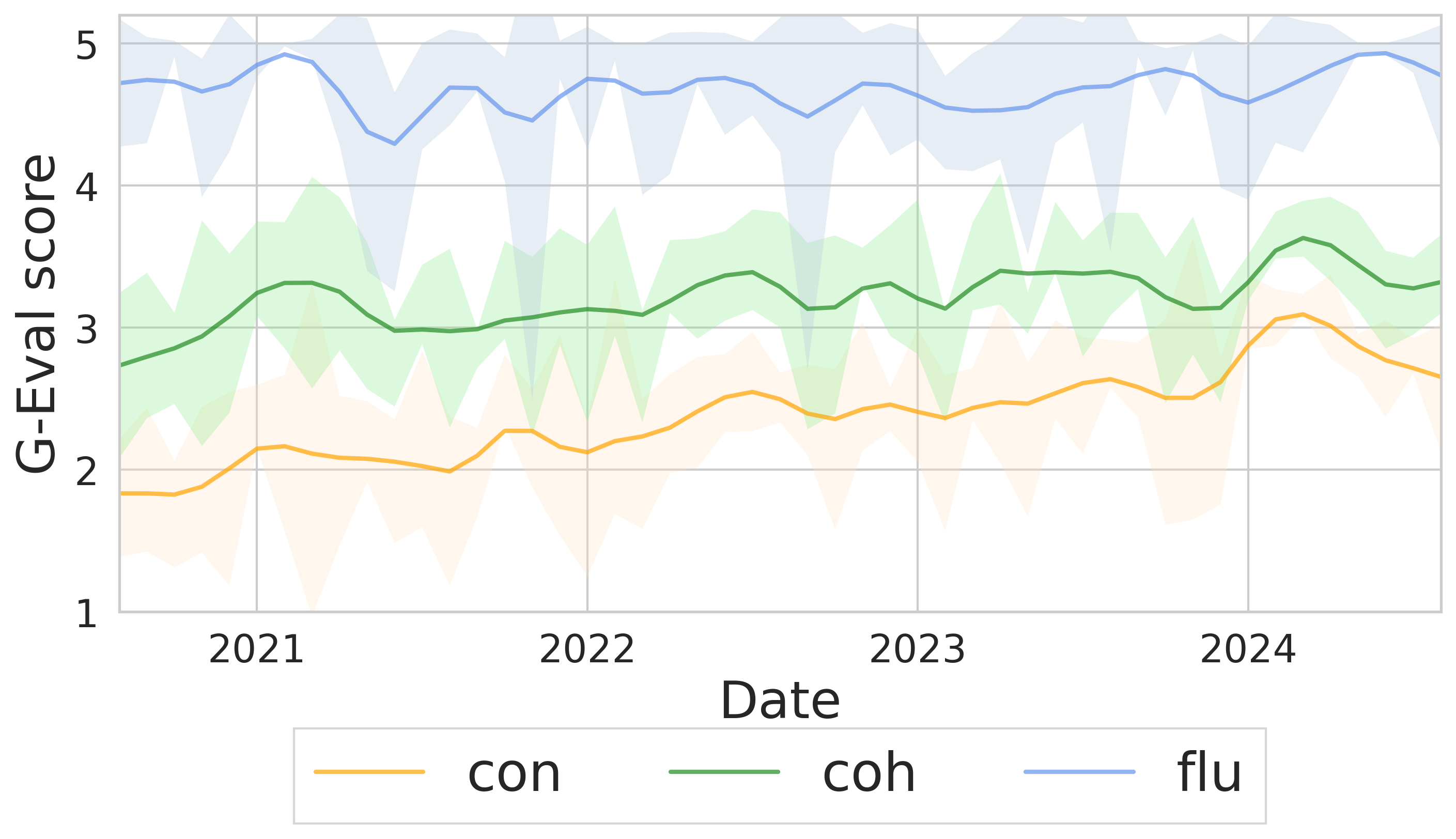}
        \caption{\llama -- real}
    \end{subfigure}
    \begin{subfigure}{0.32\textwidth}  
        \centering
        \includegraphics[width=\linewidth]{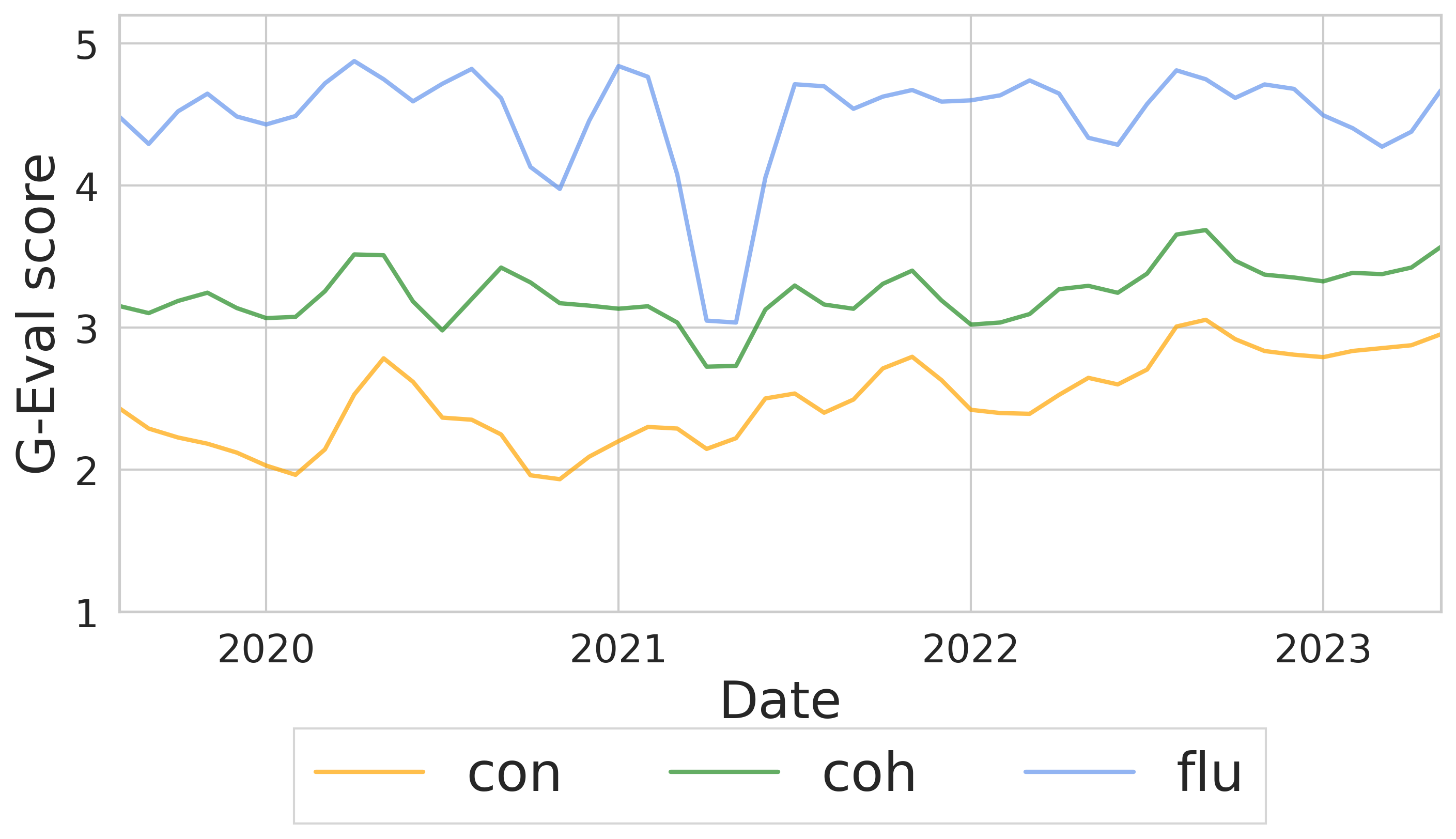}
        \caption{\llama -- Synt (2019-2024)}
    \end{subfigure}
    \begin{subfigure}{0.32\textwidth}  
        \centering
        \includegraphics[width=\linewidth]{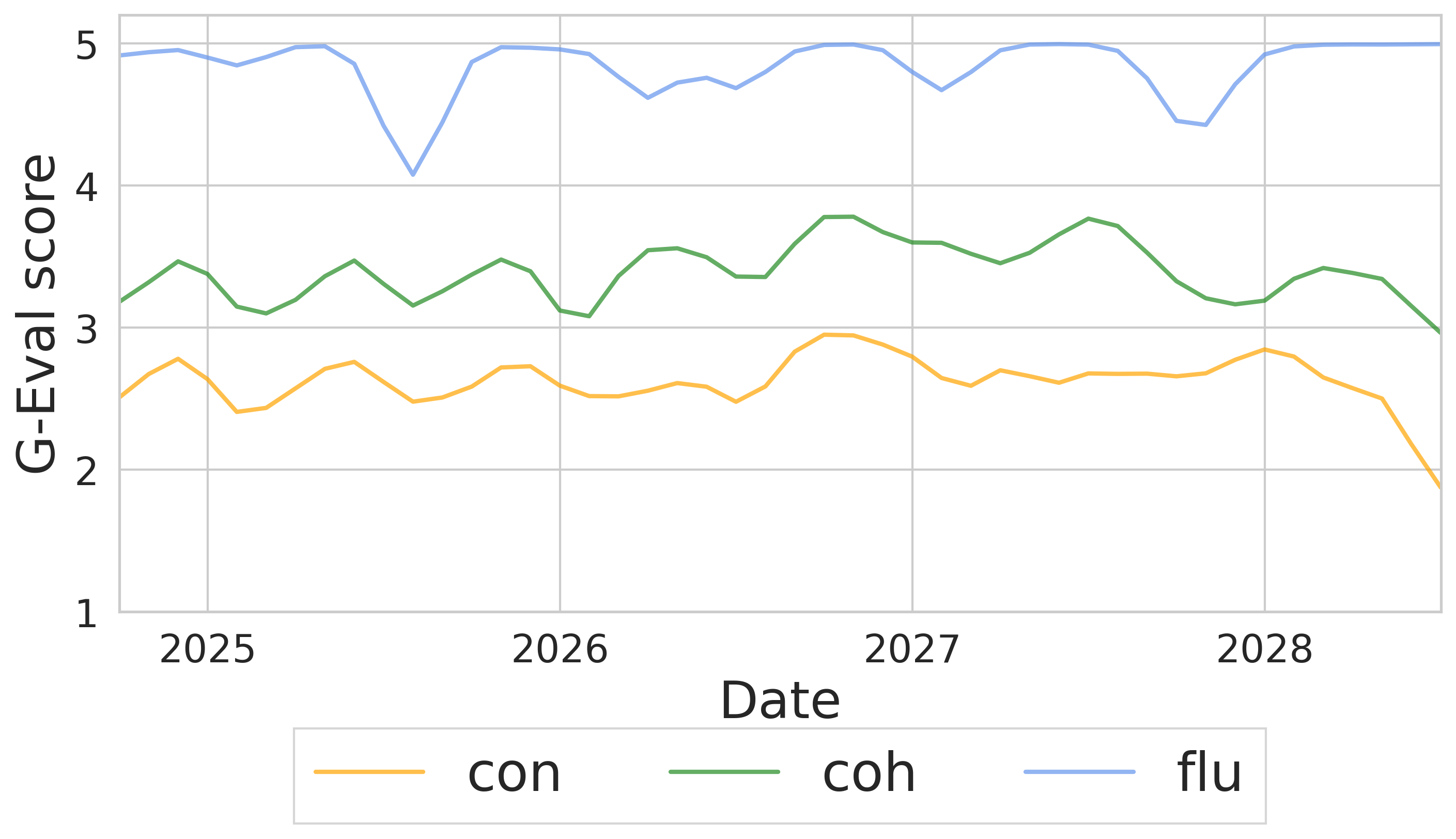}
        \caption{\llama -- Synt (2024-2029)}
    \end{subfigure}\\
    \caption{TI reports G-EVAl scores over time.}
    \label{fig:geval-time-ti}
\end{figure*}

\begin{figure*}[h!]
    \centering
        \begin{subfigure}{0.32\textwidth}  
        \centering
        \includegraphics[width=\linewidth]{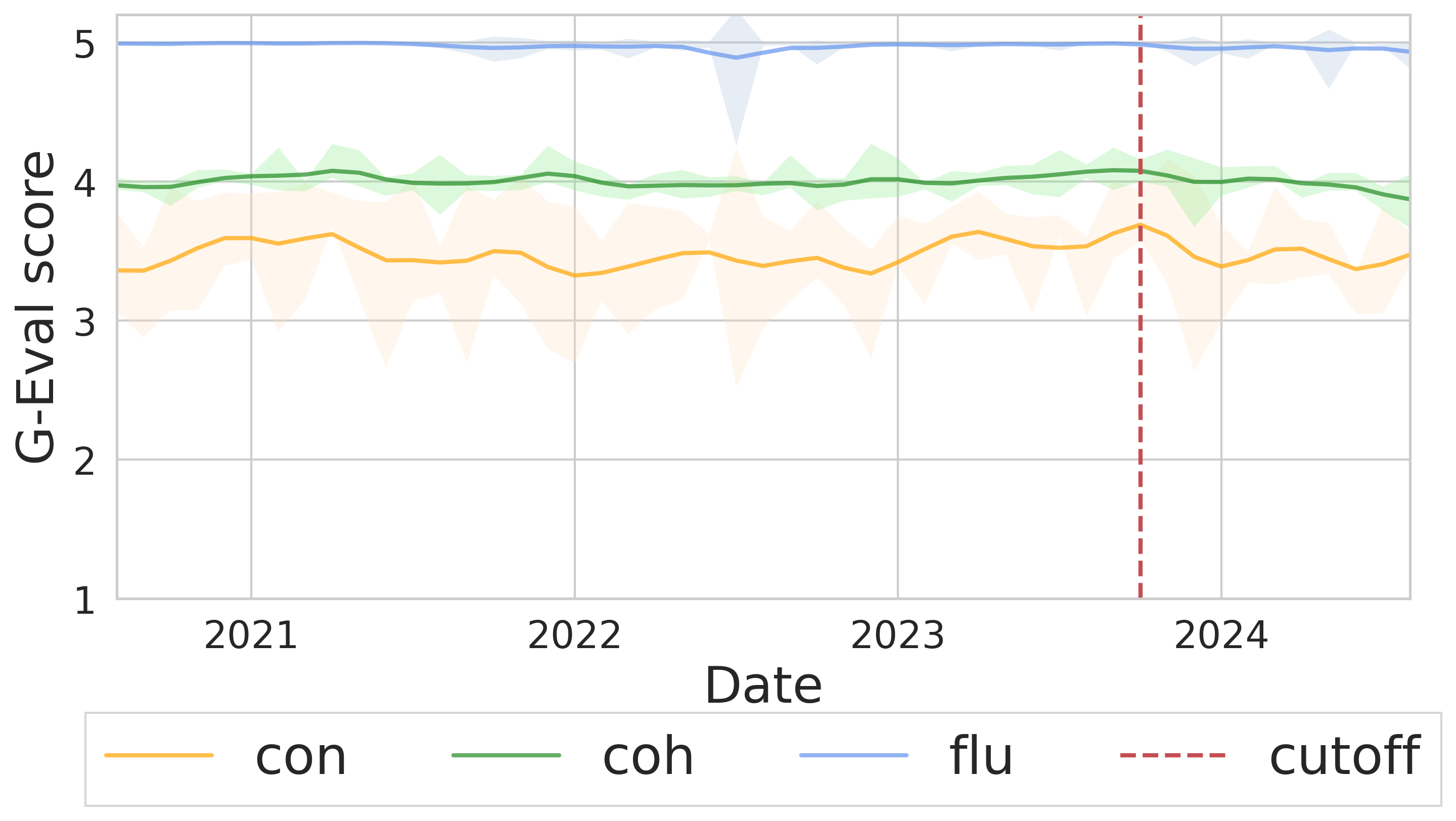}
        \caption{GPT-4o -- real}
    \end{subfigure}
    \begin{subfigure}{0.32\textwidth}  
        \centering
        \includegraphics[width=\linewidth]{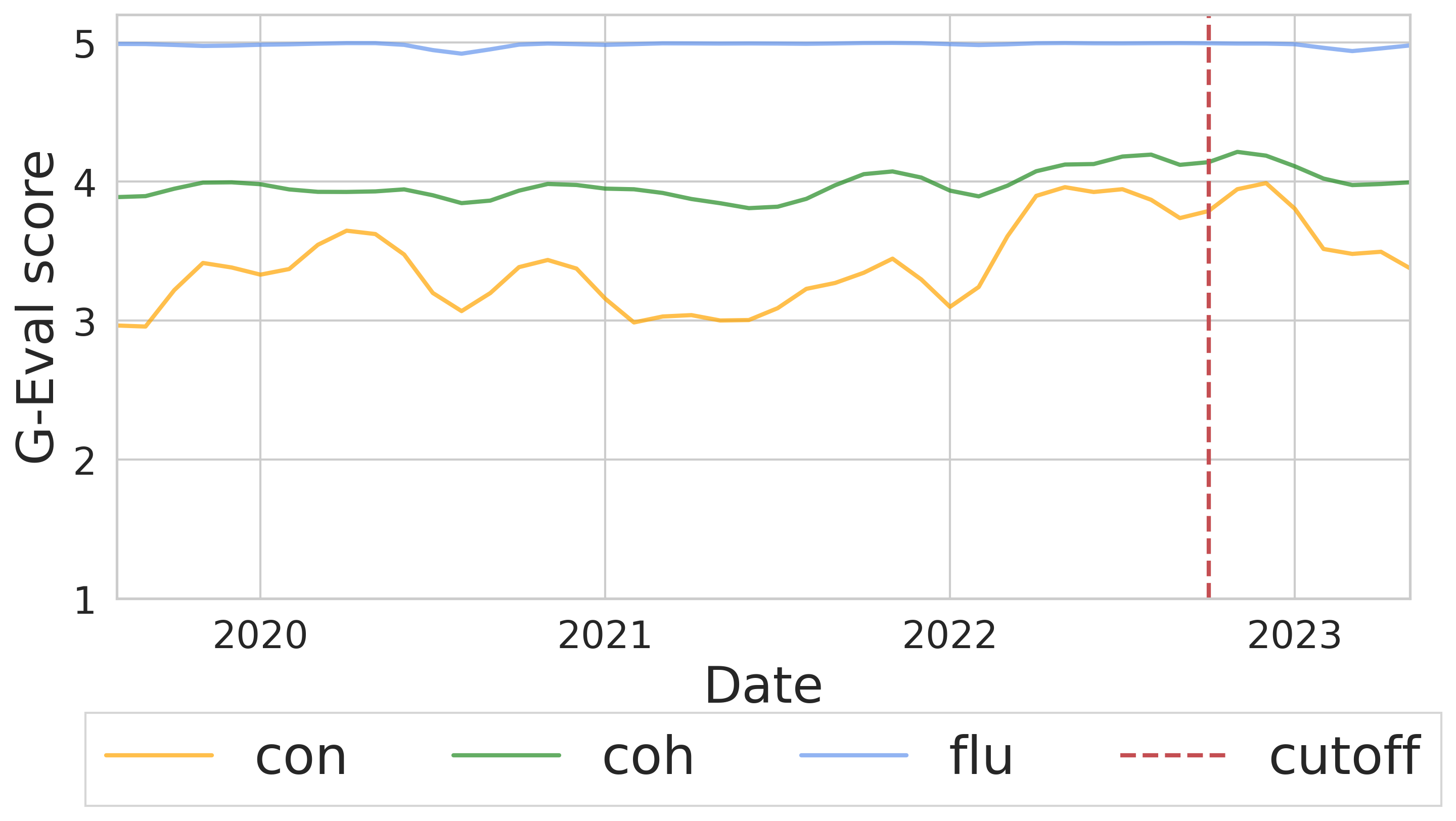}
        \caption{\gpt -- Synt (2019-2024)}
    \end{subfigure}
    \begin{subfigure}{0.32\textwidth}  
        \centering
        \includegraphics[width=\linewidth]{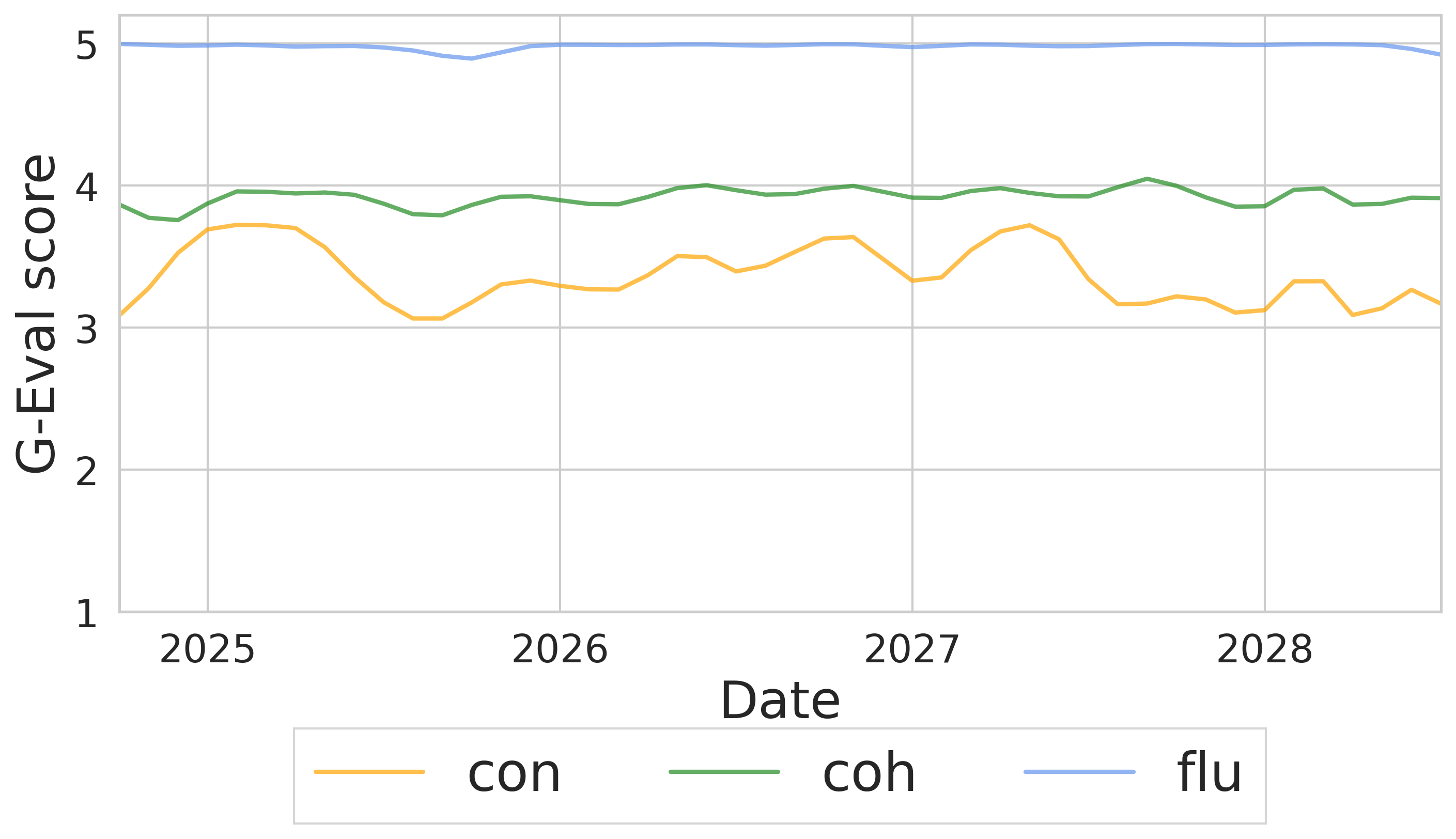}
        \caption{\gpt -- Synt (2024-2029)}
    \end{subfigure}\\
            \begin{subfigure}{0.32\textwidth}  
        \centering
        \includegraphics[width=\linewidth]{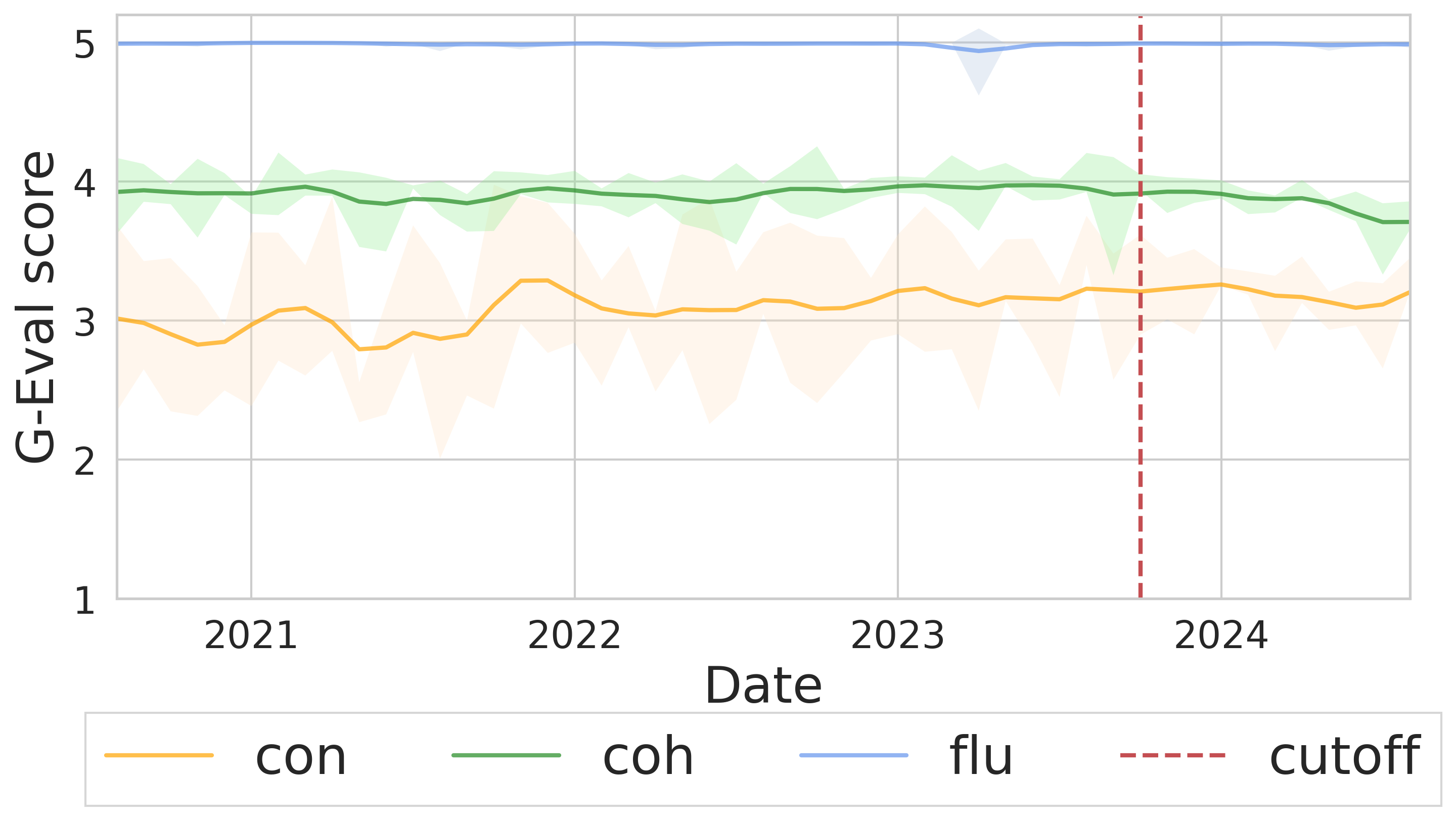}
        \caption{GPT-4o-mini -- real}
    \end{subfigure}
    \begin{subfigure}{0.32\textwidth}  
        \centering
        \includegraphics[width=\linewidth]{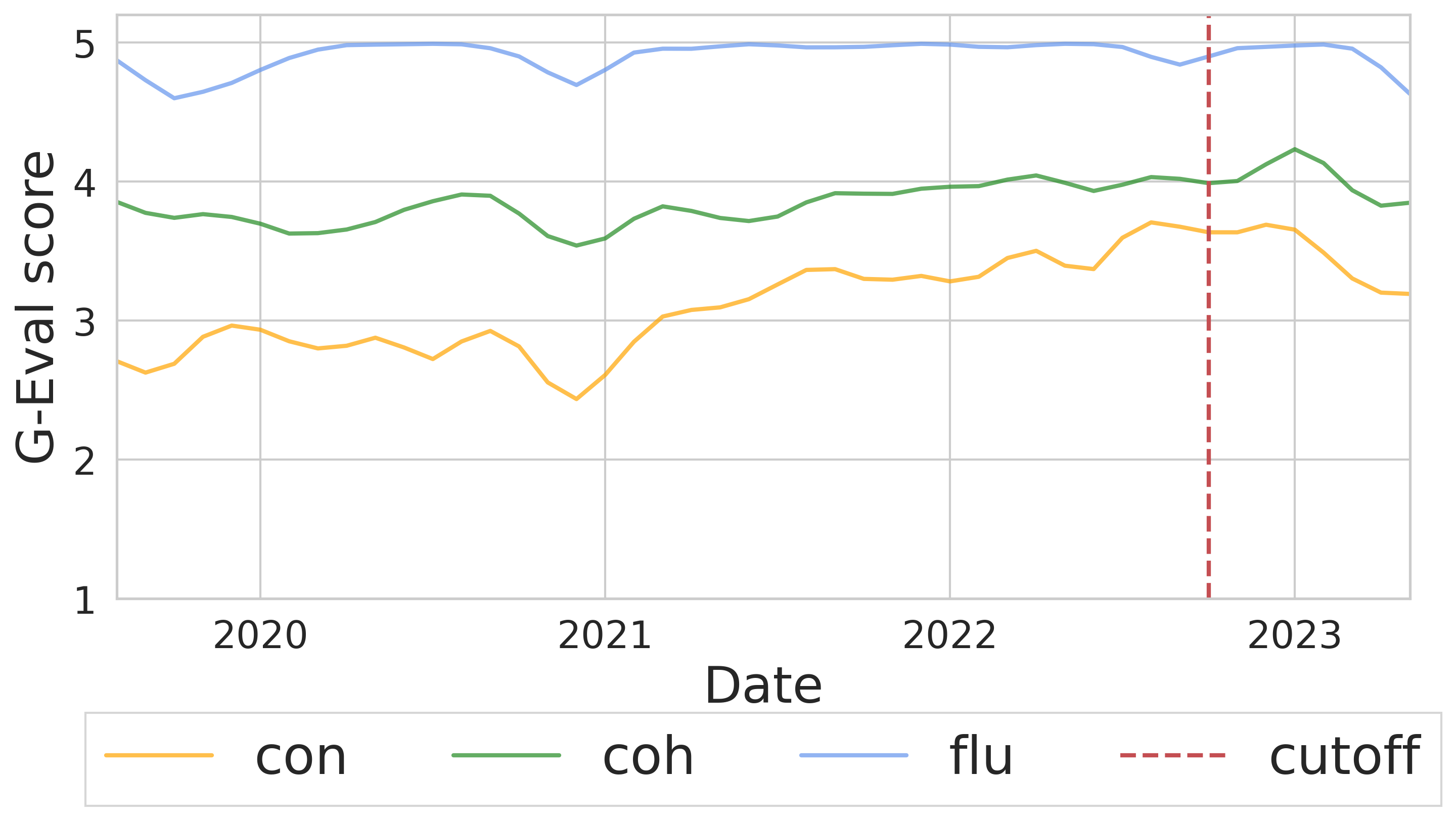}
        \caption{\gptm -- Synt (2019-2024)}
    \end{subfigure}
    \begin{subfigure}{0.32\textwidth}  
        \centering
        \includegraphics[width=\linewidth]{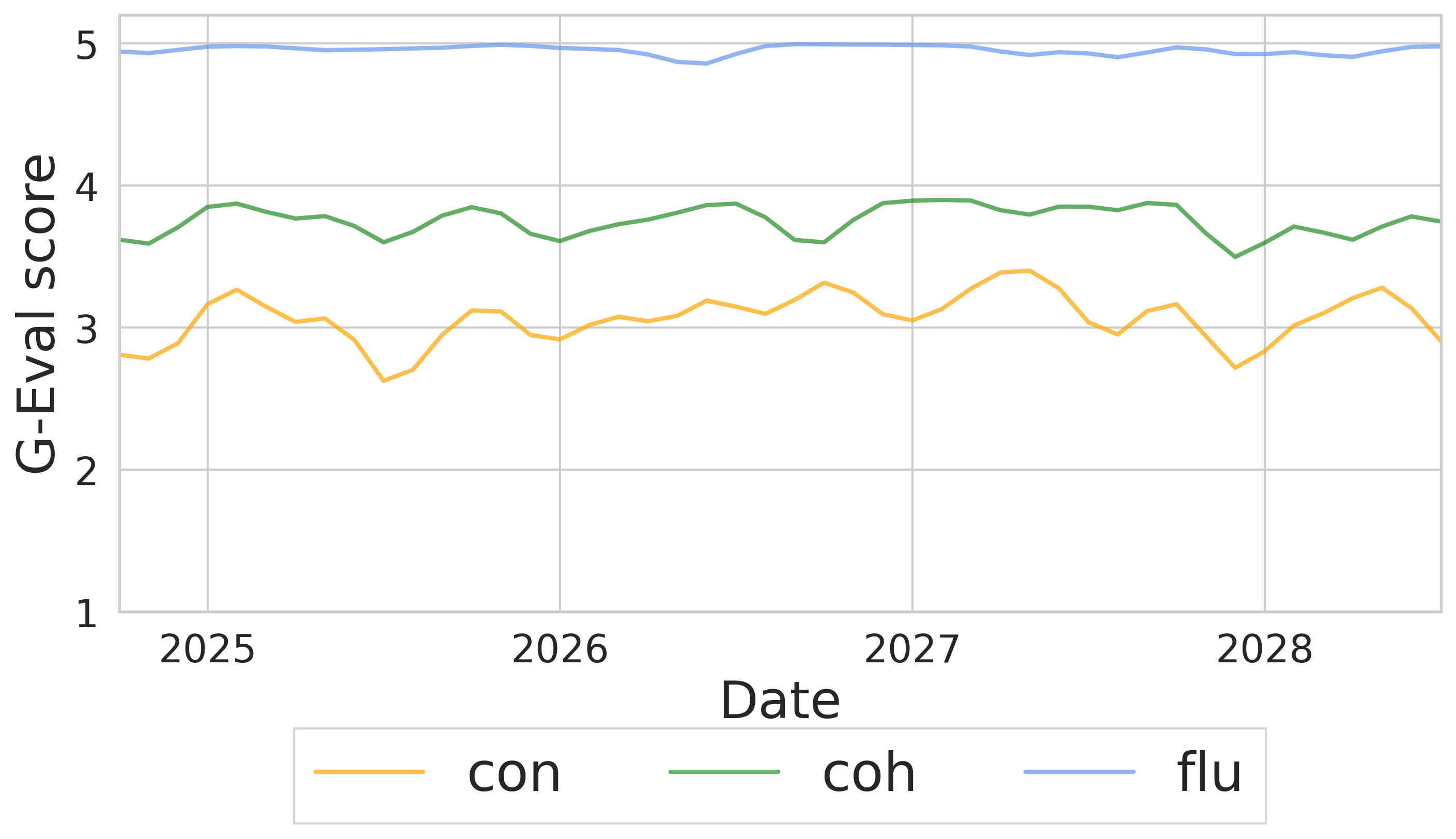}
        \caption{\gptm -- Synt (2024-2029)}
    \end{subfigure}\\
    \begin{subfigure}{0.32\textwidth}  
        \centering
        \includegraphics[width=\linewidth]{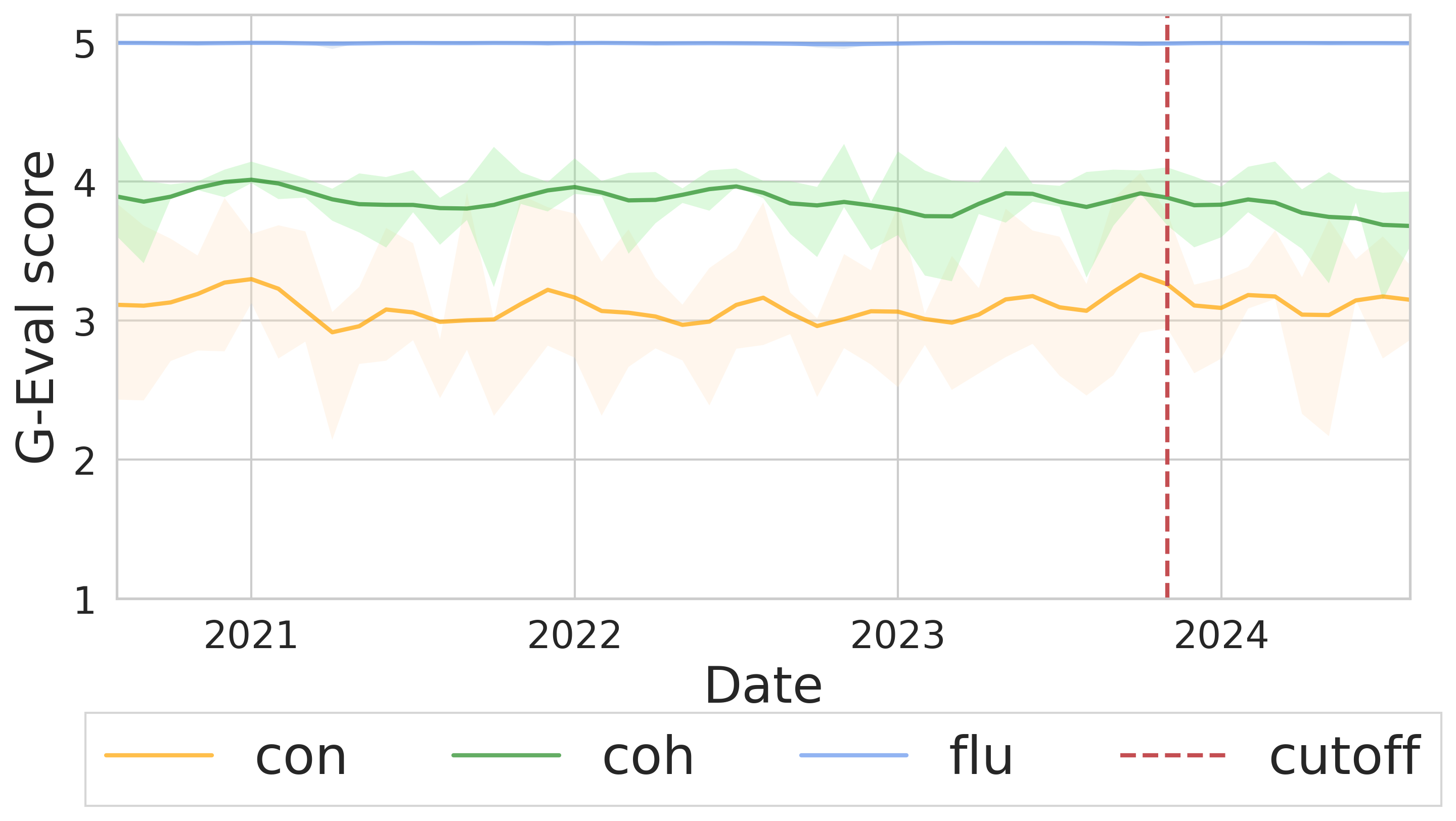}
        \caption{Gemini -- real}
    \end{subfigure}
    \begin{subfigure}{0.32\textwidth}  
        \centering
        \includegraphics[width=\linewidth]{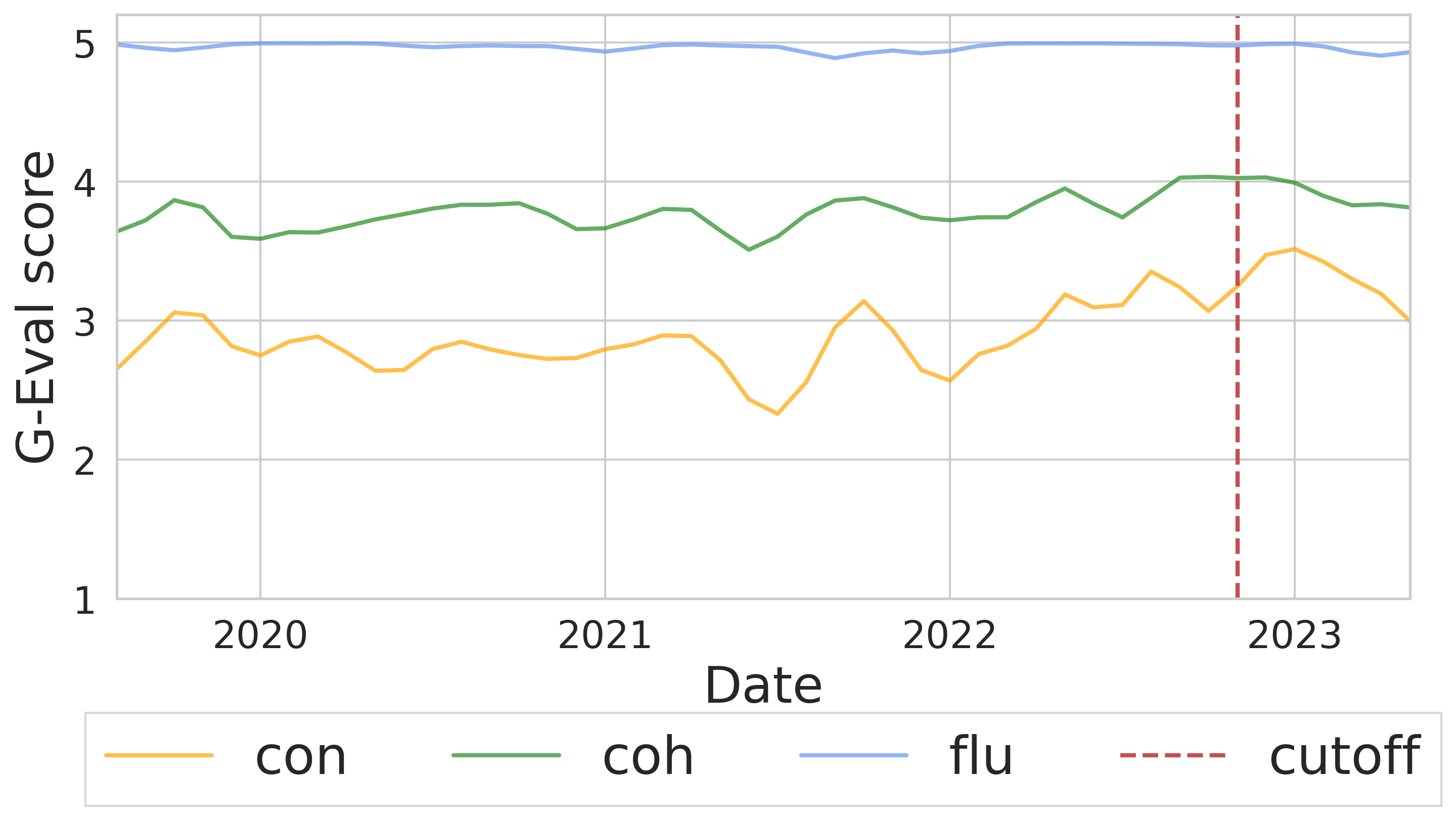}
        \caption{Gemini -- Synt (2019-2024)}
    \end{subfigure}
    \begin{subfigure}{0.32\textwidth}  
        \centering
        \includegraphics[width=\linewidth]{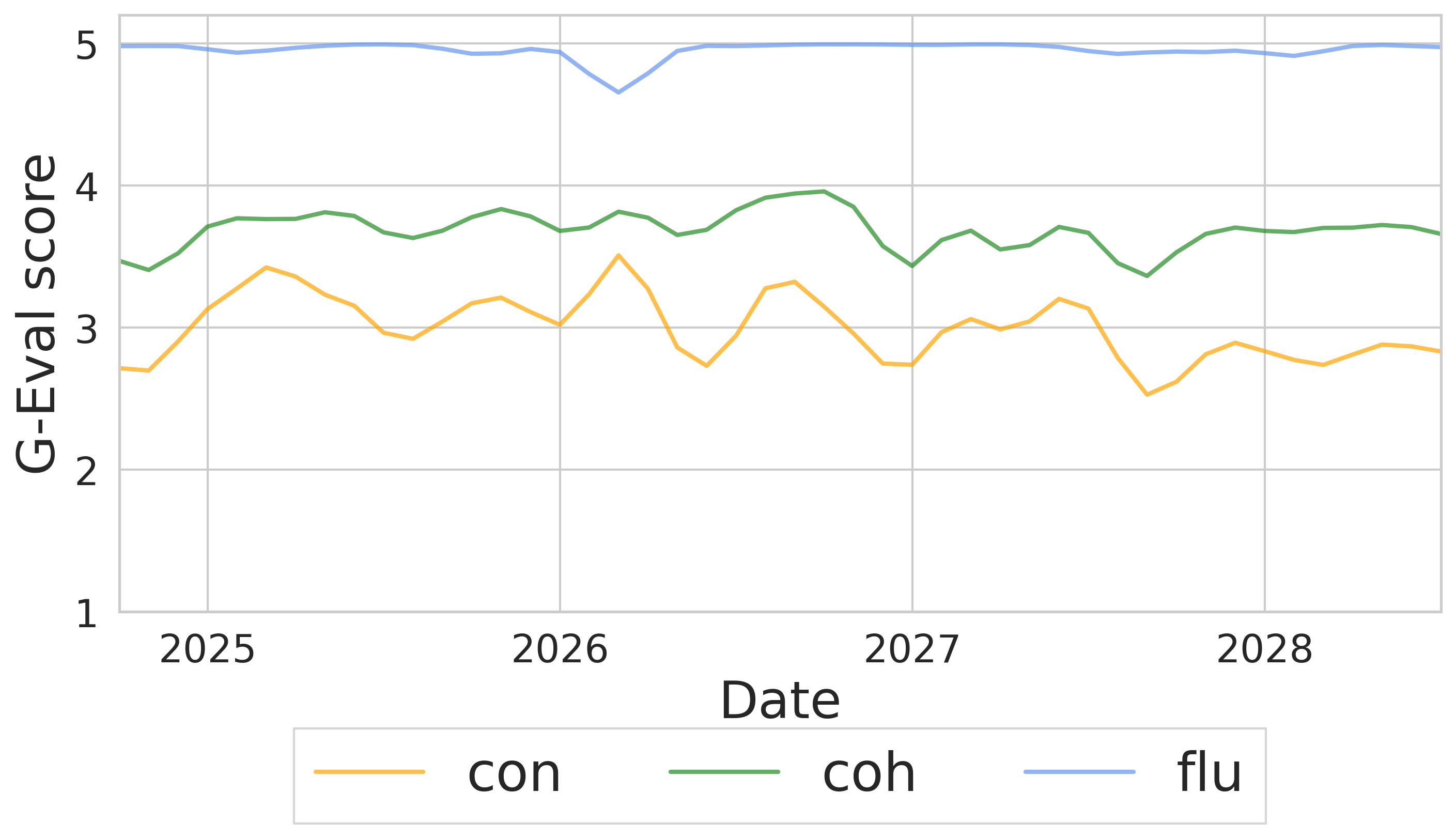}
        \caption{Gemini -- Synt (2024-2029)}
    \end{subfigure}\\
    
        \begin{subfigure}{0.32\textwidth}  
        \centering
        \includegraphics[width=\linewidth]{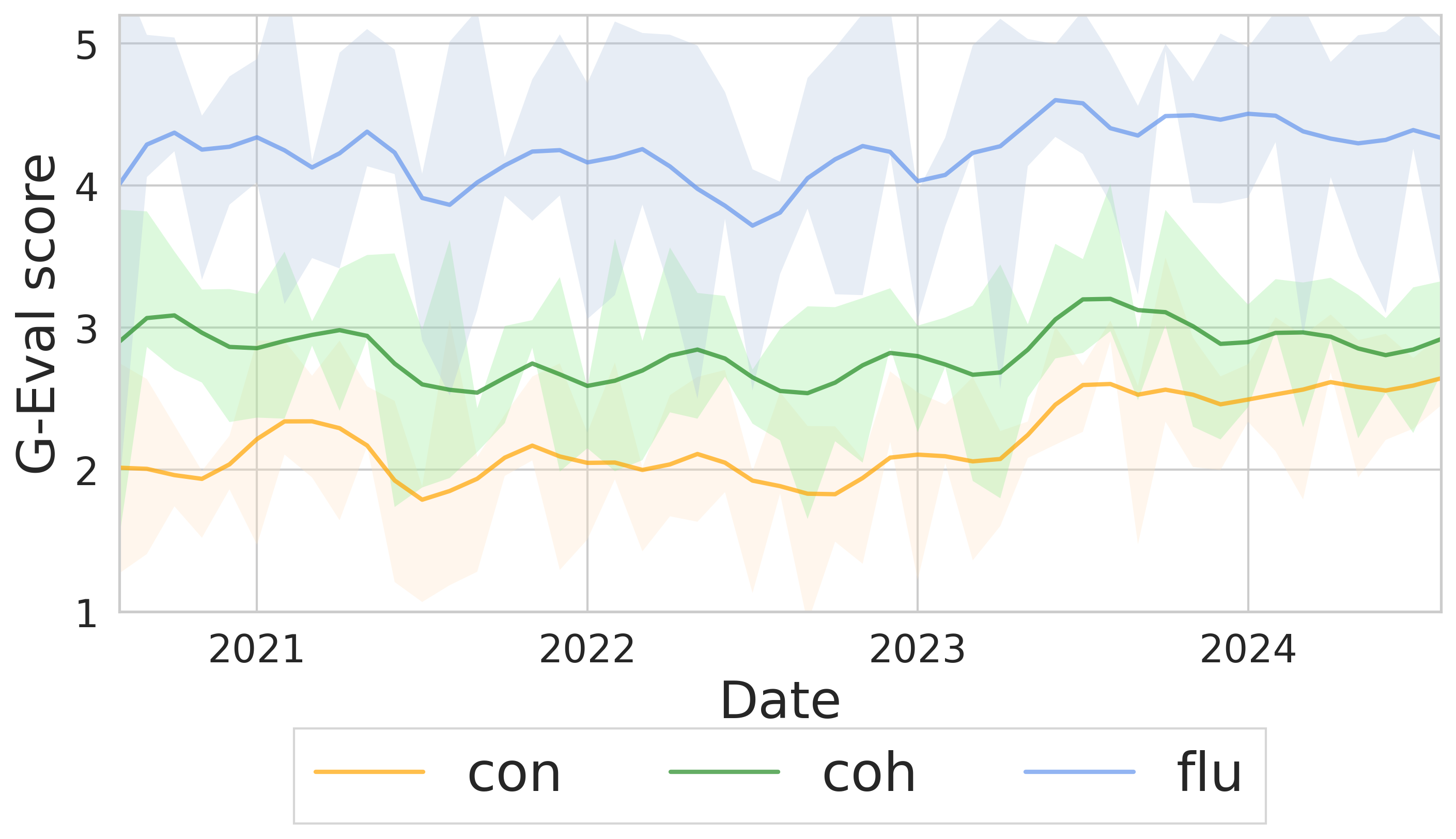}
        \caption{\llama -- real}
    \end{subfigure}
    \begin{subfigure}{0.32\textwidth}  
        \centering
        \includegraphics[width=\linewidth]{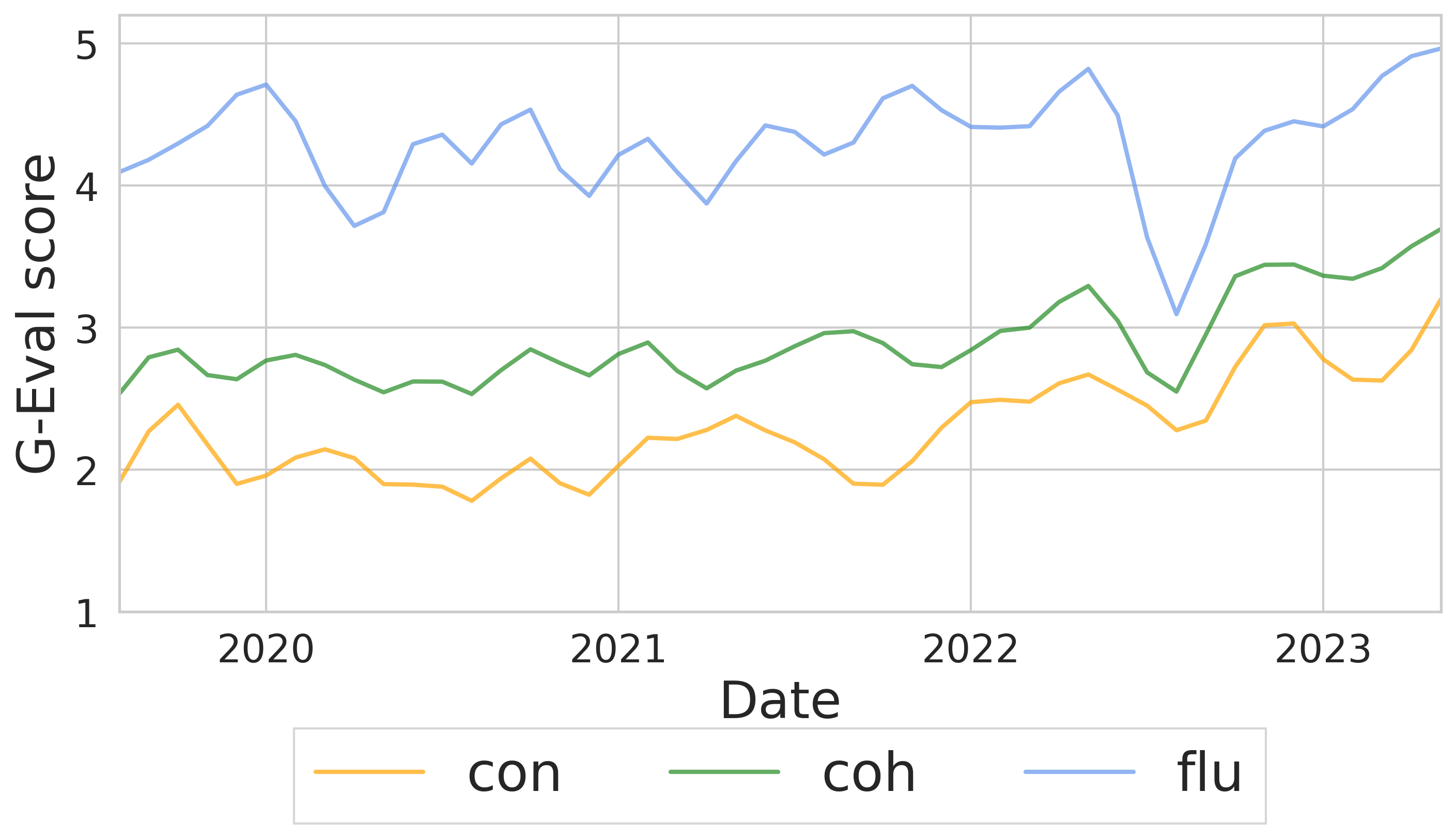}
        \caption{\llama -- Synt (2019-2024)}
    \end{subfigure}
    \begin{subfigure}{0.32\textwidth}  
        \centering
        \includegraphics[width=\linewidth]{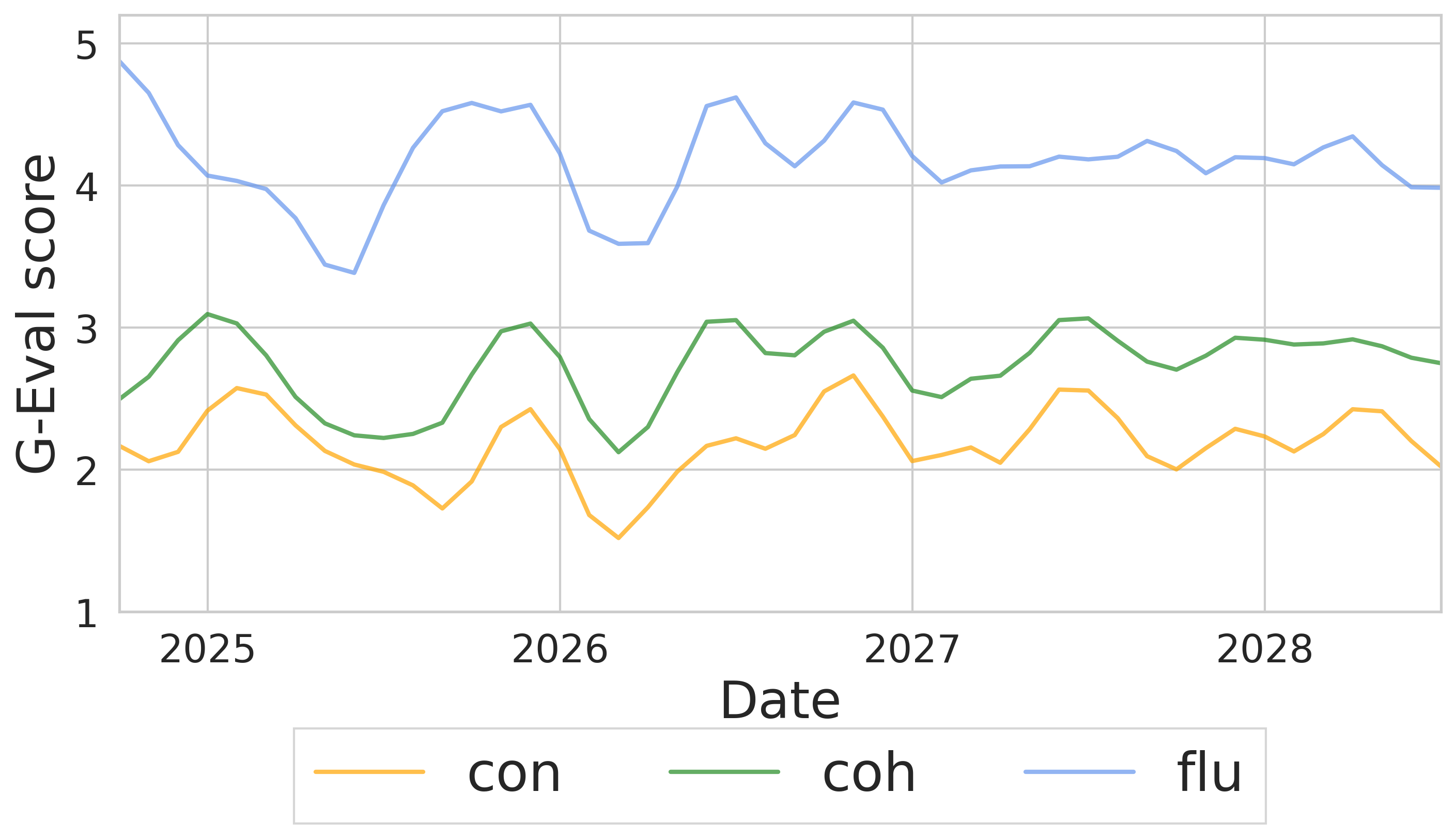}
        \caption{\llama -- Synt (2024-2029)}
    \end{subfigure}\\
    \caption{TI (plots) reports G-EVAl scores over time.}
    \label{fig:geval-time-ti-plots}
\end{figure*}

In this section we present the evolution of G-Eval scores over time for all models and categories of reports. Figure \ref{fig:geval-time-short} shows that the Consistency score, and to some extent Coherence, tends to decrease for GPT-4o and GPT-4o-mini after the cutoff date of the model training data. This is not observed in Gemini and Phi-3. For the reports generated from synthetic data, there seems to be a slight improvement in Consistency score over time for the data covering the same period of time as real data, while for the future data, scores look more constant over time. 
Figures \ref{fig:geval-time-ti} and \ref{fig:geval-time-ti-plots} show that there is a tendency of coference to decline after the cutoff date in most models, while only GPT-4o-mini shows a clear decline in consistency in Reports with TI after the cutoff date. 

\section{Report Examples}
\label{app:repexampels}
\begin{figure*}[h]
  \centering
    \includegraphics[width=\linewidth, trim={0cm 0cm 1cm 0cm},clip]{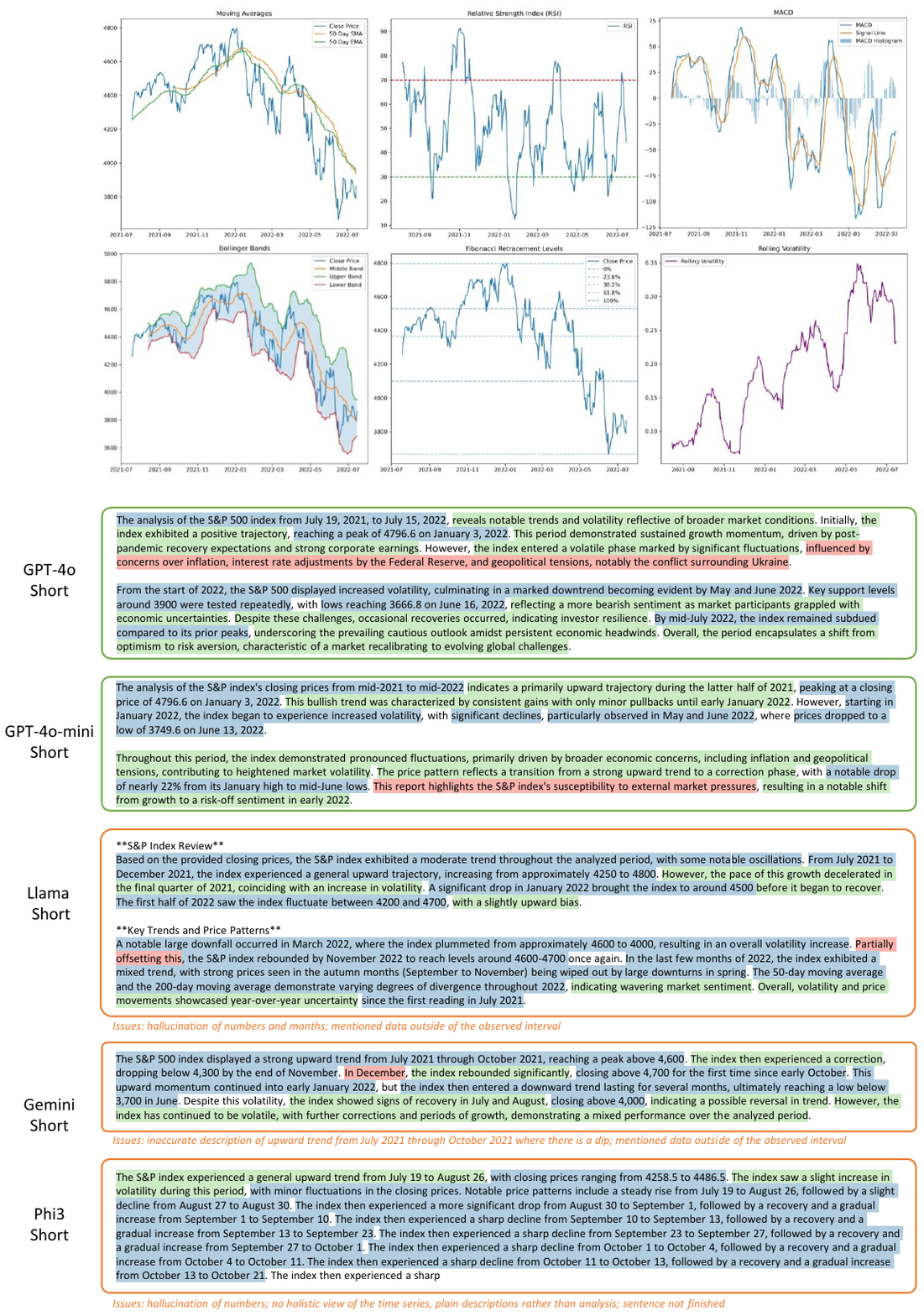}
    \caption{Example reports generated on real data: S\&P Index during 2021-07-17 and 2022-07-17. Reports in \textcolor{YellowGreen}{green} boxes are of good quality; reports in \textcolor{YellowOrange}{orange} boxes are problematic with identified issues. Colored highlights indicate \textbf{\drc{Direct Reference (blue)}}, \textbf{\fic{Financial Interpretation (green)}}, and \textbf{\ekc{External Knowledge (red)}} using our proposed highlighting system.}
    \label{fig:qual_real}
\end{figure*}
\begin{figure*}[h]
  \centering
    \includegraphics[width=\linewidth, trim={0cm 0cm 1cm 0cm},clip]{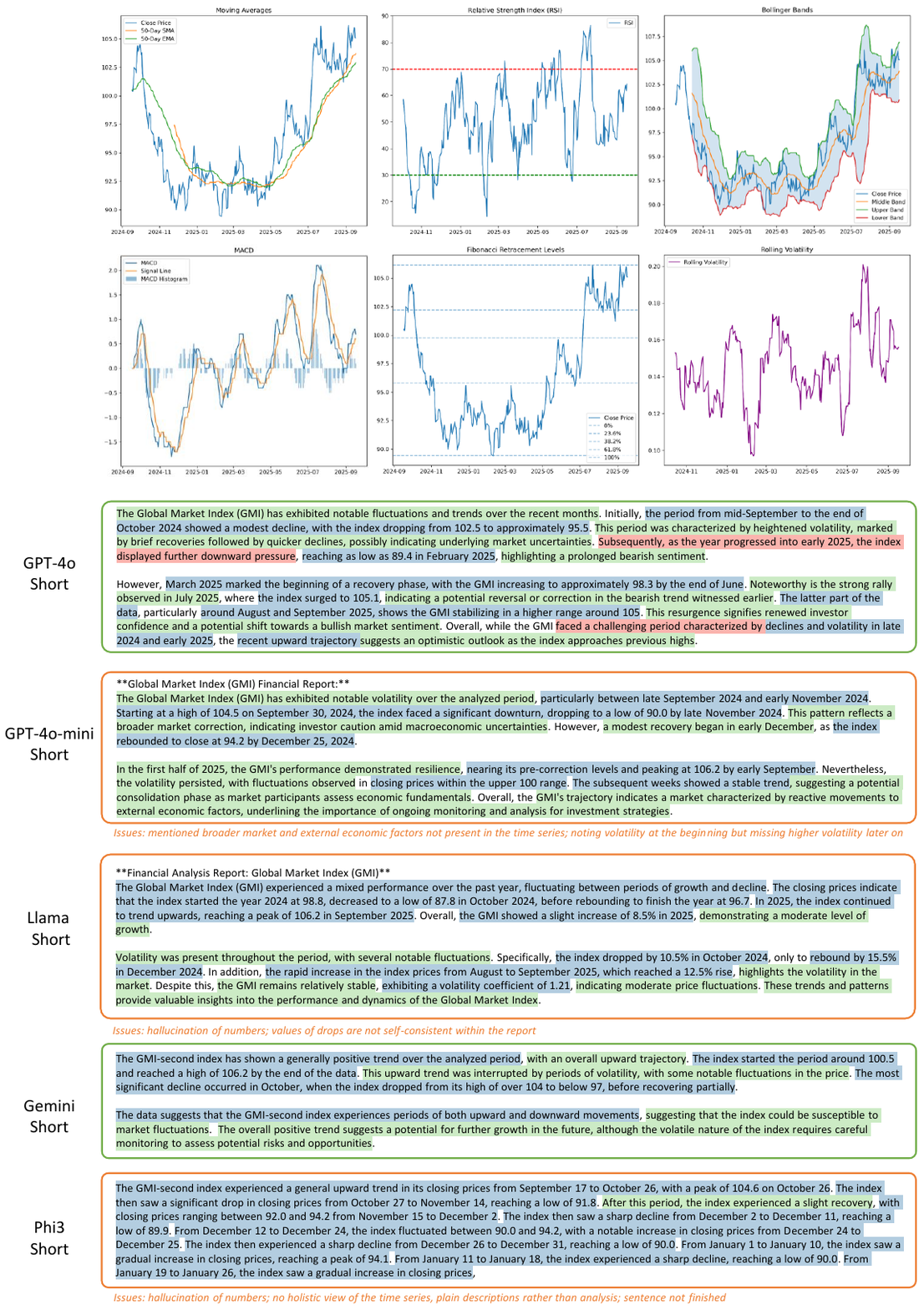}
    \caption{Example reports generated on synthetic data: GMI Index during 2024-09-17 and 2025-09-17. Reports in \textcolor{YellowGreen}{green} boxes are of good quality; reports in \textcolor{YellowOrange}{orange} boxes are problematic with identified issues. Colored highlights indicate \textbf{\drc{Direct Reference (blue)}}, \textbf{\fic{Financial Interpretation (green)}}, and \textbf{\ekc{External Knowledge (red)}} using our proposed highlighting system.}
    \label{fig:qual_syn}
\end{figure*}
\end{document}